# Unsupervised Language Acquisition: Theory and Practice

**Alexander Simon Clark**

# Declaration

I hereby declare that this thesis has not been submitted, either in the same or different form, to this or any other university for a degree.

Signature:

## Acknowledgements

First, I would like to thank Bill Keller, for his supervision over the past three years. I would like to thank my colleagues at ISSCO for making me welcome and especially Susan Armstrong for giving me the job in the first place, and various people for helpful comments and discussions including Chris Manning, Dan Klein, Eric Gaussier, Nicola Cancedda, Franck Thollard, Andrei Popescu-Beilis, Menno van Zaanen and numerous other people including Sonia Halimi for checking my Arabic. I would also like to thank Gerald Gazdar and Geoffrey Sampson for their helpful comments as part of my thesis committee.

I would also like to thank all of the people that have worked on the various software packages and operating systems that I have used, such as LaTeX, Gnu/Linux, and the gcc compiler, as well as the people who have worked on the preparation of the various corpora I have used: in particular I am grateful to John McCarthy and Ramin Nakisa for allowing me to use their painstakingly collected data sets for Arabic. Finally I would like to thank my wife, Dr Clare Hornsby, for her unfailing support and encouragement (in spite of her complete lack of understanding of the subject) and my children, Lily and Francis, for putting up with my absences both physical and mental, and showing me how language acquisition is really done – over-regularisation of past tense verbs included.

# Unsupervised Language Acquisition:
# Theory and Practice

**Alexander Simon Clark**

## Abstract

In this thesis I present various algorithms for the unsupervised machine learning of aspects of natural languages using a variety of statistical models. The scientific object of the work is to examine the validity of the so-called Argument from the Poverty of the Stimulus advanced in favour of the proposition that humans have language-specific innate knowledge. I start by examining an *a priori* argument based on Gold's theorem, that purports to prove that natural languages cannot be learned, and some formal issues related to the choice of statistical grammars rather than symbolic grammars. I present three novel algorithms for learning various parts of natural languages: first, an algorithm for the induction of syntactic categories from unlabelled text using distributional information, that can deal with ambiguous and rare words; secondly, a set of algorithms for learning morphological processes in a variety of languages, including languages such as Arabic with non-concatenative morphology; thirdly an algorithm for the unsupervised induction of a context-free grammar from tagged text. I carefully examine the interaction between the various components, and show how these algorithms can form the basis for a empiricist model of language acquisition. I therefore conclude that the Argument from the Poverty of the Stimulus is unsupported by the evidence.

# Contents

# List of Figures

# List of Tables

It is an established opinion amongst some men, that there are in the understanding certain *innate principles*; some primary notions, ... characters, as it were stamped upon the mind of man; which the soul receives in its very first being, and brings into the world with it. It would be sufficient to convince unprejudiced readers of the falseness of this supposition, if I should only show (as I hope I shall in the following parts of this Discourse) how men, barely by the use of their natural faculties, may attain to all the knowledge they have, without the help of any innate impressions; and may arrive at certainty, without any such original notions or principles.

Locke (1690, Book 1, Chapter i)

Rule 1: No more causes of natural things should be admitted than are both true and sufficient to explain their phenomena. As the philosophers say: Nature does nothing in vain, and more causes are in vain when fewer suffice. For nature is simple and does not indulge in the luxury of superfluous causes.

Newton (1687, Rules for the Study of Natural Philosophy)

# Chapter 1

# Introduction

## 1.1   The Scientific Question

Linguistic nativism or the innateness hypothesis is the claim, advanced by Chomsky (1986) and Pinker (1994) amongst others, that human beings are endowed with an innately, presumably genetically, specified domain specific knowledge of language. This knowledge is *tacit*, that is to say not accessible to conscious thought, and it specifies in some detail the nature of possible human languages, including a set of syntactic categories, a set of possible phrase structure rules, constraints on admissible transformations and so on. The primary argument for this bold hypothesis is the so-called *Argument from the Poverty of the Stimulus* (APS), that the linguistic input or evidence available to the infant child is so impoverished and degenerate that no general, domain-independent learning algorithm could possibly learn a plausible grammar without assistance.

An obvious refutation of this argument is to demonstrate the existence of an algorithm that can learn a reasonable grammar, from that amount of data. It is that issue that this thesis is intended to study. Nonetheless the algorithms presented here are I hope of general interest as pieces of computational linguistics or machine learning research.

This question is in one sense thoroughly Chomskyan: I fully accept his characterisation of linguistics as, ultimately, a branch of psychology, though for the moment it relies on very different sorts of evidence; I fully accept his argument for complete formality in linguistics, a formality that computer modelling both requires and enforces; I fully accept the idea that one of the central problems of linguistics is how to explain the fact that children manage to learn language in the circumstances that they do. On the other hand, there are many areas in which this work is not so congenial to followers of the Chomskyan paradigm. First, the work here is fully empirical; it is concerned with authentic language, rather than artificial examples. Secondly, it eschews the use of unnecessary hidden entities; far from considering this as the hallmark of a good scientific theory, the unnecessary proliferation of unobservable variables renders the link between theory and surface tenuous and unstable. Thirdly, the question I am examining is one that has been considered established beyond all reasonable doubt; and no proud home-owner likes having his foundations examined suspiciously by an outsider.

## 1.2   Basic assumptions

Throughout the thesis I make several basic assumptions. First, I concern myself as much as possible with authentic language data. Though in the field of Machine Learning, much work is done on small controlled data sets, which clearly facilitate the empirical evaluation of different learning algorithms, in the context of the purpose of this thesis, I feel this would be misguided. One of the few lessons that has been learnt from the first 30 years of Artificial Intelligence is that techniques that may work well on small *toy* domains, in general do not scale up well to realistically sized tasks. Secondly, and as a result of this first assumptiom, I shall use *statistical* models. There will inevitably be noise in the data I use: errors, misspellings, quotations in foreign languages, jokes, fragments of poems and all sorts of other deviant or ill-formed uses of language. It is essential to use techniques that will be robust in the presence of this noise; this effectively rules out a large range of different non-statistical techniques. Thirdly, I have also decided to focus on a complete language learning system: not just building a set of programs that can learn aspects of natural lan-guages but also considering how the various modules can be stuck together, to insure that nothing falls between the gaps.

Pinker (1996, p. xv) puts it well

> Anyone who has ordered a computer system à la carte from a mail-order catalogue knows the perils. Mysterious incompatibilities among the components appear and multiply, and calls to the various "help" lines just bring to mind a circle of technicians, each pointing clockwise. The same danger can arise from piecemeal research on a scientific topic as complex as language acquisition in children. What looks irresistible in a single component . . . can crash the system when plugged in with the others . . . .

Fourthly, every module I propose has a probabilistically sound interpretation: this allows the principled composition of the various parts together, and the possible global optimisation of the different parameter sets.

## 1.3 Outline of the thesis.

I will now give an outline of the rest of the thesis. Where possible I have tried to make the chapters fairly self-contained.

In Chapter 2 I discuss the innateness hypothesis and the APS in more detail, and discuss the possibility of refuting it by demonstrating the existence of a suitable algorithm. I discuss various constraints on the algorithm and on the data it operates on, and various arguments relating to this. In Chapter 3 I examine the use of machine learning techniques in natural language processing, and introduce the particular techniques I will be using in this thesis. I discuss the various levels of language such as morphology, syntax and so on, and machine learning of those areas that are not covered elsewhere in the thesis. I will also discuss the overall architecture of a language learner.

In Chapter 4 I discuss two formal issues. First, I discuss the relevance of the formal theory of learnability, as introduced by Gold (1967). In the process I shall show that if we assume that language is generated by an ergodic stochastic process (quite a weak assumption) then more or less all of the classes of languages that are linguistically interesting are identifiable in the limit *with probability one*. Secondly I consider the algebraic relationship between statistical language models, i.e. probability distributions, and formal languages, in an attempt to show that they are more similar than is generally thought.

There then follow three chapters that constitute the three principal contributions of this thesis. These are largely self-contained as pieces of statistical natural language processing, though some of the design choices and self-imposed limitations may seem eccentric if not considered within the context of the overall goal of the thesis. The first, Chapter 5 presents an algorithm for inducing a set of syntactic categories from a completely raw corpus. Though other such algorithms have been presented before, this is the first that handles ambiguity and rare words adequately. Then in Chapter 6 I present an algorithm for learning morphology in a variety of languages, English, German and Arabic and in a variety of different situations, supervised and partially supervised. This algorithm is based on using an Expectation-Maximisation based training algorithm for stochastic finite-state transducers. Then in Chapter 7 I present an unsupervised algorithm for learning syntax from tagged corpora. I present a novel criterion for determining when a sequence of tags is a constituent based on the mutual information between the symbol that occurs before it and the symbol that occurs after it. I show how this relates to the entropy of a random variable associated with a probabilistic context free grammar, and discuss the relationship between this criterion and various other criteria that have been proposed. In Chapter 8 I examine the overall result of these techniques and discuss the plausibility of the argument from the poverty of the stimulus, in the light of the techniques presented here. I include an index, primarily of the various terms and acronyms I introduce, and a few supplementary appendices.

## 1.4 Prerequisites

The intended audience for this thesis are researchers in Natural Language Processing who are interested in machine learning of natural language, or in the implications for cognitive science of some recent research. I also hope that it will be accessible to cognitive scientists interested in empiricist theories of language acquisition, though some of the technical detail may be opaque. I assume a familiarity with standard techniques of statistical estimation and machine learning. In particular Chapter 6 assumes an acquaintance with the forward-backward algorithm for Hidden Markov Models. Where I have used mathematics outside the normal range used in statistical NLP, I explain it in more detail: for example, I use some elementary measure theory in Chapter 4.

## 1.5 Literature review

Since I am trying to cover several different areas of natural language, and applying techniques from machine learning, bioinformatics and computational linguistics, my survey of previous work has necessarily been slightly cursory. I have tried to include references to basic works in each field; there are probably some grievous omissions and errors of attribution that I apologise for in advance.

## 1.6 Bibliographic note

Parts of the work presented here have been previously published, in much shorter versions. A version of Chapter 5 appeared as Clark (2000), and part of Chapter 6 appeared as Clark (2001a). A version of Chapter 7 appeared as Clark (2001c)

## 1.7 Acknowledgments

Part of this research was done while I was employed at ISSCO at the University of Geneva working on the European TMR Network *Learning Computational Grammars*.

# Chapter 2

# Nativism and the Argument from the Poverty of the Stimulus

## 2.1   Introduction

In this chapter, I shall discuss the scientific question that the work presented in this thesis is intended to address. I shall start by discussing, in Section 2.2, the Innateness Hypothesis in its various forms, and then in Section 2.3 I shall discuss the various forms of the Argument from the Poverty of the Stimulus (APS). Section 2.4 discusses how an unsupervised learning algorithm of a particular type could be a refutation of the APS. Section 2.5 discusses various objections to this approach, and how in some cases they suggest avenues for future research.

## 2.2   Nativism

The *Innateness Hypothesis* (IH) is the hypothesis that human beings have innately specified, domain specific knowledge in several areas, in particular language. This hypothesis, which until quite recently was seen as a philosophical question, is now firmly within the field of science, in particular cognitive science and must be settled empirically. Ultimately this question will be settled, one would expect, by neurological evidence of some kind, but at the moment we must be content with more indirect evidence from psycholinguistics, behavioural psychology and so on. I will concern myself exclusively with the issue of language, and neglect other hypothesised areas of innateness. The key issue here is that of the domain-specificity of this endowment; clearly the ability to learn language must have some genetic component since humans learn languages and other animals cannot, though dolphins (Tyack, 1986), apes (Brakke & Savage-Rumbaugh, 1996), and perhaps honey bees (see Wenner and Wells (1990) for a dissenting view) have some rudimentary linguistic abilities. The innateness hypothesis is that these innate abilities that humans have are *domain-specific*, that is in this case specific to language, and that they include highly detailed linguistic knowledge. Note that this is logically separate from the issue of localisation, the claim that "our ability to process language is localized to specific regions of the brain" (Bates, 1994), a claim that appears to be uncontroversial.

A first comment to make is that many species do in fact have domain-specific, innately specified abilities or behaviours: common examples (Elman, Bates, Johnson, Karmiloff-Smith, Parisi, & Plunkett, 1996) are spiders weaving complex webs on their first attempts and ungulates, such as deer, running within a few minutes of birth. In these cases nobody denies that these abilities are innate. The compelling argument is that in the absence of available experience these abilities must pre-exist in the brain of the animal in question. The situation in humans is quite different – newborn children exhibit few complex behaviours immediately after birth. Nevertheless it is claimed that in at least one field, language, particularly syntax, children acquire an ability without being exposed to sufficient stimulus, and that therefore they must have a pre-existing domain-specific innate structure that partially specifies the structure of their knowledge of language (Chomsky, 1986; Pinker, 1994).

Some have related this debate to the philosophies of Descartes and Locke: I can only follow, with perhaps more sincerity, Quine (1969) when he says:

> The connection between this indisputable point about language, on the one hand, and the disagreements of seventeenth-century philosophers on the other, is a scholarly matter on which I have no interesting opinion.

No-one denies that the brain has some innately specified structure: As Quine (1975, p.200) says,

> For, whatever we may make of Locke, the behaviorist is knowingly and cheerfully up to his neck in innate mechanisms of learning-readiness. The very reinforcement and extinction of responses, so central to behaviorism, depends on prior inequalities in the subject's qualitative spacing, so to speak, of stimulations.

The argument is between those that feel that the human brain has many individual, innately specified, domain-specific modules and those who maintain that the higher cognitive faculties develop in response to experience, and that in particular the language faculty is merely one aspect of 'general human intelligence.'(Piattelli-Palmarini, 1980)

Chomsky (1986, p. 4.) puts the central question very clearly:

> Is this a distinct "language faculty" with specific structure and properties, or as some believe, is it the case that humans acquire language merely by applying generalised learning mechanisms of some sort, perhaps with greater efficiency or scope than other organisms?

He continues

> We try to determine what is the system of knowledge that has been attained and what properties must be attributed to the initial/state of the mind brain to account for its attainment. Insofar as these properties are language-specific, either individually or in the way they are organised and composed, there is a distinct language faculty.

Cowie (1999) tries to maintain a distinction between the innateness hypothesis and domain-specificity, claiming that they are logically independent. This is slightly unusual, and not particularly relevant. As mentioned before, I do not think that the innateness of general cognitive capacities is controversial; Cowie claims that Skinnerian behaviorists do not accept this but her bibliography does not list Skinner (1957) – only Chomsky's review of it (Chomsky, 1959). In fact the only person who seems to reject domain-general innateness is Cowie herself. Chomsky (2000, p.66) says:

> . . . people who are supposed to be defenders of "the innateness hypothesis" do not defend the hypothesis or even use the phrase, because there is no such general hypothesis; rather, only specific hypotheses about the innate resources of the mind, in particular, its language faculty.

I am in full agreement with this; I use IH to refer to the specific claim about the existence of a highly domain-specific innate language faculty.

There are a number of philosophical issues associated with this debate that I shall not address at all: I shall assume that the notion of innateness is theoretically clear; I shall not discuss the difference between innate mechanisms and innate ideas; and I shall use the word knowledge to refer to whatever *mental machinery*[1] the language user has that allows him to use a language, without allowing that anything substantive follows from the choice of that word.

---

[1]Geoffrey Sampson, p.c.

## 2.3    The Argument from the Poverty of the Stimulus

The principal argument for IH, both historically and logically, is the *Argument from the Poverty of the Stimulus*(APS). In spite of its fame, it is very difficult to find a clear statement of it as Pullum and Scholz (2001) observe:

> The one thing that is clear about the argument from the poverty of the stimulus is what its conclusion is supposed to be: it is supposed to show that human infants are equipped with innate mental mechanisms specifically for assisting in the language acquisition process – in short that the facts about human language acquisition support 'nativist' rather than 'empiricist' epistemological views. What is not clear at all is the structure of the reasoning that is supposed to support this conclusion. Instead of clarifying the reasoning, each successive writer on this topic shakes together an idiosyncratic cocktail of claims about children's learning of language and claims that nativism is thereby supported.

In brief, the APS is the argument that the linguistic experience of the child is not sufficiently rich to allow a child, or anyone else, to learn the grammar of the language. The data that the child has access to, the *primary linguistic data* PLD, just does not contain enough information to allow the learning to proceed without other sources of information. Since children do in fact learn the language, the conclusion is drawn that the child must have access to some other source of information that helps or constrains the search for the correct grammar, and that this must be innate. This argument is clear and valid. It rests, however on the premise that there are no general learning algorithms that can learn grammars from the sort of linguistic evidence that children are exposed to. The inference from this seems quite solid to me. The question that must be examined, therefore, is whether there are general-purpose learning strategies that can be applied to the sort of data that is available that will produce a plausible grammar. If there are such algorithms, then the APS is false, since one of its premises is false; if on the other hand there are not, then the APS is a very strong argument for the Innateness Hypothesis.

Some are quite vehement in their support of this issue. Chomsky (1965) says

> It seems to have been demonstrated beyond all reasonable doubt that, quite apart from any question of feasibility, methods of the sort that have been studied in taxonomic linguistics are intrinsically incapable of yielding the systems of grammatical knowledge that must be attributed to the speaker of a language.

and in (Chomsky, 1988b, p.110)

> The common assumption that a general learning theory does exist, seems to me dubious, unargued, and without any empirical support or plausibility at the moment.

This is quite a radical proposal: most cognitive scientists accept that there is some general-purpose learning ability, whether or not there are also domain-specific modules. For example, people can learn to play chess: without hypothesizing a chess-specific learning module, or more generally a board-game specific module, it seems difficult to account for this without some sort of general learning ability. Nonetheless, Chomsky is right in insisting that it must be argued for. As Pinker (1996, p.359) states:

| | Nativist | Empiricist |
|---|---|---|
| Initial State | Richly structured | Unstructured |
| Primary Linguistic Data | Poor | Rich |
| Learning Algorithms | Weak, domain-specific | Powerful, general-purpose |
| Final State | Deep, Richly Structured | Shallow |

Table 2.1: Comparison of Nativist and Empiricist Theories of Language Acquisition

> It is now incumbent on skeptics of nativist learning theories to propose explicit theories that do full justice to the process whereby complex systems of linguistic rules are acquired.

This thesis can be thought of as a response to that challenge.

I have presented the general version of the APS: many specific claims have been made about particular constructions in English that are supposedly unlearnable. These are discussed in some depth in (Sampson, 1989, 1997) and (Pullum, 1996; Pullum & Scholz, 2001). Proponents of the APS have claimed that particular constructions such as the alleged non-occurrence of regular plurals in compounds (Gordon, 1985) are unlearnable because they do not occur at all in the child's linguistic environment. I will not discuss these claims as it seems quite clear their empirical foundation is very weak, namely that the events in question do occur quite frequently (Pullum & Scholz, 2001). In any event, the occurrence of particular sorts of sentences in the primary linguistic data (PLD) is neither necessary nor sufficient for learning to take place. Even though a particular sentence type appears in the input, this does not mean that a learning algorithm can correctly identify the type that it is a token of, or the rule that generated it; conversely, the mere fact that a particular sentence type does not appear in the PLD does not mean that it cannot be learned: it could be learned from other types that do occur, or from the interaction of various other rules that are well attested. So the empirical question as to whether a particular sort of sentence occurs frequently in the PLD, is of minimal importance, except perhaps to expose the gulf between rhetoric and reality in some writers.

Cowie (1999) distinguishes two versions of the APS: first we have the argument I have outlined above, which she calls the *a posteriori* APS, and secondly we have an argument based on the absence of negative information that is related to Gold's theorem. I shall discuss this second, *a priori* argument in Chapter 4.

I would like to stress two things: first, the status of the IH is highly controversial, (Pinker, 1994; Sampson, 1997) but is widely accepted inside the theoretical linguistic community, and secondly, that the APS is considered one of the strongest arguments for it. Pullum and Scholz (2001) quote Fodor (1981) as describing the APS as

> *the* existence proof for the possibility of cognitive science

Table 2.1 presents in caricature a comparison of these two competing models of language acquisition, nativist and empiricist, which I will hope clarify the various arguments that will follow. We can compare them in four contrasting areas: the initial state of the brain/mind, or the part of it that is concerned with language learning; the input to the brain, the primary linguistic data or PLD, namely all of the language that the child is exposed to during his formative years of language learning; the learning algorithms that the child uses; and the final state of the mind/brain, the hypothesised mental grammar, or whatever mental machinery the child ends up with that allows it to speak and understand the language. The nativist claims that the PLD is very weak, that there are not powerful, general-purpose learning algorithms, and that the final state is very complex and highly structured with very deep principles: he therefore concludes that the initial state must be very complex, with highly structured domain-specific knowledge that is presumably innate. The empiricist on the other hand, claims that the PLD is very rich, and full of statistical regularities that give clues to its structure; that the brain is provided with a rich set of powerful general-purpose learning algorithms, and that the final state, though complex, is not very highly structured. He therefore concludes that it is unnecessary to have a very rich initial state. Thus polemically, the nativist is keen to emphasise the poverty of the PLD, and the richness of the final state, while denying the existence of general-purpose learning algorithms; and conversely, the empiricist claims that the final state is not as rich as is claimed, that much of the supposed complexity of the final data is an artifact of the notoriously inadequate data collection habits of theoretical linguists, and that there are any number of good learning algorithms.

## 2.4 A learning algorithm as a refutation of the APS

The APS rests on the premise that there are no general purpose learning algorithms that could learn a plausible grammar for any natural language based on the small sample of data available to the infant child. It can therefore be refuted by demonstrating that there are "generalised learning mechanisms" that can do just that. The most convincing way of doing this would be to exhibit a fully implemented computer program that can perform this task, and it is this that I attempt to do later on. Note that we are not concerned here with whether the human child actually uses these mechanisms or not. There are a number of requirements for a learning algorithm to constitute a refutation of the APS. I shall mention the main criteria here, and then discuss them further below. The data that the algorithm learns from should be as close as possible to the data that is available to the child, both in quantity and type. The algorithm should not have access to any linguistic or domain specific information. It must produce a "plausible" grammar. It must be a general purpose learning algorithm. It should work on all natural languages, and not make language-specific assumptions. However, there are some criteria that it need not satisfy: in particular it need not be a cognitive model for language. This is an important point: we are not trying to find a cognitive model of language learning. All we are trying to do is demonstrate the existence of algorithms that can learn from the data available to the child. We could do this by demonstrating a cognitively plausible model, but we need not. Any cognitive plausibility that the model might have is merely icing on the cake. This can be thought of as a methodological decision: to approach a full cognitive model by looking for algorithms that work, that actually can learn from a realistic amount of raw data, rather than to approach it by looking for models that, say, produce the right sort of errors on highly simplified data sets. The end point is the same, the route is different. One wants to choose enough constraints that will guide the process of theory construction helpfully.

Pinker (1979, p. 219) has an illuminating discussion of the requirements for a formal model of language acquisition.

It is instructive to spell out these conditions one by one and examine the progress that has been made in meeting them. First, since all normal children learn the language of their community, a viable theory will have to posit mechanisms powerful enough to acquire a natural language. This criterion is doubly stringent: though the rules of language are beyond doubt highly intricate and abstract, children uniformly *succeed* at learning them nonetheless, unlike chess, calculus and other complex cognitive skills. Let us say that a theory that can account for the fact that languages can be learned in the first place has met the *Learnability condition*. Second, the theory should not account for the child's success by positing mechanisms narrowly adapted to the acquisition of a particular language. For example, a theory positing an innate grammar for English would fail to meet this criterion, which can be called the *Equipotentiality Condition*. Third, the mechanisms of a viable theory must allow the child to learn his language within the time span normally taken by children, which is in the order of three years for the basic components of language skill. Fourth, the mechanisms must not require as input types of information or amounts of information that are unavailable to the child. Let us call these the *Time* and *Input Conditions*, respectively. Fifth, the theory should make predictions about the intermediate stages of acquisition that agree with empirical findings in the study of child language. Sixth, the mechanisms described by the theory should not be wildly inconsistent with what is known about the cognitive faculties of the child, such as the perceptual discriminations he can make, his conceptual abilities, his memory, attention, and so forth. These can be called the *Developmental* and *Cognitive Conditions*, respectively.

In terms of these criteria, I suggest that the APS claims that there are no algorithms that satisfy the Learnability, Equipotentiality and Input conditions. Any implemented program will satisfy the Time condition, since the computational resources of the sort of computers that we use today are limited compared to the human brain. The remaining two conditions are required for it to be a cognitive model. Of course, even a model that does fully satisfy these requirements might not turn out to be *true* since the brain might turn out to do things in some completely unexpected way: as previously stated, the matter will ultimately be settled by some sort of neurological evidence. In particular it might be possible to produce both empiricist and nativist models that satisfy all of these criteria.

### 2.4.1 Data Limitations

Pullum (1996, p. 505) describes the ideal data set, the primary linguistic data (PLD):

> Ideally, what we need to settle the question is a large machine-readable corpus – some tens of millions of words – containing a transcription of most of the utterances used in the presence of some specific infant (less desirably, a number of infants) over a period of years, including particularly the period from about one year (i.e. several months earlier than the age at which two words utterances start to appear in children's speech) to about 4 years.

To my knowledge there is no corpus that fits this description. Failing this we can use a corpus of written language of the appropriate size. This raises several questions:

- The size of the corpus

- The use of written rather than spoken language, in particular the use of sequences of letters rather than phonemes.

- The problem of the level of the language used in the corpora in terms of genre, degree of formality, syntactic complexity size of vocabulary, and so on.

- The proportion of grammatical errors and misspellings.

The size of the corpus used should not be excessively large. Pullum says "some tens of millions of words"; back of an envelope estimates seem to bear this out – a rate of 100 words per minute, for 120 minutes a day for a period of about 1000 days gives a pessimistic lower bound of 12 million words. Other estimates can be much higher: Kelly and Martin (1992) claim that

> Given conservative assumptions . . . a typical person will be exposed to a million word sample in about two weeks.

which would give an estimate of about 100 million words for the period in question. Hart and Risley (1995) provide some interesting evidence on this point that would tend to confirm the lower estimate. The child is not limited only to speech directed at or towards himself: as Chomsky (1959) points out

> A child may pick up a large part of his vocabulary and "feel" for sentence structure from television, from reading, from listening to adults, etc..

In any event, on a practical level, this is a good match both with the size of corpus that is widely available and the size of data set that is easily manipulable on standard workstations. Most of the work done in this corpus has used the lower estimate of 12 million words.

The use of written rather than spoken language is more fundamental. My view is that at the level of syntax, the differences are not that significant, but at the morphological and phonological level they are clearly crucial. In particular in English, which is written predominantly in an alphabetic script, where the orthographic words boundaries are in general quite close to the phonological word boundaries, this assumption seems innocuous. In other languages, with very different writing systems, it might not be. Therefore, I have used a written corpus for the syntactic work presented here, and for the morphological chapter I have worked exclusively with phonetic transcriptions. A decision about how to treat aspects of language such as punctuation, that are specific to the written modality must also be taken. When working with sequences of phonemes, another decision must be taken as to whether to use the atomic phoneme symbols, or whether to decompose them into a more structured representation. So for example we could either represent the first phoneme of the word "cat" as an atomic symbol, perhaps the character 'k', or as a feature structure whose partial representation might be might be `voiced=false,velar=true,stop=true`.

With regard to the level of language, it appears that this may not be significant. General statistical properties of texts appear to be quite invariant across text genres; van Everbroeck (1999) presents an interesting comparison between the CHILDES corpus (MacWhinney, 1995) and the Wall Street Journal (WSJ) corpus, which shows that they are very similar and that in some respects the corpus of child directed speech is more complex than the WSJ corpus. Moreover, (Finch, Chater, & Redington, 1995) report experiments using techniques that are quite similar to those used here on two very different corpora, the 2.5 million word CHILDES corpus(MacWhinney, 1995), and a 10 million word corpus of USENET news. The results they reported showed that the techniques worked comparably on both corpora.

The proportion of errors is also a problem. Most publically available corpora of the right size are made of material intended for wide publication: newspaper articles, novels, and so on. These, by their nature, have been heavily post-edited. As a result the proportion of errors is probably low compared to child directed speech, though Newport, Gleitman, and Gleitman (1977, p.121) claim:

> And finally, the speech of mothers to children is unswervingly well-formed. Only one utterance out of 1500 spoken to the children was a disfluency.

Finally, in a way the most obvious point: the data should be *raw*, that is not marked up with parts of speech, or constituent structure. Thus, in Machine Learning terms we must use *unsupervised learning*.

### 2.4.2   Absence of Domain-Specific information

The algorithm may not use any linguistic resources, such as lists of parts of speech, or primitive phrase structure rules.

As Kapur and Clark (1996) argue:

> We feel that the most logical approach is to assume that the child has no access to any information unless it can be argued that without some information, learning would be impossible or at least infeasible.

Though their assumptions are very different from mine, we can see that this approach has some independent merit on completely separate methodological grounds.

### 2.4.3   Absence of Language Specific Assumptions

Another requirement is that the algorithm not make assumptions about the language input, that might be true of the language under study, but not be true in general of all human languages. For example, an algorithm that learns morphology by looking at the suffixes of words, is making an assumption that the inflectional processes in the language are primarily suffixing; an assumption that is true in the case of English but false in very many other languages.

### 2.4.4   Plausibility of output

We require the algorithm to produce a linguistically plausible grammar. This is a rather vague criterion. Clearly it is possible to write a grammar induction algorithm if it doesn't matter what the output is. We require that the grammar conform in some sense to what we know about the structure of the language. There is however, substantial disagreement about many aspects of linguistic structure, and it clearly is not necessary that the output conform to the particular scheme of annotation of a particular linguistic theory. We can distinguish evaluation at the various different levels. So for example, at the level of morphology the set of correct forms is quite well-defined in many languages, and therefore fairly standard evaluation techniques are appropriate. At the level of syntax the answers are not so well-defined, and it is difficult to define a completely satisfactory evaluation metric. I shall discuss evaluation of each part of the model in the relevant chapters.

### 2.4.5 Excluded sources of information

These requirements may seem overly onerous. In particular, the infant child learner has access to other sources of information, notably the situational context of each utterance, if you like, its semantic context. Some researchers have claimed that this could be a useful source of information and have thus written algorithms which learn from phrases which have been annotated with some sort of semantic parse (Wexler & Culicover, 1980; Pinker, 1989). This is sometimes called *semantic bootstrapping*. This assumption seems to me to be too generous. Anyone who has spent time in a foreign country where he does not speak the language, will surely agree. Consider the following situation: you are sitting at a restaurant half way through a plate of food, and the waiter approaches, and asks you something with an interrogative intonation. Is he asking you whether you are enjoying your meal? Whether you have finished? Offering you some mustard? Asking you if you mind if someone else sits at your table? The possibilities are endless, even with a very clear situational context. It thus seems clear that this sort of evidence is unreliable as (Pinker, 1979) admits. There has also been some psychological research that verifies this intuition (Gleitman, 1990; Fisher, Hall, Rakowitz, & Gleitman, 1994; Gleitman & Gillette, 1995). Moreover as Gleitman (1994, p.184) points out:

> In fact, positive imperatives pose one of the most devastating challenges to any scheme that works by constructing word-to-world pairings, for the mother will utter "Eat your peas!" if and only if the child is not then eating the peas. Thus a whole class of constructions is reserved for saying things that mismatch the current situation.

It is thus clear that semantic information, however useful it may be, can certainly not be relied on as the exclusive source of evidence for the learning of syntax, though it would of course be essential for the learning of word meaning.

Prosodic information has also been put forward as another source of information. While this clearly is relevant at a large scale, for example segmenting discourses into sentences, it appears not to be reliable for syntactic information, i.e. for dividing sentences into syntactic constituents (Gerken, Jusczyk, & Mandel, 1994), because (Steedman, 1990, p.457)

> . . . the prosodic phrase boundaries established by intonation and timing are often orthogonal to the syntactic phrase boundaries that linguists usually recognize.

In addition, I assume that we do not have access to any *negative evidence*, that is to say evidence about what sentences do not occur in the language. There has been a long debate about whether children do or do not have access to negative evidence (Bohannon, MacWhinney, & Snow, 1990). To a large extent, this debate has been motivated by concern about the negative results of learnability of formal languages by Gold (1967), a concern that I consider misplaced, as I will argue at length in Chapter 4. Moreover I consider the traditional definition of a language as being a set of grammatical utterances impossible to define in practice, given the enormous variability in language and thus the definition of negative evidence to be extremely unclear. The view put forward here is that there are merely sentences that may appear as utterances in a language with greater or lesser frequency. The role of negative evidence is then taken by the fact that particular types of sentences (ill-formed/ unacceptable/ungrammatical) occur less frequently than other types (well-formed/acceptable) as has been noted by Horning (1969) and Chomsky (1981). This is sometimes called indirect negative evidence: this is highly misleading since it is in fact positive evidence. In summary, given the polemical situation, it is clearly better to make rather pessimistic assumptions about the information available.

## 2.5 Possible Counter-arguments

Suppose there was a program that satisfied these requirements. Let us examine some arguments that could be used to deny that this was a refutation of the APS. I will first remark that the program I am going to present will not in fact satisfy all of these requirements; I shall defer discussion of the consequences of that failure until Chapter 8. Here I discuss some general objections that could be made even if it was completely successful.

### 2.5.1 Inadequate data

- The data used was not sufficiently similar to the data available to the child.

This objection is in principle valid, but it needs to be argued. What precisely are the differences that are significant? A blanket refusal to accept results of this kind until precisely the correct data is available to permit the testing is clearly unjustifiable.

Moreover, it seems reasonable that the corpus I am using here, the BNC is in fact *more* difficult in many respects than the ideal data set, since the vocabulary and range of language is much wider. If I were using a corpus from a small domain, like the ATIS corpus or even from a single source, like the Wall Street Journal corpus, this argument might have more force.

Crain and Pietroski (2001) and Fodor (2001) both claim that there is a serious objection to Pullum (1996)'s paper on the APS, but from the other side, claiming that the Wall Street Journal corpus is much more complex than the language exposed to the child and thus the fact that certain constructions occur in the Wall Street Journal corpus is not relevant to the question as to whether or not they appear in the PLD.

So clearly, if the data is too simple, this approach can be criticised for dealing with an artificially simplified situation, and if the data is too complex, then it will be criticised for having too many examples of the sorts of complex constructions that the APS is based on. So until we have a very good PLD corpus, we shall have to make do with what we can get: a large mixed corpus. As better corpora become available, we can experiment further.

Since we are not interested in modelling the actual language acquisition process, and the different stages of development it does not matter so much that the corpus is adult-to-adult rather than child-directed speech. In any event, stipulating that certain criteria must be followed, when no corpus satisfies these requirements, amounts to stipulating that the APS is irrefutable at the moment.

### 2.5.2 Incomplete learning

- Though the algorithm successfully learns a range of constructions, it doesn't learn a particular construction, such as for example parasitic gaps in English.

- Though it works for languages W, X and Y , it won't work for Z.

The form of this argument is bad: until we have a perfect model, there will always be languages and constructions that it cannot learn correctly. Allowing this argument would mean that no refutation would be valid until there was a perfect empiricist learning algorithm for every single language; clearly an untenable requirement. Simply pointing out the flaw does not constitute a rebuttal of the argument. What is needed, is an argument that explains why a particular construction or language is so different that it cannot be learned. Failing that, this objection is without force. Of course, algorithms that only learn a small fraction of the language, or algorithms that demonstrably cannot learn more than a very limited range of languages will not be very convincing as refutations of the APS. As long as the algorithm does not make language-specific assumptions, the fact that it has only been demonstrated to work on a limited subset of languages should not be a concern. Note that, even if it is shown to work on all existing human languages, in some sense this objection could still in theory be made, since the existing human languages are just a small subset of all the possible human languages that the algorithm should also work on. This argument is not completely frivolous. Consider a learning algorithm, that was provided with full lexica and grammars for ten major languages, English, Japanese, and so on. When confronted with a bit of text, it could just identify the language, which is trivial to do with high accuracy, and output the predetermined grammar for that language. A system of this type, loaded with all actual human languages, would not be subject to this objection, though it is of course making a large number of unjustified language specific assumptions. Note that Principles and Parameters models of language acquisition are rather different from this, since they do not have the lexica provided for them but must acquire them.

### 2.5.3 Undesirable Properties of Algorithm

- The particular algorithm exhibited here has an undesirable property. Therefore the argument against the APS fails.

Let us suppose the model presented has some serious flaw. This doesn't require much imagination since, as will be seen, it has several. This does not mean that all algorithms will have this flaw. Remember, our goal is merely to establish that there are algorithms that satisfy a set of criteria. One way, the most conclusive way as is always the case with existence proofs, is to actually construct or exhibit an algorithm that satisfies all of the criteria. Another, less conclusive but still valid way, is to demonstrate that there are algorithms that satisfy a less stringent set of criteria, and then hope that this offers a convincing argument for the existence of algorithms that also satisfy the further criteria. Thus criticisms of particular aspects of the preliminary models presented here will not invalidate this as a critique of the APS in the absence of some principled argument that shows that all similar models must suffer from the same flaw. Presenting a completely specific and fully implemented model as I do here provides ample opportunity for trivial criticisms. To my knowledge there are *no* implemented nativist learning programs that can learn a language from raw text. Therefore specific criticisms of this model will have little force until a direct comparison can be performed.

### 2.5.4 Incompatible with psychological evidence

- It fails to account for speakers' intuitions.

- It is contradicted by psycholinguistic evidence.

We are not trying to construct a cognitive model. We are trying to demonstrate that there are algorithms that can learn from the data available to the child.

Moreover the use of speakers' intuitions is fraught with methodological problems of great severity (Schütze, 1996). I shall talk briefly about the cognitive plausibility of this model in Chapter 8. Here I shall merely say that it is not *necessary* for the model to be psychologically plausible for the argument to go through, though it is *desirable*.

### 2.5.5   Incompatible with Developmental Evidence

- The model fails to account for developmental evidence.

- Statistical models of language acquisition must predict that child speech will match adult speech, but there are significant differences such as subject dropping in English.

Yang (1999) says, discussing various differences between the language that the child produces and the language that the child hears:

> . . . an inductive learner purely driven by corpus data has no explanation for these disparities between child and language data.

In addition to the response above, that I am not trying to produce a cognitive model of language acquisition, I will make a further comment since this is quite a general criticism of this sort of learning algorithm, and raises some interesting points about what the language model actually represents.

My argument is that this rests on an overly simplistic view of the relationship between production and comprehension. It is uncontroversial that there is a large gap with language learners, whether infant or adult second language learners, between the range of language that can be understood and the range of language that is produced. In terms of statistical models, the difference can be thought of as a change in the boundary conditions. In comprehension, we are looking for some structures that are likely given the observed utterance; in production we are looking for structures that are likely given an input semantic representation. Even a simple statistical model could produce very different sets for these, if the mapping from semantics to syntax is rather limited. Thus statistically we can consider the language not just as a probability distribution over the set of strings, but as a joint probability distribution over pairs of strings and semantic representations. The set of strings produced by the child will correspond to the most likely strings with particular semantic representations. This is discussed in Optimality theory terms in Smolensky (1996). In generative grammar terms, this could be thought of as a competence/performance distinction. A less attractive alternative would be to posit separate distributions for production and comprehension.

### 2.5.6   Storage of entire data set

- The algorithm as it stands requires operations such as re-estimation, that operate over the whole data set. The infant language-learner however clearly cannot and does not memorise every utterance he or she is exposed to. [2]

---

[2]I am grateful to John Carroll for making this point.

In general however with the sort of statistical algorithms that are used here, there are versions of the algorithms which operate in what is called *on-line* mode, that is to say those which discard each data point after it has been processed, using for example the Robbins-Monro procedure (Robbins & Monro, 1951). This is sometimes called sequential parameter estimation (Bishop, 1995, p 46.). These are normally less efficient in terms of the amount of data required than the corresponding *batch* mode algorithms. Since we are operating on a fairly pessimistic amount of data, this seems not to be a serious objection. Moreover, frequently by using sufficient statistics, we can perform calculations on-line that might at first glance appear to require the storage of the whole data set. A simple analogy might clarify this: if we want to calculate the variance of a set of numbers, the average squared distance from the mean, it might seem necessary to store the whole data set so that we can operate on it twice, first to calculate the mean, and then in the second pass to calculate the average squared distance from the mean. But it is clearly sufficient merely to calculate a running total of the sum of the values, and of the sum of the squares. This allows the calculation without having to store the data set.

Chomsky calls this *instantaneous* learning and says (Chomsky, 1975c, p.245):

> at the present stage of our understanding, I think we can still continue profitably to accept it as a basis for investigation

I shall discuss this further in Section 4.3.

### 2.5.7   Argument from truth of the Innateness Hypothesis

- There is a great deal of other evidence for the innateness hypothesis.

Garfield (1994) identifies five other arguments apart from the APS:

> (1) the argument from the existence of linguistic universals; (2) the argument from patterns of errors in early language learners; (3) the poverty of the stimulus argument; (4) the argument from the ease of first language learning; (5) the argument from the relative independence of language learning and general intelligence; (6) the argument from the modularity of language processing.

to which we could add some recent neurobiological evidence related to Specific Language Impairment (Gopnik, 1990; Gopnik & Crago, 1991; Crago & Gopnik, 1994), though this has been questioned in turn by Vargha-Khadem and Passingham (1990) and Bishop (1996).

This objection is however simply a logical fallacy. I am arguing that an argument for the IH is invalid, I am not arguing that the IH itself is invalid. The existence of other arguments for the IH, or even the truth of the IH, can have no logical relation to the validity of my argument. Logically I accept $APS \Rightarrow IH$, and am trying to prove $\neg APS$. If we have another argument for the IH, such as for example a neuro-biological argument (NB), such that $NB \Rightarrow IH$, both $NB$ and $\neg NB$ are compatible with $\neg APS$. I am not qualified to discuss any of these other arguments, so I shall leave the point there.

However there is a variant of this argument that is reasonable: given that the IH is so well-supported by all available arguments and evidence, isn't it rather pointless to quibble about the validity of just one of these many watertight arguments? As Piattelli-Palmarini (1994, p.335) puts it:

> The extreme specificity of the language system, indeed, is a fact, not just a work-
> ing hypothesis, even less a heuristically convenient postulation. Doubting that there
> are language-specific, innate computational capacities today is a bit like being still
> dubious about the very existence of molecules, in spite of the awesome progress of
> molecular biology.

Certainly if I shared this view of the merits of IH, I would probably not have examined the APS
so closely; but it is certainly worth examining the merits of the APS even if the IH is true. There
may well be domain-general parts of cognition that are applied to the task of language-acquisition
even though the core of it is domain-specific. This sort of research could fruitfully focus the
attention of researchers on particular aspects of language where the domain-specificity is more
essential; moreover, I think it is clear that at some points in the language acquisition process, even
nativists must propose some sort of statistical learning, albeit just for low-level tasks such as word
segmentation.

### 2.5.8    Argument from the inadequacy of empiricist models of language acquisition

- Empiricist models are so inadequate that it doesn't matter if the APS is so bad: the IH is
  still true.

- Neural networks don't work well, therefore all empiricist models don't work well.

Fodor (2001) presents a radical variant of this argument in his article about Cowie (1999):

> My point is that attacking this claim the way Cowie does —by attempting to
> undermine the experiments one by one— is simply not appropriate to the polemical
> situation. ... If, in short, you wish seriously to evaluate the available data about the
> poverty of the child's stimulus, the pertinent question is not 'which of them can I
> perhaps impugn?'; rather it's whether, if they aren't entirely misleading, a move in
> the direction of empiricism seems plausibly the way to account for them. Or put
> it like this: We know what facts about the PLD are alleged to argue for the face
> plausibility of the nativist picture; well, suppose all of those were to disappear. The
> question remains: What are the facts about the PLD that are supposed to argue for the
> face plausibility of the empiricist picture? Answer: As far as I know (and, certainly,
> as far as Cowie tells us) there are none.

This argument appears to be that it doesn't matter if the APS is a bad argument based on
inadequate or no data; until there is a better theory it doesn't matter, i.e. it doesn't matter how bad
my arguments are, yours are worse. This is a perfectly valid point; until good empiricist theories
of language acquisition are available, nativist theories will be the best, but nobody is going to
bother if they think that it is impossible to produce a good empiricist theory. The APS claims to
establish this impossibility; removing it is therefore a necessary preliminary to an active program
of research in this area.

Crain and Pietroski (2001, p.177) claim that the limitations of Neural Networks apply to all
statistical learning algorithms:

> There is a second inadequacy with experience-dependent learning mechanisms
> that rely on localist error-correction algorithms such as the back-propagation algo-
> rithm. In extracting information based on local connections, these mechanisms do
> not generalize beyond the training set.

They then proceed to a discussion of various flaws in Rumelhart and McClelland (1986a). I share their scepticism about the generalisation ability of neural networks, but neural networks are merely one possibility among many. As we shall see, generalisation is an important aspect of learning, but it is one that is now well understood, and the limitations of one sort of algorithm do not necessarily apply to other learning techniques.

### 2.5.9 Use of tree-structured representations

- The algorithm uses tree-structured representations for sentences – this is domain-specific.

Tree-structured representations are generally considered to be used in a number of other cognitive domains (Bloom, 1994) that have been studied extensively such as chess (Charness, 1992; Freyhof, Gruber, & Ziegler, 1992), music (Lerdahl & Jackendoff, 1983) and vision (Marr, 1982; Pylyshyn & Burkell, 1997). As Sampson (1997) points out, citing (Simon, 1962), on general evolutionary arguments we would expect tree-structured representations to be wide-spread. Miller and Chomsky (1963, p.483) recognise this when they say: [3]

> Let us accept as an instance of complicated behavior any performance in which the behavioral sequence must be internally organized and guided by some hierarchical structure that plays the same role, more or less, as a P-marker [phrase marker] plays in the organization of a grammatical sentence.

It is important to note that not all linguistic principles are structure dependent: there appear to be a number of phonological phenomena that are not (the a/an alternation in English being a simple example). Moreover prosody (Steedman, 2000) and quantifier scoping ambiguities are famously related to surface order, and have been a perennial problem for structure-dependent theories (Hobbs & Shieber, 1987) (i.e pretty much all theories); and at this point we could also point to numerous phenomena that are normally dealt with as performance matters that are also not structure dependent.

Crain and Pietroski (2001, p. 163) display this presupposition when they ask:

> Is there any guarantee that there *is* abundant evidence that lets all children learn that all linguistic principles are structure dependent?

Now I am sure they would respond by making various points about the autonomy of syntax, and the performance competence distinction and so on, but this merely establishes the point that while it *may* be possible to identify some areas of language that only use structural rules, many other areas use other sorts of rules. So the question then becomes, how can the child learn which rules are structure dependent and which are not? The simple answer is that the structure dependent rules work better to describe structure-dependent phenomena, and the surface rules work better to describe surface phenomena. Until quite recently there would have been a huge flaw in this argument, namely that the state of the art in language models were unstructured models such as smoothed tri-grams (Jelinek, 1997), but this is no longer the case and the best performing structured language models (Charniak, 2001) perform substantially better than trigram models on the same amount of data.

---

[3]It is not clear that this reflects Chomsky's current views.

In any event, leaving aside the issue of whether tree-structured representations are domain-specific psychologically, a question that in any event current cognitive research is not able to answer, it is abundantly clear that they are not domain-specific in general: they are very widely used in many areas of mathematics and computer science. [4]

Therefore I am quite willing to admit that humans have an innate structure dependent modelling capability; I deny that it is domain-specific in any meaningful sense.

### 2.5.10   Not a general purpose learning algorithm

- It is not enough to exhibit an algorithm that can learn language. For it to constitute a refutation of the APS, it must be a general-purpose learning algorithm, and that means it must also be able to learn in one or several other domains as well.

This is an objection put forward by Ramsey and Stich (1991, p. 308) in a discussion of connectionist models and the APS:

> If the only successful connectionist language acquisition devices are of a sort that require language specific architectures and/or language specific tuning, then even the rationalist version of nativism will have nothing to fear from connectionism.

This seems slightly too strict a criterion. Purely for technical reasons, one is likely to design algorithms that function well in a particular domain. Thus, the most successful language acquisition devices are likely to be those that function in the narrowest domain. If the various learning components are quite general purpose, although their organisation and configuration may have been optimised for the task at hand, then I think this objection is without force. Ramsey and Stich are quite right in my view to question the claims of the connectionist models, which invariably need a large amount of representational engineering to perform well in any domain.

Since I accept the general thrust of this argument, I have taken pains to use general-purpose machine learning techniques, or variants thereof. Thus the techniques of distributional clustering used in Chapters 5 and 7, are variants of standard clustering techniques. The Pair Hidden Markov Models used in Chapter 6 were as a matter of historical fact used first for aligning DNA sequences, and are quite promising for learning other transductions in other areas. Stochastic Context-Free Grammars might appear to be language specific, but have been used in bioinformatics as well (Sakakibara, Brown, Hughey, Mian, Sjolander, Underwood, & Haussler, 1994). Moreover considered as a certain sort of stochastic branching process, they can be seen to have links with many areas of mathematics and computer science, from multi-type Galton-Watson processes (Miller & O' Sullivan, 1992) to Pattern theory (Grenander, 1996).

### 2.5.11   Fails to learn deep structures

- The algorithm may learn the surface structure of the language, but it doesn't learn the 'deep structures' that are necessary for an adequate grammar.

- More generally, the grammar produced fails to have some particular property that is essential.

---

[4]At the risk of sounding facetious, one can also point out that tree-structured *objects* are also prevalent in the natural world, canonical examples being river deltas, corals, and, um, trees.

This point is put by Chomsky (1967, p.129)(reprinted as Chomsky (1975b))

> In the case of language acquisition, there has been much empiricist speculation about what these mechanisms may be, but the only relatively clear attempt to work out some specific account of them is in modern structural linguistics, which has attempted to elaborate a system of inductive analytic procedures of segmentation and classification that can be applied to data to determine a grammar. It is conceivable that these methods might be somehow refined to the point where they can provide the surface structures of many utterances. It is quite inconceivable that they can be developed to the point where they can provide deep structures or the abstract principles that generate deep structures and relate them to surface structures.

I think I am in agreement with Chomsky that these types of methods could not learn these highly abstract structures. The question of course is whether they are necessary. The arguments Chomsky presents for the existence of these deep structures are very inconclusive, and it is widely accepted that so-called *monostratal* theories, derived from GPSG (Gazdar, Klein, Pullum, & Sag, 1985), such as HPSG (Pollard & Sag, 1994) are adequate to represent natural languages, and in many respects are perhaps superior to current Chomskyan models, even according to Chomskyan criteria (Johnson & Lappin, 1997).

Moreover, this argument if pursued could become circular. The theories of generative syntax pursued by Chomsky and co-workers have been motivated to a large extent by nativist assumptions, justified largely by the APS: clearly then arguing from these theories to support the APS is questionable at best.

This argument is taken up again by Crain and Pietroski (2001, p.183) who conclude

> Until empiricists show how specific principles – like the Head-Movement Constraint and the Binding Theory – can be learned on the basis of the primary linguistic data, innateness hypotheses will continue to be the best available explanation for the gap between normal human experience and the linguistic knowledge we all attain.

This is incorrect: the argument is that to convince the rationalist that he is wrong, the empiricist must produce a rationalist theory. Of course he does not: what the empiricist must do is account for the *data*, not the *theory*, and the data can be accounted for in a number of different ways, either by positing deep, obscure and hard-to-learn principles as Crain and Pietroski (2001) suggest, or by positing shallow easily learnable empiricist explanations. Crain and Pietroski (2001, p.159) recognise this earlier on:

> Replies [to nativist arguments] must either challenge these descriptive claims about human grammars, by providing alternative explanations of the relevant judgments, or address the learning problems (including those concerning uncorrectable overgeneration) associated with *specific constraints*

Of course, alternative explanations abound for many of the descriptive claims they consider: for example their first example of *wanna*-contraction has an alternate morpho-lexical explanation, that is perfectly learnable as discussed by Pullum (1997).

### 2.5.12   Fails to guarantee convergence

- All you have demonstrated is that the algorithm learns from this single corpus. You need to show that it will be learned for every possible corpus that a child might be exposed to.

- You must also show that your algorithm will converge to the same grammar for every corpus of the same language.

That is true: however there are similar approaches such as Finch and Chater (1992a), Redington, Chater, Huang, Chang, Finch, and Chen (1995), Klein and Manning (2001) that have produced similar results on different corpora. There is no reason to suspect that the corpora I use has a radically idiosyncratic structure. Secondly, it is not the case that all children learn the language they are exposed to. A small proportion of children who appear to be neurologically normal do not with a wide range of deficits ranging from mild delays in the progress of their learning, to a complete failure to acquire any part of the language (Shames, Wiig, & Secord, 1998). I am not suggesting that this is caused by an unusual or misleading PLD, but merely pointing out that the empirical situation is not quite as clean-cut as it is claimed to be. Thirdly, it seems quite clear that people exposed to more or less the same linguistic environment do in fact differ from one another in terms of the grammar they acquire, that is to say they do not converge to the same grammar, a fact that causes methodological problems for traditional generative grammar data collection techniques (Schütze, 1996). The assumption that all speakers converge to the same grammar may be a "methodologically expedient counterfactual idealization" to use Newmeyer's phrase, but it is certainly not an empirical finding that must be explained.

### 2.5.13   Doesn't use semantic evidence

- Children have access to the situational context of the utterances, and learn language as a way of expressing meanings, but this algorithm completely neglects that source of evidence

I agree. What I am trying to do may in fact be impossible without using semantic evidence, and it is certainly the case that semantics drives the production of language by children. However this objection does not reduce the effectiveness of this critique of the APS, but rather strengthens it: if I can demonstrate that the grammar is learnable from this very limited evidence, then *a fortiori* it is learnable from a larger, richer source of information. The same argument applies also to other sources of information that might be available such as prosodic information.

This argument does of course weaken the validity of this model as a cognitive model, but as previously stated I am not trying to do that.

### 2.5.14   Fails to show that children actually use these techniques

- You have not provided any evidence that children actually use these techniques.

As Crain and Pietroski (2001, p.174) put it

But we would like to see some reason for thinking that any of this is true.

I shall discuss the psychological plausibility of these techniques, briefly, in Chapter 8. But in short this is a mistake about the form of the APS. The APS claims as one of its premises, that no algorithm can learn from the PLD without innate domain specific knowledge. I am refuting the APS by showing that one *can* learn from the PLD without such knowledge; I am not trying to show that children actually *do* learn using such knowledge, nor do I have to show that to refute the APS.

(Brent, 1997) has an interesting discussion of the role of computational models in the study of language acquisition. He concludes that even models that do not implement a psychologically plausible algorithm can be useful at an early stage of research.

# Chapter 3

# Language Learning Algorithms

## 3.1 Introduction

This chapter outlines the techniques I will use to implement a program that is intended to satisfy as many of the desiderata discussed in Chapter 2 as possible. I discuss the theoretical background to some of the methods I will use, and try to situate the algorithms I am going to use in their intellectual context. I also discuss various technical aspects of the methods I will use. The layout of this chapter is as follows: in Section 3.2, I briefly discuss the role of machine learning in natural language processing, and the different motivations for research in this field. In Section 3.3 I discuss the distinction between supervised and unsupervised learning techniques. In Section 3.4 I discuss statistical learning techniques, and the difference between parametric and non-parametric techniques, and then in Section 3.5 I shall briefly talk about a particular form of statistical model, Neural Networks. Section 3.6 discusses the basic technique of Maximum Likelihood Estimation, and Section 3.7 discusses the Expectation-Maximisation algorithm for Maximum Likelihood Estimation with incomplete data. Another form of estimation is Bayesian estimation which is dealt with, together with various related techniques in Section 3.8. I then address the issue of evaluation in Section 3.9. Finally I discuss the overall architecture of the learner in Section 3.10, and Section 3.11 deals with various areas or levels of language that will not be addressed in this thesis.

## 3.2 Machine Learning of Natural Languages

The study of Machine Learning is a wide field of academic study in its own right, with a large variety of different techniques being employed (Carbonell, 1990); I shall not attempt a survey here. There are two main motivations for applying Machine Learning techniques to natural languages. First, there is a pure engineering motivation. The effort involved in hand coding grammars, dictionaries and other linguistic resources is prohibitive: it is natural that researchers should explore ways of extracting them automatically from appropriate data sets (Church & Mercer, 1993; Grefenstette, 1994). Secondly, there is a cognitive science motivation: interest in modelling the acquisition of human language (Brent, 1997).

There are a variety of methods used in this area, which we can divide roughly into two classes – symbolic and statistical. Many people consider connectionist models to be a third class, but in my view they are better seen as a special type of statistical model. Symbolic methods currently used include decision trees (Quinlan, 1993), Inductive Logic Programming (ILP) (Muggleton, 1997, 1999; Muggleton & Bain, 1999) and transformation-based learning (Brill, 1992, 1993).

Statistical techniques have become very widely used in NLP recently, and in this thesis I shall be using only such methods. They have a number of desirable attributes, such as being resistant to noise and applicable in a wide number of situations. Bishop (1995) provides an excellent introduction to the use of statistical pattern recognition techniques in Machine Learning. I shall use a number of different such techniques in this thesis, including Hidden Markov Models (HMM), Distributional Clustering and Stochastic Context-Free Grammars (SCFG). Though statistical techniques are often thought of as recent arrivals some of the earliest work in natural language processing has been statistical in nature (Weaver, 1949; Locke & Booth, 1955).

### 3.3   Supervised and Unsupervised Learning

A key distinction to be made is that between *supervised* and *unsupervised* learning techniques. In supervised learning, the learning algorithm is provided with evidence about the classification of the data points. In unsupervised learning, on the other hand, the algorithm must determine the classification and structure of the data by itself. Supervised learning encompasses most of the traditional types of machine learning, and has been studied in great detail. There are however a number of motivations for using unsupervised methods. First, labelled data is hard to come by and limited in quantity, and may contain errors. Secondly, it commits you to a particular scheme of analysis that may or may not be suitable for the task in hand. Thirdly, and most importantly, with natural language data, there is very limited agreement about what the labels should be: Sampson (1995, p.4) discusses the results of a workshop in 1991 at which researchers were given sample sentences to annotate:

> Only a small fraction of the full range of grammatical structuring found in the examples was agreed on by all participants.

Though it may be the case that there is in some sense an objectively correct, psychologically real, constituent structure for natural language utterances, we just don't know what it is.

Of course, in this case we use unsupervised learning because we are interested in exploring how much the infant learner can learn *without* supervision. Thus, we have to rule out the use of supervised learning techniques except in special situations, where the supervision is provided by another part of the learning algorithm.

A standard technique of unsupervised learning is clustering; data points are grouped together based on their closeness in some suitable feature space. This can take the form of a simple *k*-means clustering where the number of clusters is decided in advance, closely related to mixture models, or a hierarchical clustering algorithm or even a neural network-based method such as a Kohonen map (Bishop, 1995). In NLP, some of these features may themselves be the results of the analysis; that is to say the features that are used are also a product of the clustering process, such as syntactic, morphological or semantic features. This leads to some difficult problems discussed in Hofmann and Puzicha (1998).

### 3.4   Statistical Learning

There are a number of different types of statistical model that can be used in machine learning. We can divide these roughly into two classes, those of parametric and non-parametric models, with perhaps a third class of semi-parametric models that would include for example the various flavours of Neural Networks, which I shall discuss in Section 3.5.

A parametric model is a model where we assume that the data is generated by, or can be approximated by, a model with a particular parametric form. So for example, if we wanted to model a data set consisting of the heights of a large number of people, one way of doing this would be to choose a parametric model, probably a normal distribution, which would have two parameters, the mean and the standard deviation. We would then choose some values for the two parameters that would make the normal distribution a good fit to the data. There are often different ways of choosing these parameters, as we shall see below. A problem with this approach is that in some cases it is not obvious on *a priori* grounds what the appropriate parameterised set of models should be. In machine learning of syntax, for example, one might propose as a suitable set of models a set of Probabilistic Context Free Grammars (PCFG), and try to find suitable parameters for them.

A distinction is often made between learning the structure of a model and learning the parameters of the model. So for example, with a PCFG, one might first learn a set of rules, and then estimate the parameters of that set of rules using the EM algorithm. However it is clearly possible to do it in a slightly different way. First of all, for any PCFG we can write a stochastically equivalent one in Chomsky Normal Form. [1] Therefore the structure of the model reduces to two things: first the number of non-terminals, and secondly which of the rules have non-zero probabilities. But the EM algorithm tends to converge to solutions where quite a lot of the rules have zero probability, so ultimately all we have to do is select the number of non-terminals, and let the EM algorithm do the rest. The size of the model can then be derived using some sort of model selection criterion, as will be discussed below in Section 3.8. Of course in many cases it will not be possible to run the EM algorithm with a fully connected model, and in other cases if we have prior information we can use this to constrain the search space, which often improves the performance. In the work in this thesis, I feel the distinction is of minimal importance since the algorithms are meant to operate without any external information.

Non-parametric models on the other hand do not assume that the data is modelled by some antecedently specified class of models. The standard example is of a histogram (Scott, 1992, Ch.3). Formal definitions of non-parametric estimators are hard to come by; as a working definition we can say that if the space of parameters is infinite-dimensional, or more formally if the number of parameters tends to infinity as the sample size tends to infinity, then it is a non-parametric estimator. In Machine Learning a common algorithm derived from non-parametric estimation is the k-nearest neighbour algorithm, which in many cases can give surprisingly accurate results. Mathematically, though, the arguments for these models are rather poor, as data is often sparsely distributed in high-dimensional spaces. In circumstances like this, these techniques are not guaranteed to work, because of the so-called curse of dimensionality (Scott, 1992), though in practice they perform well.

In NLP, two frequently used machine learning techniques that can usefully be thought of as non-parametric are Memory-Based Learning (MBL) (Lin & Vitter, 1994; Zavrel & Daelemans, 1999; van den Bosch & Daelemans, 1999), and Data-Oriented Parsing (DOP) (Bod, 1993, 1998). DOP can also be thought of as a parametric model, as a particular type of very large stochastic tree substitution grammar, that is equivalent to a PCFG (Goodman, 1996, 1998).

## 3.5 Neural Networks

A particular sort of statistical model that has been used with limited success in NLP is the so-called *Neural Network* or more formally the multi-layer perceptron (MLP), also known as Parallel Distributed Processing (PDP) (Rumelhart & McClelland, 1986b). Though they are sometimes seen as being in some sense qualitatively different from other statistical models, as Bishop (1995) is at pains to point out, they are "an extension of the many conventional techniques which have

---

[1]If the PCFG generates the empty string this cannot be done.

been developed over several decades."

Although, the use of neural networks has been quite widespread in the cognitive science and cognitive neuroscience fields, the results have in general been poor when compared to more conventional models in general NLP tasks (Reilly & Sharkey, 1989; Wermter, Riloff, & Scheler, 1996). Though they appear to be good at memorising arbitrary patterns their generalisation is often quite poor. I shall discuss some models that have been successful in Chapter 6, namely some models of the acquisition of morphology including Rumelhart and McClelland (1986a) and Plunkett and Nakisa (1997).

Though their neurological plausibility is tantalising, it appears that there are overwhelming technical difficulties with making them work effectively in practical, large-scale tasks, though there are signs that this is being overcome (Bengio & Vincent, 2000). There is an illuminating discussion of this in Steedman (1999). They are however extremely interesting as demonstrations that traditional assumptions about the sorts of representations and processes that operated in the human brain have been perhaps too narrow, and that the range of possibilities is much wider than has previously been thought.

## 3.6   Maximum Likelihood estimation

One of the most fundamental techniques in statistical modelling, which I shall use extensively, is Maximum Likelihood (ML) estimation. If we have a model with a set of parameters $\Theta$, and some data $D$ we can consider the probability of the data given the model which we can write as $p(D; \Theta)$. We can also consider this quantity as a function of the *parameters* rather than the data in which case it is called the *likelihood*. We then speak of the likelihood of the parameters with respect to the data, which is just the probability of the data given the model. Though ML estimation is intuitively very appealing and natural, ML estimators are not in general unbiased. [2] For example if we estimate a set of data with a normal distribution, the ML estimation of the variance is the variance of the sample, while as every schoolboy knows, the unbiased estimator is the *sample* variance which divides by $n - 1$ rather than $n$. However the ML estimator does have some nice formal properties to go with its intuitive appeal. In particular for large sample sizes, and subject to various regularity conditions (Silvey, 1975), ML estimators are nearly unbiased and nearly are minimum variance, i.e attain the Cramer-Rao lower bound. It many natural language applications, however, the sample sizes are small, and thus attention must be paid to the limitations of this approach.

In particular if we have a large set of models, which include an infinite sequence of models with increasing complexity, the ML estimate will often select the model that exactly memorises the data. In this case, we clearly have a poor estimate. I shall discuss this in more detail in Section 3.8.

## 3.7   The Expectation-Maximisation Theorem

The Expectation-Maximisation (EM) algorithm is a general algorithm for ML estimation. The normal citation for this is Dempster, Laird, and Rubin (1977) but the algorithm had been inde-pendently discovered several times before. The EM algorithm is used when one has incomplete information. I use this algorithm at several points in this thesis; I will therefore give a full description of it here. Suppose we have some observations which we consider as the random variable $O$, and these are generated by some other random variable $X$. We may not know what the values of $X$ are for each observation; they are *hidden variables*. We want to perform ML estimation of the parameters of the model $\Theta$. The likelihood of the model is

$$p(O|\Theta) = \sum_X p(O, X|\Theta) = \sum_X p(O|X, \Theta) p(X|\Theta) \tag{3.1}$$

---

[2]An estimator is unbiased if the expected value of the estimator is equal to the quantity to be estimated. Thus the mean of a sample is an unbiased estimator of the mean of the distribution, since the expectation of the sample mean is equal to the mean of the distribution.

We will normally be considering a sequence of such observations and the associated hidden random variable which we can denote $O_i$ and $X_i$ respectively; the likelihood will be the product of the likelihood for each observation. Here each observation might be a sentence or longer language unit, so that we can consider each observation to be independent. Then the total likelihood for all the data is

$$p(O|\Theta) = \prod_i \sum_{X_i} p(O_i|X_i, \Theta) p(X_i|\Theta) \tag{3.2}$$

Because of the product it is more convenient to work with the logarithm of this, giving the log likelihood

$$L(O|\Theta) = \sum_i \log \sum_{X_i} p(O_i|X_i, \Theta) p(X_i|\Theta) \tag{3.3}$$

The straightforward approach to optimising this would just be to differentiate this objective function with respect to each parameter, set it to zero, and solve it to find the optimal value of the parameter. Unfortunately this is not possible because of the presence of the logarithm of a sum. If we have an equation of the form

$$f(x_1, \ldots, x_n) = \log(\sum_i \alpha_i x_i) \tag{3.4}$$

then the equations to maximise this are horrible since

$$\frac{\partial f}{\partial x_i} = \frac{\alpha_i}{\sum_i \alpha_i x_i} \tag{3.5}$$

so when we try to solve these subject to various constraints we get a large set of intractable simultaneous equations. Note also that the hidden variables can take many values, since they correspond, for example in a HMM to a particular state transition sequence, of which there are exponentially many.

The EM theorem shows how this can be circumvented. Suppose we have a set of parameters $\Theta^{old}$ and we want to change them so as to improve, i.e. increase, the log likelihood. We then want to maximise the increase in log likelihood between the old and the new parameters

$$\Delta L(\Theta^{new}, \Theta^{old}) = L(O|\Theta^{new}) - L(O|\Theta^{old}) \tag{3.6}$$

$$= \sum_i \log p(O_i|\Theta^{new}) - \sum_i \log p(O_i|\Theta^{old}) \tag{3.7}$$

The EM theorem says we can increase this if we maximise

$$\sum_i \sum_{X_i} p(X_i|O_i, \Theta^{old}) \log p(O_i|X_i, \Theta^{new}) \qquad (3.8)$$

This is a lot easier because we have a sum of a log, rather than the log of a sum, which is much more straightforward to optimise, since if we have an equation of the form

$$f(x_1, \ldots, x_n) = \sum_i (\log \alpha_i x_i) \qquad (3.9)$$

then

$$\frac{\partial f}{\partial x_i} = \frac{1}{x_i} \qquad (3.10)$$

Equation 3.8 has the form of an expectation: the expectation with respect to the distribution of the hidden variables given the observed variables. So intuitively, we work out what would have happened at each observation, if the parameters were $\Theta^{old}$, and maximise the log of the new probability with that. We therefore have an iterative algorithm where we repeatedly apply this procedure getting a sequence of models that is guaranteed to have monotonically increasing log-likelihood. This will therefore converge, and we can stop when the log likelihood stops increasing by a significant amount.

There are a number of disadvantages with the EM algorithm. First, it is notorious that the EM algorithm only converges to a local maximum; unless the likelihood surface is convex, the local maximum won't necessarily be the global maximum. Whether this is a serious problem or not depends on the particular model and data set. Secondly, the EM algorithm has a tendency to favour solutions that are combinations of many different elements, as has been observed by *inter alia* Abney and Light (1999)

> . . . the EM algorithm estimates a mixture model and (intuitively speaking) strongly prefers mixtures containing small amounts of many solutions over mixtures that are dominated by any one solution.

While in some circumstances this might be an advantage, in many cases we want algorithms to commit to a particular analysis of the situation. For instance, when we divide words into syntactic classes we would like the majority of the words to be unambiguously assigned to a single class.

## 3.8   Bayesian techniques

The term *Bayesian* is used widely to mean a variety of different things in Machine Learning. Bernardo and Smith (1994) provide a detailed examination of the foundations of Bayesian theory. Here, I use it to refer to a range of techniques that trade off model complexity against the fit of the model against the data, in an attempt to reduce some of the problems of overtraining associated with ML estimation. There are a number of closely related techniques:

- Bayesian model prior.

- Minimum Description Length / Minimum Message Length.

- Structural Risk Minimisation.

I shall briefly discuss the relationships between them. Bayesian techniques derive their theoretical base from Bayes' rule (Bayes, 1763). When we have a hypothesis $H$ and some data $D$,

$$p(H|D) = \frac{p(D|H)p(H)}{p(D)} \tag{3.11}$$

The term $p(H)$ is called the prior probability and represents the degree of belief in a hypothesis before any data has been examined. The term $p(H|D)$ represents the degree of belief in the hypothesis after seeing the data $D$. Thus this relationship expresses the mathematical process of learning from data in its most fundamental form. In Machine Learning we are often interested in choosing a single model $H'$; the obvious choice is then the model that is most probable, i.e. the $H$ that maximises $p(H|D)$. Looking at Equation 3.11, we see that this is

$$H' = \underset{H}{\operatorname{argmax}} \frac{p(D|H)p(H)}{p(D)} = \underset{H}{\operatorname{argmax}} \, p(D|H)p(H) \tag{3.12}$$

since $p(D)$, the probability of the data is constant. If we assume that $p(H)$ is constant, we can see that this reduces to Maximum Likelihood estimation. Having a non-trivial prior probability will correspond to a bias for particular models.

Regardless of the theoretical motivation, in practice people choose a prior that allows the mathematics to work out cleanly: this normally takes the form of a so-called conjugate prior. This has no justification, other than the fact that it allows you to solve the equations. At this point, all of the Bayesian justification has been thrown out of the window. Moreover at times, one may wish to approximate a distribution by a set of models, for instance Hidden Markov Models (HMM), that one knows beforehand are inadequate to model the task (Stolcke, 1994). In this case, the prior probability should logically be zero, since we know beforehand that they cannot be the correct model.

Minimum Description Length (MDL) techniques (Rissanen, 1978), and the closely related Minimum Message Length (MML) (Wallace & Boulton, 1968) formalism[3] are closely related to this. If we make standard assumptions about efficient codes, using the Kraft inequality, we can show that the length of the optimal code for a hypothesis or for some data is approximately equal to the negative logarithm of the probability. If we rewrite Equation 3.12 as

$$H' = \underset{H}{\operatorname{argmax}} \, p(D|H)p(H) = \underset{H}{\operatorname{argmin}} -\log p(D|H) - \log p(H) \approx \underset{H}{\operatorname{argmin}} L(D|H) + L(H) \tag{3.13}$$

Hence the name Minimum Description Length.

The Structural Risk Minimization (SRM) principle was introduced by Vapnik and co-workers in a series of papers (Vapnik (1998) provides a useful historical note and comparison to other techniques). It provides a more rigorous mathematical treatment of many of these issues, with accurate proofs of the convergence criteria. The Empirical Risk Minimization (ERM) principle says that the objective of statistical estimation is to minimize the empirical risk functional on some data: this is equivalent to the Maximum Likelihood approach with an appropriate risk functional, or to the least-squares method in regression where the functional is the squared error. The SRM principle then incorporates, based on a worst case analysis, a term that relates to the probability

---

[3]The major difference between these is that MDL techniques first select a class of models, and then select the best model using a ML criterion, whereas MML techniques perform the two operations at the same time. (Figueiredo & Jain, 2001)

that the model will be wrong, that is fail to generalise, that derives from the amount of training data compared to the power of the model. If the model is very powerful, and able to memorise the data exactly then we cannot guarantee that it will generalise well. This term has the same effect of limiting the model size as the MDL principles. This summary is inevitably rather a caricature; I merely want to point out the similarities in goals and methods between these various techniques.

## 3.9    Evaluation

Evaluation with unsupervised methods is difficult. With supervised methods, the notion of a correct answer is coherent; in unsupervised learning of natural language, since as we saw above, there is substantial disagreement between linguists about what the correct answer is, it is not possible to choose "a right answer" to compare it against. In short, there is no *gold standard* that we can agree on. Accordingly in this thesis, I evaluate the results in two ways: one objective way and one subjective way.

The objective way to measure these sorts of stochastic models is to calculate how well the model predicts unseen data, in particular what probability it assigns to it. The cross entropy between a distribution $p$ and another distribution $q$ is

$$H(p, q) = -\sum_x p(x) \log q(x) \tag{3.14}$$

This will always be greater than the entropy of $p$, $H(p)$ since

$$H(p, q) = -\sum_x p(x) \log q(x) = H(p) + D(p||q) \tag{3.15}$$

with equality only when the distributions are equal. So if we are comparing two models we can say that the one with the lowest cross entropy is better, since it will be closer to the actual distribution. We can estimate the cross entropy by using some unseen data $x_1, \ldots, x_n$ drawn from $p$

$$H(p, q) \approx -\frac{1}{n} \sum_i \log q(x_i) \tag{3.16}$$

which is just the average negative log probability of the test data. The related measure of perplexity is just the exponent of this which is just equal to the $n$th root of the probability of the test data. Jelinek (1997) and Manning and Schütze (1999) discuss this at some length. However this measure has some disadvantages too: notably, if a single data point is assigned zero probability, then the perplexity will be infinite, or less radically a single highly improbable outlier can cause a rather misleading result.

The subjective way I will use is merely to present for scrutiny some of the linguistic structures induced by the algorithms presented. Though rather informal, this is I feel a very good way for understanding the extent to which the algorithm agrees or disagrees with traditional linguistic analyses.

In addition, where possible, I have used some more traditional styles of formal evaluation. In particular, some of the work in Chapter 6 can be evaluated by looking at its accuracy on unseen data, and I have accordingly done that. Pereira and Schabes (1992) contains some interesting data that is not discussed in the paper, which bears directly on this point. They present an algorithm for inferring context free grammars from quite small bracketted and unbracketted corpora (700 sentences in the training set). They presented two measures; first, the cross-entropy estimate, and secondly the bracketing accuracy. Table 3.1 summarises the results. Though the difference in cross-entropy is probably not significant, this result clearly establishes that these two modes of evaluation do not rank models similarly.

| Data | Cross-Entropy | Bracketing Accuracy |
|------|---------------|---------------------|
| Bracketed | 2.97 | 90.36% |
| Raw | 2.95 | 37.35% |

Table 3.1: Comparative evaluation figures from Pereira and Schabes (1992). Note that the algorithm trained on raw data outperforms the supervised algorithm according to the cross-entropy, though being vastly inferior according to the bracketing accuracy measure.

This has been confirmed by my own experiments as can be seen in Table 7.1.

## 3.10   Overall architecture of the language learner

In this section I will discuss the overall layout of the language learning algorithm. I have decomposed the task into particular modules, for software engineering reasons rather than because of a commitment to the modularity of mind (Fodor, 1983). I assume that the input of the algorithm is a corpus of language that has been tokenised, and divided into sentences. The validity of this assumption is discussed below. The program proceeds as follows:

- Categorisation using distributional information.

- Learning of Morphological processes.

- Learning of syntax.

These are discussed in Chapters 5, 6 and 7 respectively. Here I will just discuss the relationship between them.

### 3.10.1   Categorisation

Initially, the algorithm identifies a set of lexical categories based on distributional analysis. This part of the algorithm is discussed in Chapter 5. In morphologically very rich languages, such as Hungarian or Finnish, this might not be possible, because of the much lower frequency of occurrence of individual words (Kornai, 1992; Bertram, Laine, Baayen, Schreuder, & Hyönä, 2000). In these languages it would be necessary to start with a morphological learning phase, before proceeding to a categorisation phrase. Note that languages with complex inflection tend to signal the morphological class of a word in the phonology, whereas in English, for example the word *stout* could be a noun or an adjective or a verb from its phonology.[4]

### 3.10.2   Morphology

In English, given a set of automatically induced classes of distributionally similar words, that we hope will be close to syntactic categories, we can then try to learn a set of morphological relationships between the words. Note that at this point we will not know which word is the inflected form of which other word – all we will have is, for example, a set of singular nouns and a set of plural nouns. In Chapter 6 I will present algorithms for learning morphology in a supervised framework, that is to say when we know the mapping or alignment between inflected and uninflected forms, and then show how they can be extended to work in the more difficult unsupervised framework.

---

[4]It is an adjective and a noun, and there are verbs such as *pout* which have the same ending.

### 3.10.3   Syntax

The syntax algorithm takes as input a set of sentences that are just sequences of tags. This is clearly inadequate in general since a lot of syntax depends on the idiosyncratic properties of particular words. The output of this will a simple phrase structure grammar. I will discuss this at length in Chapter 7.

Marcus, Santorini, and Marcinkiewicz (1993) say

> By contrast, since one of the main roles of the tagged version of the Penn Treebank is to serve as the basis for a bracketed version of the corpus, we encode a word's syntactic function in its POS tag whenever possible.

This quote illustrates the necessity of not using manually processed tags for the input for the syntax induction algorithm. Thus I have looked at how the algorithm works when provided with the automatically derived tags from Chapter 5. I also present results of the algorithm using manual tags for comparison.

### 3.10.4   Interactions between the models

We would hope to have synergies between the models, so that, for example, the morphology component can allow re-processing of the syntactic categories based on its analysis. At the moment, the errors accumulate because I have a very linear information flow. This is not an intrinsic property of this sort of model, but just a computational convenience. It is perfectly possible to combine all of these models together and perform the estimation of all of the different levels of language simultaneously. This would mean that information from one level could be used to resolve ambiguities at another level. This is one of the advantages of trying to use principled statistical models: they can be combined cleanly as Miller, Stallard, Bobrow, and Schwartz (1996) discuss.

Moreover the use of the EM algorithm permits this in quite a principled way. Given a product of two stochastic processes, one can train two models using the EM algorithm on both of them simply by combining the hidden variables of each process together: in effect by taking a Cartesian product of the dynamic programming trellises. Though sometimes this is not feasible, I have used this in Chapter 6 to show how one can go from supervised to unsupervised learning of morphology.

## 3.11   Areas of language not discussed in this thesis

In this section I shall briefly review some of the areas of language learning that I am *not* going to cover in this thesis, and discuss some relevant research.

### 3.11.1 Acoustic processing

The infant learner is presented not with a string of phonemes, but with a set of sounds. de Marcken (1995) presents a program for the unsupervised acquisition of a lexicon from a speech signal. For our purposes, this is not a relevant task. As is well known, animals can be trained to recognise phonemes in speech (see the references in Holt, Lotto, and Kluender (1998) for example). Since animals do not have language, we can safely assume that any abilities they do have will not be part of a language-specific innate ability. Since they do have these abilities, we can conclude that they are not a language-specific ability but rather part of a more general low-level perceptual analysis ability.

Moreover, there are forms of language such as sign languages, that do not require this ability, and yet these sign languages exhibit all of the complexity of spoken natural languages (Neidle, Kegl, MacLaughlin, Bahan, & Lee, 1999). There is also some interesting empirical work on how children learn to distinguish phonemes (Saffran, Aslin, & Newport, 1996; Maye & Gerken, 2000) that suggests that some sort of statistical learning algorithm, similar to those used in this thesis, accounts very well for how children actually acquire the phonemes of a language.

### 3.11.2 Phonology

There are important areas of phonology to be learned but they appear to be in situations where there is plenty of data and are easily learnable using similar techniques to those I employ in Chapter 6. For example the sort of learning discussed in Gildea and Jurafsky (1996) fits naturally into that framework. Moreover this is not an area of language which is commonly used in the APS.

### 3.11.3 Segmentation

This task involves segmenting the stream of phonemes into a sequence of words. This is an area that has been studied extensively in the cognitive science community. There have been several techniques suggested. Brent and Cartwright (1997) presents a technique using contemporary methods of statistical modelling based on distributional and phonotactic constraints, following work done by Harris (1955). In languages other than English, this could be much more difficult. For example, in French, there is the phenomenon of *liaison* and *enchainement* where phonemes at the end of one word, end up being attached to the beginning of the next word in the syllable structure. However, Christophe, Dupoux, Bertoncini, and Mehler (1994) show that there appears to be phonological evidence available to resolve this problem.

### 3.11.4 Semantics and Discourse

Here we are dealing largely with the interface between language and the real world. This clearly falls outside the aims of this thesis since we are concerned with learning purely from a set of strings of the language, not from a set of (string, meaning) pairs. That said there are two relevant areas of research. First, there is a lot of work on identifying topics, and other semantically flavoured aspects of text. Brown, Della Pietra, de Souza, Lai, and Mercer (1992) show that it is possible to cluster words based on their long-distance distributions, which produces semantic groupings. Latent Semantic Analysis (Deerwester, Dumais, Landauer, Furnas, & Harshman, 1990) can induce semantic structure from a set of diverse documents. Similar techniques could certainly classify words into semantic classes. These techniques could be useful for identifying semantic relationships between, for example, verbs and their arguments. Of course, these approaches are not really learning semantics in the traditional sense, since they are not linking the words with any references or meanings. There has, however, been some very interesting work in the area of Miniature Language Acquisition (Feldman, Lakoff, Stolcke, & Weber, 1990; Feldman, Lakoff, Bailey, Narayanan, Regier, & Stolcke, 1996). This task involves learning from a set of pairs of simple pictures, and true sentences about those pictures in a fragment of English.

# Chapter 4

# Formal Issues

## 4.1 Introduction

The point of this chapter is to rebut two possible arguments: the first is an *a priori* version of the
APS that claims that in the absence of negative evidence (being given some examples of ungram-
matical sentences, and being told that they are ungrammatical) it is not possible to learn which
sentences are not in the language. The second argument is that the algorithms I shall present do
not produce the right *sort* of grammar; in particular that since they produce a distribution over the
set of all strings of words, a Language Model, rather than a set of strings that are grammatical,
a Formal Language, that these algorithms fail to *explain* anything, they just summarise the data
in some meta-theoretically uninteresting way. This second argument is rather vague, and I shall
answer it in a rather vague way, by showing that Language Models and Formal Languages are in
fact rather more similar that is generally thought.

The structure of this chapter is as follows. In Section 4.2 I discuss this *a priori* version of the
APS; and point out some of its flaws. In Section 4.3 I discuss the formal issue of learnability, in
particular the concept of *identification in the limit* introduced by Gold (1967) and how it relates to
the feasibility of the algorithms I use here. Then in Section 4.4 I introduce a related notion that of
*measure-one learnability*, that I claim is much more relevant to the issue of learnability of natural
languages. Section 4.5 presents a slight extension of a result of Horning (1969), that shows that
under quite mild assumptions all interesting classes of languages are measure-one learnable. In
Section 4.6 I attempt to rebut the arguments using Gold's theorem that purport to establish that the
unsupervised learning of natural language is impossible. Then, in Section 4.7 I discuss the use of
statistical models in linguistics, and in Section 4.8 I discuss a formal issue relating to the choice of
distributions over strings as the output, rather than sets of strings. I argue that these are formally
very similar, at an appropriate level of abstraction.

## 4.2 The A Priori APS

The reason that the issue of negative evidence in language acquisition has been so heated is because
of a widespread belief that Gold's theorem established that without negative evidence, language
learning is impossible. Though Cowie (1999) rightly rejects Gold's theorem, she fails to reject this
second argument that follows from it. I shall discuss the applicability of Gold's theorem below;
here I shall discuss some arguments, that seem to derive from the same insight, but that do not
actually call upon Gold's theorem for support.

For example, Crain and Thornton (1998, p.20) say

> It is conceivable that constraints could be learned by children, assuming the usual
> mechanisms of induction, only if the relevant kind of evidence is available. This evi-
> dence is called *negative evidence* (or *negative data*). Negative evidence is the presen-
> tation of ungrammatical sentences, marked as such. ... If Hornstein and Lightfoot are
> correct in asserting that children lack access to negative evidence, then it follows that
> children's knowledge about the ungrammaticality of sentences (i.e. constraints) is not
> learned. Hence, this knowledge is known independently of experience; presumably,
> it is innately specified.

or in Pinker (1995, p.153)

> ...it is very important for us to know whether children get and need negative
> evidence, because in the absence of negative evidence, children who hypothesize a
> rule that generates a superset of the language will have no way of knowing that they
> are wrong (Gold, 1967; Pinker, 1979, 1989).

Geurts (2000) points out the flaw with great clarity:

> It is obvious, I take it, that the following argument is no good:
> (A) Most people know that there are no three-legged animals.
> (B) The knowledge that there are no three-legged animals is acquired in the absence
> of negative evidence. (Surely none of us have ever observed that there are no three-
> legged animals, and most of us haven't been told about this, either)
> so: (C) The knowledge that there are no three-legged animals is innate.
> This is a patent howler, but the remarkable thing is that the argument seems to
> improve if it refers to linguistic knowledge instead of knowledge of the world;

Crain and Pietroski (2001, p.166) state, with complete vacuity,

> Only negative evidence, or some substitute for it, can inform learners that they
> have overshot the target language.

A simple example will demonstrate the flaw in this argument when applied to learning. Suppose we have an alphabet of a single letter $a$, and the language L consists of all the strings of even length, $aa, aaaa, \ldots$. The argument is that no learning algorithm could learn this without being shown some negative examples, i.e. being told that some odd-length strings are not in L, because there is no way it could recover from an overly general hypothesis such as for example that L is $a^*$. In fact, all learning algorithms that I am aware of would be able to do this. Many learning algorithms start from overly general models anyway and gradually shrink them; others start from very specific models and gradually expand them.

I won't discuss this sort of argument any further since it is clear that this sort of *a priori* argumentation has very little to do with the real behaviour of actual machine learning algorithms.

## 4.3   Learnability in Formal language theory

Though we are interested in the performance of these algorithms on a finite, strictly limited set of data, it is nonetheless useful to examine the limiting behaviour, as the amount of data tends to infinity. However it is important to remember that we are only interested in this as a side-issue: our primary concern remains the behaviour on finite data sets. Moreover, as I shall argue below the formal theory of learnability (Osherson, Stob, & Weinstein, 1986) has very little to do with learnability in the normal sense. One reason for this is that the motivation for the work is very different. Learnability theorists want to find interesting, formally definable classes of languages; they are not interested in criteria that are too easy or too hard. The holy grail would be to find a definition of learnability that cut the classes of languages at a linguistically interesting point: say the class of linear indexed grammars. At the moment however, the criteria are either much too strong (allowing no interesting classes to be learned) or much too weak (allowing the class of recursively enumerable languages to be learned).

### 4.3.1 Gold

The starting point for formal theories of learnability is Gold's 1967 paper (Gold, 1967). In this paper, Gold laid the foundation for much of modern learnability theory, and proved some key negative results. Gold introduced the paradigm of *identification in the limit* (IIL). In this paradigm, the learning algorithm is presented with a sequence of strings, and after each string must make a guess. A language is *identified in the limit* for all presentations of the language, there is an *n* such that after *n* strings the language is guessed correctly. The problem here is the definition of a *presentation* of a language. A sequence of strings is a presentation of a language if it is composed only of strings of the language, and all of the strings in the language occur at least once in the sequence. Note that these are two separate, complementary criteria – the first says the presentation must not contain too wide a range of strings, the second says that it must not contain too narrow a range of strings. These restrictions are clearly absolutely necessary for any sort of learning to take place, but the requirement that each string only occur once in an infinite sequence is clearly very weak, and the criterion requires that the language is identified in the limit for every single one of these presentations, no matter how bizarre or misleading. Gold was then able to show that in this framework, any class of languages that included all finite languages and at least one infinite language (a *suprafinite* class of languages) was unlearnable. In particular, the class of regular languages and *a fortiori* the class of context-free languages are unlearnable.

Gold also showed that other sources of information would allow more interesting classes to be identified. In particular he discussed the use of *negative evidence*, that is to say examples of sentences that are not in the language, as well as the use of membership queries, i.e. allowing the algorithm to inquire as to whether a string is in the language or not. Unsurprisingly, these additional sources of information allow learnability of a wider range of classes.

## 4.4 Measure-one Learnability

### 4.4.1 Horning

Horning (1969), published only two years later, presented some techniques that have since been extended (Kapur, 1991; Kapur & Bilardi, 1992), that showed that a slightly different definition of presentation allows all recursively enumerable classes of recursive languages to be learnable, under very mild assumptions. I will outline the proof here, since it appears not to be widely known, and where known it seems to have been slightly misinterpreted. The presentation here is taken from Osherson et al. (1986), with a few changes to simplify the notation.

I shall use a particular result from measure theory (Halmos, 1950) that may not be familiar: one of the Borel-Cantelli lemmas. This says that if the sum of the measures of a set of events is finite then the measure of the set of events that happen infinitely often is zero. In terms of probability this means that if we have a set of random variables, then the probability that we have something happen infinitely often is zero if we can show that the sum of the probabilities is finite. The other Borel-Cantelli lemma says that if the sum is infinite then the probability that it will happen infinitely often is one. So suppose we toss a fair coin infinitely often. What is the probability that we will get infinitely many heads? The sum of the probabilities is $\sum_{i=1}^{\infty} 0.5$ which is clearly infinite, so the probability of infinitely many heads is one. Suppose we toss a sequence of increasingly unfair coins, so that the probability of a head with the $n$th coin is $2^{-n}$. Now the sum of the probabilities is $\sum_{i=1}^{\infty} 2^{-n} = 1$ which is finite, so the probability of infinitely many heads is zero. A third, less obvious, example is where the probability of a head with the $n$th coin is $1/n$. Here the sum of the probabilities is again unbounded, so we will have infinitely many heads with probability one.

Let $\mathcal{L}$ be a recursively enumerable set of recursive languages over some finite alphabet $A$. So in particular, the classes of context free grammars, regular grammars and so on are suitable classes. Then if we can prove this is learnable, then any suitable subset will also be learnable. If we have a learning function $\phi$, a function from finite sequences of strings to languages. We assume that each language $L$ has a measure $M_L$ associated with it – i.e. a probability distribution over it. We need to have some mild constraints on this measure (see Kapur and Bilardi (1992) for details); for our purposes we can just assume that it must be computable, which is an unnecessarily strong assumption. We assume that the support of the probability distribution and the language coincide exactly. The support of the distribution is just the set of strings that have non-zero probability.

We then extend this probability distribution to a probability distribution over all (sets of) sequences over the strings. We do this by assuming that each one is independent and identically distributed (iid), so the probability of a sequence of sentences is the product of the probabilities of each sentence with respect to the measure. We start by defining the probability distribution over the cylinder sets. Given a language $L \in \mathcal{L}$, which has an associated probability distribution function (pdf), $M_L$, and given a sequence, $s_n$ of strings from $A^*$, of length $n$, $x_1 x_2 \ldots x_n$ we can define the cylinder set of $s_n$, which we can call $C(s_n)$, as the set of all sequences that begin with $s_n$. We can then define the probability of this cylinder set as being

$$M_L'(C(S_n)) = \prod_{i=1}^{n} M_L(x_i) \tag{4.1}$$

So we now have a measure on the set of cylinder sets. If we consider the closure of the set of cylinder sets under the operations of countable union and complement we get what is called a $\sigma$-algebra, and we can extend this measure to the $\sigma$-algebra in the natural way (Halmos, 1950). I shall use $M_L$ rather than $M_L'$ to refer to this measure without I hope causing any confusion. This is a standard technique for measure theory, similar to how one defines the measurable sets of the real line: you start by defining the measure of intervals on the line in the natural way, and then you extend it to a larger set of sets. Given this measure on the set of all sequences, which will be a pdf since the measure of the set of all sequences is 1, we define, using $\mathcal{T}$ to mean the set of all infinite sequences of strings from $A^*$:

**Definition 1** $\phi$ *measure one identifies $L$ iff $M_L(\{t \in \mathcal{T} | \phi \text{ identifies } t\}) = 1$*

**Definition 2** $\phi$ *measure one identifies $\mathcal{L}$ iff $\phi$ measure one identifies all $L \in \mathcal{L}$*

**Definition 3** *$\mathcal{L}$ is measure one identifiable if there is some $\phi$ that measure one identifies $\mathcal{L}$*

So here we don't require the algorithm to identify the correct language every time, because there may be pathological sequences, such as the ones used in Gold's proof, that can mislead it. All we require is that the probability of choosing one of these weird sequences is zero. Note also that we have dispensed with the requirement that these sequences are presentations of the

language – by the Borel-Cantelli lemmas, with probability one all of the strings in the language appear infinitely often, and all the strings not in the language do not appear. Thus we have replaced the two rather arbitrary constraints on presentations in Gold's theorem with a single rather simpler requirement: that the support of the distribution is the same as the language.

I will now prove that $\mathcal{L}$ is measure-one identifiable. Let $L_1, L_2, \ldots L_l, \ldots$ be a listing of all the languages in $\mathcal{L}$, and let us further suppose that we have an ordering of all of the strings in $A^*$, shortest first. If we have a finite sequence of strings $S$, and a language $L$ we say that $S$ agrees with $L$ through $n$ iff for the first $n$ strings $s$, $s \in L \Leftrightarrow s \in S$. Intuitively this means that the sequence $S$ correctly defines all the strings at the 'beginning' of the language.. Now we define a sort of error set:

**Definition 4** $A_{L,n,m}$ *is the set of all sequences such that the first m elements of the sequence do not agree with L through n.*

So intuitively this is a set of sequences that will in some sense mislead you about $L$; the first $m$ elements do *not* tell you what the elements of the language $L$ in the first $n$ strings are. Consider the measure of this set, as we let $m$ the length of the sequence we are looking at, go to infinity.

$$\lim_{m \to \infty} M_L(A_{L,n,m}) = 0 \tag{4.2}$$

Let us prove this carefully. The set $A_{L,n,m}$ could be misleading in two ways: it could contain strings in the first $n$ strings that are not in $L$, but in this case they will have probability zero, and so the set of all of the sequences that contain at least one string not in the language will have measure zero. So we just need to show that the probability it will not contain sentences that it should contain, tends to zero as the number of samples tends to infinity. Intuitively this is obvious: there are at most $n$ strings we have to consider; as we consider longer and longer sequences, sooner or later all of these $n$ must turn up.

Let the rarest of the strings in L, that are less than $n$ have probability $p$. So the probability that a particular string $s$ does not occur is

$$M_L(\{t | s \text{ does not occur in the first } m \text{ of } t\}) \leq (1-p)^m \tag{4.3}$$

The probability that one of the first $n$ strings does not occur must be less than $n(1-p)^m$, [1] so

$$M_L(A_{L,n,m}) \leq n(1-p)^m \tag{4.4}$$

which establishes Equation 4.3. Note that we used the i.i.d assumption here.

So for a given $L$ and $n$ we define

$$d_L(n) = \text{least } m \text{ such that } M_L(A_{L,n,m}) < 2^{-n} \tag{4.5}$$

So for any $L$ and $n$ we just choose an $m$ big enough that the probability is small – intuitively $d_L(n)$ is the amount of data we need to see so that we are fairly sure about the membership of the first $n$ strings with respect to $L$. We choose the limit $2^{-n}$ so when we sum them up they are bounded, and we can apply the Borel-Cantelli lemma. Then we define

$$d(n) = \max_{h(L) \leq n} d_L(n) \tag{4.6}$$

given the enumeration $h$ of the languages; so this has a diagonal feel to the argument. What we have done here is to remove the $L$ from the picture: we look at all of the first $n$ languages, and choose the $d_L(n)$ of the "most difficult" language $L$. So we know that if we have $d(n)$ samples we are fairly sure about the first $n$ strings for all of the first $n$ languages. Now for all languages $L$ we can sum up over all $n$ as follows

$$\sum_{n=1}^{\infty} M_L(A_{L,n,d(n)}) < \sum_{n=1}^{\infty} 2^{-n} = 1 \tag{4.7}$$

---

[1] $P(A \cup B) \leq P(A) + P(B)$

Now we use the Borel-Cantelli lemma which states that if the sum of the measures of a set of events is finite then the measure of the set of events that happen infinitely often is zero. We define the set $X_L$ which intuitively is the set of all sequences where you will make infinitely many mistakes:

$$X_L = \{ t | t \in A_{L,n,d(n)} \text{ for infinitely many } n \} = \limsup_n A_{L,n,d(n)} \tag{4.8}$$

and by the aforementioned lemma $M_L(X_L) = 0$ for all $L$.

So we can now define the learning function $\phi$. Given the first $m$ elements of a text, choose the largest $n$ we can such that $d(n) \leq m$. This means that we are fairly sure that we have the right set for the first $n$ strings. Given this $n$, we choose the first language that agrees with the text through $n$.

We can now show that this learning procedure does in fact give the right answer with probability one. If the language we are learning is $L$, we will not identify it in the limit if and only if we make infinitely many mistakes; i.e. if and only if the sequence we are presented with is in the set $X_L$; we have established that the measure of this set with respect to $M_L$ is zero; therefore with probability one we identify $L$ in the limit. We have therefore demonstrated that $\mathcal{L}$ is measure one identifiable.

Horning's result is normally cited as proving that stochastic context-free grammars can be learned, and this was in fact what he proved in his thesis. As we have seen, this rather understates the generality of the result.

## 4.5 Extension to Ergodic Processes

The requirement that the sequence of strings is i.i.d is too strong, and somewhat unsatisfying, since in general sentences occur in sequences of related utterances. Osherson says (p. 186)

> It should be noted that children's linguistic environments do not typically exhibit stochastic independence in the foregoing sense.

Horning mentions the possibility of relaxing this as well (Horning, 1969, p.80)

> Strictly speaking we do not need the existence of a stochastic presentation. A virtually identical proof of Theorem V.7 can be based on convergence of the information sequence rather than on stochastic presentation. In the limit, therefore, the successive strings of the sample need not be independent, so long as the relative frequencies converge properly,

We can make it weaker by requiring that it is merely a *stationary ergodic* stochastic process. Intuitively this is the weakest criterion we can apply such that the strong law of large numbers applies. Suppose we have a stochastic process, i.e. a set of random variables $X_i$ where $i \in \mathbb{Z}$. The requirement that it is ergodic means that the time average will equal the space average (Rosenblatt, 1974). Formally, a random process is *strictly stationary* if the random variables $X_t, X_{t+1}, \dots, X_{t+l}$ have the same joint probability distribution as the random variables $X_{t+d}, \dots, X_{t+d+l}$ for all values of $d$ and $l$. We will suppose now we have a real-valued process. The particular process we will be looking at uses indicator variables, so for each string $s \in A^*$ we have a real-valued random process $I^s = \{\dots, I_t^s, \dots\}$ which takes the value 1 if the process produces $s$ at time $t$ and takes the value 0 otherwise. The requirement that it is stationary merely means that though there maybe time dependence between the variables, there is not dependence on the absolute value of the time. For example a process that has new words entering and leaving the language at particular times is not a stationary process. We clearly need to exclude this possibility, because there will be a non-zero probability that a word in the language may not appear at all. A strictly stationary process is *ergodic* if

$$\lim_{n \to \infty} \frac{1}{n} \sum_{k=0}^{n-1} f(T^k w) =^{a.e.} E_w[f(w)] \tag{4.9}$$

where $T^k$ shifts the sequence forwards by $k$.

This means that for almost all sequences $w$, (i.e except on a set of measure 0), the time average equals the average over all sequences. In this case this means that the average number of times a particular process generates a particular string will tend to the probability of that string, except for a set of weird sequences that has measure zero. For example in a process generating English, we might have a sequence that just generates the word *spam* at every time frame. Clearly in this case, the time average of $I^{spam}$ is constantly one, which is not equal to the actual probability of spam in English [2]. The measure of the set of all these unusual sequences is zero.

Given a particular language with a particular distribution, an example of a non-ergodic process would be one which for every string $s$ with probability $p(s)$, generates an infinite sequence of $s$. Clearly the expectation of $s$ over all sequences is $p(s)$ but the time average of each sequence will be 1 or 0. In this case, it is clear that no learning algorithm could learn the language, since with probability 1, the learner would only ever see a single string. Ergodicity is thus quite a natural requirement. In terms of the infant child's linguistic environment, while the requirement for i.i.d. was clearly too strong, and in fact is demonstrably false, the requirement for ergodicity seems much more plausible. Though language is not stationary in any meaningful sense over the time scales that we are concerned with we can assume that the linguistic environment is stable.

All we need to do is to prove Equation 4.2, reproduced here

$$\lim_{m \to \infty} M_L(A_{L,n,m}) = 0 \tag{4.10}$$

---

[2]65 times in the 100 million word British National Corpus. Many of these occur in a single article about spam. Note that spam is a good example of the non-stationarity of English, since it is currently used quite frequently to refer to bulk unsolicited commercial email.

Define $A_m^s$ to be the set of all infinite sequences that do not have the string $s$ in the first $m$ places. Clearly $A_{m+1}^s \subseteq A_m^s$ so they form a decreasing sequence, and therefore by the continuity theorem for decreasing sequences

$$\lim_{m \to \infty} M_L(A_m^s) = M_L \left( \bigcap_{m=1,2,\dots} M_L(A_m^s) \right) \tag{4.11}$$

The set on the right hand side here is just the set of sequences where $s$ never occurs. Now we appeal to the ergodicity to show that this set has measure zero, if $s$ is in the language.

Ergodicity says that the time average must be equal to the total expectation almost everywhere i.e. except on a set of measure zero. But the time average of $I_s$ of all the sequences in $\cap_{i=1,2,\dots} M_L(A_m^s)$ is zero, since it never occurs. But since $s$ is in the language it has non zero probability, so this set violates the ergodicity condition, and must therefore be contained in the set of measure zero that is the exception. Therefore it has measure zero. Therefore for all $s \in L$

$$\lim_{m \to \infty} M_L(A_m^s) = 0 \tag{4.12}$$

For completeness, we should also define $B_m^s$ to be the set of all sequences such that $s$ does occur in the first $m$. If $s$ is not in $L$ then the measure of this set is zero, since it is stationary.

$$A_{L,m,n} = \left( \bigcup_{s_i \in L, i \le n} A_m^{s_i} \right) \cup \left( \bigcup_{s_i \notin L, i \le n} B_m^{s_i} \right) \tag{4.13}$$

and this gives us a finite sum of sequences that tend to zero, so it tends to zero, which establishes this slight extension of the result.

Note that all of these results are concerned with exact identification at some finite point in time. If all we are concerned with is that we tend to it in the limit, we can just estimate the language as being the empirical distribution of the sample, which will tend to the correct probability by the law of large numbers.

## 4.6 Arguments from Gold's Theorem

It has sometimes been argued that Gold's theorem has implications for the learnability of natural languages by children. This argument has been put most forcefully by Matthews (1989), and also in Pinker (1979), Wexler and Culicover (1980), and is still being cited as a proof that context-free grammars, and natural languages are not learnable (Juola, 1998; Villavicencio, 2000; van Zaanen & Adriaans, 2001).

A typical statement is this (Clark, Gleitman, & Kroch, 1997):

> In this form, the "poverty of the stimulus" argument is equivalent to Gold's famous mathematical result (Gold, 1967), which Bates and Elman (Bates & Elman, 1996) misinterpret in their Perspective. Gold showed that, for even simple classes of languages, no procedure (statistical or other) exists that could learn a language without nontrivial a priori assumptions.

However, as we have seen above, the results of Gold's theorem arise only because of an overly restrictive definition, namely requiring that they correctly identify it on all texts. The more serious objection to my mind is that we are not interested in the limiting behaviour but in the behaviour on small finite data sets. Clearly any algorithm that learns a grammar on the basis of a finite amount of data will make mistakes. There are an infinite number of possible grammars compatible with any finite data set: any algorithm that chooses one rather than another has an inductive bias, and will make an error if that choice is incorrect. Clearly infants learning language fall into this situation; therefore they have an inductive bias.

Given this inductive bias, we can then define a set of possible human languages to be precisely that set which can be learnt by humans with this inductive bias. This set is learnable.

A further problem concerns the notion of "the set of sentences in a language". This is the natural way to define a formal language, but there are insuperable difficulties in defining this with a natural language. First of all, whether or not a sentence is in a natural language is an empirical matter, that can even in principle be verified for only a finite number of sentences of limited length. Secondly, even for an individual sentence determining whether it is in the language, i.e. grammatical, is fraught with methodological complexities. Even Chomsky admits (Chomsky, 1988a, p. 106)

> In fact the notion language might turn out just to be a useless notion. For example, if we fix a certain level of acceptability then this internally represented system of grammar generates one set, and we say OK that is the language. If we fix the level a bit differently, the same grammar generates a different set, and we can say that is the language. There is no meaningful answer to the question: which is the real language.

It is not in general possible to make inferences about natural languages based on the properties of formal languages without a keen awareness of the differences between the two (Kornai, 1998).

I think it is worth spending some time to dispose of this argument. It is difficult to find a very precise definition of the argument but I can sketch it roughly here:

- Gold showed that the class of context free grammars is unlearnable.

- Natural languages are in general context free. [3]

- Therefore natural languages are unlearnable.

Among the errors in this argument are: first, ignoring the distinction between formal languages, which are well-defined mathematical objects, and natural languages which are as their name implies natural objects. Secondly, ignoring the distinction between Gold's concept of identifiability in the limit, a well-defined mathematical property, and the learnability of languages; and thirdly, ignoring the fact that Gold-style learnability is a property of *classes* of language, while being context-free is a property of individual languages. Bearing these errors in mind, we can try to provide a more precise characterisation of the argument.

First we need to define what formal object corresponds to a particular natural language such as English. The set of grammatical sentences is a traditional definition (Chomsky, 1975a). But as Chomsky (1975a, p.129) points out:

> Thus we must project the class of observed sentences to a larger, in fact, infinite class of grammatical sentences.

---

[3]Leaving aside a few cases such as Bambara and Swiss German (Savitch, 1987).

But this projection is fraught with complexity as Kornai (1998) and numerous other people have pointed out. In fact even in English, there is still very substantial disagreement, not just about what the form of grammar for English is, nor even as to what the set of grammatical sentences should be, but as to what the criteria for deciding which sentences are in the language are.

Second we need to define the class of natural languages. There are a few possible candidates arranged in order from smallest to largest:

- Actual natural languages

- Historically possible languages

- Biologically possible languages

- Logically possible languages

First of all we have the smallest class of languages: the class of actual natural human languages, of which there are a small number, perhaps 4,000 on a conservative estimate. The exact number depends of course on the granularity one uses, the degree of closeness that particular idiolects or dialects must be in to count as variants of the same language. As a maximally fine-grained upper bound we could say that each person that has ever lived has spoken a different language, which would give us an upper bound of some billions, but in any event a finite number. We can next consider the class of historically possible languages: these can be thought of as all languages that might have come about given the constraints of normal human life, and language evolution. The larger class of biologically possible languages includes languages that children could learn in the normal way, but that could never have arisen, perhaps because it does not satisfy some necessary communicative goal, or has some complex property that while learnable tends to die out very quickly. Finally we have the class of logically possible languages; we could identify this with some fairly large family of formal languages, say the class of recursively enumerable languages. We shall be concerned here with the class of biologically possible languages (BPL).

The form of the argument is this: we take some natural objects which we map into formal objects, we then prove something about the formal objects and then we show that the property of the formal objects relates to some property of the natural objects. So we take the class of biologically possible languages, we map this onto a set of formal languages, we establish that the formal languages have some property (in this case the property of not being IIL), and then we attempt to show that this formal property tells us something about the natural objects.

The first step in the argument from Gold's theorem is to show that BPL contains a set of languages that is unlearnable in Gold's sense. Immediately we have a problem. The most obvious way to do this is to show that BPL is *suprafinite*; but it clearly does not contain all finite languages. For example, there are finite languages whose smallest grammatical string is longer than the number of atoms in the universe: this is not a member of BPL. One might think that this is a fixable problem: but it is not; all variants of Gold's proof require an infinite sequence of languages of increasing complexity, and BPL though an infinite class, is of bounded complexity in a sense that can be made precise. The issue is to reconcile the essentially finite nature of BPL, related to the finite computational capacities of the human brain, with the infinite requirements of the Gold paradigm. Note that here we are not making any assumptions about BPL other than the truism that if they can be generated by a human mind they must be of bounded complexity. In particular I am not assuming that they must form some algebraically definable class, say context-free languages with a bounded number of non-terminal symbols.

Let us suppose that we have overcome this problem in some way. We then can show that this class of languages is not IIL. We now have to show that this implies something about the learnability in the natural sense. In particular we need to show that if a class of languages is not IIL then it cannot be learned. The problem is that IIL is just one criterion for learnability, and that it is very strong. It requires that the class is learnable from every single presentation for the language. As we have seen, under minimal assumptions we can show that it is learnable with probability one, i.e. except on a vanishingly small set of pathologically strange texts, Moreover the criterion of identifiability in the limit itself is extremely difficult to justify, whether over all texts or just with probability one. Surely it is enough to establish that the learner in some sense comes close to the target language. After all, it is not the case that all children converge to the same grammar. There is plenty of evidence that they do not (Chomsky, 1979) and that even adults disagree substantially about the acceptability of unusual and complex sentences (Schütze, 1996, pp.99-107).

There is a further point I would like to make here. Matthews (1989, pp 60–61) states

> Context-sensitive (or context-free) languages of infinite cardinality could however, be acquired on the basis of a text sample drawn from the language to be acquired if there were further constraints *known to the learner*, on either the class of context-sensitive (or context-free) languages or the ordering of the data composing the text sample. . . . rather what is important is that a learner would be able to acquire a language of infinite cardinality on the basis of a text sample only if he had (and could use) information about these constraints. In the case of human learners such information would presumably be innate. (Emphasis is the author's)

The claim here is that constraints must be *known*. This is such an absurd idea that it is difficult to argue against it. Any learning program I write will have limitations; some I will know beforehand some I will not. None of them will be 'known' by the program. In fact some of these constraints might in fact be unknowable in quite a precise sense, just as the constraints on Turing machines computational ability are unknowable: the halting problem is not decidable. The characterisation of the limits of a learning program or learning organism as knowledge is quite unjustified, but this idea is echoed by numerous other writers. The reason for this is of course that it raises the question as to where this knowledge comes from. Seen as limitations however, it is clear they do not need much explanation. All organisms have limits, and operate within those limits. Indeed the recognition of this allows a simple characterisation of the set of natural languages as the set of languages that is within the capabilities of the human learning ability. No innate knowledge required. We do not need to claim as Chomsky (1965, p.27) does that:

> the child approaches the data with the presumption that they are drawn from a language of a certain antecedently well-defined type

The child merely approaches the data with a particular limited learning device, albeit one of great power and flexibility.

We can thus see that there are several major problems with this argument from Gold's theorem.

- Failure to identify a class of languages that satisfies the requirements of Gold's theorem

- Measure-one learnability seems a much more natural notion of learnability, but it is ignored.

- We are concerned with the behaviour of the algorithms after a finite amount of time, not the limiting behaviour.

- Confusion of formal with natural languages.

The "No Free Lunch" Theorems (NFL) (Wolpert & Macready, 1997) provide another set of limitations on the theoretical performance of learning algorithms, but I shall not discuss them here, as they appear to be general limits on learnability not related to languages, and thus apply equally to all cognitive domains.

The arguments presented here are not intended to rule out all arguments from learnability theory, merely those from Gold's theorem. I think learnability results can provide interesting guidance to computational linguistics, but perhaps more in the PAC-learnability paradigm, rather than the IIL paradigm.

## 4.7    Statistical Methods in Linguistics

As is suggested by the result above, the use of frequency information in a learning algorithm formally allows a larger class of grammars to be learned. In practice this appears also to be the case, as we shall see in Chapter 7. However the use of statistical methods in linguistics, in particular the use of probabilistic models rather than set based models (Sampson, 1987; Atwell, 1987), raises some foundational problems. This subject has been dealt with very thoroughly in (Abney, 1996; Gazdar, 1996; Goldsmith, 1998; Pereira, 2000); I shall merely discuss some additional matters after a brief summary of the arguments.

Historically, linguists have been rather antagonistic to the use of statistics. For example, Chomsky (1966) says:

> Dixon speaks freely throughout about the 'probability of a sentence' as though this were an empirically meaningful notion. ...We might take 'probability' to be an estimate of relative frequency, .... This has ...the disadvantage that almost no 'normal' sentence can be shown empirically to have a probability distinct from zero. That is, as the size of a real corpus ...grows, the relative frequency of any given sentence diminishes, presumably without limit.

and Newmeyer (1983, p. 79)

> However there are two specific criticisms of variable rules [i.e., rules with a probability attached] that, in my opinion, discredit them as motivated additions to grammatical theory. First, and most fundamentally, there is no sense in which such rules could be said to explain anything.

While at the time these arguments might have appeared plausible, to the modern ear they are clearly ridiculous. I think there are several important events that have taken place since then. First, what has been called the Bayesian revolution: a set of technical and philosophical changes in statistics incorporating a change in the view of probability as defined in terms of the *frequency* of events, to a view defined as a *subjective degree of uncertainty*. Secondly, the widespread availability of large corpora has made many people realise that useful information can in fact be gleaned from the observed distribution of sentences in a corpus, and that the problems of data sparseness to which Chomsky alludes in the quote above can be overcome with the appropriate techniques.

Thirdly, the realisation that classical, symbolic methods of Artificial Intelligence (AI), that are sometimes called GOFAI, for *Good Old Fashioned AI*, are brittle and intractably complex on large-scale real world problems (Boden, 1990), and that more robust methods are necessary, that can work with noisy and occasionally incorrect data – criteria which statistical methods satisfy.

## 4.8   Formal Unity of Statistical Grammars and Algebraic grammars

Abney (1996) says:

> Gradations of acceptability are not accommodated in algebraic grammars: a structure is either grammatical or not.

This is not quite correct. It is quite true that a simple binary distinction between grammatical and ungrammatical sentences is too blunt an instrument to separate out varying degrees of acceptability, but algebraic models need not make such a simple distinction.

The technical apparatus has been in place for some time and is discussed in depth in Kuich and Salomaa (1986), and involves an extension to the normal algebraic techniques of formal language theory (Moll, Arbib, & Kfoury, 1988, Ch.6). Broadly speaking, if we have a language, i.e. a subset of $A^*$ for some finite alphabet $A$, we can consider it instead as its *characteristic function*, a function from the set of strings to the set $\{0,1\}$. The set $\{0,1\}$ can be considered a semi-ring, which I shall define formally below, the Boolean semi-ring, $\mathbb{B}$, under operations of logical-and, and logical-or. If we then replace this semi-ring by a larger, more complex, semi-ring, and consider the functions from $A^*$ into this new semi-ring, we then have the base for an algebraic grammar that can be used to model these subtler distinctions. Any ring or field is also a semi-ring, so we can also consider the semi-ring of non-negative real numbers, in which case we get, with a bit of manipulation, the familiar distributions of statistical grammars.

A *semi-ring* is a simple algebraic object that is roughly a ring without subtraction. Formally it is a tuple $< R, +, ., 0, 1 >$, a set $R$ with two binary operations (addition and multiplication) , a zero and a unit, which are identity elements for addition and multiplication respectively. We require the two operations to be associative, addition must be commutative (but multiplication need not be), and the two operations must be distributive $a(b+c) = ab + ac$ and $(a+b)c = ab + bc$, and finally multiplication by zero must be zero $a0 = 0a = 0$. The equivalent of languages now are the functions from the set of strings over $A$ (the free monoid of $A$) into this semi-ring $R$. In the literature these are discussed in terms of formal power-series, but I shall use a simpler notation. The important thing now is that we can define a semi-ring over these languages, i.e. we can define some operations of "addition" and "multiplication" that satisfy these semi-ring axioms. Given two functions $f_1, f_2$ from $A^*$ to $R$, we define their sum to be the function that maps each $x \in A*$ to

$$(f_1 + f_2)(x) = f_1(x) + f_2(x) \tag{4.14}$$

which is pretty trivial. So in the boolean semi-ring, the sum of two languages is the union of the two languages. For multiplication we define the *Cauchy product* to be the function

$$(f_1 f_2)(x) = \sum_{x_1, x_2 : x_1 x_2 = x} f_1(x_1) f_2(x_2) \tag{4.15}$$

Here for every string $x$ we sum over all substrings that concatenate to $S$, and add up their products. In the Boolean semi-ring this is just the normal product of languages; a string is in the product is it is the concatenation of some string in the first language with some string in the second language. Kuich and Salomaa (1986, p.302) point out:

> In considerations dealing with context-free grammars, the choice of the semi-ring of the corresponding algebraic system reflects the point of view we want to emphasize. If we are interested only in the language $L(G)$ we choose the semi-ring $\mathbb{B}$. The semi-ring $\mathbb{N}$ is chosen if we want to discuss ambiguity. Also modifications of context-free grammars, such as weighted and probabilistic grammars, can be taken into account. For probabilistic grammars, the natural choice of semi-ring is $\mathbb{R}_+$ the semi-ring of nonnegative reals.

Similar points are also made by Goodman (1998).

Consider a context-free grammar that has two productions expanding a non-terminal $N$, say $N \to AB$ and $N \to CD$. Algebraically we can write this as $N = AB + CD$. This means that the set of strings that can be derived from $N$ is the union of the products of $A$ and $B$ and of $C$ and $D$. In a stochastic context free grammar, where the semi-ring is $\mathbb{R}_+$, we attach a weight to each rule, to form the equation $N = \alpha_{N,AB} AB + \alpha_{N,CD} CD$. If we want to count numbers of violations of a finite set of principles we would use the semi-ring $< \mathbb{N} \cup \{\infty\}, \min, +, \infty, 0 >$, since we want to have the minimum number of violations amongst possible derivations, and as is pointed out above we want to use $\mathbb{N}$ with the normal operations for counting the number of ambiguities.

It is worth noting that if the semi-ring has non-zero elements that sum to zero, then the generative power of the models can be much greater than expected, because selection operations can be encoded in the algebra of the ring. For example (Kuich & Salomaa, 1986, p.124), if we operate with the semiring of the integers, we can find rational power series i.e. the equivalent of regular languages, whose support is the language

$$\{a^i b^j \mid i, j \geq 1 \text{ and } j \neq i^2\} \tag{4.16}$$

which is non context-free.

Typically, Chomsky had noted the possibility of degrees of acceptability quite early (Chomsky, 1964). His proposed solution is to consider a hierarchy of categories, which gives rise to a hierarchy of utterances depending on the degree to which constraints are violated. He says (Chomsky, 1964, p.387)

> By adding a refinement to the hierarchy of categories, we simply subdivide the same utterances into more degrees of grammaticalness, thus increasing the power of the grammar to mark distinctions among utterances.

Though lacking in formal detail, a hierarchy of the sort he describes is a poset, which if it satisfies certain closure properties is a semi-ring. Thus his approach could be naturally incorporated in this framework.

However the use of $\mathbb{R}$ for the semi-ring has a large number of advantages, including a large technical apparatus of estimation techniques, and a well-defined theoretical interpretation in terms of information theory and probability. Moreover, it has a direct empirical interpretation, as the probability, which allows the use of statistical tests to evaluate comparative theories.

As noted by Alshawi (1996, p.30)

> It is not, of course, necessary for the quantities of a quantitative model to be probabilities. For example, we may wish to define real-valued functions on parse trees that reflect the extent to which the trees conform to, say minimal attachment and parallelism between conjuncts. . . . Nevertheless, probability theory does offer a coherent and relatively well understood framework for selecting between uncertain alternatives, making it a natural choice for quantitative language processing. The case for probability theory is strengthened by a well-developed empirical methodology in the form of statistical parameter estimation.

The reason for going into the issue at such length is to show that though there may appear to be a great difference between the formal devices used in statistics and in formal linguistics, at an appropriate level of abstraction they are very similar.

# Chapter 5

# Syntactic Category Induction

## 5.1 Introduction

In this chapter I present an algorithm that induces a set of syntactic categories in English. A version of this chapter appeared as Clark (2000). There have been a number of previous approaches to this problem, some concerned with cognitive science (Finch & Chater, 1992a, 1992b), and some applied to smoothing statistical models (Brown et al., 1992; Ney, Essen, & Kneser, 1994). The advantages of the approach presented here in comparison with previous work is that it can cope with words that are ambiguous or rare, and that the techniques used here can be extended to the domain of syntax acquisition, as we shall see in Chapter 7.

The rest of this chapter is organised as follows. I start by briefly discussing in Section 5.2 what syntactic or lexical categories are and whether it is necessary to use them. Then, in Section 5.3 I discuss previous work in the area of learning of syntactic categories. Section 5.4 defines the context distributions that I use in this chapter, and presents the algorithm. Then in Section 5.5, I show how this model can be applied to learning the syntactic class of ambiguous words, and Section 5.6 deals with the treatment of rare words. In Section 5.7 I present some results on an experiment with data from the British National Corpus. Section 5.8 concludes the chapter with a discussion of various limitations of the algorithm, and proposals for future work.

The set of categories produced in this chapter will then be used both in Chapter 6, where they will be used as the input for a system of learning morphology, and in Chapter 7 where they will be used in the induction of a context-free grammar.

## 5.2 Syntactic Categories

### 5.2.1 Necessity of Syntactic Categories

The notion of syntactic or lexical categories, such as Noun, Verb, Adjective and so on, is as old as linguistics itself, and appears first in the work of the Greek linguist Dionysius Thrax (c. 170 - c. 90 BC). All the words of a language can be divided into certain classes, called lexical categories or parts of speech such that words with the same or similar syntactic functions are in the same class. It is sometimes claimed that there is a well-defined set that is the same for all languages. This breaks down into two claims; first that the notion of the set of syntactic categories of a particular language is well-defined, and secondly that this set is identical for all human languages. Neither of these assumptions seem to be supported by the evidence. First of all, even in English, there is little agreement about the exact set of syntactic categories (Jespersen, 1924, Ch. 4–6). There is a general problem with choosing the level of granularity to describe them at: for example, the class of adverbs can be further sub-divided into sub-classes depending on the category of the word or phrase it modifies. In addition there are idiosyncratic lexical items – like the infinitive particle "to" in English – which can be shoe-horned into a class of auxiliary verbs, but which really have a unique syntactic function. Secondly, many languages have radically different systems: in many polysynthetic languages it is even difficult to find a good way of defining a word let alone a syntactic class that corresponds to the classes in European languages. It is of course possible to force all of these into the same Procrustean bed, but only at the cost of removing all empirical content from the proposition.

Nonetheless, it is in my opinion necessary to use some similar idea to alleviate the problems of data sparseness. The number of different types in a corpus is too large with respect to the number of tokens to allow one to express significant generalisations except over very frequent words. Since many words are infrequent, it is necessary to group them together into classes of similar words. The work of van Zaanen (2000), which I shall discuss in Chapter 7, seems to be an exception to this, in that he tries to learn syntax without using parts of speech. He uses a very small corpus with a limited vocabulary, the ATIS corpus, which may account for his success.

### 5.2.2 Learning categories

We are therefore looking for an algorithm that will take as input a sequence of phonemes or letters, segmented into words, and will divide these words into various classes. Since words are frequently ambiguous, we must allow words to be in more than one class.

Within the context of this thesis we must ask what sort of information is available for the learning algorithm to work with. This takes place at a very early phase of language learning so we have to use superficial properties of the language stream.

**Local Distributional Information** We can look at the pattern of words that each word occurs in.

**The form of the word itself** We can look at the sequence of letters or phonemes that make up the word.

**Morphological information** We can look at the existence of other words that are related to it.

**Frequency information and Burstiness** We can look at how frequently it occurs, and how its occurrences are distributed in different discourses.

Morphological information is one of the defining characteristics of the notion of syntactic class, as McCawley (1988, p. 183) points out:

> Lexical category is as much a notion of morphology as one of syntax, in that what an item is inflected for provides sufficient grounds for assigning it to a lexical category, e.g. an English word that is inflected for tense and for agreement with its subject in person and number is a verb regardless of any other facts about its behavior.

The techniques used in this chapter use just a single source of information: local distributional information. As has been noted before, this very limited source of information is sufficient to allow the algorithm to identify the various syntactic classes with high accuracy, at least in English.

It has been suggested that semantics can account for learning syntactic categories – the so-called semantic bootstrapping theory (Pinker, 1996; Braine, 1988, 1992). This seems implausible for various reasons. First, there are numerous theoretical difficulties with defining syntactic categories in terms of semantics (Jespersen, 1924). Words with similar or identical semantics can be in different syntactic classes (execute, execution, executing). Secondly the sort of semantic information available to the child must be by assumption cross-linguistically valid, but the syntactic information we want to acquire is specific to a particular language. Finally identifying the role of closed class words is clearly impossible using this method, and must be performed using some other technique.

Even if we hypothesise that these closed class categories are innate, a difficult assumption given the high cross-linguistic variability in the set of lexical categories, the infant learner is still faced with the difficulty of working out which words correspond to which classes – the so-called linkage problem.

It is important to note that the fact that local distributional information is relevant can be learnt by a simple principal component analysis on the sequences of words. Thus this does not constitute a piece of domain specific knowledge – it is a directly observable property of the input signal, that words have an effect on which words follow them.

## 5.3   Previous Work

There has been a certain amount of previous work on using local distributional information to identify syntactic categories. Previous work can be divided into two broad categories. A number of researchers have obtained good results using pattern recognition techniques. Finch and Chater (1992a, 1992b) and Schütze (1993, 1997) use a set of features derived from the co-occurrence statistics of common words together with standard clustering and information extraction techniques. For sufficiently frequent words this method produces satisfactory results. On the other hand, Brown et al. (1992) use a very large amount of data, and a well-founded information theoretic model to induce large numbers of plausible semantic and syntactic clusters. Both approaches have two flaws: they cannot deal well with ambiguity, though Schütze addresses this issue partially, and they do not cope well with rare words. Since rare and ambiguous words are very common in natural language, these limitations are serious. I will now discuss this work in more detail. Brill (1991) presented some similar work, but with a limited evaluation.

### 5.3.1   Harris and Lamb

Harris (1954) introduced the idea of using distributional analysis to identify syntactic classes. The first concrete application of this work that I have found is a remarkably prescient paper, Lamb (1961, p.679) which shows this approach in full detail.

> In the course of the analysis, groupings of two kinds will be made. These may be referred to as *horizontal* or *vertical* groupings, or H-groups and V-groups for short. A vertical grouping or V-group is a grouping of items (and/or sequences of items) into a distribution class or an approximation to a distribution class. An H-group or horizontal grouping is a grouping of constituents of a construction (or tentative construction) into a constitute.

> But how is the machine going to make these V-groups and H-groups? Zellig Harris, in his procedure-oriented *Methods in Structural Linguistics* set up distribution classes of morphemes before considering horizontal groupings. To do so in a meaningful way requires that items grouped together be found in identical environments, extending several items on either side. It would be futile to attempt such an approach even with a machine because a corpus of truly collossal[sic] size would be required, and even the computer has limits with regard to the volume of data that can be processed at high speed.

This paper shows all of the key elements of this approach. Obviously at that time (1961), neither the computational power nor the linguistic data were available, and the relevant mathematical apparatus, though well-defined, was not widely known.

### 5.3.2 Finch and Chater

Finch and Chater in a series of papers (Finch & Chater, 1992a, 1992b) showed that it was possible to induce a set of syntactic categories from unlabelled data. Their approach was to define a set of features derived from the co-occurrence statistics of frequent words. They considered the context of a word to be the two words before and after each word. They defined a set of features corresponding to the number of times various frequent context words occurred in particular positions relative to the target word that the features are being calculated for. They then calculated the similarity between words by using Spearman's rank correlation, and clustered the results using a standard hierarchical clustering algorithm. They only performed this for the one or two thousand most frequent words, for which they get good results – their algorithm successfully identified various clusters corresponding to traditional notions of syntactic category. This approach has two weaknesses: first it only works for words that occur very frequently as the estimates of the features require many data points, and secondly, this model cannot cope with ambiguous words. However these are important papers because they established that local distributional evidence *alone* is sufficient to identify syntactic categories in English. In later work, they showed (Redington et al., 1995) that the same techniques were also applicable to the learning of syntactic categories in Chinese. Similar work has also been presented on Japanese by other researchers (Mori, Nishimura, and Ito (1997) cited in Mori and Nagao (1998)). A similar approach was also discussed in Brill (1991).

### 5.3.3 Schütze

Schütze (1993, 1997) is an extension of this work that is technically rather more sophisticated. Schütze's principal innovation was the use of a Singular Value Decomposition (SVD) to alleviate the sparseness problems. The SVD is a linear algebra operator that is a way of approximating high dimensional spaces with low dimensional spaces. This has the effect of allowing better performance with rarer words. He also evaluated it more carefully.

### 5.3.4 Brown et al.

Brown et al. (1992) present a different set of techniques. They consider the problem of finding a partition of the words into classes that will maximise the likelihood of a class-based markov model. They show that under certain assumptions this is equivalent to finding the partition that has the maximum mutual information amongst clusters. They use this technique to identify 1000 classes of words that are syntactically similar. They also show how a similar technique can be used to identify semantically similar words. Ney et al. (1994) contains amongst much else, a very similar approach. These two approaches are designed to maximise the likelihood of a model: the plausibility of the classes produced is irrelevant for their work. There has been quite a lot of similar work in language modelling (Niesler, 1997). Grünwald (1996) presents a very similar algorithm, but using a MDL approach to control the number of clusters.

### 5.3.5 Pereira, Tishby and Lee

Pereira, Tishby, and Lee (1993), Pereira and Lee (1999) present an technically interesting algorithm for clustering nouns and verbs into subclasses based on their distributional contexts. They use the Kullback-Leibler divergence (KLD) to measure the similarity between clusters defined as

$$D(p||q) = \sum_x p(x) \log \frac{p(x)}{q(x)} \tag{5.1}$$

Of course what they are actually using is the KLD between the *empirical distributions*, which is not at all the same thing. Estimating the KLD between two distributions by calculating the KLD between the empirical distributions (the ML estimator) is rather misguided – an error that is repeated at greater length in Lee (1999). They discuss smoothing before calculating the KLD, but this again produces a poor estimator. Wolf and Wolpert (1992, 1993) discuss the various more well-founded algorithms for performing these calculations. That said, I will now proceed to make *exactly* the same errors below, when I use the KLD on the smoothed empirical distributions.

### 5.3.6 Dagan et al.

Dagan, Marcus, and Markovitch (1993) do not classify words into classes, rather they define a notion of similarity between words and between word-pairs, and use this to smooth information about particular word pairs by averaging them over similar pairs. This can be thought of as a sort of non-parametric clustering, that they use to perform various tasks including word-sense disambiguation.

### 5.3.7 Li and Abe

Li and Abe (1996, 1998) use an MDL approach to cluster nouns and verbs into separate classes, to model their co-occurrence, and use the resulting model to perform structural disambiguation on data extracted from the Wall Street journal corpus.

### 5.3.8 Brent

Brent (1997) is in a slightly different vein. He uses a small amount of lightly-edited child-directed speech, together with a template grammar to induce some very accurate syntactic clusters. In some respects this is similar to the approach of van Zaanen (2000) in syntax. However it is not clear how well this approach will work with noisy, more complex data.

## 5.4 Context Distributions

All of these methods share the same basic intuition, i.e. that similar words occur in similar contexts. I formalise this in a slightly different way: each word defines a probability distribution over all contexts, namely the probability of the context given the word. Thus if we restrict ourselves to sentences, contexts can be considered to be sentences with a single gap. For a given word $w$ we can look at the probability of a context given that the gap is filled by $w$. Thus possible contexts are for example

- The cat $\triangle$ on the mat.

- The $\triangle$ sat on the mat.

where we use the symbol $\triangle$ to designate the gap. Intuitively the probability of the first sentence is much greater than the second if the gap is filled by a finite verb such as *sat*, and vice-versa if the word is *cat* or some other noun.

We can formalise the context as an ordered pair of strings from the vocabulary $V$, the string before the gap and the string after. The set of contexts $\mathcal{C}$ is then $V^* \times V^*$. For each word $w$ we then have a probability distribution over $\mathcal{C}$ which we can denote by $p^w$. We can then measure the similarity between the words by measuring the similarity between these distributions. The difficulty is to find a measure of similarity that can be calculated based on the meagre evidence available.

A first step is to divide the set of contexts into equivalence classes so that we can consider only local context. This places some constraints on the form of the distance function we should use. So we consider two contexts to be equivalent if the words immediately to the left of the gap are equal and the words immediately to the right are equal. Here I will consider a narrow context of just one word to the left and one word to the right. So the contexts " The cat $\triangle$ on the mat" and "The dog and a cat $\triangle$ on the table over there." are equal and we can consider each equivalence class to be an ordered pair of words, namely the word before and the word after. So now we can define the context distribution to be a distribution over all ordered pairs of words. Thus the context distribution of the word *cat* will have a higher probability for contexts such as $< the, is >$, and lower for contexts like $< dog, the >$. The context distribution of a word can be estimated from the observed contexts in a corpus. We can then measure the similarity of words by the similarity of their empirical context distributions,

There are a number of possible choices for a distance function. The Kullback-Leibler divergence (KLD) is a natural choice, and as we shall see below, given some simplifying assumptions, the divergence between the distributions is equal to the divergence between the distributions over the equivalence classes. Informal experiments established that other distance measures produced similar results. A drawback of using the KLD is that it is not a metric as it is not symmetric and does not satisfy the triangle inequality: this means that some clustering algorithms will not necessarily converge (Lance & Williams, 1967). Another potential drawback is that it is undefined, or rather produces an infinite value if there is some $x$ such that $p(x) > 0$ while $q(x) = 0$. This is only a problem if we confuse the empirical with the true distributions, and fail to smooth.

To reduce this problem, we will generally be estimating the KLD between the distribution of a particular sequence, and the distribution of some cluster, that will be estimated from rather more data. If we have $n$ counts to estimate $p$, we can write $p(x) = c(x)/n$. In this case the closest cluster will be the maximum likelihood cluster as we can see from rewriting the KLD thus,

$$D(p||q) = \sum_x p(x) \log \frac{p(x)}{q(x)} = -H(p(x)) - \frac{1}{n} \log(\prod_x q(x)^{c(x)}) \qquad (5.2)$$

The first term is just the entropy and the second term is the average negative log-likelihood Thus since the entropy of $p$ is constant we would merely select the most likely model.

That noted, the solution I use here is far from correct. In general if we use a smoothing technique based on Bayesian estimation, and we are interested in calculating a non-linear function of the distribution such as the entropy, or the mutual information, then it is incorrect to estimate this by smoothing the empirical distribution, and then calculating the function of that. We have in effect two operators; one which smooths, and one which calculates the function. What we want is a smoothed estimate of the function, not the function of the smoothed distribution. The two operators do not commute: if the function is linear, such as the mean, then they do commute, and the Bayesian estimate of the mean is the same as the mean of the Bayesian estimate of the distribution. See Wolf and Wolpert (1992, 1993) for a detailed analysis and discussion.

### 5.4.1 Clustering

Unfortunately it is not possible to cluster based directly on the context distributions for two reasons: first the data is too sparse to estimate the context distributions adequately for any but the most frequent words, and secondly some words which intuitively are very similar (Schütze's example is 'a' and 'an') have radically different context distributions. Both of these problems can be overcome in the normal way by using clusters: approximate the context distribution as being a probability distribution over ordered pairs of clusters multiplied by the conditional distributions of the words given the clusters :

$$p(<w_1, w_2>) = p(<c(w_1), c(w_2)>)p(w_1|c(w_1))p(w_2|c(w_2))$$

where we write $c(w)$ for the cluster that word $w$ is in, or for brevity $c_1$ for $c(w_1)$, and so on. Note that under this model, when we have conditional distributions that are the same for $p$ and $q$, we can simplify the calculation of the KLD to

$$
\begin{aligned}
D(p_1||p_2) &= \sum_{w_1, w_2} p_1(<w_1, w_2>) \log \frac{p_1(<w_1, w_2>)}{p_2(<w_1, w_2>)} \\
&= \sum_{w_1, w_2} p_1(<c_1, c_2>)p(w_1|c_1)p(w_2|c_2) \log \frac{p_1(<c_1, c_2>)p(w_1|c_1)p(w_2|c_2)}{p_2(<c_1, c_2>)p(w_1|c_1)p(w_2|c_2)} \\
&= \sum_{c_1, c_2} \sum_{w_1 \in c_1} \sum_{w_2 \in c_2} p_1(<c_1, c_2>)p(w_1|c_1)p(w_2|c_2) \log \frac{p_1(<c_1, c_2>)}{p_2(<c_1, c_2>)} \\
&= \sum_{c_1, c_2} p_1(<c_1, c_2>) \log \frac{p_1(<c_1, c_2>)}{p_2(<c_1, c_2>)}
\end{aligned}
\tag{5.3}
$$

that is, it is just the divergence between the context distributions over the clusters.

### 5.4.2 Algorithm Description

I use an iterative algorithm, starting with a trivial partial clustering, with each of the $K$ clusters filled with the $k$th most frequent word in the corpus. At each iteration, I calculate the context distribution of each cluster, which is the weighted average of the context distributions of each word in the cluster. The distribution is calculated with respect to the $K$ current clusters and a further "ground" cluster of all unclassified words: each distribution therefore has $(K+1)^2$ parameters. For every word that occurs more than 50 times in the corpus, I calculate the context distribution, and then find the cluster with the lowest KL divergence from that distribution. I then sort the words by the divergence from the cluster that is closest to them, and select the best as being the members of the cluster for the next iteration. This is repeated, gradually increasing the number of words included at each iteration, until a high enough proportion has been clustered, for example 80%.

After each iteration, if the distance between two clusters falls below a threshold value, the clusters are merged, and a new cluster is formed from the most frequent unclustered word. Since there will be zeroes in the context distributions, they are smoothed using Good-Turing smoothing (Good, 1953) to avoid singularities in the KL divergence. At this point we have a preliminary clustering – no very rare words will be included, and some common words will also not be assigned, because they are ambiguous or have idiosyncratic distributional properties.

## 5.5 Ambiguous Words

Ambiguity can be handled naturally within this framework. The context distribution $p^{(w)}$ of a particular ambiguous word $w$ can be modelled as a linear combination of the context distributions of the various clusters. We can find the mixing coefficients by minimising $D(p^{(w)} || \sum \alpha_i^{(w)} q_i)$ where the $\alpha_i^{(w)}$ are some coefficients that sum to unity and the $q_i$ are the context distributions of the clusters.

We want to minimise

$$D = D(p || \sum_i \alpha_i q_i) \tag{5.4}$$

subject to the constraint $\sum_i \alpha_i = 1$. If we proceed in the normal way using the method of Lagrangian multipliers (see e.g. (Bishop, 1995, Appendix C)) we minimise

$$L = D(p || \sum_i \alpha_i q_i) + \lambda(\sum_i \alpha_i - 1) \tag{5.5}$$

Differentiating we get

$$\frac{\partial L}{\partial \alpha_i} = \lambda + \sum_x p(x) \frac{-q_i}{\sum_i \alpha_i q_i} = 0 \tag{5.6}$$

which is a polynomial of order the number of non-zero elements of $p(x)$, which it is not practical to solve in closed form. Here, $x$ is a variable which ranges over all $K^2$ contexts. There are a number of methods in non-linear optimisation that one can use to find a minimum – I chose the EM algorithm (Dempster et al., 1977) which gives good results, and has a simple form since this reduces to a mixture model estimation problem. We use an iterative algorithm and the update equation is:

$$\alpha_i^{new} = \sum_x p(x) p^{old}(i|x) = \sum_x p(x) \frac{\alpha_i^{old} q_i(x)}{\sum_i \alpha_i^{old} q_i(x)} \tag{5.7}$$

So in summary we weight each cluster by the posterior probability that each one is in the cluster. When the $p$ distribution is unsmoothed, this is the same as maximising the probability of the contexts with respect to a mixture of the cluster distributions.

There are often several local minima, that the EM algorithm will get stuck in, since it is a hill-climbing algorithm – in practice this does not seem to be a major problem. I chose to initialise the models with a uniform distribution for the $\alpha$s, which seemed to give the best results.

## 5.6 Rare words

The observed context distributions of rare words may be insufficient to make a definite determination of its cluster membership. In this case, under the assumption that the word is unambiguous, which is only valid for comparatively rare words, we can use Bayes's rule to calculate the posterior probability that it is in each class, using as a prior probability the distribution of rare words in each class.

$$P(w \text{ is in cluster } i) = \frac{1}{C} p(\text{contexts of } w | w \text{ is in cluster } i) p(w \text{ is in cluster } i | w \text{ is a rare word}) \tag{5.8}$$

This incorporates the fact that rare words are much more likely to be adjectives or nouns than, for example, pronouns. Dermatas and Kokkinakis (1995) and Baayen and Sproat (1996) discuss this issue. Here I use a slightly different approach based on the frequency of *types* rather than *tokens*.

Figure 5.1 shows for comparison three possible priors.

$$p_{\text{token prior}}(c_i) = \frac{\sum_{w \in c_i} n_w}{\sum_w n_w} \tag{5.9}$$

$$p_{\text{type prior}}(c_i) = \frac{\sum_{w \in c_i} 1}{\sum_w 1} \tag{5.10}$$

$$p_{k\text{-sparse prior}}(c_i) = \frac{\sum_{w \in c_i \wedge n_w \leq k} n_w}{\sum_w n_w} \tag{5.11}$$

where $n_w$ is the number of times word $w$ appears in the corpus.

In this work I used the type prior. The only slight practical difference is that this prior assigns non-zero probability to all of the clusters, whereas some of the closed-class clusters have zero probability with respect to the sparse prior. Surprisingly not all closed class words appear in the 12 million word subcorpus that these figures were calculated from. In particular the very rare pronoun "ourself" does not. Moreover I feel that the type prior is somehow more natural.

## 5.7 Results

I used 12 million words of the British National Corpus as training data, and ran this algorithm with various numbers of clusters (77, 100 and 150). All of the results in this chapter are produced with 77 clusters corresponding to the number of tags in the CLAWS tagset used to tag the BNC, plus a distinguished sentence boundary token. In each case, the clusters induced contained accurate classes corresponding to the major syntactic categories, and various subgroups of them such as prepositional verbs, first names, last names and so on. The precise clusters produced obviously depend on the number of clusters.

Tables 5.1, 5.2 , 5.3 and 5.4 show various sets of classes. Each line shows the five most frequent words in a particular class. I have separated the classes manually into various categories for clarity. In general, as can be seen, the clusters correspond to traditional syntactic classes.

In Table 5.1 we see various classes of orthographic marks. The `&SENTENCE` token is the distinguished sentence boundary token I inserted between each sentence during preprocessing. `&HELLIP` is an ellipse. Note that the separator marks comma, semi-colon and so on are in a different class from the terminator marks. This is because the terminator marks have a strong tendency to occur before the sentence boundary marker.

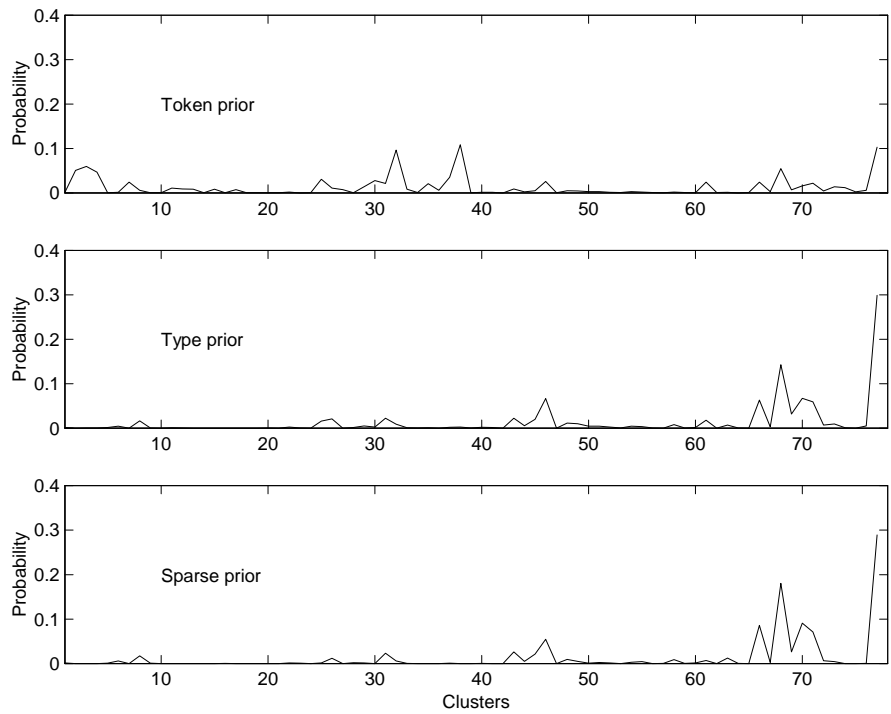

Figure 5.1: Three prior cluster probability distributions. The sparse prior is for $k = 1$. Note that the type prior and the sparse prior are almost identical.

| Clusters |
|---|
| `&SENTENCE` |
| `, &MDASH ( : ;` |
| `. ? !` |
| `&BQUO` |
| `&EQUO` |
| `'S '` |
| `&HELLIP ..` |

Table 5.1: Various clusters of orthographic marks

Table 5.2 shows the various clusters of closed class words. There are a few errors – notably, the right bracket is classified with adverbial particles like "UP", but it successfully separates the pronouns into nominative and accusative classes. Note that "YOU" appears in the accusative class. "WHICH" and "WHO" appear in a separate class, which is quite encouraging, since their syntactic roles are rather long-distance in effect.

Table 5.3 shows various open class clusters. There are a couple of comments to make: first that each syntactic category as traditionally conceived is here split into several classes. As can be seen, the splits in the classes correspond largely to differences in the subcategorisation of the words: thus the various classes of verbs in their base form are divided into a class of verbs such as "THINK", which often are followed by "THAT", verbs such as "GO" that are often followed by a preposition, and so on. Secondly, some ambiguous words are at this stage assigned unambiguously to a particular class: so the word "US" is here assigned to the class of proper nouns. In addition, the words in these classes often have some morphological properties in common, even though this was not used at all in the algorithm. So we have a class of comparative adjectives "LONGER", "BIGGER", a class of verbs in the present participle form and so on.

Table 5.4 shows two classes that are a little unusual. The first class is just a class of various rather rare words that occur in unusual distributions, and the second class is a class of words that form the first word of a very fixed pair of words: "the Nez Perce", a tribe of Native Americans, and "the Khmer Rouge".

To examine the behaviour of the algorithm with regard to ambiguous words, for each word $w$, I then calculated the optimal coefficients $\alpha_i^{(w)}$. Table 5.5 shows some sample ambiguous words, together with the clusters with largest values of $\alpha^i$. Each cluster is represented by the most frequent member of the cluster. Note that "US" is a proper noun cluster. As there is more than one common noun cluster, for many unambiguous nouns the optimum is a mixture of the various classes. Figures 5.2 and 5.3 each show some graphs for four ambiguous words, that show the values of $\alpha$ for each cluster. In Figure 5.2 the words "MAY", "THIS" , "ROSE" and "VAN" are shown. "MAY" is ambiguous primarily between the modal auxiliary and the month of the year, "THIS" is ambiguous between the determiner (this dog . . . ) and the demonstrative pronoun (this is . . . ), and "VAN" is ambiguous between the common noun, and the honorific in surnames in

```
THE A HIS THIS AN
OF IN FOR ON WITH
IS WAS HAD HAS DID
) UP OUT BACK DOWN
TO
BECOME
BUT WHEN IF WHERE BECAUSE
AND AS OR UNTIL SUCH␣AS
WILL WOULD CAN COULD MAY
IT EVERYBODY GINA
NOT BEEN N'T SO ONLY
ONE ALL MORE SOME TWO
WHICH WHO
WHAT HOW WHY HAVING MAKING
THERE
THOSE
HE I THEY SHE WE
YOU THEM HIM ME THEMSELVES
SOMETHING ANYTHING SOMEONE EVERYTHING EVERYONE
NOTHING NOWHERE RISEN
THAT BEFORE ABOVE OUTSIDE BELOW
THAN
```
```
'M AM
'D
CA WO AI
'RE
'LL
'VE
```

Table 5.2: Closed class words

| |
|---|
| HOWEVER OF␣COURSE FOR␣EXAMPLE INDEED |
| NEVERTHELESS |
| FAR INFINITELY |
| LATER AGO EARLIER THEREAFTER |
| ENOUGH |
| IMPORTANT POSSIBLE CLEAR HARD CLOSE |
| NEW OTHER FIRST OWN GOOD |
| LAST NEXT GOLDEN FT-SE |
| BETTER WORSE LONGER BIGGER STRONGER |
| YEARS PER␣CENT DAYS TIMES MONTHS |
| GROUP NUMBER SYSTEM OFFICE CENTRE |
| PEOPLE WORK LIFE RIGHT END |
| WORLD GOVERNMENT PARTY FAMILY WEST |
| US BRITAIN LONDON GOD LABOUR |
| TIME WAY YEAR DAY MAN |
| PART SORT THINKING LACK NONE |
| NEED NEEDS SEEM ATTEMPT OPPORTUNITY |
| FACT IMPRESSION ASSUMPTION IMPLICATION |
| MR MRS DR HONG MR. |
| CHARLES MARK PHILIP HENRY MARY |
| JOHN SIR DAVID ST DE |
| KLERK CLOWES HOWE COLI GAULLE |
| ARE WERE |
| BE HAVE DO MAKE GET |
| WANT WANTED TRIED WISH WANTS |
| MADE USED FOUND LEFT PUT |
| BASED RESPONSIBLE COMPARED INTERESTED ASSOCIATED |
| SAID SAYS WROTE EXPLAINED REPLIED |
| THOUGHT FELT KNEW DECIDED HOPE |
| ASKED LIKED WATCHED SMILED INVITED |
| CAME WENT LOOKED SEEMED BEGAN |
| TOOK TOLD SAW GAVE MAKES |
| LOOK RUN LIVE MOVE TALK |
| USE HELP FORM CHANGE SUPPORT |
| THINK BELIEVE SUPPOSE INSIST RECKON |
| GO COME TRY CONTINUE APPEAR |
| SEE SAY FEEL MEAN REMEMBER |
| KNOW UNDERSTAND REALISE |
| GOING ABLE LOOKING TRYING COMING |
| COPE DEPEND CONCENTRATE SUCCEED COMPETE |
| SUCH USING PROVIDING DEVELOPING WINNING |

Table 5.3: Open class words

```
RO HVK AMEN
NEZ KHMER
```

Table 5.4: Errors. The second class here is the class of words that appear invariably as the first word in a two word compound – Nez Perce and Khmer Rouge, respectively.

| Word | Clusters | | |
|------|------|------|------|
| ROSE | CAME | CHARLES | GROUP |
| VAN | JOHN | TIME | GROUP |
| MAY | WILL | US | JOHN |
| US | YOU | US | NEW |
| HER | THE | YOU | |
| THIS | THE | IT | LAST |

Table 5.5: Ambiguous words. For each word, the clusters that have the highest $\alpha$ are shown, if $\alpha > 0.01$.

Dutch. "ROSE" I have specially selected as it is an example of the worst case scenario in English: it is massively ambiguous between a variety of open-class words. It is (at least) a common noun, a colour word, a surname, a first name, and the past tense of the verb "to rise". Nonetheless as the graph shows, this algorithm does a good job of separating out the various forms.

Figures 5.4 and 5.5 show the behaviour of the algorithm for two rare words that occur ten times each in the corpus, "PRE-EMINENTLY" and "BUSINESSWOMAN". The graphs show the posterior probability of the word being in each cluster, varying with respect to the number of occurrences of the word, from zero, when it is purely the prior, up to ten, the full number of appearances of each word in the corpus.

Table 5.6 shows the accuracy of cluster assignment for rare words. For two CLAWS tags, AJ0 (adjective) and NN1 (singular common noun) that occur frequently among rare words in the corpus, I selected all of the words that occurred $n$ times in the corpus, and at least half the time had that CLAWS tag. I then tested the accuracy of my assignment algorithm which I call CDC for Context Distribution Clustering, by marking it as correct if it assigned the word to a 'plausible' cluster – for AJ0, either of the clusters "NEW" or "IMPORTANT", and for NN1, one of the clusters "TIME", "PEOPLE", "WORLD", "GROUP" or "FACT". I did this for $n$ in $\{1, 2, 3, 5, 10, 20\}$. For comparison, I proceeded similarly for the Brown clustering algorithm, (Brown et al., 1992) selecting two clusters for NN1 and four for AJ0. This can only be approximate, since the choice of acceptable clusters is rather arbitrary, and the BNC tags are not perfectly accurate, but the results are quite clear; for words that occur 5 times or less the CDC algorithm is more accurate. Note that the Brown algorithm outperforms the CDC algorithm on the AJ0 class for higher frequency words. This is probably because my choice of four Brown clusters for AJ0 was overly generous; conversely, my choice of five CDC clusters for NN1 is probably also too generous, and overstates the merits of the CDC algorithm on that class.

Evaluation is in general difficult with unsupervised learning algorithms. Previous authors have relied on both informal evaluations of the plausibility of the classes produced, and more formal statistical methods. Comparison against existing tag-sets is not meaningful – one set of tags chosen by linguists would score very badly against another without this implying any fault as there is no 'gold standard'. I therefore chose to use an objective statistical measure, the perplexity of a very simple finite state model, to compare the tags generated with this clustering technique against the BNC tags, which uses the CLAWS-4 tag set (Leech, Garside, & Bryant, 1994) which had 76 tags. This is by no means an ideal measure, since the perplexity does not directly relate to what I am trying to achieve here.

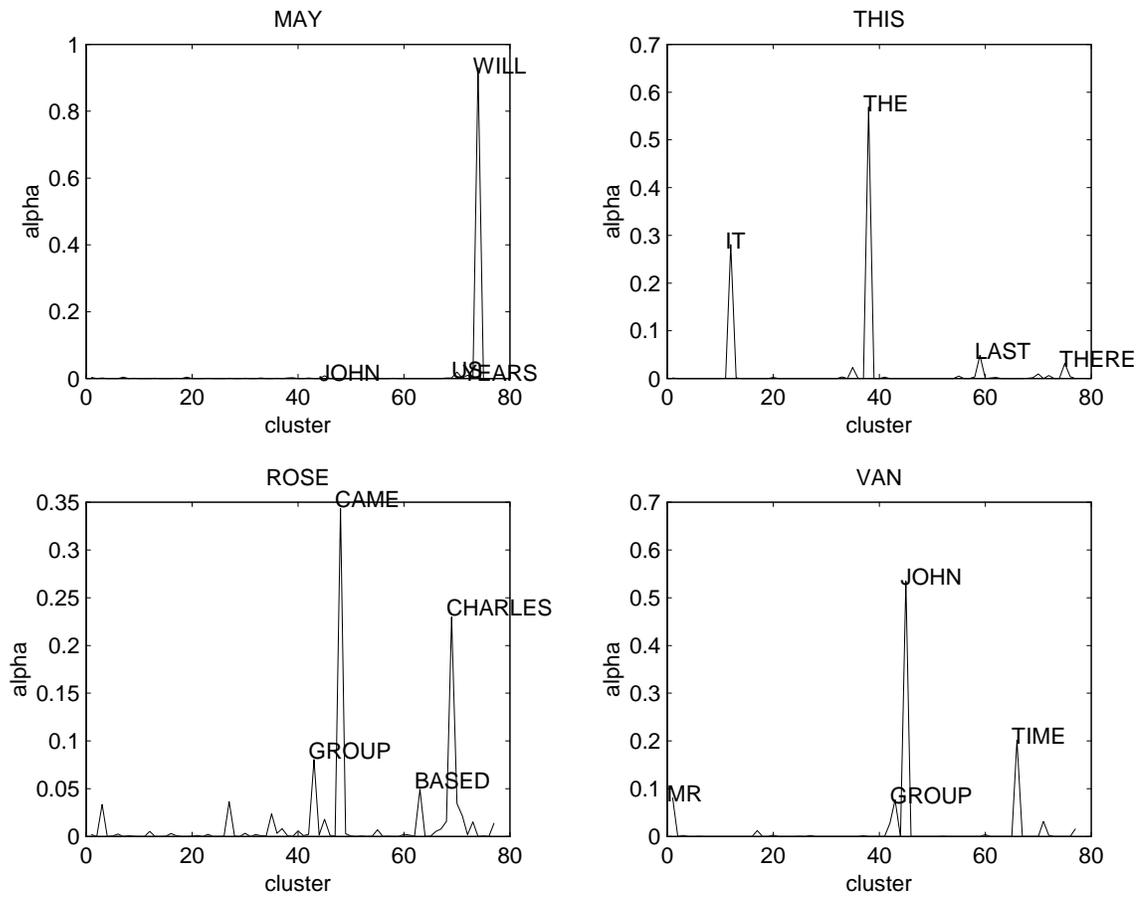

Figure 5.2: Some ambiguous words

| Model | CDC | Brown | CDC | Brown |
|-------|-----|-------|-----|-------|
| Freq | NN1 | NN1 | AJ0 | AJ0 |
| 1 | 0.66 | 0.21 | 0.77 | 0.41 |
| 2 | 0.64 | 0.27 | 0.77 | 0.58 |
| 3 | 0.68 | 0.36 | 0.82 | 0.73 |
| 5 | 0.69 | 0.40 | 0.83 | 0.81 |
| 10 | 0.72 | 0.50 | 0.92 | 0.94 |
| 20 | 0.73 | 0.61 | 0.91 | 0.94 |

Table 5.6: Accuracy of classification of rare words with tags NN1 (common noun) and AJ0 (adjective).

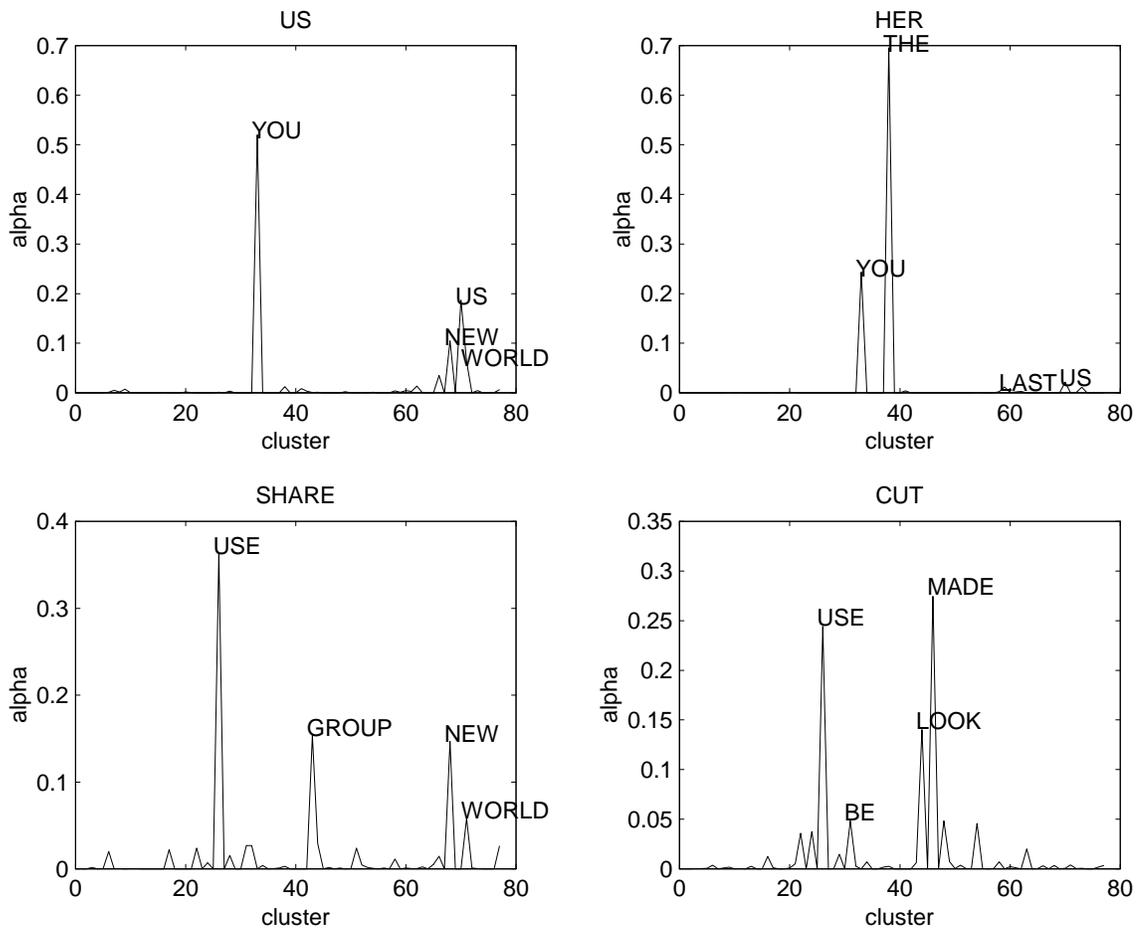

Figure 5.3: Some ambiguous words

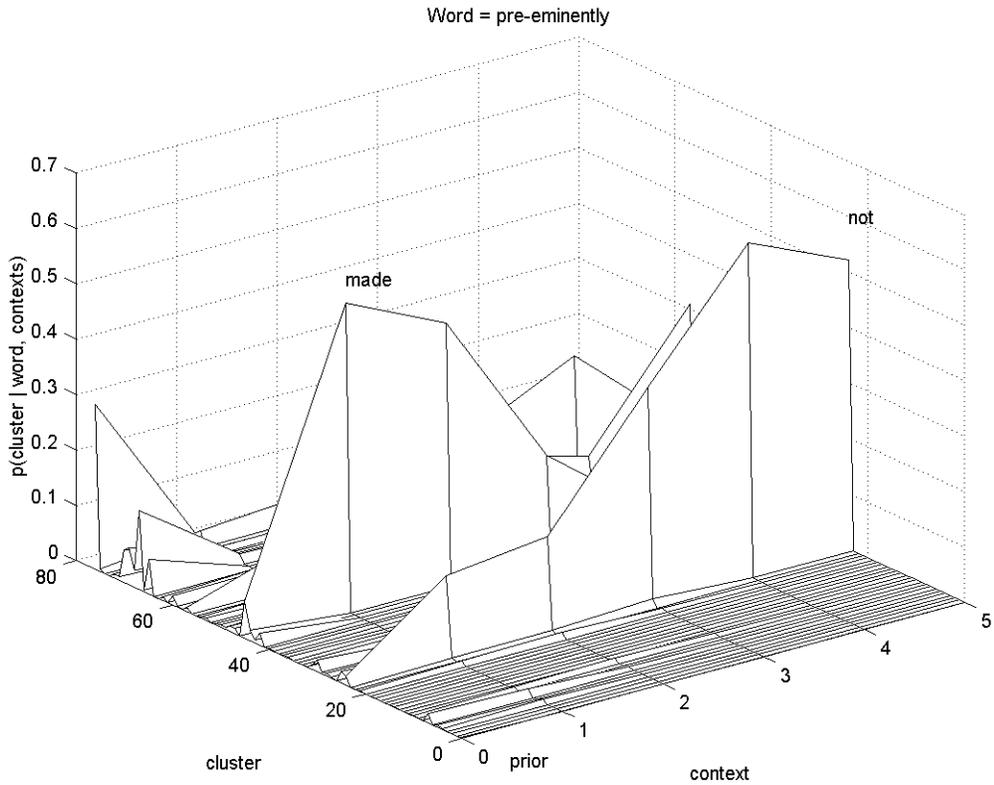

Figure 5.4: A rare word, showing how the cluster membership probabilities change after each context

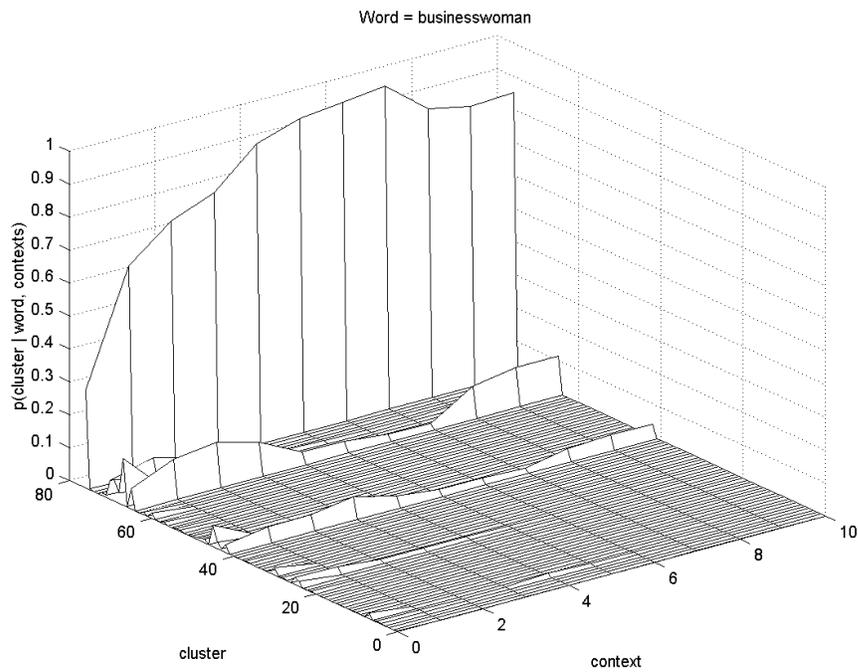

Figure 5.5: Another rare word, showing how the cluster membership probabilities change after each context

| Test set | 1 | 2 | 3 | 4 | Mean |
|---|---|---|---|---|---|
| CLAWS | 411 | 301 | 478 | 413 | 395 |
| Brown et al. | 380 | 252 | 444 | 369 | 354 |
| CDC | 372 | 255 | 427 | 354 | 346 |

Table 5.7: Perplexities of class tri-gram models on 4 test sets of 100,000 words, together with geometric mean.

I tagged 12 million words of BNC text with the 77 tags, assigning each word to the cluster with the highest *a posteriori* probability given its prior cluster distribution and its context. I then trained 2nd-order Markov models (equivalently class trigram models) on the original BNC tags, on the outputs from my algorithm (CDC), and for comparison on the output from the Brown algorithm. These models have the form of a Hidden Markov Model (HMM), where each state corresponds to a pair of classes. We decompose the probability

$$P(w_n|c_{n-1}, c_{n-2}) = P(c_n|c_{n-1}, c_{n-2})P(w_n|c_n) \tag{5.12}$$

The perplexities on held-out data are shown in Table 5.7. As can be seen, the perplexity is lower with the model trained on data tagged with the new algorithm. This does not imply that the new tagset is better; it merely shows that it is capturing statistically important generalisations. [1] In absolute terms the perplexities are rather high; I deliberately chose a rather crude model without backing off and only the minimum amount of smoothing, which I felt might sharpen the contrast. The Brown algorithm is a maximum likelihood model, so its performance compares well.

## 5.8 Discussion

I have presented an algorithm that when given a text in English, can identify some classes that correspond well to different parts of speech, and can also identify when ambiguous words fall into more than one class. In addition this algorithm can correctly classify very rare words with quite high accuracy. I will now discuss some of the problems with this approach, and how I hope to address them in future work.

### 5.8.1 Limitations

It is worth considering the limitations of this approach. Given arbitrarily large amounts of data, what is the degree of precision we could expect, and how fine-grained could the divisions be?

We can see some of the limitations already: the algorithm produces a class that consists of nouns and verbs. [2] Though this does not correspond exactly to a distinction that is made in traditional linguistics, this is not a problem that need concern us. At the very worst it will multiply the number of phrase structure rules required – in addition to rules like $NP \rightarrow Det\,N$ we will also need a rule like $NP \rightarrow Det\,(N|V)$. This is quite representative of the errors that this approach will make. For example, if we look at the subclassifications it induces in verbs, we see that it does not identify individual subcategorisation frames, but rather separates them into sets of verbs that participate in the same diathetic alternations (Levin, 1993). This will obviously affect the way that the phrase structure rules are defined. So given a particular alternation in English, say the dative alternation,

> John gave the cake to Mary

and

> John gave Mary the cake

rather than saying "gave" is in some sense ambiguous between these two, and expressing this using some sort of lexical redundancy rule or other mechanism in the lexicon, we would have a single entry in the lexicon, expressing the fact that 'gave' is a member of a particular class, and two phrase-structure rules that introduce the two alternations. In this case this seems not to be a serious problem; with the Noun-Verb alternation something more serious is happening. But there is a simple technique that could remove this if desired. If we look at the distribution of the Noun-Verb class, it will be a combination of the distributions of the noun class and the verb class. We can thus use exactly the same technique for ambiguous words on the ambiguous classes.

---

[1] I have not performed an analysis of the statistical significance of the results in Table 5.7.

[2] The class whose most frequent members are USE, HELP, FORM, CHANGE, and SUPPORT.

### 5.8.2 Independence assumptions

The work of Chater and Finch can be seen as similar to the work presented here given an independence assumption. We can model the context distribution as being the product of independent distributions for each relative position; in this case the KL divergence is the sum of the divergences for each independent distribution.

If $p_0(u, v) = a_0(u)b_0(v)$ and similarly for $p_1(u, v) = a_1(u)b_1(v)$ then

$$
\begin{aligned}
D(p_0||p_1) &= \sum_{u,v} p_0 \log \frac{p_0}{p_1} \\
&= \sum_{u,v} a_0(u)b_0(v)\left(\log a_0(u)\log b_0(v) - \log a_1(u) - \log b_1(v)\right) \\
&= \sum_{u} a_0(u)(\log a_0(u) - \log a_1(u))\sum_{v} b_0(v) \\
&\quad + \sum_{v} b_0(v)(\log b_0(v) - \log b_1(v))\sum_{u} a_0(u) \\
&= D(a_0||a_1) + D(b_0||b_1)
\end{aligned}
\tag{5.13}
$$

This independence assumption is most clearly false when the word is ambiguous; this perhaps explains the poor performance of these algorithms with ambiguous words. I will discuss this in depth in the chapter dealing with syntactic induction.

### 5.8.3   Similarity with Hidden Markov Models

A problem with the current approach is that I assume the words are unambiguous during the early part of the algorithm, and then at the end calculate the ambiguity. A more principled way of dealing with this would be to use the EM algorithm, treating the ambiguity as a hidden variable. This would mean effectively treating the model as a Hidden Markov Model (Murakami, Yamatomo, & Sagayama, 1993). This raises the possibility of identifying the set of syntactic categories by simply training a HMM on the data, and identifying the hidden states with the lexical categories. The problem with this is that there are two weaknesses in the training algorithm for HMMs: first, as has been widely noted, the EM algorithm used to train HMMs converges to a local not necessarily to a global optimum, and secondly, as previously noted, the EM algorithm has a tendency to favour solutions that are combinations of many different elements This was confirmed by some informal experiments using this approach. However there are possibilities of slightly more complex models that could alleviate this problem.

### 5.8.4   Hierarchical clusters

It is clearly inadequate to fix a number of classes in advance on an *a priori* basis. As mentioned previously several researchers (Grünwald, 1996; Li & Abe, 1998) have used MDL approaches to determine the "right" number of clusters, by trading off the size of the model against its predictive power. Others produce hierarchical clusterings or dendrograms, which together with a measure of the cohesion of the various clusters might provide a way of determining the "correct" number of categories in each language.

### 5.8.5   Use of orthography and morphology

The new algorithm currently does not use information about the orthography of the word, an important source of information. In English, with its trivial morphology, this is not a problem, but the situation would be very different with a language like Hungarian or Finish.

As Bertram et al. (2000, p. 5) point out:

> Of the 1,022,944 distinct noun types on our database [of Finnish text] ... only 2.6% is accounted for by monomorphemic nominative singulars. More than 95% of the morphologically complex nouns appear with no more than 1 occurrence per million. Thus the bulk of Finnish words in running text is polymorphemic (fairly often even tri- or quadro-morphemic) and of very low surface frequency.

Unfortunately I do not have access to a database of Finnish at the moment, but Figure 5.6 shows some graphs for various languages taken from the European Corpus Initiative (ECI) CD-rom. The graphs show Zipf-ian graphs of the rank of words against frequency, for 100,000 token samples taken from a variety of languages. As can be seen, morphologically complex languages such as Czech have a much lower frequency across a wide range of words. These graphs were prepared with minimal pre-processing so they are very sensitive to the particularities of the different corpora, so the details should not be examined very closely, but it is clear that English is in some sense very 'easy'.

Clearly, a learning algorithm would have to learn morphology at the same time as, or prior to, learning the set of syntactic categories. It is this subject that the next chapter will address.

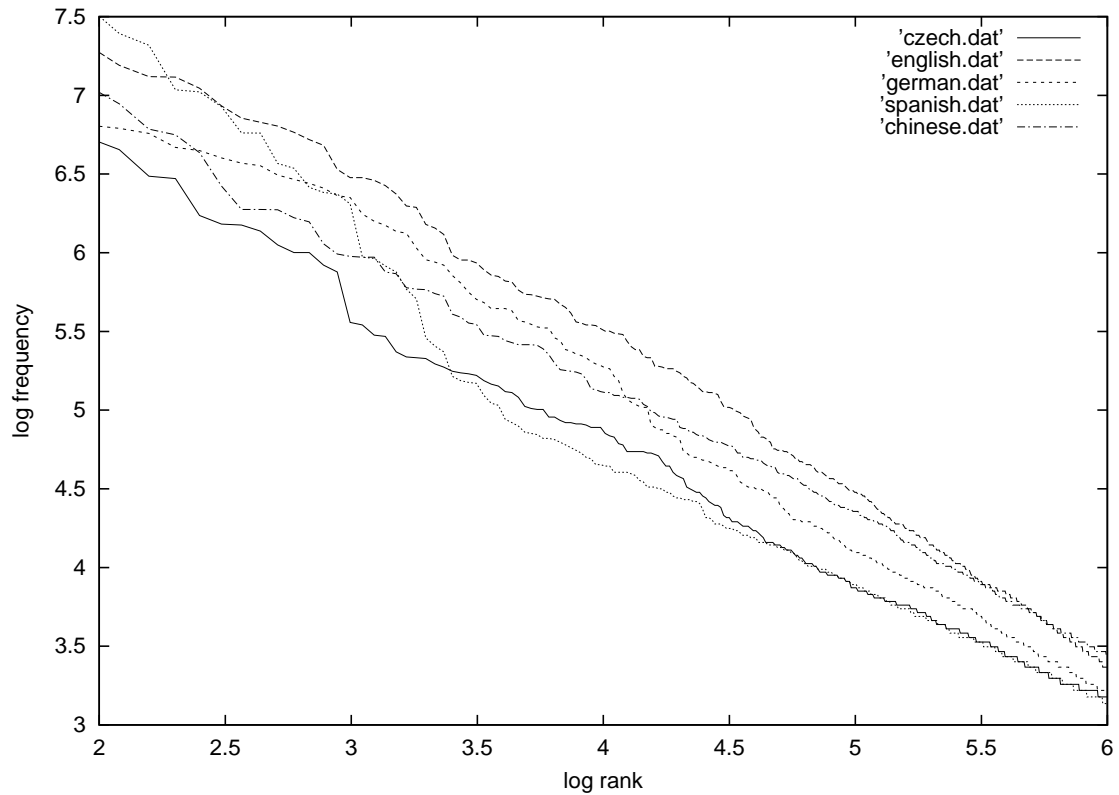

Figure 5.6: Log-log plots of word frequencies for various languages.

Languages with comparatively rich morphology tend to have rather free word order, which might cause problems with distributional induction techniques. However these languages tend to signal the part of speech in the surface form of words, so it would be possible to use that information to learn. What would create serious problems is a very free word order language with very limited morphology: fortunately such languages seem not to exist.

### 5.8.6 Multi-dimensional classes

The assumption I have been making is that we can partition the set of words into a single set of classes. While more or less adequate for English, in other languages this is clearly not suitable: many languages also divide words into subclasses according to other grammatical features such as gender or number or case. There are two ways to handle this. Either we could just divide the set of words into very fine-grained classes, (NOUN-SING-NEUT) or we could split them into various cross-cutting partitions, one of which might be according to the category, another according to gender, and so on. If a word is a noun then that will have a definite effect on its local context; if it is plural, then in many languages this will have a slightly less local effect on the agreement features on, or choice of, nearby adjectives, verbs or determiners depending on the agreement rules. This could provide the basis for this approach.

# Chapter 6

# Morphology Acquisition

## 6.1 Introduction

In this chapter I shall discuss some algorithms for learning morphology. The main innovation here is the use of an algorithm for learning stochastic finite-state transducers based on a fairly straightforward application of the Expectation-Maximisation algorithm. This is the first time this technique has been used to learn morphology. The algorithm works well in supervised learning settings, and because of its principled probabilistic interpretation it can be adapted to learning in more difficult, less supervised frameworks, that are in my opinion more plausible as models of how children learn morphology. I shall also discuss their application to learning morphological relations between some of the part of speech clusters derived in Chapter 5.

The structure of this chapter is as follows. I first discuss very briefly the current state of computational morphology in Section 6.2. I then discuss, in Section 6.3, various frameworks for learning morphology, namely supervised, partially supervised and unsupervised, and previous work in each of those frameworks, together with the particular requirements that I have for a morphology learning algorithm. I then present, in Section 6.4, the technique I will use to do this, based on the use of Pair Hidden Markov Models (PHMMs hereafter), a model that was introduced in bioinformatics, together with associated algorithms for training them and so on. Section 6.5 briefly discusses previous work relating to PHMMS as well as other application of PHMMs since they are a general algorithm for learning string transductions. Section 6.6 introduces the use of mixtures of these models to model morphological processes. I then present various experimental results on the supervised learning of morphological processes in three languages, English, German and Arabic, in Section 6.7. Section 6.8 deals with how PHMMs can be used to learn in partially supervised situations. I present a rather artificial situation which has a very clean mathematical solution, and then show how it can be modified to deal with a more realistic situation, and I then give in Section 6.9 some experimental results, in English and Arabic, together with some preliminary results based on the syntactic categories induced in Chapter 5. I then discuss some possibilities for future work in Section 6.10 and conclude with some general discussion in Section 6.11 about the place of this work in the overall argument of the thesis.

## 6.2 Computational Morphology

The progress of morphology in computational linguistics has in the past been influenced heavily by the dominance of English. The fact that English has a very simple morphological system has had two consequences: first, many researchers have been able to ignore morphology completely, and work with a fully expanded lexicon. Secondly, when researchers have worked with English, they have been able to use very trivial techniques that are not sufficiently general to work with a wide range of languages.

The situation has changed greatly and there is now concern with working with systems of morphological analysis that can work with a wider range of languages, though there are still some languages whose morphology is poorly understood. Languages have traditionally been grouped into four classes according to their morphological systems (Spencer, 1991, p.38): *isolating*, *agglutinating*, *inflectional* and *polysynthetic*. Isolating languages are those such as Vietnamese and Chinese which effectively have no morphology at all. Agglutinating languages like Hungarian have words which are polymorphic, a sequence of more or less unchanging morphemes in sequence. Inflectional languages, of which Latin and Russian are good examples, have morphemes which combine several different functions such as number, person and tense. Polysynthetic languages like many North American languages, can combine many different words together so that the words correspond almost to entire sentences.

If we consider the range of morphological phenomena that occur in these various languages we can again provide various classifications. Most trivially, some processes involve adding prefixes or suffixes. Thus we have a simple concatenation, which may be combined with various slight changes to the stem, such as voicing or de-voicing, lengthening or shortening of vowels, and so on. A slightly more complicated situation is where an infix is inserted within the stem, or a circumfix, perhaps best thought of as a combination of a prefix and a suffix, placed around the stem. Next we have various changes that might be made, rather than material that is added. So for example, we have the way that some forms of the plural in German are formed by changing a vowel in the stem, the umlaut process. More radically we can have a complete change of the stem, or suppletion, as with the English go/went, though this is comparatively rare. Combinations of any of these are quite common. A much more complex process is that of reduplication (Raimy, 2000) where some phonologically specified material is duplicated. The most extreme case of this is unbounded full-stem reduplication which occurs in Malay and Indonesian where the plural, for instance, is formed by complete reduplication of the singular, thus *orang* (man), *orangorang* (men). A further class of processes, that I shall discuss more later, occur in the Semitic languages, such as Arabic and Hebrew, where morphological processes operate on a tri-consonantal root, changing the vowels in quite a radical way.

Theories of morphology can themselves be grouped into various types (Hockett, 1958). I shall provide a very brief, and unavoidably rather inaccurate, description of each of these so I can place the techniques presented here into some sort of context. *Item and Arrangement* theories model morphology as essentially a process of joining together words out of their component parts. *Item and Process* theories, on the other hand, consider the process of word formation to start from a root, that is then changed by various processes, that add elements or change the stem in some sense. *Word and Paradigm* models focus on the contrast between different forms of the same word, and do not try to analyse the word into constituent parts. This relates to the contrast between morphemes as things and morphemes as functions. As Spencer (1991, p.12) says:

> There are two persistent metaphors which are used by linguists to conceptualise this mapping. One is to regard morphemes as things which combine with each other to produce words. In this metaphor, a morpheme is a bit like a word, only smaller, and the morphology component of a grammar is a bit like syntax in that its primary function is to stick the morphemes together. The other metaphor regards morphemes as the end product of a process or rule or operation. Here it is not the existence of the morphemes that counts but rather the system of relations or contrasts that morphemes create.

In line with general principles of parsimony, I shall not use the notion of morpheme at all. This is not devoid of empirical content. It amounts to assuming that morphemes must be intimately tied to a particular process. If there was a language, and there may well be, such that a particular morpheme, with a stable morphosyntactic function, could be used either as a prefix, or a suffix, then this would be evidence for the morpheme having an existence independent of the operation of prefixation or suffixation.

I shall now discuss Computational Morphology. Morphology has not been a central concern of Computational Linguistics. Partly this is to do with the excessive emphasis on English, as mentioned before, so that for many applications the morphology can simply be ignored. Space does not permit a survey of all of computational morphology – I shall merely mention a few of the key concepts and techniques.

The first thing to observe is that in general finite-state techniques are adequate to model morphological processes in a very wide range of languages. Kaplan and Kay (1994) [1] discuss this in some detail. Notwithstanding certain theoretical debates about the adequacy of finite-state models, (Carden, 1983; Langendoen, 1981), it appears that with the exception of unbounded reduplication in some languages (Raimy, 2000), of which Malay/Indonesian is perhaps the most widely-spoken example, regular transductions are adequate for this task. However, *partial* reduplication, while not formally requiring non-finite-state models, may nonetheless be more naturally handled in a slightly more powerful framework. By partial reduplication, I mean everything from gemination of consonants and lengthening of vowels to the reduplication of syllables or segmental sequences such as, to use an example from an Indo-European language, the formation of the perfect in classical Greek (grapho, gegrapha).

Two-level morphology as developed by (Koskenniemi, 1983b, 1983a, 1984), uses as the name suggests, a lexical level and a surface level, without any intermediate layers. Phonological rules are expressed as constraints that operate in parallel, and can refer to both the lexical and surface level. Though this is very widely used, I have little to say about it, largely because though powerful it is difficult to see how it might be learnable, since it is necessary to learn not just the transductions from lexical to surface forms, but also the lexical forms themselves.

## 6.3 Computational Models of Learning Morphology

There has been quite a lot of interest in learning morphology, perhaps because unlike many natural language tasks, it is possible to get reasonably good results with quite simple techniques. For learning of morphology we can distinguish three learning frameworks, according to the amount of supervision used, namely *supervised*, *unsupervised* and *partially supervised* learning.

### 6.3.1 Supervised

The most common form of learning, and the easiest, is supervised learning where the learning algorithm is presented with pairs of strings of symbols. Most of the time this has involved surface to surface transductions, i.e. between inflected and uninflected forms.

The earliest work I have come across of this type is Golding and Thompson (1985), who present an algorithm LEXICRUNCH, that takes pairs of inflected and uninflected words for particular morphological processes in English, Finnish and French, and learns a set of simple re-writing rules.

Much of the other work focusses on the formation of the English past tense starting with the seminal paper of Rumelhart and McClelland (1986a), who present a neural network that can learn from suitably encoded representations of the base and past form. This gave rise to a very active debate about the relative merits of symbolic and connectionist models for language processing (Pinker & Prince, 1988). Various further models have been proposed that improve on the original connectionist model, (Plunkett & Marchman, 1990; MacWhinney & Leinbach, 1991) and various models have been proposed that learn in a symbolic way using decision trees (Ling, 1994) or variants of Inductive Logic Programming (Mooney & Califf, 1995). However the triviality of the English past tense make it an unsuitable test bed for the comparison of different models. There have been a few models that learn other languages, notably German (Westermann & Goebel, 1995) and Arabic (Plunkett & Nakisa, 1997) and Manning (1998) presents a symbolic learning model that he tests on Anmajere, an Australian language, rightly criticizing the reliance on English of previous work. He also notes the problem that Semitic languages pose for these approaches. van den Bosch and Daelemans (1999) present a slightly unusual morphological segmentation algorithm, using Memory-Based Learning (MBL) to segment words into morphemes. Theron and Cloete (1997) present an algorithm to learn sets of two-level rules in a variety of languages, English, Xhosa and Afrikaans. They provide an algorithm for segmenting surface forms into morphemes, and aligning them with lexical forms using Levenshtein edit distance, which has some relation to the work presented here. Daelemans, Berck, and Gillis (1997) present an algorithm for learning classifications of Dutch diminutives, and there have also been various other techniques for applying other non-parametric techniques such as analogical reasoning (AML) (Derwing & Skousen, 1994).

---

[1]The date on this citation is misleading as versions of this manuscript had circulated widely beforehand.

### 6.3.2   Unsupervised Learning

Unsupervised learning is when the algorithm is presented merely with a single set of words, and must work out what the morphological relationships are. There have been a number of different approaches to this task, many of them sharing the same limitation, namely an assumption that the morphological processes are exclusively suffixational, and limited also to words with regular inflections. As Goldsmith (2001) says:

> It is not difficult to construct algorithms to produce an initial morphological analysis of a corpus of languages whose morphology is as simple as those of Indo-European languages.

Déjean (1998) is a good example of this work: he essentially generates various stemming rules in a variety of languages. Gaussier (1999) presents an algorithm that learns derivational morphology when provided with an inflectional lexicon, assuming that suffixation is the relevant process. A less arbitrary solution is presented by Goldsmith (2000, 2001) who uses an MDL framework to induce a set of endings in various languages. Schone and Jurafsky (2000) present an interesting innovation: rather than looking solely at the forms of the words, they also filter the results based on an analysis of the semantics of the words, using a SVD in the form of latent semantic analysis to limit the results. Snover and Brent (2001) is another attempt in the MDL framework, with results in English and French. All of these techniques assume that the problem is to segment words into sequences of morphemes. This is however not a cross-linguistically valid assumption, therefore I decided not to pursue this line of reasoning.

### 6.3.3    Partially supervised

In addition to these two classes, which are generally recognised, I will add a third one, of interme-
diate difficulty: that of *partially supervised* algorithms. Here the algorithm is presented with a pair
of sets of words, and must find a partial matching between the two sets, as well as a transduction
that models that alignment.  My reasons for introducing this class are several.  First, I think the
task of learning in a completely unsupervised setting is not *necessary* for a complete unsupervised
language learning algorithm.  As we saw in Chapter 5, it is possible to induce a set of syntactic
classes from a text without using information about morphology.  Thus, we can always induce a
set of syntactic classes, and then use a partially supervised algorithm to learn the morphological
relationships between them, as I shall demonstrate below in Section 6.9.  Secondly, I think com-
pletely unsupervised learning is perhaps too difficult to do in general, i.e.  with languages that
have a much more complex morphological system than English. Partial supervision seems to give
enough information for the algorithm to learn, while still being cognitively plausible in a way
that completely supervised learning is not.  The justification for using supervised learning is that
the infant child will already have worked out the alignment between the inflected and uninflected
forms using semantic information.  Though semantic information is a useful aid, as Schone and
Jurafsky (2000) demonstrate, it is not, I think sufficient to generate an alignment by itself, even
in English. In more complex languages, the token frequencies of the individual words may be so
low, that semantic information cannot be extracted until after a certain amount of morphological
analysis has taken place.

Minimally supervised learning, (Yarowsky & Wicentowski, 2000), which is quite similar, uses
a variety of different sources of information to learn both regular and irregular inflections.  As they
say:

> But for many languages, and to a quite practical degree, inflectional morphologi-
> cal analysis and generation can be viewed primarily as an alignment task on a broad
> coverage word list.

In addition they use a weighted edit distance to help in the alignment. This is a topic I will return
to at greater length later on.

### 6.3.4    Desiderata

Given the overall aim of this thesis, we can sketch out the desirable properties of our morphology
learning algorithm.

- It must work at least in partially supervised settings.

- It must be robust in the presence of noise.

- It must have a principled probabilistic interpretation, so we can combine it with other parts
  of a system.

- It must work on a wide range of languages, and must work without making assumptions
  about the form of the morphology.

My approach is thus to start with an algorithm for learning stochastic finite state transducers
in a supervised setting, and then show how it can be extended to work in a partially supervised setting. An important point to note at the outset is that in many languages, even in English, the morphological processes that a word undergoes are not specified purely phonologically, but in general are specified lexically. Thus in the formation of the English past tense, just because we know that the phonology of the stem of a verb is, for example, *ring*, or more strictly sounds like *ring*, this does not tell us what the past tense is. It could be *rang*, *wrung*, or *ringed* depending on which particular word it is, and there are numerous other examples in many other languages that confirm this point. Thus modelling the morphological processes simply as a single surface string to surface string transduction is inadequate. This was one of the criticisms levelled against the Rumelhart and McClelland (1986a) model by Pinker and Prince (1988).

## 6.4   Pair Hidden Markov Models

The formal device that I shall use to learn string transductions is the Pair Hidden Markov Model. As its name suggests it is a variant or extension of the Hidden Markov Model (HMM). HMMs have been used widely in the speech recognition community for many years (Rabiner, 1989), and as a result both their formal properties and their behaviour in practical applications are well understood.

Pair Hidden Markov Models were first introduced by Durbin, Eddy, Krogh, and Mitchison (1998) in the field of bioinformatics. They present them as a generalisation of an edit distance. With the growth of biotechnology and the ability to sequence DNA, which culminated in the completion of the Human Genome Project (IHGMC, 2001), there has been a vast amount of data, primarily in the form of DNA sequences, to be analysed. This has driven the very rapid development of the discipline called bioinformatics or computational biology, which has adopted wholesale many of the techniques of statistical modelling originally used for speech recognition and computational linguistics. Some of these techniques seem to be better suited to their new role than they were to the task they were originally designed for (Gusfield, 1997). This work perhaps returns the favour – stealing some ideas back from bioinformatics to apply to computational linguistics.

### 6.4.1   Definition

I shall start by defining a normal HMM, albeit not fully general, and then make the changes needed to convert it to a PHMM.

Let us consider a stochastic process, i.e. a sequence of random variables $X_t$, that at any given moment is in one of a finite number of $k$ states, $s_0, \ldots, s_{k-1}$. At each time step, it can change state according to a transition function that depends only on the state. The process has a defined start state, $s_0$ and a defined end state $s_1$. At time step 0, the process is in state $s_0$ ($p(X_0 = s_0) = 1$), and when it moves to state $s_1$ the process terminates. This is a point of difference with the traditional definition of HMMs, which tend to be used without a terminating state to model indefinitely long sequences of symbols – when they have a defined end state, they are sometimes called anchored HMMs. This process satisfies a Markov independence property, namely that the probability that the process changes from state $i$ to state $j$ depends only on the state $i$ and not on any previous states. So we can write $p(X_{t+1} = s_j | X_t = s_i)$. At each state the process outputs a symbol from a finite alphabet $A$. Again we assume that the symbol depends only on the current state. We can denote the symbol output at time $t$ by another random variable $O_t$, and we can write $p(O_t = a | X_t = s_i)$.

We can consider these two probabilities, the transition probabilities, and the output probabilities as parameters of the model which we can write as $p(s_j|s_i)$ the probability of going from state $i$ to state $j$, and $q(a|s_i)$ the probability of outputting $a$ from state $i$. This will define a probability distribution over $A^*$, the set of strings from $A$. [2]

With PHMMs, the situation is exactly the same except that instead of outputting a single stream of symbols, it outputs two streams of symbols. At each state it can emit symbols on one or both of two streams, or *tapes* in the Turing machine idiom. The more general way to formalise this is to suppose we have two alphabets, $A, B$ one associated with each tape. Then each state can output either a symbol from $A$ on the left tape only, or a symbol from $B$ on the right tape only, or simultaneously output a symbol from $A$ on the left tape, and from $B$ on the right tape. These three sorts of output I call $q_{10}$, $q_{01}$ and $q_{11}$ outputs respectively, and we denote the probability of each output by $q_{10}(a|s)$ and so on. We could also allow null outputs, where nothing is emitted on either tape, a $q_{00}$ output, but this causes mathematical complications, and I shall ignore this possibility. Clearly for each state $s_i$ these output probabilities must sum to one, so we require

$$\sum_{a \in A} q_{10}(a|s_i) + \sum_{b \in B} q_{01}(b|s_i) + \sum_{a \in A} \sum_{b \in B} q_{11}((a,b)|s_i) = 1 \tag{6.1}$$

I will assume that we have a distinguished start and end symbol, and that the initial and final states, output this symbol with probability 1.

For the applications presented in this chapter, I will assume that we have only one alphabet, that is $A = B$, which I shall denote by $A$. In addition I will limit the $q_{11}$ outputs so that they are zero unless the two symbols are the same; that is to say the $q_{11}$ always outputs the same symbol on the left and right streams. This reduces the number of parameters of the model, and gives the algorithm a bias towards producing similar strings on the left and right. I will then write $q_{11}(a|s)$, for $q_{11}((a,a)|s)$. This is probably not necessary for the supervised algorithm, but becomes necessary for the unsupervised algorithms.

As van Noord and Gerdemann (2001) say:

> Expressing identity between input and output is crucial: This notion of identity can be seen as a consequence of the linguistic principle of *Faithfulness*: corresponding input and output segments tend to be identical.

This model will then define a probability distribution over pairs of strings. Where $u, v \in A^*$, we will write $p_m((u,v))$ to denote the probability of the pair of strings being produce by a model $m$. This probability will be the sum of the probabilities of all transition sequences that produce that pair of strings. Since we are interested in stochastic string transductions, we are in the end interested in the conditional probability, not the joint probability. I will write $p_m^L(u)$ for the probability that the model produces the string $u$ on the left tape, and $p_m^R(v)$ for the probability it produces $v$ on the right tape. Clearly,

$$p_m^L(u) = \sum_{v \in A^*} p_m((u,v)) \tag{6.2}$$

The conditional probability $p(v|u)$ is thus

$$p_m(v|u) = \frac{p_m((u,v))}{p_m^L(u)} \tag{6.3}$$

---

[2] As defined, for some values of the parameters, there may be a non-zero probability of the process never terminating. In practice, given the initialisation I use, and the training regime, this never happens.

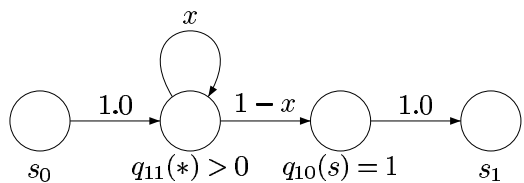

Figure 6.1: Example of a PHMM that adds an s to the end of any non-empty string.

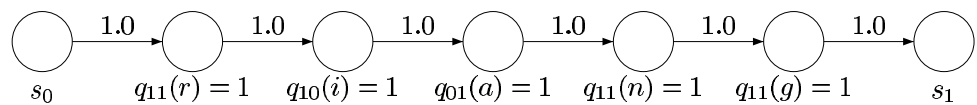

Figure 6.2: Example of a PHMM that changes *ring* to *rang*.

The decision to have a model for the joint rather than the conditional probability has a number of implications. Firstly, the joint distribution is clearly more informative: it is easy to recover the conditional probability from the joint but not vice versa. Secondly, it is easy to combine models with joint probability but very difficult, in a principled way, with conditional probabilities. On the other hand, we are ultimately interested in the conditional probability and this can cause some mild computational difficulties, as discussed further below.

### 6.4.2 Examples

I will now give some simple examples to illustrate the functioning and power of these models.

Figure 6.1 shows a small PHMM with four states that appends an *s* to any non-empty string. More formally we can say that the joint probability distribution is non-zero if and only if the right string is the the concatenation of the left string with *s*. In this case, the conditional probability will be 1.

Figure 6.2 is a much more specific model. It generates a particular pair of strings with probability 1, namely (*ring*,*rang*). This illustrates the use of the joint probability: we can model either very general transductions that take an arbitrary string as input, or alternatively a very specific transduction that takes only a single string, and obviously we can have transductions that select a phonologically specified set of strings as input. This will be useful to model morphology as we shall model a particular morphological process as a mixture of general transductions that model the regular words, and more specific transductions that model the irregular words.

Figure 6.3 illustrates a highly non-concatenative transduction, which occurs as one of the forms of the Arabic broken plural. This maps strings of the form *CaCC* where *C* stands for any consonant, to *CuCuuC* where the corresponding consonants are the same, thus *bank* will be mapped to *bunuuk*.

### 6.4.3 Algorithms

Just as with the normal HMM, PHMMs have efficient polynomial algorithms for the "three fundamental questions":

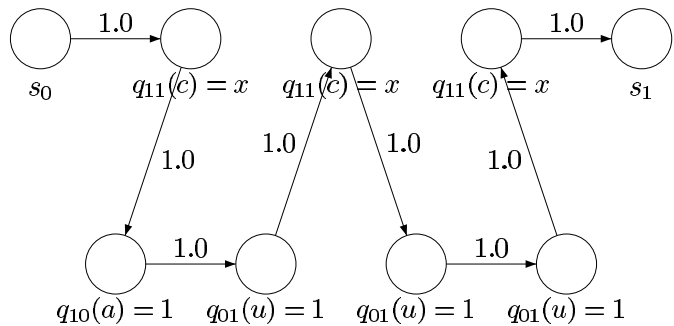

Figure 6.3: Example of a PHMM that changes a word with template *CaCC* to *CuCuuC*. The variable $c$ in the diagram ranges over the set of consonants $C$.

- Calculating the probability of an observation, i.e. a pair of strings

- Finding the most likely state sequence given an observation

- Choosing parameters that maximise the likelihood of some training data

In addition we will want to be able to

- Find the most likely output given an input.

- Find the probability of a string being produced on the left.

In general, an observation, $(u, v)$, could be generated by exponentially many state sequences. Thus to calculate the probability of that observation one needs to add up the probabilities of all of the sequences of transitions that might have produced it. Just as with simple HMMs, we can perform this computation in polynomial time by using dynamic programming techniques to avoid repetitive calculation. With simple HMMs, the dynamic programming data structure is called a *trellis*, which stores certain probabilities that are called the forward and backward probabilities. I refer the reader to the discussion in Manning and Schütze (1999, pp. 325–339).

With PHMMs the process is more complicated. We define the *forward* and *backward* probabilities as follows. Given two strings $u_1, \ldots u_l$ and $v_1, \ldots v_m$ we define the forward probabilities $\alpha_s(i, j)$ as the probability that it will start from $s_0$ and output $u_1, \ldots, u_i$ on the left stream, and $v_1, \ldots, v_j$ on the right stream and be in state $s$, and the backward probabilities $\beta_s(i, j)$ as the probability that starting from state $s$ it will output $u_{i+1}, \ldots, u_l$, on the right and $v_{j+1}, \ldots, v_m$ on the left and then terminate, i.e. end in state $s_1$.

We can calculate these using the following recurrence relations:

$$
\begin{aligned}
\alpha_s(i, j) = &\sum_{s'} \alpha_{s'}(i, j-1) p(s|s') q_{01}(v_j|s) \\
&+ \sum_{s'} \alpha_{s'}(i-1, j) p(s|s') q_{10}(u_i|s) \\
&+ \sum_{s'} \alpha_{s'}(i-1, j-1) p(s|s') q_{11}(u_i, v_j|s)
\end{aligned}
\tag{6.4}
$$

We start the recursion at $(0, 0)$ by noting that $\alpha_{s_0}(0, 0) = 1$ and zero otherwise.

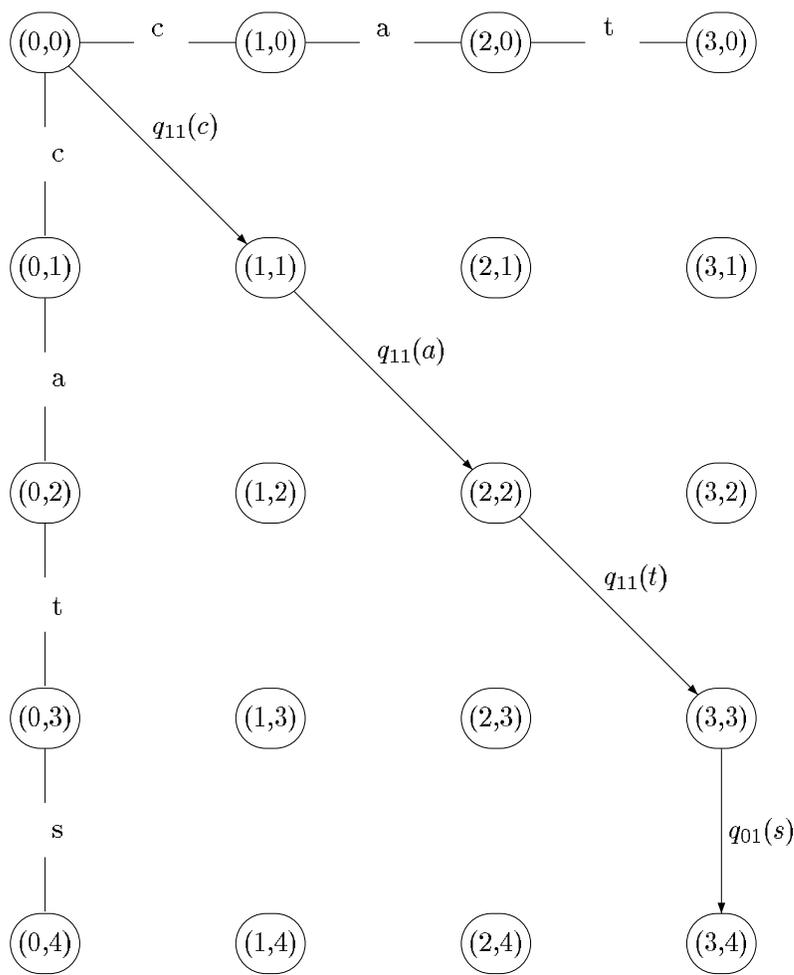

Figure 6.4: Dynamic programming trellis showing how the pair *cat,cats* is generated. Each node represents a vector of states. $q_{10}$ transitions are horizontal, $q_{01}$ transitions are vertical, and $q_{11}$ transitions are diagonal. Note the resemblance to the methods of calculating the Levenshtein edit distance.

$$\beta_s(i,j) = \sum_{s'} \beta_{s'}(i,j+1)p(s'|s)q_{01}(v_{j+1}|s')$$
$$+ \sum_{s'} \beta_{s'}(i+1,j)p(s'|s)q_{10}(u_{i+1}|s')$$
$$+ \sum_{s'} \beta_{s'}(i+1,j+1)p(s'|s)q_{11}(u_{i+1},v_{j+1}|s') \qquad (6.5)$$

Here we start the recursion at the end, $(l,m)$ with $\beta_{s_1}(l,m) = 1$ and zero otherwise.

Figure 6.4 outlines the dynamic programming trellis for the pair of words *cat* and *cats*. The trellis has one more dimension than the traditional HMM trellis. The most likely path, which gives the alignment between the two strings is also shown.

To train the model, we use the EM algorithm. We therefore want to calculate the expected number of times each transition or output function is taken. This is just a sum over the expected number of times each transition is taken between each link in the trellis, with respect to the probability of the hidden variables given the observations. I will sketch the derivation from the EM theorem.

It may be recalled that if we maximise this quantity, we will improve the log likelihood of our model.

$$\sum_z P^{old}(z|o) \log P^{new}(z,o) \tag{6.6}$$

Here $z$ is the hidden variable and $o$ is the observation, and $P^{old}$ and $P^{new}$ are the probabilities with respect to the old and new parameter values respectively: i.e. we want to maximise the *new* parameters while the old parameters are constant.

In this case $z$ will be the sequence of transitions and outputs through the PHMM, and $o$ will be the pair of strings $(u,v)$. Let $z_{11}(s,s',a)$ denote the event of going from state $s$ to state $s'$ and emitting an $a$ on both channels, and similarly for $z_{10}(s,s',a)$ and $z_{01}(s,s',a)$, and use $p_{11}(s,s',a)$ etc. to denote the probabilities of each of these. If we use $c_{11}(s,s',a)$, as counting variables, to denote the number of times each of these is used in any given state sequence, then the new values of the probabilities will be

$$p'_{11}(s,s',a) = \frac{1}{H_s}\sum_Z P(z|o)c_{11}(s,s',a) = \frac{1}{H_s P(u,v)}\sum_Z P((u,v),Z)c_{11}(s,s',a) \tag{6.7}$$

and so on, where $H_s$ is a normalisation constant for the state $s$.

But clearly the sum over all state sequences $Z$ is exponential. How can we calculate the sum $\sum_Z P((u,v),Z)c_{11}(s,s',a)$ effectively? We consider the subset of these state transitions where it makes a transition from $s$ to $s'$ emitting $a$ on both sides for the $i$th element of $u$ and the $j$th element of $v$, i.e. between $(i-1,j-1)$ and $(i,j)$ in the trellis. The sum over this subset, which we can call $Z_{ij}$ is easy to calculate – note we have removed $c_{11}$:

$$\sum_{Z \in Z_{ij}} P((u,v),Z) = \alpha_s(i-1,j-1)p(s'|s)q_{11}(u_i,v_j|s')\beta_{s'}(i,j) \tag{6.8}$$

The sum of the probability of all these transition sequences is just the probability it will generate the first parts of the two strings, namely the forward probabilities, times the probability it will make the appropriate transition, times the probability it will generate the remaining bits of the sentence.

So if we sum over all $i$ and $j$, the number of times each sequence will turn up will be, obviously, equal to the number of these particular transitions it has in it, ie $c_{11}(s,s',a)$ so

$$\sum_Z P((u,v),Z)c_{11}(s,s',a) = \sum_{i=0}^{l-1}\sum_{j=0}^{m-1}\sum_{Z \in Z_{ij}} P((u,v),Z) \tag{6.9}$$

So using these equations we can compute the new values as follows. We define various accumulators corresponding to the expected number of times each transition is used: $C_{11}(s,s',a)$ is the expected number of times we will make a transition from $s$ to $s'$ and output a symbol $a$, and so on. Then we can calculate these as the sums:

$$C_{11}(s, s', a) = \sum_{u,v} \frac{1}{p(u,v)} \sum_{i=0}^{|u|-1} \sum_{j=0}^{|v|-1} \alpha_s(i,j) p(s'|s) q_{11}(u_{i+1}, v_{j+1}|s') \beta_{s'}(i+1, j+1) \delta(a, u_{i+1}) \quad (6.10)$$

$$C_{10}(s, s', u_{i+1}) = \sum_{u,v} \frac{1}{p(u,v)} \sum_{i=0}^{|u|-1} \sum_{j=0}^{|v|} \alpha_s(i,j) p(s'|s) q_{10}(u_{i+1}|s') \beta_{s'}(i+1, j) \delta(a, u_{i+1}) \quad (6.11)$$

$$C_{01}(s, s', v_{j+1}) = \sum_{u,v} \frac{1}{p(u,v)} \sum_{i=0}^{|u|} \sum_{j=0}^{|v|-1} \alpha_s(i,j) p(s'|s) q_{01}(v_{j+1}|s') \beta_{s'}(i, j+1) \delta(a, v_{j+1}) \quad (6.12)$$

$$(6.13)$$

Here the delta function $\delta(a,b)$ is 1 if $a = b$ and zero otherwise: this merely restricts the sums to those parts of the trellis which have the right outputs. The complexity of the formulae above hides their simplicity. For example, the first of the three is just a sum over $\alpha_s(i,j) p(s'|s) q_{11}(u_{i+1}, v_{j+1}|s') \beta_{s'}(i+1, j+1)$ where $a$ is output. This just represents the probability of taking the transition from stage $i, j$ to stage $i+1, j+1$, from state $s$ to state $s'$ and of course outputting $u_{i+1}$ on the left and $u_{j+1}$ on the right. Note the presence of the factor $1/p(u,v)$. This is because we are calculating the expectation with respect to the conditional probability. In traditional applications of HMMs where all of the data is in a single observation sequence, this is a constant and can be neglected. Here however, we are dealing with multiple observation sequences, one for each pair of words, and this varies and must be included.

The new parameters will then be the appropriately normalised sums.

$$\hat{p}(s'|s) = \frac{\sum_a C_{11}(s, s', a) + C_{10}(s, s', a) + C_{01}(s, s', a)}{\sum_{s'} \sum_a C_{11}(s, s', a) + C_{10}(s, s', a) + C_{01}(s, s', a)} \quad (6.14)$$

$$\hat{q_{11}}(a|s') = \frac{\sum_s C_{11}(s, s', a)}{\sum_{s'} \sum_a C_{11}(s, s', a) + C_{10}(s, s', a) + C_{01}(s, s', a)} \quad (6.15)$$

$$\hat{q_{10}}(a|s') = \frac{\sum_s C_{10}(s, s', a)}{\sum_{s'} \sum_a C_{11}(s, s', a) + C_{10}(s, s', a) + C_{01}(s, s', a)} \quad (6.16)$$

$$\hat{q_{01}}(a|s') = \frac{\sum_s C_{01}(s, s', a)}{\sum_{s'} \sum_a C_{11}(s, s', a) + C_{10}(s, s', a) + C_{01}(s, s', a)} \quad (6.17)$$

### 6.4.4 Finding the most likely output

As mentioned previously, given this model of the joint distribution $p(u,v)$, for a particular input string $u$ we will want to find the most likely output, ie

$$\underset{v}{\operatorname{argmax}}\, p(u,v) \quad (6.18)$$

The obvious way to find this is to use a Viterbi-style algorithm to find the most likely state sequence given $u$, and then to select the most likely output given that state sequence (Ristad, 1997). Unfortunately this does not in general give the right answer, and in practice I have found that it does in certain circumstances give incorrect answers. We can see that this by considering a simple example shown in Figure 6.5.

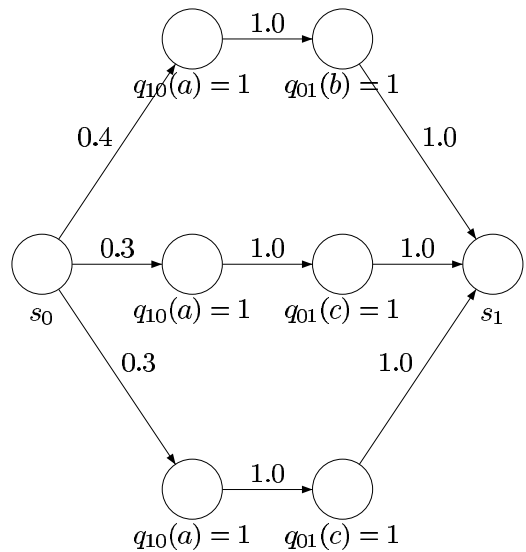

Figure 6.5: A simple PHMM where the most likely state sequence does not give the most likely output. The most likely output given the input of *a* is clearly *c* with probability 0.6, but the most likely state transition outputs *b* with probability 0.4.

In general the problem is that the most likely output can be the output of exponentially many state transitions. This is equivalent to finding the most likely output of a HMM which Goodman (1998) and Casacuberta and de la Higuera (2000) showed is NP-hard. There will not therefore be an efficient algorithm for performing the computation precisely. However, in this specific application, we are concerned with distributions where, ideally, there is a string $v$ where $p(u, v) = p^L(u)$ i.e. where $p(v|u) = 1$. So if we can sample from the distribution $p(v|u)$, we can then find such a $v$ rapidly.

More specifically, we sample from the conditional distribution, keeping track of two quantities: first the highest value of $p(v|u)$ we have so far found, say $p_{max}$ and of course the corresponding string $v_{max}$, and secondly the sum of $p(v|u)$ for all the distinct strings we have so far examined, $p_{sum}$. Since $p_{sum}$ will always be less than or equal to 1, when $1 - p_{sum} < p_{max}$ we will know that $v_{max}$ is the correct answer since any new string must have $p(v|u) \leq 1 - p_{sum} < p_{max}$. In particular, if we find a string that has $p(v|u) > 0.5$ we immediately know that it is the most likely output.

### 6.4.5 Sampling from the Conditional Distribution

For a given PHMM $m$, we can define two simple HMMs corresponding to the left and right outputs of the PHMM as follows. Considering the left outputs, we can define a HMM with exactly the same states and transition probabilities as the original PHMM, but with the following output probabilities

$$q(\varepsilon|s) = \sum_a q_{01}(a|s) \tag{6.19}$$

where $q(\varepsilon|s)$ is the probability of not outputting anything,

$$q(a|s) = q_{10}(a|s) + q_{11}(a|s) \tag{6.20}$$

This gives a HMM $h$, which will output exactly the same strings with the same probabilities as the original PHMM will on the left. That is

$$\forall u \in A^* \, p_m^L(u) = p_h(u) \tag{6.21}$$

This HMM will have various null transitions which will complicate the computation, but it is possible to remove these using a standard technique (Jelinek, 1997). We define a new HMM, that won't have any null transitions that will generate the same strings with the same probabilities. To do this we proceed as follows. Define $z^n(s,t)$ to be the probability that starting from state $s$ the HMM makes $n$ transitions ending in state $t$ and doesn't output anything. That is at each state, it doesn't produce anything. We can calculate this recursively as follows. Clearly when $n = 1$ this is just the probability it will go from $s$ to $t$ and output nothing at state $t$.

$$z^1(s,t) = p(t|s)q(\epsilon|t) \tag{6.22}$$

For generality we can also define

$$z^0(s,t) = \delta(s,t) \tag{6.23}$$

We can then define a recurrence relation. If

$$z^n(s,t) = \sum_r z^{n-1}(s,r)p(t|r)q(\epsilon|t) \tag{6.24}$$

If we define a matrix $Z$ whose elements are $z^1(s|t)$, we can see that $z^n$ is just $Z^n$, the matrix $Z$ raised to the $n$th power. Define $y(s)$ to be the probability that at state $s$ it outputs something

$$y(s) = \sum_a q(a|s) \tag{6.25}$$

Then we can define the transition probabilities of the new model to be

$$p^{new}(t|s) = \sum_{n=1}^{\infty} z^n(t|s)y(t) \tag{6.26}$$

The interpretation of this is that we are collapsing all sequences of null transitions that end in a non-null transition into a single transition.

The new output probabilities are

$$q^{new}(a|s) = \frac{1}{y(s)}q(a|s) \tag{6.27}$$

It is clear that this model produces exactly the same strings. The problem is to compute the infinite sum in 6.26. This is possible by using the matrix representation $Z$

$$p^{new}(t|s) = \sum_{n=1}^{\infty} z^n(t|s)y(t) = \sum_n Z^n Y = (I - Z)^{-1} ZY \tag{6.28}$$

So at the cost of a matrix inversion we can calculate this. We then have the *left HMM* of the PHMM.

Given this HMM we can use this to sample from the conditional distribution of the PHMM. For a given $u$ of length $l$ we calculate the backward probabilities of this left HMM.

$$\beta_s^L(i) = P(\text{starting from } s \text{ generates } u_{i+1}, \dots, u_l) \qquad (6.29)$$

There are now two ways to explain the sampling algorithm. I will start with the more comprehensible but less precise approach, and then later sketch the more precise method. If we consider the set of all sequences of transitions and outputs, each of which has a probability, if we want to sample from the conditional distribution $p(v|u)$, we want to sample at random from the subset of sequences that generates $u$ on the left. The left backward probabilities tell you the proportion of sequences that are in that subset. As we generate we want to weight the selection process by that proportion to ensure that we remain in that subset.

So if we are in state $s$ and we have output the first $i$ characters of $u$, and by hypothesis $\beta_s^L(i) > 0$, we want to select from the various possibilities. We can take one of the three different transitions types.

If we make go to state $s'$, and do a $q_{11}$ output, we must then output $u_{i+1}$ on both channels. We accordingly weight this transition by $\beta_{s'}^L(i+1)$, since this represents the probability of outputting the rest of the string, starting from state $s'$.

If we make a $q_{01}$ transition to state $s'$, we must weight it by $\beta_{s'}^L(i)$, and if we output a $q_{10}$ transition we weight it by $\beta_{s'}^L(i+1)$. The sum of all these probabilities must be equal to $\beta_s^L(i)$.

$$\beta_s^L(i) = \sum_{s'} p(s'|s) q_{11}(u_{i+1}|s') \beta_{s'}^L(i+1) \qquad (6.30)$$

$$+ \sum_{s'} p(s'|s) q_{10}(u_{i+1}|s') \beta_{s'}^L(i+1) \qquad (6.31)$$

$$+ \sum_{s'} p(s'|s) \sum_a q_{01}(a|s') \beta_{s'}^L(i) \qquad (6.32)$$

This follows from the definition of the left backward probabilities.

Note that $\beta_s^L(i)$ appears also on the right hand side of this equation when $s' = s$. This is why we need to use the matrix inversion to find its value. So to sample from the conditional distribution, at state $s$, having output the first $i$ characters of $u$, we do one of the following three things.

With probability

$$\frac{p(s'|s) q_{11}(u_{i+1}|s') \beta_{s'}^L(i+1)}{\beta_s^L(i)} \qquad (6.33)$$

output $u_{i+1}$ on both streams and move to state $s'$.

With probability

$$\frac{p(s'|s) q_{10}(u_{i+1}|s') \beta_{s'}^L(i+1)}{\beta_s^L(i)} \qquad (6.34)$$

output $u_{i+1}$ on left stream only and move to state $s'$.

With probability

$$\frac{p(s'|s) q_{01}(a|s') \beta_{s'}^L(i)}{\beta_s^L(i)} \qquad (6.35)$$

output $a$ on the right stream only and move to state $s'$. It may not be clear that this actually samples correctly from the conditional distribution. Intuitively, we restrict the sampling algorithm to the paths that generate the correct string on the left. Note that a particular path will have a probability that is the product of a sequence of terms, each one has the form of one of the three possibilities above. The ratios that we adjust the normal probabilities by are of the form,

$$\frac{\beta_{s_{t_{k+1}}}(j_{t_{k+1}})}{\beta_{s_{t_k}}(j_{t_k})} \tag{6.36}$$

where we go from state $s_{t_k}$ to state $s_{t_{k+1}}$. Therefore for an entire path, we will multiply the joint probability of the path by a factor

$$\frac{\beta_{s_{t_1}}(j_{t_1})}{\beta_{s_0}(0)} \times \frac{\beta_{s_{t_2}}(j_{t_2})}{\beta_{s_{t_1}}(j_{t_1})} \times \ldots \frac{\beta_{s_1}(l)}{\beta_{s_{t_k}}(j_{t_k})} = \frac{1}{\beta_{s_0}(0)} = \frac{1}{p^L(u)} \tag{6.37}$$

So the probability of the path is the correct conditional probability. A more formal way of defining the same thing is to treat this as sampling from a more complex HMM. We can consider a HMM where each state is a pair $< s, i >$ where $s$ is a state of the PHMM, and $i$ is an index in the first string $u$. We can then define the HMM transitions exactly according to the equations above. This is closely related to a proposal by Eisner (2001) to compose a string and a transducer.

### 6.4.6    Complexity of the algorithms

If we have two strings $u$ and $v$ of length $|u|$ and $|v|$ respectively, $n$ states and an alphabet of size $|A|$, let us consider the requirements in space and time of these algorithms. Each model has $n(n+3|A|)$ parameters. The trellis will have $n(|u|+1)(|v|+1)$ nodes in it. The training process for each iteration requires a calculation for each transition in the trellis, and so the overall complexity of each iteration is $O(n^2|u||v|)$. To calculate the conditional probabilities we need a matrix inversion at a cost of $O(n^3)$ but we only need to do this once, so this is not a significant factor for these models.

### 6.4.7    Initial Experiments

We can use this training algorithm to learn transductions by taking randomly initialised models of a given size, and training them. The question here is how powerful is the training algorithm. It is a variant of the EM algorithm and thus suffers from the problem of local minima. Will this algorithm suffice, or does one need to use more sophisticated techniques, such as MAP-estimation, conditional maximum likelihood estimation or forms of stochastic optimisation?

I experimented with various artificial data sets with small alphabets, to determine how the algorithm performs in outline. The data set for the first experiment is shown in Table 6.1. The results, shown in Table 6.2, demonstrate that this simple suffixation could be rapidly learned.

Table 6.3 shows the data used for the second experiment. This transduction accepts strings that consist of a string of $a$s optionally preceded by a $b$. If the string is preceded by a $b$ then a $b$ is appended, otherwise an $a$ is appended.

Again on these trivial, clean examples the model learns it accurately and generalises correctly.

An interesting issue is whether this learning algorithm is sufficient to learn non-deterministic transductions, that is to say transductions that cannot be performed by a deterministic transducer.

| U | V |
|---|---|
| a | ab |
| aa | aab |
| aaa | aaab |
| aaaa | aaaab |
| aaaaa | aaaaab |
| aaaaaa | aaaaaab |
| aaaaaaa | aaaaaaab |

Table 6.1: Training data for Experiment 1.

| U | V | $p(u,v)$ | $p(u)$ | $p(v|u)$ |
|---|---|---|---|---|
| a | ab | 0.22 | 0.22 | 1 |
| aaa | aaab | 0.13 | 0.13 | 1 |
| aaaaa | aaaaab | 0.08 | 0.08 | 1 |
| aaaaaaaaa | aaaaaaaaab | 0.029 | 0.029 | 1 |

Table 6.2: Results for Experiment 1 with 4 state model.

| $U$ | $V$ |
|---|---|
| ba | bab |
| baa | baab |
| baaa | baaab |
| baaaa | baaaab |
| aa | aaa |
| aaa | aaaa |
| aaaa | aaaaa |
| aaaaa | aaaaaa |

Table 6.3: Training data for Experiment 2: non-locally testable data set.

| $u$ | 5 states | 10 states | 15 states |
|---|---|---|---|
| aa | 0.625 | 1.0 | 1.0 |
| ba | 1.0 | 1.0 | 1.0 |
| aaaaaaa | 0.625 | 1.0 | 1.0 |
| baaaaaa | 0.375 | 1.0 | 1.0 |

Table 6.4: $p(v|u)$ of correct answer for varying sizes of model, for the non-locally testable data set.

| U | V |
|---|---|
| a | c |
| aa | bb |
| aaa | ccc |
| aaaa | bbbb |
| aaaaa | ccccc |
| aaaaaa | bbbbbb |
| aaaaaaa | ccccccc |
| aaaaaaaa | bbbbbbbb |

Table 6.5: The data set for testing whether a non-deterministic transduction is learnable.

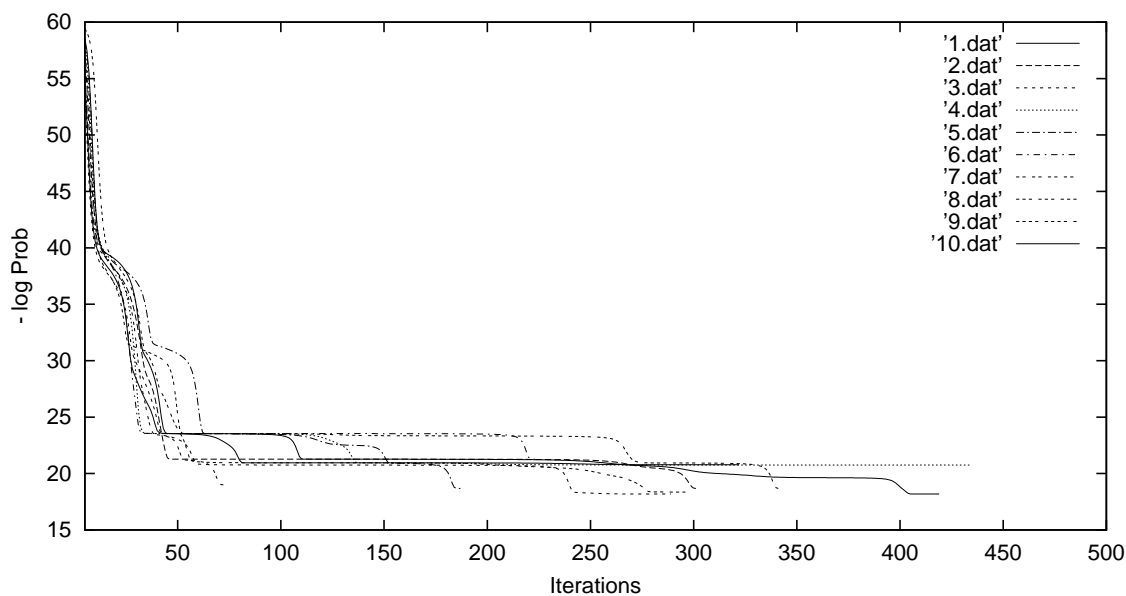

Figure 6.6: Log probability of 10 runs on the non-deterministic data set. Note that there are three distinct levels of log probability that the models make quite rapid transitions between.

A standard example (van Noord & Gerdemann, 2001) is a transduction that maps $(aa)^+$ to $(bb)^+$ and $a(aa)^*$ to $c(cc)^*$. This maps all even length strings to a sequence of $b$s and all odd length strings to a sequence of $c$s. This is non-deterministic since you cannot decide which rule you should use until you have got to the end of the input string, and you can't delay the output until then because that would require storing an unbounded number of symbols.

Table 6.5 shows the data set I used to test whether this was possible. A PHMM would require at least 12 states to model this correctly; I accordingly experimented with 20 state models. I trained 10 of these models: Figure 6.6 shows a graph of the log probability for these 10 models over the course of the training. This was rather unusual since instead of converging in comparatively few iterations, with this data the models often settled into long "plateaux" for a hundred iterations before suddenly dropping down to a better place in the parameter space. We can see three plateaux at levels of log probability of roughly -23, -20 and -18.

Of these 10 models, 8 dropped down to the bottom plateau which corresponds to the correct transduction. These ones modelled the transduction perfectly, performing it correctly on strings of arbitrary length with conditional probability 1. The remaining 2 which got stuck at the middle plateau performed less well but still produced the correct answers with a conditional probability of greater than a half, mapping for example $a^{10}$ to $b^{10}$ with conditional probability 1, but $a^{11}$ to $c^{11}$ with conditional probability 0.58. In fact these results got slightly *better* as the strings got longer. [3] It is nonetheless clear that this training algorithm can learn simple non-deterministic transductions, which I personally find rather surprising.

## 6.5   Discussion of Pair Hidden Markov Models

In this section I will discuss PHMMs in general and previous work relating to them. In particular I shall discuss various other statistical models that are related in some sense to PHMMs.

### 6.5.1   Previous Work

The literature on this topic is, quite frankly a mess. PHMMs clearly fall into the class of regular syntax-directed translation schemes (Aho & Ullman, 1969a, 1969b). As previously mentioned PHMMs were introduced in the context of bioinformatics; there they generally have a very few states with manually selected meanings; the earliest work here I have found has been by Allison, who uses them with 1,3 or 5 states (Allison, Wallace, & Yee, 1990, 1992; Allison, 1993; Allison & Wallace, 1994; Allison, Powell, & Dix, 1999). They are viewed as an extension of the Levenshtein edit distance (Gusfield, 1997); they present the training algorithm here for a single alphabet with slightly different boundary conditions. The relationship to the edit distance can be seen more clearly if we take the negative logarithm of all of the parameters; then the cost of an alignment is the sum of the various costs associated with the elementary operations – insertion ($q_{01}$), deletion ($q_{10}$) and copying ($q_{11}$). Indeed one of the best ways of understanding PHMMs is to think of them as a mixture of the edit distance and an HMM. They are also equivalent to a finite-state version of Ristad's memory-less stochastic transducers (Ristad & Yianilos, 1998), who also discuss a similar formalism in (Ristad, 1997), and present the EM algorithm for an identical model. Bengio and co-workers (Bengio & Frasconi, 1996; Bengio & Bengio, 1996; Bengio, 1999) present *Asynchronous Input/Output Hidden Markov Models* (IOHMM), which have many points in common. The principal difference is that they model the conditional probability rather than the joint probability.

Eisner (2001) has a much more general algorithm for doing EM over finite-state transducers that can handle the full power of the semi-ring calculus used by Mohri, Pereira, and Riley (2000). The complexity of the algorithm appears to be higher though. Casacuberta (1995, 1996), Pico and Casacuberta (2001) present a very similar algorithm, together with a variety of different estimation techniques, but without citing any other work. Their algorithm is slightly more general, and can handle states that output nothing on both streams – $q_{00}$ transitions in my terminology.

The PHMM is formally identical to certain parameterisations of weighted finite state transducers; the choice of this non standard terminology deserves explanation. Consider the normal Hidden Markov Model; these can be described alternatively as non-deterministic stochastic finite-state automata, or as stochastic regular grammars. The choice of terminology depends primarily on the techniques that are being used. The term 'HMM' tends to be used when one emphasises the training and smoothing aspects, such as in (Tjong Kim Sang, 1998). I have accordingly used a similar terminology here. In addition this emphasises the fact that there has been a lot of prior work on this in bio-informatics. The basic idea of sticking together the Levenshtein edit distance algorithm and forward-backward algorithms has been discovered independently several times, [4] but the earliest work is definitely in bioinformatics (Allison et al., 1990, 1992; Allison, 1993).

---

[3]I do not have a good explanation for this. It is possible that the model had not converged fully.

### 6.5.2   Joint versus Conditional probabilities

The choice to model the joint rather than the conditional probability is one of the key points of this model. The advantages are

- The joint probability is more informative than the conditional probability.

- It is possible to combine models of the joint probability to make mixture models.

- The two streams are treated symmetrically, though the transductions are asymmetrical.

The disadvantages are the additional complexity in converting to the conditional probability, and that sometimes the models do not model the transduction accurately, since they are modelling the joint probability, though this problem could be alleviated by using a more discriminative training regime (Normandin, 1996; Krogh & Riis, 1999).

In the NLP community, Knight and Graehl (1997) present as part of a system for transliterating Japanese versions of English words back into English a simple EM training algorithm for finite state transducers but without a version of the correct dynamic programming algorithm – i.e. they actually sum over the exponentially many alignments directly, in an application where the growth rate is not too high.

### 6.5.3   Extension to Context-Free Transductions

Just as Hidden markov Models can be extended from finite state to context free, giving rise to SCFGs and requiring a change in the training algorithm from the forward-backward algorithm to the inside-outside algorithm (Baker, 1979), so PHMMs can be extended to CFGs, in which case they are very close to the Stochastic Inversion Transduction Grammars (SITG) of Wu (1995, 1997). Naturally this gives rise to an EM based training algorithm, that can be modified to train SITGs as well. I will sketch the algorithm here. Instead of the extensions of the forward and backward probabilities we have extensions to the inside and outside probabilities. Given a string $u_1, \ldots, u_l$ and a string $v_1, \ldots, v_l$:

**Definition 5** $\alpha_s(i, j, i', j')$ *the outside probabilities are defined as the probability of starting from the start symbol, generating on the left the non-terminal $N_s$ from $i$ to $j$ and on the right the same non-terminal goes from $i'$ to $j'$, and everything outside it on both sides.*

**Definition 6** $\beta_s(i, j, i', j')$ *the inside probabilities are defined as the probability that starting from $N_s$ you can generate the strings on the left from $i$ to $j$ and the strings on the right from $i'$ to $j'$*

Given these definitions the algorithm proceeds in the normal way.

---

[4]Allison, Ristad, Bengio et al., Casacuberta; and the current author

### 6.5.4 Other applications

This is a general algorithm for learning finite state string transductions. PHMMs are clearly unsuitable for machine translation since they are *monotonic*: there is a monotonic alignment between the sequence of state transitions and the sequence of symbols in the two streams. This was part of the motivation for Wu's move to context-free models for bilingual alignment (Wu, 1997). However there are two interesting applications of these models. First, if we consider a slightly richer set of output functions, we could use them as an error model for spelling correction. Thus we could add sets of parameters corresponding to particular sorts of spelling errors. We already have the output functions that correspond to inserting extra letters and omissions, but we could also add ones for duplication, such as an output that produces *aa* on one stream and *a* on another, or for reversals such as an output that produces *ab* on one stream and *ba* on the other. This could be integrated into the morphological transducers themselves to produce a robust morphological processor (Oflazer, 1996). It seems likely that conditioning the probabilities of errors on a hidden state would improve the performance of the error-correction.

Another interesting application is that of grapheme-phoneme conversion (Divay & Vitale, 1997; Rentzepopoulos & Kokkinakis, 1996). In English, for example, the mapping between graphemes and phonemes is not completely straightforward (Torkkola, 1993; Daelemans & van den Bosch, 1997). as sequences of more than one grapheme may be realised as a single phoneme, such as *th*, and a single letter *j* may correspond to more than one phone. In other languages, notably Japanese which has several different scripts that interact in non-trivial ways the task can be even more difficult (Baldwin & Tanaka, 1999, 2000). I have done some preliminary experiments on this which are very promising.

### 6.6 Mixture Models

I have so far shown an algorithm that can learn a single surface to surface transduction. As previously noted, since morphological processes are in general lexically specified as well as phonologically specified, we need an extension to handle morphology. I will use the traditional idea of morphological classes (Cahill & Gazdar, 1997, 1999). Just as we can consider the words of a language to be divided into syntactic classes according to the range of syntactic constructions they can participate in, so too we can consider them to be further subdivided into morphological classes depending on the particular morphological transductions that can operate on them.

Let us now consider slightly more complex models, namely mixtures of PHMMs. A *mixture* model as its name implies is a combination of simple models. Given $k$ PHMMs, $m_1, \ldots, m_k$, and $k$ mixing parameters $p_1, \ldots p_k$ that sum to 1, we can define a mixture model $M$ which has this distribution

$$p_M((u,v)) = \sum_{i=1}^{k} p_i p_{m_i}(u,v) \qquad (6.38)$$

It is important to realise that this sort of simple mixture is exactly equivalent to a larger PHMM with a particular structure. If we remove the initial and final states from each sub model, and combine all of them together, and add initial transitions, weighted by the parameters $p_i$, we will have a PHMM with a block diagonal transition matrix, that generates exactly the same strings as the mixture, with exactly the same probabilities.

### 6.6.1 Modelling Morphological classes

If we divide the data set into various classes according to which inflection it takes, we can then model each morphological class separately. I use morphological class in a rather broad sense here, to refer both to large productive sets of words which are quite productive, right down to very small, even singleton classes, that model highly idiosyncratic words.

This raises two questions. First, how can we separate the set of all words into various classes automatically, and secondly, how do we choose which class to assign a new word to.

The overall model will then be a mixture of submodels, and for each submodel we have a set of lexical items that are part of that submodel. Let us define some parameters: let $p(u|i)$ denote the probability that the uninflected form $u$ is generated by class $i$. It is perhaps more natural to think of the converse parameter $p(i|u)$ which tells you which class the word is in, but it is less convenient mathematically. The probability distribution of the combined model is then

$$p_M(u,v) = \sum_i p(i) p(u|i) p_{m_i}(v|u) \qquad (6.39)$$

To re-iterate $p(i)$ and $p(u|i)$ are parameters of the model, and $p_{m_i}(v|u)$ is the conditional probability with respect to the $i$th PHMM. Since the parameters $p(u|i)$ sum to 1 over the words that appear in the corpus, this model allows no probability mass for new words: i.e. it is overtrained. We will solve this problem by smoothing later on.

### 6.6.2 Splitting the data into various classes

A principled way of splitting the data into separate classes is to use standard techniques of mixture modelling using the EM algorithm and treating the class membership as a hidden variable. This has one highly undesirable property though; it requires the number of classes to be fixed beforehand. So we proceed in a slightly more *ad hoc* way. Rather than training on all of the data, we allow the algorithm to select a subset of the data that it will model. We do this by weighting each training pair $(u,v)$ by its conditional probability $p(v|u)$. Thus a pair that is correctly modelled by the model will receive full weight, and a pair that is part of a different class will receive weight of zero. This allows the algorithm to select a subset of the data, gradually closing in on that part of the data that it can model correctly. Unfortunately, the use of this weighting means that the algorithm loses some of its attractive properties – guaranteed convergence and so on – but in practice it does seem to converge quite rapidly. There are a number of alternative machine learning techniques that might be applicable here.

The algorithm then keeps track of which pairs it has already modelled. At each iteration, it creates a new model, training it on the unmodelled pairs, weighting by the conditional probability at each iteration of the EM algorithm. When it has converged it will cover a subset of the data. If that subset is empty, then this means that the model is insufficiently complex. We then increment the number of states by one, and start again. If the subset is non-empty, then we train a larger model on this subset. We repeat this process until we have modelled all of the data. At this point we will have a number of PHMMs of different sizes, starting with small models, and going up to larger models. Each word in the training set will be assigned to one of these models. In my opinion the *ad hoc* nature of this algorithm is not a cause for concern. A more natural algorithm is one which is less supervised; such an algorithm will have to have its own way of identifying subsets. A slight artificiality in the solution to this problem of supervised learning can be tolerated, since this is in my view a rather artificial problem.

*Determining the morphological class of new words*

Given a model of this type, we are then faced with the second problem; how to determine which class a new word goes in and thus which inflection a new word comes in.

The first thing to say is that according to equation 6.39 there is no probability mass left for any new words. In other words the model is over-trained. We need to allow some space to account for new words. We can do this by smoothing the parameters $p(u|i)$ with the distribution $p_{m_i}^L(u)$. The parameters $p(u|i)$ are effectively a Maximum Likelihood model for exactly those words in a morphological class that have appeared in the corpus. $p_{m_i}^L(u)$ is a marginal distribution that represents what the input to the class $i$ is expected to be. We interpolate the two of them to get good generalisation.

If for each model we define a parameter $\lambda_i$ we have

$$p_M(u,v) = \sum_i p(i) \left( (1 - \lambda_i) p(u|i) + \lambda_i p_{m_i}^L(u) \right) p_{m_i}(v|u) \qquad (6.40)$$

Intuitively the $\lambda_i$ parameters represent how productive the paradigm is. If it is zero, then no new words will be assigned to that paradigm; if on the other hand it is greater than zero, then new words may be assigned to it, depending on the phonology.

We can find an optimal value for these parameters using the EM algorithm on held-out data. We take a guess for the values of $\lambda_i$ for each $i$; an obvious starting point is the ratio of unseen words in the held-out data say 0.01. Using these values as a starting point we estimate the expected number of times we will either use the specific $p(u|i)$ or general $p_{m_i}^L(u)$ distribution.

In some cases the solution will be degenerate; any value of $\lambda_i$ will give the same result. If we have a PHMM that produces a particular word pair with probability 1 – say *go/went*, then the specific and the general distributions will be the same – namely $p(go) = 1$. In this case we are trying to maximise $\lambda + (1 - \lambda)$. This can also happen with more than one word at a time, if the PHMM is large enough to memorise the data exactly.

This is quite close to the proposal of Baayen (1991), who proposes a quantitative measure of the productivity of a particular morphological paradigm.

### 6.6.3 Smoothing Models

These models are Maximum Likelihood models, and are thus susceptible to over-training. Accordingly we need to smooth them. We can use standard techniques of linear smoothing (Jelinek, 1997) to do this. In this work I have only attempted to smooth the output functions, and not the transition functions.

The idea is to combine the very specific per-state output distributions, which may have been based on rather few pieces of data, with less specific output distributions, that we calculate by tying the outputs from many different states.

The smoothing distributions I calculated by tying all of the outputs of each of the three types – namely the $q_{11}$, $q_{10}$ and $q_{01}$ outputs. This will result in three distributions $q_{11}^*$, $q_{10}^*$, and $q_{01}^*$, that have very good estimates. For each state $s_i$ we will then define four mixing parameters $\lambda^i, \lambda_{11}^i, \lambda_{10}^i, \lambda_{01}^i$ which sum to 1. $\lambda^i$ represents the original component of the output distribution, ie before smoothing, and the other three parameters represent the amounts of the smoothing distributions that we smooth in. For each state we then define the interpolated output distributions

$$\hat{q}_i = \lambda^i q_i + \lambda_{11}^i q_{11}^* + \lambda_{10}^i q_{10}^* + \lambda_{01}^i q_{01}^* \tag{6.41}$$

or in the notation I have used previously

$$\hat{q}_{11}(a|s_i) = \lambda^i q_{11}(a|s_i) + \lambda_{11}^i q_{11}^*(a) \tag{6.42}$$

$$\hat{q}_{10}(a|s_i) = \lambda^i q_{10}(a|s_i) + \lambda_{10}^i q_{10}^*(a) \tag{6.43}$$

$$\hat{q}_{01}(a|s_i) = \lambda^i q_{01}(a|s_i) + \lambda_{01}^i q_{01}^*(a) \tag{6.44}$$

We can find the optimal values of these $\lambda$ coefficients using the EM algorithm on held-out data.

I will give a concrete illustration. In the portion of training data I used for the experiments on the English past tense below, there are no occurrences of the segment $z$ in the portion that takes the suffix $t$ (in UNIBET notation). Therefore the PHMM that models this transduction will assign zero probability to any string that contains the phoneme $z$. Thus, if one tries to calculate the past tense of *zap*, that submodel will not generate any answer, and the answer the model as a whole gives will be wrong. However with smoothing this problem will be obviated. The $q_{11}^*$ distribution will contain a non-zero value for outputting $z$, and mixing in a small proportion of this to the appropriate states will allow the model to generalise correctly.

## 6.7 Experiments on Supervised Learning

I shall now present the results of three experiments in supervised learning on the English past tense, the German plural and the Arabic broken plural.

### 6.7.1 English Past Tense

The English past tense, notwithstanding its triviality, has become something of a test case for models of language acquisition. As a result, I decided to start my experiments with this algorithm, since there are standard data sets, and thus it is possible to compare results with other approaches. For this experiment, I used a standard data set (Ling, 1994), available on-line which consisted after processing of 1394 pairs of uninflected and inflected verb forms in the UNIBET phonemic transcription. I use normal orthography for the examples in this paper to improve readability. I used the associated Kucera-Francis frequency information to define a simple distribution, and sampled 20,000 tokens each for the training and held-out data. To test the models, I used the full data set, and a selection of 120 infrequent words from the BNC, which I manually transcribed into UNIBET notation. This evaluation corresponds to the situation the language learner is in – the training data is likely to have nearly all the irregular verbs in, and thus the generalisation ability of the algorithm is measured mostly on regulars. It is appropriate to use this evaluation method in English, because the system is very regular. In other languages this is not the case, as I discuss below. This amount of data is very small compared to the linguistic experience of even a very young child, but it is adequate in this case.

| Data | Number of Tokens | Number of Types |
|------|------------------|-----------------|
| Complete | 24,810 | 1396 |
| Training | 20,000 | 1289 |
| Held out | 20,000 | 1287 |
| BNC Test Data | 120 | 120 |

Table 6.6: Summary of the Ling English past tense data set, and the additional BNC test set.

After training on the training set, the algorithm produced 27 classes, as shown in Table 6.7. The first class produced was the irregular class of words with invariant pasts, such as *beat*, *shed* and so on. Next came the three regular classes that take the three suffixes of *d*, *t* and *Id*. Note that these appear as three separate classes, rather than a single class. This is slightly an artifact of the selection procedure. Since I model the productivity of each class separately, it is important that the granularity of the division into classes is quite fine: if a regular and an irregular class are joined together then this mixed class will be productive and there will be errors of over-application of the irregular rule. I therefore used a method that errs on the side of dividing it into too many classes.

The values of $\lambda$ presented in the table are rather misleading: since for many of them the solution is degenerate. The only ones which are truly productive are the three regular classes 1, 2 and 3 and the irregular classes 7 (slightly) and 13.

Then we have a sequence of classes that correspond to the various irregular verbs; in general these are divided into classes. Sometimes these correspond nicely to how a linguist would define them, but sometimes they were mixtures of different sorts of irregular verb.

Cluster 8 is fairly typical of the sort of confusion – it covers eleven verbs, which we can group into three classes:

- creep sleep sweep weep

- come become

- bleed feed lead mislead read

Note that *overcome* and *keep* do not form part of this class.

The data sets used here are quite small, so we find that there are still irregular words occurring in the held out data. In particular class 13, as we can see, is mildly productive which is an excellent match with psychological data (Bybee & Slobin, 1982; Bybee & Moder, 1983; Prasada & Pinker, 1993). Table 6.8 shows the output of the whole model on two nonce words, *sping* and *spling*. We can see from the probabilities that the model produces *spung* and *splinged* as the respective past tenses. Though I am not trying to create a cognitive model, this is nonetheless extremely interesting, since for some speakers these irregulars are mildly productive. The fact that *spling* is given the regular form, whereas *sping* gets the irregular, highlights the fact that this is to a certain extent a result of rather random properties of the data set and the way I selected it.

| i | States | Words | First word | λ | Entropy |
|---|--------|-------|------------|---|---------|
| 0 | 8 | 13 | beat | 2.10856e-08 | 3.80351 |
| 1 | 9 | 607 | abandon | 0.0709904 | 14.8788 |
| 2 | 9 | 248 | accomplish | 0.0608113 | 12.4003 |
| 3 | 11 | 309 | abet | 0.0752991 | 14.7088 |
| 4 | 11 | 16 | befall | 1.83297e-20 | 8.369 |
| 5 | 11 | 20 | arouse | 2.52683e-14 | 6.21204 |
| 6 | 11 | 1 | eat | 0.000992045 | 0.00353328 |
| 7 | 11 | 13 | drink | 0.0874433 | 6.66736 |
| 8 | 11 | 11 | become | 1.18605e-09 | 5.12286 |
| 9 | 11 | 4 | bend | 0.000985733 | 1.38629 |
| 10 | 11 | 4 | bind | 2.53593e-07 | 2.50897 |
| 11 | 11 | 2 | hide | 9.65926e-07 | 1.38988 |
| 12 | 11 | 4 | catch | 9.67531e-07 | 2.21603 |
| 13 | 11 | 8 | cling | 0.298878 | 4.78478 |
| 14 | 11 | 5 | deal | 2.68994e-07 | 2.56084 |
| 15 | 11 | 2 | sell | 0.000987128 | 0.699749 |
| 16 | 11 | 3 | buy | 1.67762e-08 | 2.37265 |
| 17 | 11 | 1 | grind | 0.000983235 | 0.00343301 |
| 18 | 11 | 2 | give | 0.00099005 | 0.693147 |
| 19 | 11 | 1 | seek | 0.000987171 | 1.52531e-10 |
| 20 | 13 | 2 | flee | 9.61631e-07 | 1.39126 |
| 21 | 11 | 1 | run | 0.00098919 | 0.00686603 |
| 22 | 12 | 1 | overcome | 0.00367263 | 2.77359 |
| 23 | 13 | 2 | stand | 3.80346e-10 | 3.30199 |
| 24 | 13 | 3 | bring | 2.89545e-10 | 2.66922 |
| 25 | 13 | 3 | go | 5.49077e-11 | 4.37597 |
| 26 | 13 | 2 | build | 9.64597e-07 | 1.39489 |
| 27 | 15 | 1 | understand | 8.81937e-16 | 2.8662 |

Table 6.7: Classes derived from the English past tense.

| sping | $p(v|u)$ | spling | $p(v|u)$ |
|-------|----------|--------|----------|
| spung | 0.930 | splung | 1.92 e-10 |
| spang | 0.070 | splang | 3.4e-8 |
| spinged | 1.6e-4 | splinged | 1.0 |

Table 6.8: Mild productivity of a strong verb class.

| Data | Token accuracy | Type accuracy | Type Errors |
|------|----------------|---------------|-------------|
| Complete Ling Data set | 99.96% | 99.57% | 6 |
| BNC test data | 99% | 99% | 1 |

Table 6.9: Results on the English past tense data set. The single error in the BNC test set was the pair forbid/forbad which was over-regularised.

| Suffix | Joint probability |
|--------|-------------------|
| +Id    | 9.4 e-13          |
| +d     | 3.4 e-6           |
| +t     | 1.0 e-14          |

Table 6.10: Probabilities of the different suffixes for aid / *ed*

| Word | +d       | +t       | +Id      |
|------|----------|----------|----------|
| sem  | **2.0 e-4** | 4.7 e-15 | 1.6 e-16 |
| sep  | 1.3 e -12 | **3.6 e-4** | 1.6 e-8  |
| sed  | 2.7 e-6  | 7.3 e-15 | **5.9 e-4** |
| sek  | 1.7 e -12 | 4.4 e-4  | 6.9 e-8  |

Table 6.11: $p(u)$ for the three regular sub-models, evaluated for four test words.

Table 6.9 summarises the results of the evaluation. On a separate test set of 120 words selected at random from comparatively rare words (frequency of 10) from the BNC and manually transcribed, the algorithm made a single error: *forbid* was given the regular plural instead of the correct irregular plural *forbad*. On the full Ling data set, there were 6 errors out of 1388 pairs. Note that most of the data had already been seen during the training process. *withhold, wet* , *overtake overhear* , *bid* and *aid*. It over-regularised all of the irregular ones. For the (regular) verb aid (UNIBET *ed*), it produces the completely incorrect form *aidd* , in the UNIBET notation *edd*. It is worth examining this error. On close inspection, Table 6.10, it appears that the PHMM , (model 3 above) responsible for the +*Id* generates it with a much lower probability than the default +*d* inflection. The problem here is caused by the fact that the +*d* model applies to almost every string. It is not blocked by the more specific +*Id* model because in the training data there are no words that start with *e* and take the +*Id* ending – the only other words in the data that start with *e* are aim and ache, which take *d* and *t* respectively. Thus in spite of the smoothing it is still over-trained. In other runs with the same data set, normally the +*d* model learns that it only applies to words with a particular ending. In this case, even though it has not, it is still generally blocked by the other more specific rules.

Table 6.11 shows how the various regular classes compete to produce the right answer. The table shows the values of $p(u)$ for the three regular sub-models, calculated for four test words. Note that here, the values of $p(u)$ allow the overall model to select the correct form.

### 6.7.2 German Plural

German has a much more interesting and complex system of inflectional morphology (Clahsen, Rothweiler, Wöst, & Marcus, 1992; Marcus, Brinkmann, Clahsen, Wiese, & Pinker, 1995; Cahill & Gazdar, 1999). I will concern myself with the formation of the noun plural. This has a number of interesting properties, two of which present special challenges for learning algorithms. The first is the presence of productive non-concatenative processes such as the umlaut, which changes a vowel (changes back vowels to front) in the root, which may be followed by for example a consonant cluster. Thus this cannot be modelled naturally by the sorts of techniques that have been used in symbolic learning paradigms. The second is the troubled status of the default status of the $+s$ inflection. As Clahsen et al. (1992, p. 225) say:

> The noun plural system in German is particularly interesting, because most nouns have irregular plurals in German, and the regular (default) plural is less frequent than several of the irregular plurals. Thus it is unclear how a language learner determines whether German even has a regular plural and if so what form it takes.

This will require a little bit of background to explain properly. As previously mentioned, there is a great deal of controversy over whether people use a single method for processing morphology, the single-route hypothesis, or whether they use two methods, the first a symbolic rule based approach and the second a pattern associator, being used for regular and irregular morphology respectively, the dual-route hypothesis (Pinker, 1991). Rumelhart and McClelland (1986a) proposed a model where both regular and irregular words were processed by a single model. Dual route advocates have therefore spent a great deal of time showing that certain types of models cannot cope with certain constructions, in particular the German $+s$ plural inflection. This is considered to cause particular problems because it is claimed that it is the *default* inflection, but is comparatively rare. Cahill and Gazdar (1999) agree, stating

> This latter [the $+s$ inflection] may seem a curious choice for the default plural suffix for German since it occurs much less frequently than the other plural suffixes. However, we are persuaded by the extensive linguistic evidence given by Clahsen and his collaborators that $+s$ is indeed the default plural suffix for German: it is the suffix that standardly occurs with surnames, product names, acronyms, truncated nouns, unassimilated borrowings, foreign words, derived forms, neologisms, onomatopoeic nouns and nouns formed from VPs and APs.

However while default inflections must be regular, non-default inflections need not be irregular. In fact there are several other German noun inflections that are also perfectly regular. There appears to be a deliberate attempt by proponents of the dual-route hypothesis to blur the distinction between default processes and regular processes, which we can see in the quote by Clahsen et al. above. Empirically it is not the case that new words are put into this default class. Köpcke (1988) (discussed also by Bybee (1995)) presents a fascinating study on this issue. He selected a group of German native speakers, without dialect variation, and asked them to produce the plurals of 50 nonce words. The results were surprisingly divergent. Only for some types of words, nouns ending in a full vowel, was the $+s$ inflection used frequently (69%). For other types of words, there was almost unanimous agreement on the plural form, but for many types of nonce word there was substantial disagreement among the speakers as to what the plural should be. For example, neuter nouns with a suffix $-lein$, for example *das Poftlein*, produced

- 6% $+en$,

- 19%, +e

- 51% unchanged

- 20% +s

- 3% +r

| $i$ | states | words | example | | $\lambda$ |
|---|---|---|---|---|---|
| 0 | 8 | 1389 | `a:` | A | 0.259059 |
| 1 | 9 | 1557 | `apbIt@` | Abbitte | 0.261867 |
| 2 | 9 | 1415 | `a:l` | Aal | 0.226159 |
| 3 | 9 | 384 | `abOn@mA~:` | Abonnement | 0.231288 |
| 4 | 12 | 2663 | `apb@ru:fUN` | Abberufung | 0.221994 |
| 5 | 12 | 4 | `aNst` | Angst | 2.39063e-10 |
| 6 | 12 | 252 | `apt` | Apt | 0.216562 |
| 7 | 12 | 139 | `apbrUx` | Abbruch | 0.251388 |
| 8 | 14 | 127 | `apSti:k` | Abstieg | 0.246547 |
| 9 | 13 | 1 | `ham@r` | Hammer | 0.000985667 |
| 10 | 12 | 92 | `a:b@nt` | Abend | 0.263093 |
| 83 | 24 | 1 | `atlas` | Atlas | 0.000978075 |
| 84 | 24 | 2 | `blu:m@nStraus` | Blumenstraus | 3.22659e-09 |
| 85 | 24 | 3 | `atvErp` | Adverb | 0.354987 |
| 86 | 24 | 1 | `mo:s` | Moos | 0.000980299 |
| 87 | 25 | 2 | `o:m@n` | Omen | 0.000962894 |
| 88 | 27 | 1 | `vas@rgla:s` | Wasserglas | 1 |
| 89 | 26 | 1 | `Strantba:t` | Strandbad | 0.000963431 |
| 90 | 27 | 1 | `Sta:tso:b@rhaupt` | Staatsoberhaupt | 2.60003e-09 |
| 91 | 26 | 2 | `artri:tIs` | Arthritis | 1.47099e-18 |
| 92 | 26 | 1 | `bErkman` | Bergmann | 0.000969702 |
| 93 | 28 | 1 | `vErb@faxman` | Werbefachmann | 1 |
| 94 | 27 | 1 | `IndEks` | Index | 0.000970578 |

Table 6.12: First ten and last ten classes induced from German plural data set.

So as far as I am concerned, if a model trained on the German plural, produces the $+s$ form for new words uniformly, this would be a point against it rather than a point in favour.

The data set here was generated from the CELEX lexical database on CDROM (Baayen, Piepenbrock, & van Rijn, 1993). I generated all pairs of singular and plural nouns from the database, which gave a data set of 17076 pairs. Using the associated frequency information, I defined a simple distribution. I then sampled 20,000 tokens for a training set which has 4709 distinct pairs, and then sampled 100,000 tokens for a held-out data set, which gave 9346 distinct pairs.

The algorithm produced 94 classes, with many classes corresponding to individual irregular words, such as class 83, which produces the pair *atlas* and *atlanten*. Table 6.12 shows the first and last ten classes produced. A noticeable difference from the English results are the larger number of classes with high values of $\lambda$ i.e. productive classes.

Table 6.13 shows 10 pairs of strings randomly generated from the model for class 7. Note that this model produces only a single vowel change; since the phonemes are treated as atomic symbols, the algorithm produces a separate model for each vowel change. It would certainly be possible to alter the model to accommodate structured symbols. Note also that the model changes *all* occurrences of `U` (u) to `Y` (ü), not just the terminal ones, and adds a schwa with certain endings.

| u | v | p(u, v) |
|---|---|---|
| grUtr | grYtr | 1.4441e-05 |
| z@:nmUSt | z@:nmYSt@ | 1.6213e-09 |
| au:aurmUm | au:aurmYm@ | 3.75777e-10 |
| kUs | kYs@ | 0.00150417 |
| EsplUrxr | EsplYrxr | 3.92098e-10 |
| bu:lSSUrUrprUrinta:mkUxt | bu:lSSYrYrprYrinta:mkYxt@ | 6.43179e-23 |
| lUrx | lYrx@ | 0.000141695 |
| SrUs | SYrYs@ | 0.000165491 |
| @sSmUs@ | @sSYmYs@@ | 1.8472e-08 |
| tsllUx | tsllYx@ | 3.96769e-07 |

Table 6.13: 10 pairs of strings randomly generated from model 7 of the German plural model.

In spite of the reservations expressed above, I did evaluate it in the straightforward way. On the complete data set, it scored on types 14222 right, 2854 wrong for an accuracy of 83.2%, and in terms of tokens, 944135 right, 19438 wrong, an accuracy of 98.0%. Many of these had in fact been seen before, either in the original data set, or in the held-out data. These results are objectively rather poor; this is partly due to the fact that the algorithm does not have access to the gender of the words, which is an important factor in determining the correct plural form. In addition, the prevalence of compound nouns complicates matters. This could be resolved in a number of different ways: most simply, the data could be restricted to monomorphemic nouns. Alternatively, one could insert some segmentation symbols into the sequence of phonemes to indicate the boundaries.

### 6.7.3  Arabic Plural

Arabic has a complex system of morphology based on the system of triconsonantal roots that is common in Semitic languages. It has three numbers, the singular, the dual and the plural. The plural comes in two forms, the sound plural or *pluralis sanus*, and the broken plural or *pluralis fractus* because (Wright, 1967):

> ... it is more or less altered from the singular by the addition or elision of consonants, or the change of vowels.

The Arabic plural is an extremely complex system. McCarthy and Prince (1990) analyse it in the framework of Prosodic morphology capturing significant generalisations and greatly reducing the apparent complexity. Here our task is not so much to reduce the complexity of the system, but rather to model the process using as much complexity as we need. The only motivation for reducing complexity would be to improve the generalisation of the model.

There are a couple of preliminary points I want to make before I proceed to a description of the plural itself; here I am paraphrasing some standard reference works on Arabic (Comrie, 1987; Versteegh, 1997). First, Arabic does not really exist as a single language: in Arab speaking countries there is a comparatively rare system of diglossia where two variants of the language are spoken. Each country or geographical region has a distinctive dialect, which is often incomprehensible to the speakers of another dialect. The dialects are normally grouped into five families, (Versteegh, 1997, p.145) Arabian, Mesopotamian, Syro-Lebanese, Egyptian and Maghreb. These are the dialects that are spoken in the shops, at home and in other informal situations. In addition there is a standard more formal dialect, called Modern Standard Arabic, based on the classical form of the language, that is more or less the same throughout the Arab speaking world, which is learnt later on through formal instruction and is heavily constrained by religious and political factors. Thus in general Arab speakers are not *native* speakers of the standard dialect; they are native speakers of the local dialect. It is traditionally thought that the Bedouin speak the purest form of Arabic, (Versteegh, 1997, pp. 63–64) though the Cairo dialect of Egyptian Arabic has become widely understood because it is the centre of the film industry. The results I am presenting here are based on dictionaries, primarily the Hans Wehr (Wehr, 1979) dictionary of Modern Standard Arabic. Secondly, Arabic is normally written without most of its vowels. Since the roots of words are based on consonants, this is possible in a way it wouldn't be in English. For obvious reasons, I will be working with fully vocalised transcriptions. Thirdly, the broken plural is also productive: recent loan words that satisfy various phonological criteria will take the broken plural (McCarthy & Prince, 1990).

As previously mentioned, the plural comes in two forms. The first, the sound plural is primarily formed by suffixation, with minimal changes to the stem. The second type, the broken plural, is formed by a variety of different, sometimes quite radical, changes to the vowels and consonants in the stem. Moreover the selection of which modification is used is dependent, among other things on overall prosodic properties or patterns of the whole stem. I will give some concrete examples now to illustrate this. I will use $C$ to denote consonant and $V$ to denote vowel. There are three vowels in Arabic, $a$, $i$ and $u$; I denote the long vowels with capital letters. One process takes stems of the pattern $CaCC$ to $CuCUC$, where the three consonants map to each other in sequence; so $nafs$ becomes $nufUs$, and $bank$ becomes $bunUk$. Another process takes stems of the pattern $CVCCVC$ to $CaCACiC$ so $jundub$ becomes $janAdib$ and $zalzal$ becomes $zalAzil$. There are a number of other patterns like this, but the situation is complicated by the fact that many nouns have several possible broken plurals as well as a sound plural. The sound plural on the other hand is a suffix of $Un$ or $At$ depending on the gender.

I used a data set prepared by Ramin Nakisa (Plunkett & Nakisa, 1997), to whom I am very grateful. It does not include frequency information, and is quite small, 859 singular plural pairs, so I was not able to examine the productivity of the various classes. It contains a mixture of sound and broken plurals taken from the Wehr dictionary, with some errors either in the transcription process, or according to a native informant, in the dictionary. The data is in a non-standard ASCII transcription. Capital vowels (A, I and U) correspond to long vowels.

On this data set the algorithm produced 56 classes, the most frequent of which are shown in Table 6.14. As can be seen, the algorithm separated the data into various classes with reasonable accuracy. Table 6.15 shows ten randomly generated pairs of words from the model for class 7. This submodel deals with a particular non-concatenative transduction mapping singulars of the form `CaCC` where `C` is a consonant, to `CuCUC`. Some of the other models model single transductions. Others, unfortunately, model mixtures of different transductions. This is perhaps to do with the size of the data set.

| i | states | words | first | plural | description |
|---|--------|-------|-------|--------|-------------|
| 0 | 9 | 31 | Euqda | Euqad | CVCCa -> CuCaC |
| 1 | 10 | 72 | racIfa | racIfAt | +At or +a -> +At |
| 2 | 11 | 73 | *aHl | a*HAl | CVCC -> aCCAC |
| 4 | 11 | 9 | ijtimAE | ijtimAEAt | +At |
| 5 | 12 | 56 | tarika | tarikAt | +At or +a -> +At |
| 7 | 12 | 47 | fatq | futUq | CaCC -> CuCUC |
| 10 | 12 | 10 | qinAE | qunuE | CVCVC -> CuCuC |
| 11 | 12 | 17 | =A'ira | =awA'ir | CVCVC or CVCVCa -> CawACiC |
| 13 | 12 | 9 | fAtik | futAk | CACiC -> CuCAC |
| 14 | 14 | 36 | mavhara | mavAhir | CaCCaC(a) -> CaCACiC |
| 15 | 14 | 15 | =amar | =imAr | mixture |
| 16 | 14 | 21 | zA'ir | zA'irUn | +Un |
| 17 | 14 | 14 | vay | avwA' | mixture CauC -> aCwAC |
| 22 | 14 | 17 | bur=un | barA=in | mixture |
| 24 | 14 | 17 | sakra | sakarAt | CaCCa -> CaCaCAt |
| 26 | 14 | 9 | su'Al | as'ila | CVCVC -> aCCiCa |
| 27 | 14 | 21 | faqIr | fuqarA' | CaCIC -> CuCaCA' |
| 28 | 14 | 12 | xAtUn | xawAtIn | CVCVC -> CawACIC |
| 29 | 14 | 5 | maulan | mawAlin | CauCa(C) -> CawACi(C) |
| 30 | 14 | 11 | mikyAl | makAyIl | CVCCVC -> CaCACIC |
| 34 | 15 | 10 | wadIEa | wadA'iE | CaCICa -> CaCA'iC |
| 39 | 15 | 6 | vAGin | vuGAh | CAU->CuAh or CACiC-> CuCAh |
| 40 | 16 | 5 | mijmara | majAmir | mixture |
| 44 | 15 | 7 | qaEIda | qaEA'id | mixture |

Table 6.14: Classes with five or more words produced by the algorithm on Plunkett and Nakisa data set. A, I and U are long vowels, a i and u are short vowels. All other symbols are consonants. There were 56 classes in all. Where possible, I have tried to describe the transduction using the CV-template. Here C refers to any consonant, and V refers to any vowel, though there are often restrictions on the range of vowels or consonants that can be used in each position. Where this is not possible I have just put 'mixture'.

| $u$ | $v$ | $p(u,v)$ |
|------|------|----------|
| falE | fulUE | 0.000516428 |
| zamH | zumUH | 5.16428e-05 |
| tabb | tubUb | 6.88571e-05 |
| kamr | kumUr | 0.000413142 |
| mafd | mufUd | 0.000361498 |
| valj | vulUj | 6.88571e-05 |
| warl | wurUl | 0.000619712 |
| qamd | qumUd | 0.000903745 |
| cacm | cucUm | 6.88571e-05 |
| faHq | fuHUq | 8.60713e-05 |

Table 6.15: 10 pairs of words randomly generated from model 7 from the Arabic plural model.

## 6.8 Partially Supervised Learning

Partially supervised learning refers to the learning paradigm where the learning algorithm is presented with two lists of words, rather than with lists of pairs of words. The algorithm must then work out which word is aligned with which other word. In a realistic learning situation, the two lists will be of different lengths with various forms of noise errors and omissions. I will first present a algorithm which will work in an artificial situation where there is an exact match between the two sets. This is mathematically quite clean, but hopelessly unrealistic. I will then show how these can be extended to deal with a more imperfect and realistic situation.

### 6.8.1  Perfect Partially Supervised Learning

In this situation we are concerned with the artificial situation where we have a list of words and their inflected forms, but we do not know the one-to-one mapping between them. For example, we might have two lists (cat,dog,cow) and (dogs,cows,cats), and we would have to learn the alignment between the two sets, namely the one to one mapping (cat,cats), (dog,dogs), and (cow,cows). If we already know the morphology, this is straightforward, under very reasonable assumptions. If we know the form the morphology takes, e.g. suffixation, prefixation, then it is again trivial – we can merely sort the strings alphabetically, or sort the reversed strings alphabetically, and this will give us a quick result. But in general it is not trivial to learn the alignment and simultaneously the morphological process that gives rise to that alignment. In this section I will present an algorithm based on the EM algorithm that can do this.

Let us suppose we have 2 sets, each of which has $n$ words, $U = \{U_1, \ldots, U_n\}$ and $V = \{V_1, \ldots, V_n\}$. We want to find a PHMM, $M$, and an alignment $\pi$ that maximises the probability of the observed data. The alignment $\pi$ is an element of the set of permutations of $n$ elements, that we can consider as $n \times n$ permutation matrices. Recall that a permutation matrix is a square matrix all of whose elements are one or zero, such that each row and each column has exactly one one, the rest being zeros. Alternatively we will write $\pi(i)$ for index of the element in $V$ that the $U_i$ is mapped to - so the mapping will be $(U_1, V_{\pi(1)}), \ldots, (U_n, V_{\pi(n)})$. We can consider this permutation as a hidden variable; we can train a model using the EM algorithm and then find the value with the highest posterior probability given the data.

If we denote the hidden variable, the permutation, as $X$, as is usual we will have

$$p(U, V) = \sum_X p(X) p(U, V | X) \tag{6.45}$$

There are $n!$ possible values for $X$ so we can set $p(X) = 1/n!$, The probability of the data for a particular permutation will just be the product of the joint probability of the $n$ pairs

$$p(U, V | X = \pi) = \prod_i p_M(U_i, V_{\pi(i)}) \tag{6.46}$$

The overall joint probability will then be

$$P(U, V) = \sum_\pi p(\pi) \prod_{i=1}^n p_M(U_i, V_{\pi(i)}) \tag{6.47}$$

Now this sum of the $n!$ products of permutations of a matrix is called the *permanent* of a matrix (Bhatia, 1996). It is similar to the determinant of the matrix, but without the alternating minus signs. I will use $Q(M)$ to denote the permanent of an arbitrary square matrix. Using this notation we can write the probability more succinctly

$$P(U, V) = \frac{1}{n!} Q(P_M(U, V)) \tag{6.48}$$

where $P_M(U, V)$ is the matrix of joint probabilities of $U$ and $V$ with respect to the model, i.e. the matrix whose $(i, j)$ element is $p_M(U_i, V_j)$.

When we train our model using the EM algorithm, we weight each transition by the posterior probability of the hidden variable given the data $p(X | U, V)$. Thus in this case this reduces to weighting each pair $(U_i, V_j)$ by the posterior probability that $U_i$ is aligned with $V_j$. Intuitively, as the PHMM more closely models the transduction, the posterior probability matrix will become closer to the correct permutation matrix, and thus the model will be trained on a more correct alignment. This is guaranteed to converge by the EM theorem.

The $(i, j)$ element of the posterior matrix will be

$$P(\pi_{ij} | U, V) = \frac{\sum_{\pi : \pi_{ij} = 1} P(\pi, U, V)}{P(U, V)} \tag{6.49}$$

So on the top we have the sum of the $(n-1)!$ permutations that match $i$ with $j$, and on the bottom we have the sum of all $n!$ permutation. This will just be

$$P(\pi_{ij} | U, V) = \frac{P_M(U_i, V_j) Q(P^{ij}(U, V))}{Q(P(U, V))} \tag{6.50}$$

Where we use $P^{ij}$ to mean the $i j$th minor of the matrix, i.e. the $n-1$ by $n-1$ square matrix formed by removing the $i$th row and the $j$ the column from the probability matrix. So this has quite a nice interpretation as the product of two terms $P_M(U_i, V_j)$ which is the effect of how likely the matching is between $U_i$ and $V_j$, and the remaining term which represents how likely it is to match the remaining $n-1$. This is an important factor. For example, suppose we had two sets of strings (a, aa, aaa) and (aa, aaa, aaaa). Clearly the correct matching is the order they are in, together with a transduction of appending $a$. It is however also possible to match the *aa* with *aa* and *aaa* with *aaa*, but then one has to match *a* with *aaaa*; the constraint that it is a one to one matching would penalise this erroneous mapping. Note that the matrix of posterior probabilities will be doubly stochastic – that is to say all of its rows and columns will sum to 1, just like the permutation matrices. This matrix is the posterior estimate of the permutation matrix.

However there is a large problem that I have so far neglected to mention: equation 6.50 does not help since it is (almost certainly) not possible to calculate the permanent efficiently (in polynomial time), as the calculation is #*P* complete (Valiant, 1979). The determinant can be calculated in $O(n^3)$ time, which gives rise to various techniques for approximating the permanent (Jerrum & Sinclair, 1989; Barvinok, 1999). This is a very active area of research, with a number of open questions with deep relations to other areas of mathematics. One straightforward way of estimating the permanent, the Godsil-Gutman estimator, is to randomly flip the signs of all the elements in the matrix, after having taken the square root of all of the elements first. Then the expectation of the square of the determinant of this randomised matrix is the permanent of the original matrix. More sophisticated techniques involve lifting the matrix to a more complex algebra, to the complex numbers, or quaternions, or even higher dimensional Clifford algebras.

I decided to use a much simpler method. Since all of the elements of the matrix are non-negative, the permanent will clearly be non-negative, and it will be less than or equal to the product of the sum of the rows. If we expand out the product of the sum of each row, we will clearly see every permutation in there. Similarly, the permanent will be less that the product of the column sums. I therefore estimate the permanent of a matrix $M$ by

$$Q(M) \approx \min \left\{ \prod_i \sum_j M_{ij}, \prod_j \sum_i M_{ij} \right\} \tag{6.51}$$

We can use the same technique to calculate the permanents of all the minors of the matrix. Naively this might take $O(n^4)$ time ($n^2$ minors, times $(n-1)^2$ for each minor), but if we cache the total row and column sums, and then subtract for each minor, we can do it in $O(n^3)$, which makes it quite practical.

This is the technique I have used here, but I have since identified a much better technique. The matrix map that takes each element

$$p_{ij} \rightarrow \frac{p_{ij} Q(P^{ij})}{Q(P)} \tag{6.52}$$

is known as the Bregman map, (Bregman, 1967; Beichl & Sullivan, 1999), and though it is intractable to compute exactly, the technique of Sinkhorn scaling (Sinkhorn, 1964) produces a doubly stochastic matrix that is under reasonable conditions an approximation to the Bregman map. The method of Sinkhorn balancing is intuitively quite straightforward; we want to scale a positive matrix so it is doubly stochastic. If we normalise the row sums, we will have a matrix that is row-stochastic, i.e. has its row sums equal to unity, but not necessarily column-stochastic. If we then normalise the column sums, we will have a matrix that is column-stochastic but probably not row-stochastic. If we continue in this way, alternating normalising the rows and normalising the columns, we converge to a doubly stochastic matrix. Soules (1991) established that it converges, under reasonable assumptions, linearly which gives an overall complexity of $O(n^3)$. This has relations to some other techniques for finding graph matchings, used to solve the travelling salesman problem (Gold & Rangarajan, 1996). In future work I will use this technique (Clark, 2001b).

So, given a model, we can compute the posterior probabilities, and then use those to weight the contributions of the models. The algorithm then proceeds as follows. Start with a random model. Calculate the $n$ by $n$ matrix of probabilities $P_M(U_i, V_j)$. Calculate the $n$ by $n$ matrix of posterior probabilities. Train the model $M$ using these as weights. Repeat until the matrix of posterior probabilities has converged to a permutation matrix.

This algorithm is rather slow since it requires a computation of the training algorithm for each of the $n^2$ pairs. I tested it on the BNC test data set for the English past tense, which had 120 pairs, and it converged in 14 iterations to the correct alignment. As mentioned previously, this is an artifical situation; I shall not investigate it further.

### 6.8.2   Imperfect Partially Supervised Learning

In a more realistic situation, there will be numerous errors and omissions. We thus want to align a subset of one set, with a subset of the other set. We can adapt the previous technique to this more complex situation.

Given two sets of words, $U$ and $V$, and models for each set, $p_U$ and $p_V$, we could model the set of data, assuming that there is no relationship between them, with probability function

$$p(U, V) = \prod_{u \in U} p_U(u) \prod_{v \in V} p_V(v) \tag{6.53}$$

If we assume there is a relationship $\pi$ between two subsets $U' \subseteq U$ and $V' \subseteq V$ which have the same size, that are generated by a joint model $M$, we could write this as

$$p(U, V | X) = \prod_{u \notin U'} p_U(u) \prod_{v \notin V'} p_V(v) \prod_{u \in U'} p_M(u, \pi(u)) \tag{6.54}$$

As long as $p_M(u, v) > p_U(u) p_V(v)$ this will be an improvement. This is similar to the approach of Allison et al. (1999). It is important to take into account the regularities in both sets, by explicitly modelling them, in order to avoid spurious correlations.

Proceeding as before we can define an indicator function $\pi(i, j)$ which indicates if $u_i \in U', v_j \in V', \pi(u_i) = v_j$. We also define $\phi(i)$ and $\psi(j)$ to represent not being members of $U'$ and $V'$ respectively. Therefore $\phi(i) + \sum_j \pi(i, j) = 1$ and $\psi(j) + \sum_i \pi(i, j) = 1$.

Using these indicator functions we can write

$$\log p(U, V | X) = \sum_i \phi(i) \log p_U(u_i) + \sum_j \psi(j) \log p_V(v_j) + \sum_{i,j} \pi(i, j) \log p_M(u_i, v_j) \tag{6.55}$$

We can rewrite this as

$$\log p(U, V | X) = \sum_i \log p_U(u_i) + \sum_j \log p_V(v_j) + \sum_{i,j} \pi(i, j) \log \frac{p_M(u_i, v_j)}{p_U(u_i) p_V(v_j)} \tag{6.56}$$

We will want to weight the training algorithm by the expectation of $\pi(i, j)$ with respect to the posterior distribution. That is, we will want to train the model with the pair $(u_i, v_j)$ with the weight equal to $E[\pi(i, j)]$ where the expectation is taken with respect to the distribution $p^{old}(X | O)$ using the notation of the previous section.

Since this is either 1 or 0 this expectation is just $p(\pi(i,j)|O)$. We can estimate this

$$p(\pi(i,j)|O) = \frac{\sum_{X:\pi(i,j)} p(X)p(U,V|X)}{\sum_X p(X)p(U,V|X)} \qquad (6.57)$$

I neglect $p(X)$, assuming it is constant. If we define a matrix $M$ such that the $ij$th-element is

$$M_{ij} = \frac{p_M(u_i,v_j)}{p_U(u_i)p_V(v_j)} \qquad (6.58)$$

We can estimate the top and bottom as sums of products of elements of $M$. In this case they do not correspond to permanents, but we can nonetheless approximate in a similar way. The sum is over any permutation of subsets ; so as long as we pick at most one from each row and column we can take any product. However we do not have to pick one from each row or column, so instead of multiplying the row sums or column sums, we take the product of each row sum plus one. If we denote this function by $R$ we can say

$$\sum_X \prod_{ij} \pi(i,j)M_{ij} = R(M) \leq \min\left\{\prod_i(1+\sum_j M_{ij}), \prod_i(1+\sum_j M_{ij})\right\} \qquad (6.59)$$

where $X$ ranges not just over all 0-1 matrices with at most one 1 in each row or column. and therefore

$$p(\pi(i,j)|0) \approx \frac{\Pi_{ij}R(\Pi^{ij})}{R(\Pi)} \qquad (6.60)$$

where $\Pi^{ij}$ is the $ij$-minor of the matrix $\Pi$.

This still requires us to calculate $p_m(u_i,v_j)$ for each pair of words, which is very computationally expensive. It is also unnecessary; if two words have radically different sequences of letters we don't need to compare them. For example, if we have some singular and plural nouns, we don't need to consider the possibility that the plural of *catapult* is *oxen* (or that the past tense of *go* is *went*?). Accordingly, for each string I calculate the letter distribution, a simple vector of occurrence counts of each letter, and use a cosine metric with a manually chosen cutoff to accelerate the algorithm. This is a pure optimisation that should have no effect on the algorithm other than speed.

There is a slightly cleaner way of formalizing this which maintains the connection with the permanents. I shall only sketch this here as I do not use this to produce the results later on. Suppose $U$ has $m$ elements, and $V$ has $n$ elements, then we have an $m \times n$ matrix of the joint probabilities, which we can call $M_J$ We can also define a $m \times m$ diagonal matrix, corresponding to the probabilities according to the model of $U$, which we can call $M_U$, and a $n \times n$ diagonal matrix for the probabilities of $V$, $M_V$. If we also create a matrix of size $n \times m$ with every element 1, $M_1$, then we can form an $(m+n)$ square matrix thus:

$$M = \begin{pmatrix} M_U & M_1 \\ M_J & M_V \end{pmatrix} \qquad (6.61)$$

Then every permutation of this matrix corresponds to a particular choice of the alignment, and we can use the same permanent techniques on this matrix, as we did before. There is one substantive difference which is that we end up with an additional factor of $k!$ in the joint probability, where $k$ is the number of words aligned. This is caused because there are $k!$ paths through the submatrix 1 in the upper right of the composite matrix. Formally, we can accommodate this by adjusting the prior probability of the alignment $p(X)$, and informally it is fine, because it will give a bonus to alignments that align large numbers of words, which is something that we want to encourage.

| | |
|---|---|
| Total | 2163/1394 |
| Total to be aligned | 1394 |
| Total aligned | 1286 |
| Correctly Aligned | 1276 |
| Incorrectly Aligned | 10 |
| Unaligned | 108 |
| $p(u|v)$ accuracy on BNC test set | 119/120 |

Table 6.16: Results of alignment on unaligned Ling data set. This gives a precision of 99.2% and a recall of 91.5%. The results on the BNC test set are evaluated with $p(u|v)$, the probability of getting the right base form given the inflected form.

| Suffix | Aligned | Total | Recall |
|---|---|---|---|
| d | 657 | 661 | 99.3 |
| t | 257 | 263 | 97.7 |
| Id | 324 | 335 | 96.7 |

Table 6.17: Performance on regular verbs. Precision was 100% – no regular past was aligned with an incorrect base.

## 6.9 Experiments on Partially Supervised Learning

### 6.9.1 Ling data set

Before processing the Ling data set consists of an alphabetical list of English verbs with their morphology and phonology marked. Extracting all base forms and all past-tense forms produced two sets of words with 2163 and 1394 members respectively. I trained two 7-state HMMs on each set of words. I then ran this algorithm for 10 iterations, at the end of which the vast majority of the posteriors were either zero or very close to one. I considered a word in the set of base verbs to be aligned with another word if the posterior probability was greater than 0.1. No word in the base set was aligned with more than one word in the past tense set.

Table 6.16 summarises the results. As can be seen, the alignment proceeded with very high precision, and good recall. To further evaluate the resulting model, I tested to see how accurate the transduction modelled the morphological process. On the BNC test set of 120 infrequent words, it scored 119 correct, with one error on the single irregular words, evaluated according to the less stringent criterion of $p(u|v) > 0.5$. On regular verbs, shown in Table 6.17, it performed with 100% precision; no regular past form was aligned incorrectly, and the recall was very high as well. It aligned a respectable number, 38, of irregulars correctly.

The 10 errors it made are shown in Table 6.18. Three of these errors are attributable to errors in the data set. *tore* was incorrectly transcribed to have the same vowel as *toe*, *did* was not included in the data set, and *fort* is not a verb in my dialect of English. All of the other errors are of regular verbs, whose past tense does not occur in the data set, aligned with an irregular past tense that sounds similar. Note that if we use a stricter measure of alignment based on the conditional probability we can eliminate these errors, at the cost of not aligning irregular verbs. Note this model only has seven states, five excluding the initial and final states.

| Base | Past |
|------|------|
| brand | ran |
| do | drew |
| fort | fought |
| group | grew |
| lend | led |
| ready | read |
| send | said |
| soak | spoke |
| sponge | spun |
| toe | tore |

Table 6.18: Incorrectly aligned pairs on English data set.

| | |
|------|------|
| Total | 2450/3101 |
| Total to be aligned | 2450 |
| Aligned | 277 |
| Correctly aligned | 196 |
| Incorrectly Aligned | 81 |
| Unaligned | 2173 |

Table 6.19: Result of first iteration of alignment on Arabic data set. Precision was 70.8%.

### 6.9.2 Partially Supervised Arabic

I then performed some experiments with Arabic, using a data set kindly provided by John Mc-Carthy for McCarthy and Prince (1990). This was a much more difficult test than the previous experiment. First of all, the data set is much more noisy with various errors and omissions, as well as numerous multiple plurals for particular singular forms. Secondly, the morphological system is very much more complex, as previously discussed. This is a larger and more complex data set than the one prepared by Ramin Nakisa that I experimented on above.

Table 6.19 summarises the results of the algorithm on this data set. These are rather disappointing. First of all, the very low recall is not an issue: we can run the algorithm repeatedly, excluding at each step the words previously aligned, in a similar manner to the technique used for supervised learning earlier. The poor precision is a more serious concern. Examination of the 81 erroneously aligned pairs, see Table 6.20, showed that many of them clearly have the same consonantal root. My expertise in Arabic is insufficient to identify the problem, but it appears to be related to the structure of the data set.

| Singular | Plural |
|----------|--------|
| 9abd | 9abada |
| 9u*r | 9u*ur |
| ?a9jam | ?a9jaam |
| ?ab9ad | ?ab9aad |
| ?aHdab | ?aHdaab |
| ?aHlas | ?aHlaas |
| $akl | $ukul |
| $aqq | $aqqa |
| $armuuTa | $uruuT |
| ?asmar | ?asmaar |

Table 6.20: First ten incorrectly aligned singular-plural pairs from Arabic data set.

### 6.9.3 Experiments on automatically induced classes.

In this section I will discuss experiments made with the automatically induced classes from Chapter 5. I selected two classes as the basis for my experiment which corresponded to singular and plural nouns. After selecting only those types that occurred at least 5 times in each cluster, I had two sets of words with 2931 and 1946 words in respectively. A crude measure suggested that there were only 304 pairs of words to be aligned, where a word W occurred in the singular set and a word W with the suffix S or ES occurred in the plural set. Note that these words are all in normal orthography rather than in phonemic transcription. I excluded all other possible plural pairs, such as irregulars (MAN, MEN) or words ending in Y where it is changed to IES, from the calculation of the possible alignments.

Thus the data set is extremely noisy; there are numerous errors and omissions. Table 6.21 shows a sample from the data set.

Table 6.22 summarises the results. As can be seen, even in this extremely noisy situation, the algorithm performed reasonably well. The incorrectly aligned ones included *SQUID*,*QUID* (my favourite), *BEDSIDE*,*BESIDES* and *CHANCEL*, *CHANCE*, as well as various pairs where the singular form was in the plural class and vice-versa. Having the value of the exponent at 1 improved the recall but the precision was much lower, it included many pairs where one words was roughly an anagram of the other, for example *YELLOW* and *SLOWLY*. Effectively, the only thing making it work at all was the prefiltering with the letter-distributions. Nonetheless, this validates the approach, and demonstrates that it can be used in very noisy situations. In a complete system, we would take all of the clusters, and try this algorithm on each pair of them. This should suffice to identify the major inflectional paradigms, at least in as simple a language as English.

## 6.10 Applications and Extensions

In this section I discuss various further extensions of this approach to learning morphology using PHMMs.

| Cluster 70 (plural) | Cluster 71 (singular) |
| --- | --- |
| PARAGRAPHS | PARABASIS |
| PARENT | PARACHUTE |
| PARENTS | PARALLEL |
| PARALLEL | PARASITE |
| PARISHES | PARENT |
| PARTICIPANTS | PARISH |
| PARTIES | PARK |
| PARTISAN | PARLIAMENTARIAN |
| PARTNERS | PARMA |
| PARTS | PARQUET |
| | PARTICIPANT |
| | PARTICLE |
| | PARTISAN |
| | PARTITION |
| | PARTNERSHIP |
| | PARTY |

Table 6.21: Randomly selected portion of the data.

| Aligned | 323 |
| --- | --- |
| Errors | 35 |
| Correct | 288 |
| Unaligned | 16 |
| Possible | 304 |

Table 6.22: Results of the alignment on the two induced classes. Exponent of 0.7, and cosine threshhold of 0.9. This gives a precision of 89% and a recall of 95%.

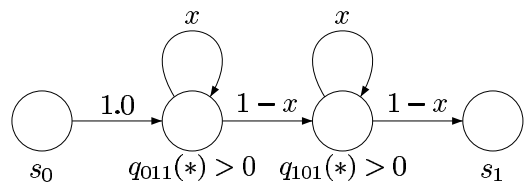

Figure 6.7: Diagram of a triple Hidden markov Model that concatenates two strings together to make a third.

### 6.10.1   Number of Channels

Up till now I have considered models that have only two streams of data. It is of course straightforward to extend these models to any arbitrary number of streams of data, replacing the three sets of output parameters with $2^n - 1$ sets, where we have $n$ streams. This might be very useful to deal with reduplication. Reduplication in morphology is quite widespread; see the discussion in Raimy (2000). We could handle the Malay data with a Triple Hidden Markov Model, using three streams, if we stipulated that two of the strings were identical, and functioned as input strings. Then extending the notation in the obvious way, Figure 6.7 shows a Triple Hidden Markov Model that concatenates two strings together to make a third. $q_{101}$ is an output that emits the same symbol on the first and third streams, $q_{011}$ outputs the same symbol on the second and third stream. Thus this model will only generate triples of strings where the third string is the concatenation of the first with the second.

### 6.10.2   Deep stems

As noted above, all of the algorithms here are concerned with learning surface to surface transductions. Given a set of lexical strings, and the corresponding surface forms it would certainly be possible to apply the same algorithms. A more interesting alternative however presents itself.

Given a different alphabet of deep symbols $B$ (phonetic symbols augmented with various additional symbols) we could define the transductions from some unknown string of symbols thus

$$p(u,v) = \sum_{w \in B^*} p_1(w,u) * p_2(w,v) \qquad (6.62)$$

By using this technique we could learn the lexical strings in an unsupervised framework. We could then dispense with the morphological categories used here, since words with identical surface strings could have different lexical strings. There is a technical difficulty here – how to sum over all the exponentially many strings in $B^*$. This is quite close to some of the models used in bioinformatics, and some of the techniques used there might be suitable.

### 6.10.3   Complete paradigms

One of the areas of complexity in morphology is the study of complete inflectional paradigms. In many languages, the full range of morphosyntactic possibilities is patchily covered by available inflections. If one considers the full range of possible combinations of number, gender, person, and case features, possibly including agreement features with several other words or phrases, the number of different morpho-syntactic possibilities can grow exponentially. It is often the case that not every possible combination has a distinct phonological realisation. The interaction between the various possibilities is complex, and is completely ignored in the approach I use here, where I take each pair in isolation.

Moreover in morphologically rich languages, the number of examples for each individual pair may be extremely low. It is thus necessary to model the entire inflectional paradigm of a word class at the same time. Kornai (1992), discussing Hungarian, notes that in a corpus of 500,000 words, the most frequent noun occurs in only 51 of the 714 possible forms for nouns. He calculates that in order to saturate the paradigms, i.e. have every different possibility attested, of the first 30 nouns, would require a corpus of 33 billion words. Here however the various forms are produced by the composition of rather unchanging morphemes. We can model this in a straightforward way, since our models can perform modifications at specified parts of the stem. Schematically, if we have a word that is built up from say a root plus four morphemes that express, say, definite/indefinite, number, gender and case, where there are, let us suppose, two possibilities for each, giving $2^4$ possible inflected forms. So the input to the transduction would look something like `Root-Def-Sing-Masc-Nom`. We can then define four transductions, each of which would change one of these morphemes. So we would have a singular to plural transduction, a masculine to feminine transduction and so on. The singular to plural transduction would map `Root-Def-Sing-Masc-Nom` to `Root-Def-Plur-Masc-Nom`, and also `Root-Def-Sing-Masc-Acc` to `Root-Def-Plur-Masc-Acc` and so on. Each of the $2^4$ possibilities could be reached by a composition of the appropriate transductions, so instead of requiring $2^4$ transductions, we could get by with just 4.

The German noun system, on the other hand, has eight possible combinations of syntactic features; four cases, and two numbers. These eight possibilities are realised by at most four distinct forms. Each of the various morphological classes realises the eight possibilities with various transductions, many of which are the same. A straightforward implementation of this would have for each class a set of eight transducers, or perhaps seven since we can assume one of them will be the input to all of the others. This might be too complex to be learnable. In that case we could have a smaller set of transducers that could be shared between each of these class-specific transducers.

Modelling complete inflectional systems would also allow a more thorough evaluation. We can take words where the training algorithm has only had access to a subset of the eight possibilities, an *unsaturated* paradigm in Kornai's terminology, and ask it to produce the remaining ones. In many cases knowing only a few of the eight will enable a determination of the inflectional class.

## 6.11 Discussion

In summary, I have demonstrated a novel algorithm for learning finite-state transductions, that appears well suited to the task of learning morphology.

### 6.11.1 Comparison with other learning approaches

The algorithm presented here has a number of advantages compared to other techniques of learning morphology in a supervised setting. First, compared to Neural Network MLP techniques, (Rumelhart & McClelland, 1986a; Plunkett & Nakisa, 1997), it does not require the extensive rep-resentational engineering and tweaking that these models need, nor does it have the fixed length limits that these models require. In addition MLP models often show poor generalisation, whereas with appropriate smoothing PHMMs generalise very well. Furthermore, PHMMs can only learn a limited range of transductions, whereas the MLP models can learn implausible transductions such as reversals.

Secondly compared to ILP techniques, (Mooney & Califf, 1995; Manandhar, Dzeroski, & Erjavec, 1998; Muggleton, 1999), PHMMs do not require pre-specification of the range of possible transductions. In addition PHMMs can be used in unsupervised learning environments, and are robust in the presence of noise. Moreover it is not clear whether these ILP techniques can learn complex non-concatenative transductions such as the Arabic plural system. Thus the techniques presented here appears to have substantial advantages compared to other models.

However, strictly as an engineering solution to the problem of morphology, the models that this system produces have a number of disadvantages: the models for highly irregular words effectively memorise the input and output forms in an extremely inefficient form. It would probably be preferable to have a separate system for these, and use the PHMM for the regular and sub-regular processes.

### 6.11.2 Single versus Dual-route

With respect to the single route versus dual route controversy, the current work is of marginal relevance. Which of these two hypotheses is correct is an empirical question about human psychology, which will be settled by psychological evidence. Clearly this approach incorporates both irregular and regular inflections in a single system, which are learned successfully; this refutes the claim that single route learning models *cannot* learn morphology correctly. However it in no way establishes that humans use a similar system, or even if they did, that the two sorts of transduction, regular and irregular, are neurologically instantiated in a single system.

### 6.11.3 Conclusion

With regard to the overall argument and goals of this thesis, I feel these models present a satisfactory solution to the problem of learning morphology. Though they obviously have a strong innate bias to learn particular transductions, they are a general purpose learning algorithm for transductions; as a matter of historical fact, they are not domain specific, since they were originally introduced in the different domain of bio-informatics.

# Chapter 7

# Syntax Acquisition

---

## 7.1 Introduction

In this chapter I present an unsupervised algorithm for the induction of a grammar from tagged text. The unsupervised learning of syntax is one of the most difficult areas of natural language learning, and many researchers believe it to be impossible on *a priori* grounds. As discussed in Chapter 4 these arguments are of limited validity. The algorithm presented here learns from a sequence of tags. This is clearly inadequate; part of speech tags are too coarse to allow a completely effective syntax induction algorithm, since many syntactic constructions depend on quite specific properties of words. Nonetheless, as we shall see, using part of speech tags is both necessary in terms of efficiency and data sparseness, and sufficient, in terms of allowing learning of the basic phrase structure of English. I shall present results with two sets of tags; first, I shall use the tagged data from the British National Corpus (BNC) (Burnard, 1995; Aston & Burnard, 1998). Using a set of tags that have been assigned by linguists raises a number of issues, that I shall discuss below, but allows a cleaner exposition of the principles and techniques used. I shall then present some more informal results using the set of automatically assigned tags from Chapter 5; this will then be a completely unsupervised algorithm.

The rest of this chapter is organised as follows: I discuss in Section 7.2 previous work on the unsupervised learning of syntax; which I divide rather arbitrarily into three overlapping categories: likelihood-based, compression-based and distribution based. The next section, Section 7.3 discusses the Inside-Outside algorithm for training PCFGs, and presents the results of some simple experiments on using this on the small ATIS corpus, together with an analysis of why it doesn't work very well. In Section 7.4 I discuss how the technique of distributional clustering could be used as part of an algorithm. In Section 7.5 I present a mutual information (MI) criterion for identifying constituents, which I justify in a number of ways. Section 7.6 provides a mathematical justification for this criterion. Section 7.7 shows that empirically this criterion does in fact select constituents, and then in Section 7.8 I compare it to various other criteria that have been proposed. I show how this criterion can be incorporated in an MDL algorithm in Section 7.9 and evaluate it in Section 7.10. I then present some more informal results based on automatically derived tags in Section 7.11, and conclude with a discussion and some proposals for future work in Section 7.12.

## 7.2 Previous Work

There has been quite a lot of work on the unsupervised learning of syntax – too much for me to produce an exhaustive survey. I shall just choose some representative papers, except with the distributional techniques that are closest to my own, where I have tried to be more thorough. I shall ignore the large literature on the inference of particular classes of context-free languages, and restrict myself to the learning of natural languages or artificial approximations.

I will discuss this not chronologically but grouped into various types:

- Likelihood based.

- Compression based.

- Distribution based.

### 7.2.1 Likelihood-based

The first class of systems are likelihood-based: they select the maximum likelihood model using a PCFG. There have been a number of experiments using the EM algorithm – often called the inside-outside algorithm (IO) (Baker, 1979). Early experiments were performed by Lari and Young (1990), Briscoe and Waegner (1992) and a number of other researchers. Pereira and Schabes (1992), Carroll and Charniak (1992) and Charniak (1993) produced some rather discouraging research that seemed to indicate that the fact that the IO algorithm converged to a local optimum meant that it would almost always converge to a linguistically implausible grammar. Other researchers have tried to use more advanced search techniques such as genetic algorithms (Keller & Lutz, 1997) to avoid this problem. I will discuss this approach in more depth in Section 7.3.

### 7.2.2 Compression based

This line of work rather than using the ML criterion, uses an MDL criterion or some equivalent. This is an idea that has been around for some time, but has not really produced satisfactory results. Gerald Wolff in a series of papers over a number of years (Wolff, 1977, 1980, 1988, 1991) has argued for a notion of learning based on compression. Stolcke (1994) advocates the use of a Bayesian model selection criterion for HMMs and PCFGs but restricts his consideration to artificial languages. Chen (1995) presents an MDL based algorithm that for technical reasons is limited to regular rather than context-free languages. He presents results on a number of artifical languages that approximate English. He starts from a simple grammar and expands it; Kit (1998) advocates an MDL based measure for guiding a model that looks for frequent strings in a corpus; i.e. starts with a very large grammar and gradually compresses it. The problem with these techniques is that hypothesizing as constituents sequences of tags that occur together more frequently than would be expected (i.e. have high mutual information) does not work: in particular sequences such as verb preposition, or preposition article have high mutual information but are clearly not constituents. For example, in the Penn tree-bank, the sequence IN DT has pointwise mutual information 1.3675, and the sequence DT NN has MI 1.266. Thus using this technique to divide a sequence IN DT NN will give the intuitively wrong answer.

This MDL gain criterion is in some cases very closely related to the mutual information of the sequence itself under standard assumptions about optimal codes (Cover & Thomas, 1991). Suppose we have two symbols $x$ and $y$ that occur $n_x$ and $n_y$ times in a corpus of length $N$ and that the sequence $xy$ occurs $n_{xy}$ times. We could instead create a new symbol that represents $xy$, and rewrite the corpus using this abbreviation. Since we would use it $n_{xy}$ times, each symbol would require $\log N/n_{xy}$ nats. [1] The symbols $x$ and $y$ have codelengths of $\log N/n_x$ and $\log N/n_y$, so for each pair $xy$ that we rewrite, under reasonable approximations, we have a reduction in code length of

$$\Delta L \approx -\log N/n_{xy} + \log N/n_x + \log N/n_y$$
$$\approx \log \frac{p(xy)}{p(x)p(y)}$$

which is the point-wise mutual information between $x$ and $y$.

---

[1] A nat is $\log_2 e$ bits.

I therefore include here the work of Magerman and Marcus (1990); they use an algorithm that rather than proposing that constituents are those sequences that have high mutual information, proposes that the constituent boundaries are those places where a (generalised) mutual information measure dips, together with a specific distituent grammar that stipulates that a few crucial sequences cannot be constituents.

The MDL objective function itself is fine – the problem is that a greedy bottom up implementation of it will make errors by failing to note the long-distance dependencies that characterise natural language.

### 7.2.3   Distribution based

The third class of algorithm uses *distributional* evidence to identify constituent structure. The idea here is that sequences of words or tags that are generated by the same non-terminal will appear in similar contexts. As previously discussed one of the first attempts to learn syntax was an attempt using distributional clustering (Lamb, 1961). He works with a small data set of 5000 tokens, and uses an *ad hoc* measure, following a suggestion of Harris (1955), related to the conditional entropy to identify constituents. Clearly with these extreme limitations his results are poor, but the basic method is sound.

Brill and Marcus (1992) use a measure based on the KL divergence applied to distributional contexts to identify possible syntactic rules. A peculiarity of their approach is that they compare only sequences of two tags to sequences of a single tag. Since single tags are by definition constituents, this has the effect of biasing the algorithm to producing sequences that are in fact constituents. A weakness of this is that it means that the algorithm cannot hypothesise non-terminal sequences that do not correspond distributionally to a single non-terminal. This is not, I think cross-linguistically valid, and even in English there are certainly non-terminals, such as non-finite clauses, that we would want, but that are not distributionally equivalent to a single word. In addition they hypothesise a particular entropy criterion to filter out some common errors the algorithm produces. This is related to the criterion of Lamb (1961).

Finch et al. (1995) present some intriguing preliminary results showing how distributional clustering algorithms can be used to find sets of tag sequences that occur in similar contexts. Their techniques produce some linguistically plausible clusters, but many implausible ones, and they do not demonstrate a grammar induction algorithm. Nevertheless, they show that distributional clustering can work with syntactic constituents.

The work of Mori and Nagao (1995) is in a similar vein. They propose three hypotheses:

> 1. Part-of-speech sequences on the right-hand side of a rewriting rule are less constrained as to what part-of-speech preceded and follows them than non-constituent sequences.
> 2. Part-of-speech sequences directly derived from the same non-terminal symbol have similar environments.
> 3. The most suitable set of rewriting rules makes the greatest reduction of the corpus size.

They use these three principles in an algorithm that derives a context free grammar from sequences of tags from the Penn Tree-Bank. The algorithm clusters part-of-speech sequences together based on their similarity using Euclidean distance. In addition they use a criterion based on the ratio of delimiters that appear on either side of the POS sequence in question. Theeramunkong and Okumura (1997) use distributional clustering together with a bracketed corpus along the lines of Pereira and Schabes (1992). Klein and Manning (2001) present a pair of distributional learning algorithms one of which is very similar to the technique I use. I shall compare their work more fully at the end of this chapter.

All of these techniques use *local* distributional context. There are also two techniques that use whole sentence contexts. Adriaans (1999) presents EMILE, which initially used a form of supervision (it could query an oracle about the grammaticality of example sentences), but in later work (Adriaans, Trautwein, & Vervoort, 2000; van Zaanen & Adriaans, 2001) is modified to be completely unsupervised. A similar approach is presented by van Zaanen (2000): a very interesting algorithm called Alignment-Based Learning. Both of these techniques look for *minimal pairs*; a specific form of distributional learning, where the contexts are the rest of the sentence. That is to say, the algorithm looks for pairs of sentences that are identical except for a particular phrase being changed. Unfortunately, this requires a large number of examples of each sentence type, to guarantee that there will be suitable minimal pairs. Though both of these algorithms produce reasonable results on the ATIS corpus, which is a very small and repetitive corpus with a limited lexicon, and a very restricted set of syntactic structures, they do not seem to work so well on more realistic data. Adriaans et al. (2000) present some very limited results on the Bible, which only learns a few phrases, and they estimate (van Zaanen & Adriaans, 2001) that it would require 50 million sentences to learn a grammar for English. However, they work exclusively with words not tags, and is possible that suitable modifications might make their algorithms more practical.

## 7.3 Inside-Outside algorithm

The inside-outside algorithm is the name for the application of the EM algorithm to the estimation of the parameters for a PCFG. It was first presented by Baker (1979) and later applied by a number of researchers (Lari & Young, 1990; Pereira & Schabes, 1992). It can usefully be considered a generalisation of the forward-backward (Baum-Welch) algorithm for HMMs.

I won't give a full exposition of the inside-outside algorithm; the reader is referred to Manning and Schütze (1999) for a more detailed explanation. Just as with the forward-backward algorithm we define a dynamic programming trellis with two related sets of probabilities, the forward and backward probabilities, for the inside-outside algorithm we define the inside and outside probabilities. If we have a sentence of length $l$, with word $w_0$ in the span $(0, 1)$, and the final word $w_{l-1}$ in the span $(l-1, l)$, we define:

**Definition 7** $\beta_s(i, j)$, *the inside probability, to be the probability that starting from the non-terminal symbol s we will generate all of the words in the span* $(i, j)$.

**Definition 8** $\alpha_s(i, j)$ *the outside probability to be the probability that starting from the root symbol we will generate over the span* $(i, j)$ *the symbol s, and all of the words in the spans* $(0, i)$ *and* $(j, l)$.

We can calculate these recursively, first doing the inside probabilities and next the outside probabilities. Then in the EM algorithm we accumulate the expected number of times each rule $r \rightarrow s, t$ is taken by summing expressions like this:

| Iterations | NT + T | UR | UP | F-score | CB | 0 CB | $\leq$ 2 CB | -LP |
|---|---|---|---|---|---|---|---|---|
| 20 | 5 + 35 | 26.60 | 31.61 | 28.89 | 3.55 | 18.25 | 40.43 | 8159 |
| 100 | 5 + 35 | 25.27 | 30.03 | 27.45 | 3.85 | 17.17 | 37.39 | 7853 |
| 20 | 15 + 35 | 34.93 | 41.50 | 37.93 | 2.64 | 24.51 | 55.64 | 7152 |
| 100 | 15 + 35 | 35.38 | 42.03 | 38.42 | 2.53 | 22.36 | 59.03 | 6480 |
| 500 | 15 + 35 | 35.51 | 42.19 | 38.56 | 2.48 | 22.90 | 60.47 | 6427 |
| 20 | 35 + 35 | 36.22 | 43.03 | 39.33 | 2.57 | 19.50 | 58.32 | 6471 |
| 100 | Left 15 + 35 | 16.74 | 19.89 | 18.18 | 4.70 | 7.16 | 33.45 | 6550 |
| 100 | Right 15 + 35 | 37.31 | 44.33 | 40.52 | 2.39 | 24.87 | 62.61 | 6703 |

Table 7.1: Results of evaluation of inside-outside algorithm on tagged ATIS corpus. NT + T is the number of non-terminals and the number of terminals. The rules were of the form $NT \rightarrow (NT \cup T) \times (NT \cup T)$ UR is unlabelled recall, UP is unlabelled precision, CB is average number of crossing brackets, $\leq$ 2 CB is percentage with two or fewer crossing brackets, -LP is the negative log probability of the training data at the final iteration. Since the IO algorithm produces a binary tree the crossing bracket scores are rather poor.

$$\alpha_r(i,j) \, p(r \rightarrow s,t) \beta_s(i,k) \beta_t(k,j) \qquad (7.1)$$

So the expectation arises out of the interaction between the inside and outside probabilities.

The algorithm then consists of selecting a number of non-terminals, and the structure of the grammar; normally all rules are allowed, randomly initialising the parameters and running the algorithm to convergence.

I first experimented with using the tagged version of the ATIS corpus. This consists of 559 sentences, generally rather short and predominantly orders and questions. I used various parameter settings, with random initialisation of the models, produced the Viterbi parse of each sentence in the corpus and evaluated the results using the standard (unlabelled) PARSEVAL structural consistency metrics using the EVALB program. Table 7.1 summarises the results. Note first that since I use Chomsky normal form grammars, the model produces purely binary branching trees and thus the crossing bracket measures are poor compared to techniques that can produce flatter trees. First, the lower two rows of the table show the scores for two base-line models which are restricted to left and right branching. Note that the right branching model scores highly on the recall and precision metrics, but that the negative log likelihood of the left branching model is lower (better) than that of the right branching model, and that the right branching model has worse perplexity than several of the other models, in spite of its good score on the structural consistency metrics.

Regardless of these scores, what we are primarily interested in is the plausibility of the grammars. They aren't that bad – certainly they don't conform to the particular scheme used in the ATIS corpus, but that isn't necessarily bad. To give an idea of how well they perform, I selected two sentences, shown in Figures 7.1 and 7.2 where the IO algorithm performs poorly and well respectively.

Table 7.2 shows the 20 most frequent rules that rewrite two terminal symbols that are used in the Viterbi parse of the whole corpus. As can be seen while several of these rules are perfectly sensible there are quite a few that are not sensible such as IN DT and DT JJ, preposition determiner and determiner adjective respectively, although they do occasionally form constituents. This is not disastrous but it is not good enough; and we can identify two possible causes. Either we are using the wrong objective function, or we are using the wrong search algorithm.

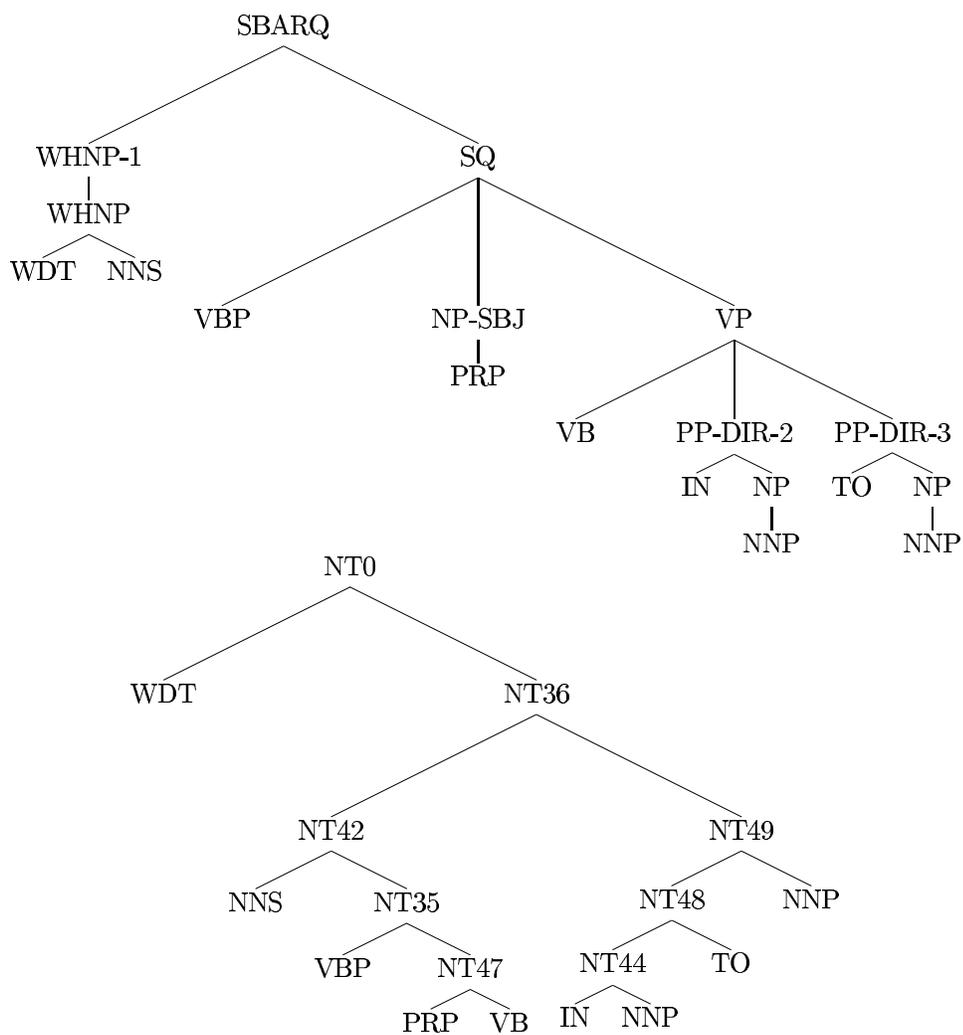

Figure 7.1: Sentence where the inside outside algorithm performs badly (below) compared to gold standard parse (above). Sentence is "What flights do you have from Milwaukee to Tampa".

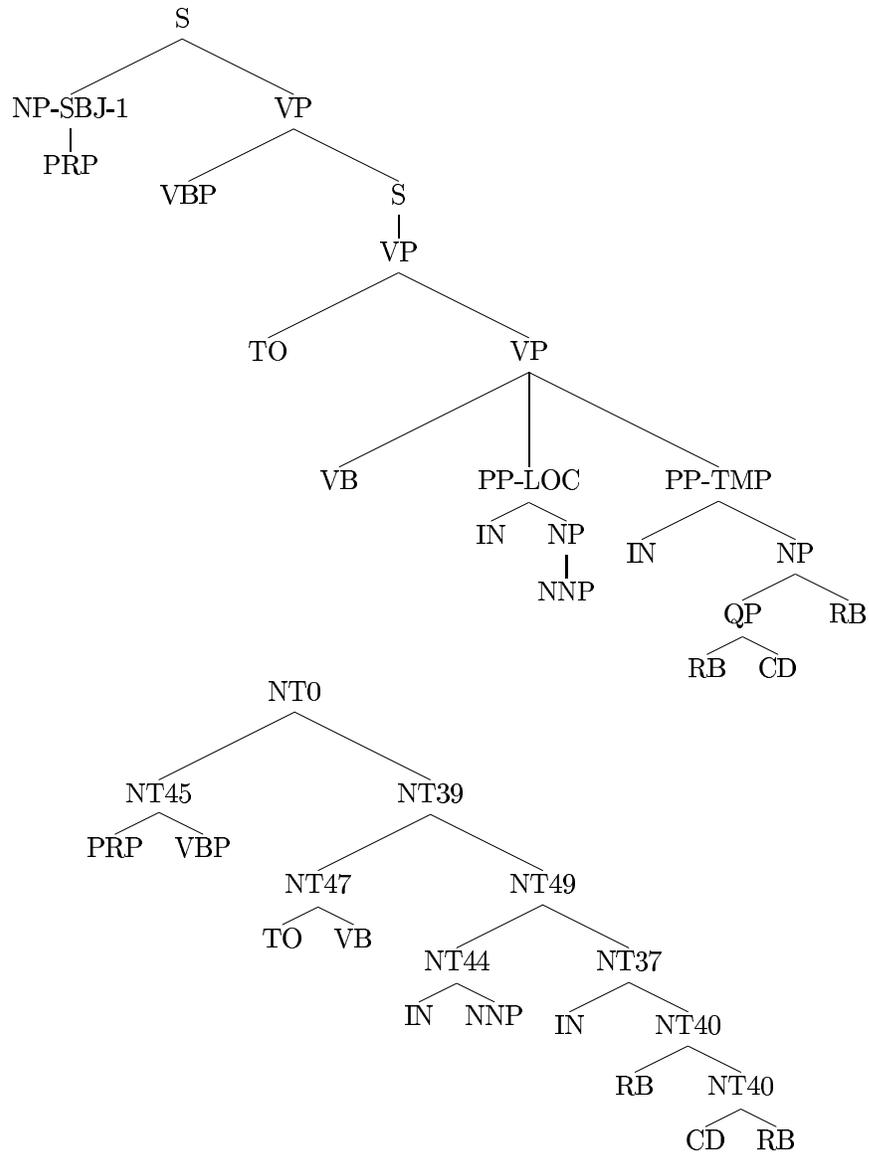

Figure 7.2: Sentence where the inside outside algorithm performs well (below) compared to gold standard parse (above). Sentence is "I need to arrive at Charlotte at around five p.m.".

| Count | Right Hand Side | Left Hand Side |
|-------|-----------------|----------------|
| 416 | NT44 | IN NNP |
| 100 | NT45 | VB PRP |
| 59 | NT47 | DT NNS |
| 45 | NT40 | DT NN |
| 40 | NT47 | DT NN |
| 38 | NT40 | CD RB |
| 32 | NT42 | NNS VBP |
| 31 | NT41 | IN DT |
| 30 | NT38 | MD VB |
| 29 | NT46 | JJ NN |
| 27 | NT47 | TO VB |
| 26 | NT45 | WP VBZ |
| 24 | NT46 | NN NN |
| 24 | NT45 | PRP VBP |
| 22 | NT35 | VBP EX |
| 19 | NT43 | CD CD |
| 17 | NT41 | DT JJ |
| 15 | NT41 | DT JJS |
| 15 | NT39 | DT NN |
| 14 | NT43 | NNP NNP |

Table 7.2: 20 most frequent rules with two terminals.

### 7.3.1 Objective Function

I have assumed up to now that the most likely model will be linguistically plausible, and that therefore trying to find the maximum likelihood model is a good plan. As others have noted (de Marcken, 1999; Klein & Manning, 2001) this is not necessarily the case. Klein and Manning (2001) note that traditional arguments for phrase structure have nothing to do with the independence assumptions of the PCFG, and that

> it could be that the ML and linguistic criteria align, but in practice they do not always seem to, and one should not expect that, by maximizing the former, one will also maximise the latter.

I agree up to a point; but if the data were generated by a context-free grammar, then asymptotically, as the amount of data increases, the Maximum Likelihood model is guaranteed to give you the right answer. There are a number of questions this argument raises: first, is the size of the ATIS corpus large enough to guarantee that the ML model is in some sense right? Secondly, though the set of grammatical sentences of English, however this is defined, is probably weakly context-free, is the distribution over actual English sentences stochastically context free? That is to say, are there statistical dependencies in English that cannot be captured, even in principle by a PCFG? Thirdly, even if English is not stochastically context-free, does that mean that the ML model is wrong? Fourthly, does the structure of this hypothetical optimal PCFG correspond to a linguistic grammar? I have few answers to these questions, except to say that grammars written by linguists disagree frequently.

The experiments here are inconclusive, so there is still an interesting open question: with the ATIS corpus, is the most likely model of a given size (say 15 non-terminals) linguistically plausible? There is an indication in Pereira and Schabes (1992) that this is not the case. They trained two models using the IO algorithm, one on a partially bracketed version of the ATIS corpus, and one on the raw corpus. The model trained on the raw corpus had *lower* perplexity (more likely) that the one trained on the bracketed corpus, though its linguistic plausibility as measured by structural consistency measures with the bracketed corpus was much worse.

### 7.3.2 Search Algorithm

The alternative explanation is that the problem lies with the search algorithm. As a number of researchers have pointed out (Carroll & Charniak, 1992; Charniak, 1993; de Marcken, 1999) the inside-outside algorithm is highly sensitive to initial conditions, as it only converges to a local optimum that may globally be very sub-optimal. Though there are a huge number of different techniques of non-convex optimisation that could be applied in this case, the computational burdens appear to make them impractical at the present.

Klein and Manning (p.c.) claim that part of the problem is that the outside probabilities are very diffuse at the beginning of the algorithm and thus the IO algorithm is driven predominantly by the inside probabilities: i.e. it tends just to build sequences with high mutual information like the other algorithms we saw earlier.

So there are some possible solutions:

- Try a richer formalism that takes account of lexical dependencies.

- Try a flatter formalism where the outside probabilities give more help.

- Don't try and maximise the likelihood directly.

- Try to find some substitute for the outside probabilities.

There are other problems with the algorithm. The ATIS corpus is an unreasonably easy corpus; it is very small and very repetitive, and consists largely of short simple sentences. This hides the fact that the IO algorithm is very slow. In fact the algorithm is cubic in the length of the sentences, and linear in the number of rules of the grammar which is cubic in the number of non-terminals. My solution here has therefore been to use distributional techniques.

## 7.4 Distributional Clustering

My approach then is not to directly maximise the log-likelihood. Distributional clustering has been used widely in a number of NLP areas – see for example (Brown et al., 1992; Pereira et al., 1993) as discussed in Chapter 5. In these approaches, one considers the set of contexts that a word occurs in; words that are similar will occur in similar contexts. This can form the basis for an algorithm that forms syntactic or semantic categories, by clustering based on the similarity or distance between the contexts. In this work we are interested in the behavior of strings of words or tag sequences. If two sequences of tags occur mostly forming the same non-terminal, then we would expect the contexts that those strings occur in to be similar. So for example, we would often expect to find the string "the big cat" either at the beginning of the sentence and followed by a finite verb, or immediately after a finite verb and before a preposition or many other possibilities, and we would expect other "noun phrase" sequences to have similar distributions. If we clustered sequences according to their distributions we would thus expect to find clusters corresponding to various syntactic constituents. However, we do not know in advance which sequences are constituents and which are not, so we will have to cluster all frequent sequences. We will thus also have clusters corresponding to frequent sequences that are not constituents such as "to the", or "ran up the high". Moreover sequences of words and even sequences of tags are quite sparse so we may not have enough counts of them to estimate their distributions accurately.

In this work I will use as the distributional context the terminal symbols occurring immediately before and immediately after. The distribution is thus a distribution over ordered pairs of terminal symbols – words or in this case syntactic tags. This is the most local distributional context; we could also define the global distributional context which is the distribution over all sentences with holes in.

Clearly in the infinite data limit, all unambiguous sequences derived from the same non-terminal will have exactly the same distribution; it is not necessarily the case that different non-terminals will have different *local* distributions. Clearly if we had two non-terminals whose *global* distributions were the same we could merge the two together without causing any change to the probability distribution function (pdf); also if two non-terminals produce the same sequences of symbols with the same probabilities we could merge them without any change in the predictions of the model. So we will have problems with this algorithm if we have two non-terminals with the same local distributions but different global distributions. A simple context-free language that has this property would be something like

$$L = \{acWaW^Rc \cup bcWbW^Rc | W \in \{d, e\}^*\} \tag{7.2}$$

where we have a palindrome language with a symbol embedded in the middle (a or b) that must agree with a symbol outside.

I assume we have an additional terminal symbol that represents a sentence boundary. If there are $k$ symbols then each distribution has $k^2$ parameters. In the data sets used here, there are 77 tags giving 5929 parameters for each distribution. Note that I am not making the independence assumption that the distribution of words before is independent of the distribution of words after, which would reduce the number of parameters to $2k$. In fact, it turns out that the divergence from this assumption is a key quantity.

The data set for these preliminary results consisted of 12 million words of the British National Corpus, tagged according to the CLAWS-5 tag set, with punctuation removed. There are 76 tags; I introduced an additional tag to mark sentence boundaries. I operate exclusively with tags, ignoring the actual words. My initial experiment clustered all of the tag sequences in the corpus that occurred more than 5000 times, of which there were 753, using the $k$-means algorithm with the $L_1$-norm or city-block metric applied to the context distributions. Thus sequences of tags will end up in the same cluster if their context distributions are similar; that is to say if they appear predominantly in similar contexts. I chose the cutoff of 5000 counts to be of the same order as the number of parameters of the distribution, and chose the number of clusters to be 100. To identify the frequent sequences, and to calculate their distributions I used the standard technique of suffix arrays (Gusfield, 1997), which allows rapid location of all occurrences of a desired substring.

| Cluster 1 | Cluster 2 |
|---|---|
| `ATO AJO NN0` | `AJO AJO` |
| `ATO AJO NN1` | `AJO CJC AJO` |
| `ATO AJO NN2` | `AVO AJO` |
| `ATO AVO AJO NN1` | `AVO AVO AJO` |
| `ATO NN0` | `ORD` |
| `ATO NN1 PRP ATO NN1` | |
| `ATO NN1` | |

Table 7.3: Some of the more frequent sequences in two good clusters

| Cluster 1 | Cluster 2 |
|---|---|
| `AJO NN1 ATO` | `CJC ATO AJO` |
| `AJO NN1 PRF ATO` | `CJC ATO` |
| `AJO NN1 PRP ATO` | `CJC CRD` |
| `NN1 ATO AJO` | `CJC DPS` |
| `NN1 ATO` | `CJC PRP ATO` |
| `NN1 CJC AJO` | `PRF AJO` |

Table 7.4: Some of the sequencess in two bad clusters

As expected, the results of the clustering showed clear clusters corresponding to syntactic constituents, two of which are shown in Table 7.3. Of course, since we are clustering all of the frequent sequences in the corpus we will also have clusters corresponding to parts of constituents, as can be seen in Table 7.4. We obviously would not want to hypothesise these as constituents: we therefore need some criterion for filtering out these spurious candidates.

The problem is how to identify which ones are constituents, and how to incorporate this into a complete system. In the next section I discuss an approach to this problem.

## 7.5 Mutual Information Criterion for Constituents

The problem with this approach is that there will be many clusters that do not correspond to constituents in the traditional sense. Consider, for example the common string of tags "PREP DET". This has quite a characteristic distribution, and will form large coherent clusters, consisting of prepositional phrases missing the final noun. Moreover there is as pointed out before, high mutual information between the two symbols. Clearly, we do not want to hypothesise this as a constituent in the early phases of the algorithm. We need a criterion which will filter out these non-constituents. In this section I will present such a criterion and motivate it in three ways, intuitively, mathematically and empirically.

The criterion is this: a constituent will exhibit high mutual information between the symbol that occurs before it, and the symbol occurring after it. So it is precisely the deviation from the independence assumption mentioned in the previous section that is the criterion since the mutual information is equal to the Kullback-Leibler divergence between the joint distribution and the product of the two marginal distributions – ie the joint distribution modelled as if the two variables were independent.

Consider a true constituent such as a noun phrase. If a noun phrase occurs at the beginning of a sentence, it is likely to have a sentence boundary before it, and a finite verb after it. If it occurs in a prepositional phrase it is likely to have a preposition before it and perhaps a sentence boundary after it. It is much less likely to have a sentence boundary both before and after it. Thus the symbol that occurs before it is highly correlated with the symbol that occurs after it; i.e. it has high MI. On the other hand if we have a spurious constituent such as "PREP DET" there will be essentially no correlation between what occurs before and after. On the right we must have a completion of the noun phrase. This is completely independent of the context that the prepositional phrase appears in, and thus has no relation to the symbols that appear before the phrase. Intuitively, we are looking for sequences that allow us to capture long range dependencies. More formally, we can derive limits on this mutual information derived from the parameters of a PCFG generating the data.

## 7.6   Mathematical Justification for the Criterion

We can gain some insight into the significance of the MI criterion by analysing it within the framework of SCFGs. We are interested in looking at the properties of the two-dimensional distributions of each non-terminal. The terminals are the part of speech tags of which there are $T$. For each terminal or non-terminal symbol $X$ we define four distributions, $L(X), P(X), S(X), R(X)$, over $T$ or equivalently $T$-dimensional vectors. Two of these, $P(X)$ and $S(X)$ are just the prefix and suffix probability distributions for the symbol (Stolcke, 1995): the probabilities that the string derived from $X$ begins (or ends) with a particular tag. The other two $L(X), R(X)$ for left distribution and right distribution, are the distributions of the symbols before and after the non-terminal. Clearly if $X$ is a *terminal* symbol, the strings derived from it are all of length 1, and thus begin and end with $X$, giving $P(X)$ and $S(X)$ a very simple form.

If we consider each non-terminal $N$ in a SCFG, we can associate with it two random variables which we can call the *internal* and *external* variables. The internal random variable is the more familiar and ranges over the set of rules expanding that non-terminal. The external random variable, $Z_N$, is defined as the context in which the non-terminal appears. Every non-root occurrence of a non-terminal in a tree will be generated by some rule $r$, that it appears on the right hand side of. We can represent this as $(r, i)$ where $r$ is the rule, and $i$ is the index saying where in the right hand side it occurs. The index is necessary since the same non-terminal symbol might occur more than once on the right hand side of the same rule. So for each $N$, $Z_N$ can take only those values of $(r, i)$ where $N$ is the $i$th symbol on the right hand side of $r$.

The independence assumptions of the SCFG imply that the internal and external variables are independent, i.e. have zero mutual information. This enables us to decompose the context distribution into a linear combination of the set of marginal distributions we defined earlier.

Let us examine the context distribution of all occurrences of a non-terminal $N$ with a particular value of $Z_N$. We can distinguish three situations: the non-terminal could appear at the beginning, middle or end of the right hand side. If it occurs at the beginning of a rule $r$ with left hand side $X$, and the rule is $X \rightarrow NY \ldots$ then the terminal symbol that appears before $N$ will be distributed exactly according to the symbol that occurs before $X$, i.e. $L(N) = L(X)$. The non-terminal symbol that occurs after $N$ will be distributed according to the symbol that occurs at the beginning of the symbol that occurs after $N$ in the right hand side of the rule, so $R(N) = P(Y)$. By the independence assumption, the joint distribution is just the product of the two marginals.

$$D(N|Z_N = (r, 1)) = L(X) \times P(Y) \tag{7.3}$$

Similarly if it occurs at the end of a rule $X \rightarrow \ldots WN$ we can write it as

$$D(N|Z_N = (r, |r|)) = S(W) \times R(X) \tag{7.4}$$

and if it occurs in the middle of a rule $X \rightarrow \ldots WNY \ldots$ we can write it as

$$D(N|Z_N = (r, i)) = S(W) \times P(Y) \tag{7.5}$$

The total distribution of $N$ will be the normalised expectation of these three types with respect to $P(Z_N)$. Each of these distributions will have zero mutual information, and the mutual information of the linear combination will be less than or equal to the entropy of the variable combining them, $H(Z_N)$. To simplify the notation if we have a linear combination of independent distributions thus:

$$P(X = x, Y = y) = \sum_i \alpha_i p_i(x) q_i(y) \tag{7.6}$$

where each of the individual distributions if of the form $p_i(x)q_i(y)$ and thus has zero mutual information, and the $\alpha_i$s are the mixing parameters, then using Jensen's inequality we can prove that

$$I(X;Y) \leq \sum -\alpha_i \log \alpha_i \tag{7.7}$$

with equality when the $p_i$'s and the $q_i$s are sufficiently distinct. For example, suppose that each of the $p_i$ distributions are completely different, and similarly for the $q_i$ distributions. Then if you know the value of the symbol on the left, then that will completely identify which of the distributions you are in, and this will reduce the entropy of the right distributions. Recall that $I(X;Y) = H(Y) - H(Y|X)$, so the MI is equal to the reduction in entropy of $Y$ when you know $X$.

Therefore

$$MI(N) \leq H(Z_N) \tag{7.8}$$

Thus a non-terminal that appears always in the same position on the right hand side of a particular rule, will have zero MI, whereas a non-terminal that appears on the right hand side of a variety of different rules will, or rather may, have high MI.

This is of limited direct utility, since we do not know which are the non-terminals and which are other strings, but this establishes some circumstances under which the approach won't work.

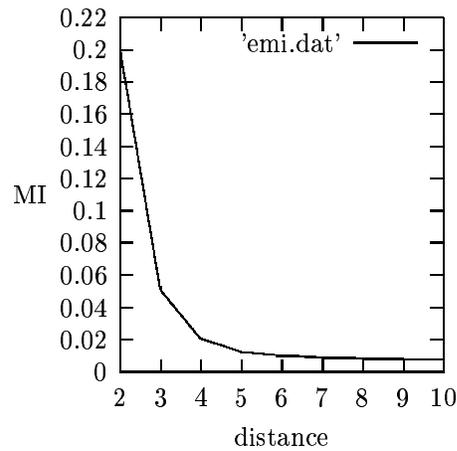

Figure 7.3: Graph of expected MI against distance.

Some of these are constraints on the form of the grammar, namely that no non-terminal can appear in just a single place on the right hand side of a single rule. Others are more substantive constraints on the sort of languages that can be learned.

## 7.7 Empirical Justification for the Criterion

To implement this, we need some way of deciding a threshold which will divide the sheep from the goats. A simple fixed threshold is undesirable for a number of reasons. One problem with the current approach is that the maximum likelihood estimator of the mutual information is biased, and tends to over-estimate the mutual information with sparse data (Li, 1990). A second problem is that there is a "natural" amount of mutual information present between any two symbols that are close to each other, that decreases as the symbols get further apart. Figure 7.3 shows a graph of how the distance between two symbols affects the MI between them. Thus if we have a sequence of length 2, the symbols before and after it will have a distance of 3, and we would expect to have a MI of 0.05. If it has more than this, we might hypothesise it as a constituent; if it has less, we discard it. In term of the graph, we can say that if it falls above the line, we accept it, and if it is below the line we reject it.

In practice we want to measure the MI of the clusters, since we will have many more counts, and that will make the MI estimate more accurate. We therefore compute the weighted average of this expected MI, according to the lengths of all the sequences in the clusters, and use that as the criterion. Table 7.5 shows how this criterion separates valid from invalid clusters. It eliminated 55 out of 100 clusters

In Table 7.5, we can verify this empirically: this criterion does in fact filter out the undesirable sequences. Clearly this is a powerful technique for identifying constituents.

Figure 7.4 shows a graph of actual versus expected MI. The dividing line shows the criterion I apply; I have plotted the four good clusters and the four bad clusters on it. These clusters are from a different run of the algorithm: I selected the clusters that contained the eight examples in Table 7.5. Their classification as good or bad remained the same in both runs of the algorithm, and in general the classifications appeared quite robust.

| Cluster | Actual MI | Exp. MI | Valid |
|---|---|---|---|
| `ATO NN1` | 0.11 | 0.04 | Yes |
| `ATO NPO NPO` | 0.13 | 0.02 | Yes |
| `PRP ATO NN1` | 0.06 | 0.02 | Yes |
| `AVO AJO` | 0.27 | 0.1 | Yes |
| `NN1 ATO` | 0.008 | 0.02 | No |
| `ATO AJO` | 0.02 | 0.03 | No |
| `VBI ATO` | 0.01 | 0.02 | No |
| `PRP ATO` | 0.01 | 0.03 | No |

Table 7.5: Four valid clusters where the actual MI is greater than the expected MI, and four invalid clusters which fail the test. The four invalid clusters clearly are not constituents according to traditional criteria.

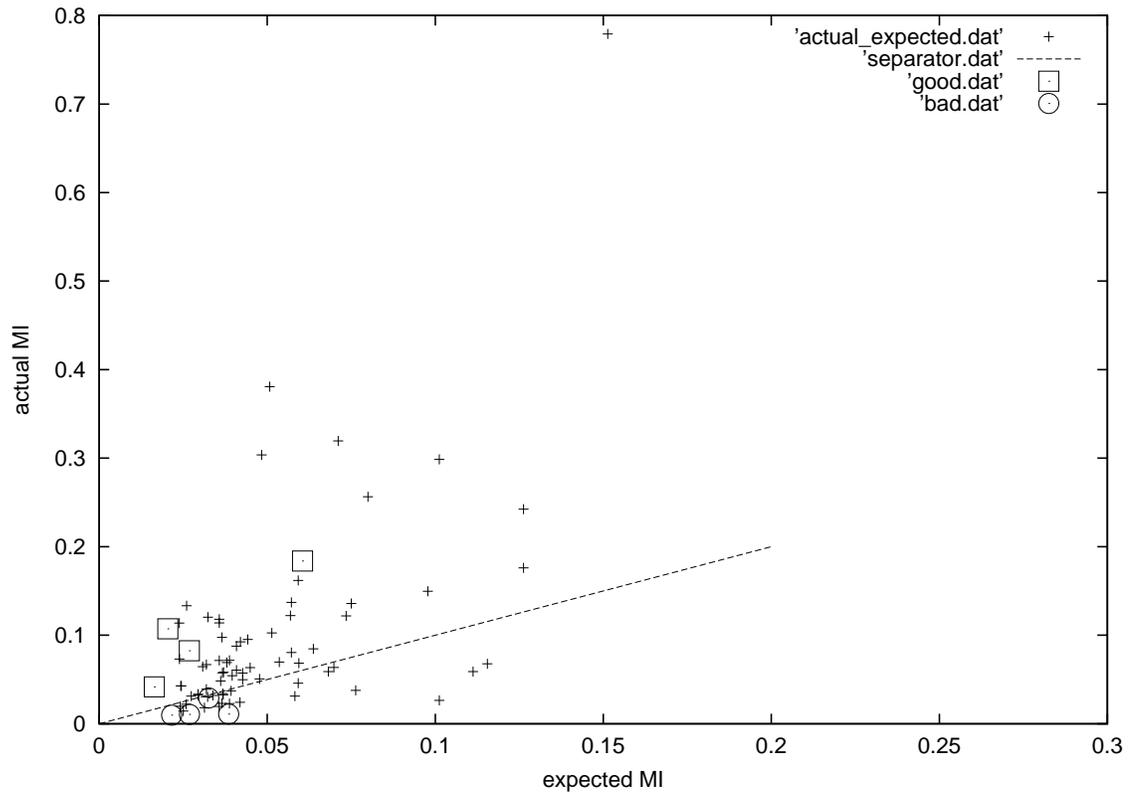

Figure 7.4: Graph of expected MI against actual MI.

## 7.8    Comparison to other criteria

As previously discussed, purely internal (inside) properties of the tag sequence are insufficient to determine whether that sequence is likely to be a constituent. I will now compare various other criteria, based on external properties of the sequence, that have been proposed by other researchers.

Lamb (1961) proposes a criterion he calls the token/neighbour ratio. This is the ratio between the number of tokens in the corpus and the number of distinct types that appear on the right or left of the item in question; the reciprocal of this quantity can be thought of as an approximation to the conditional entropy of the symbols before or after the item. As he puts it (p. 680)

> The highest T/N ratios [i.e. lowest conditional entropy] identify the points of maximum restriction on freedom of combination, insofar as such identification can be made without prior information about the structure of the language.

This is related to a proposal of Harris (1955).

Brill (1992) suggests that given a sequence of tags $tag_x, tag_y$ we should only consider this as a constituent if the conditional entropy of the tag that follows the sequence $tag_x, tag_y$ is greater than the conditional entropy of the tag that follows $tag_x$. The idea here is that at a constituent boundary the uncertainty goes up. Similarly, Mori and Nagao (1995) require that the conditional entropy of the tag before and the tag after are above a certain threshold, that they manually select to get a good result. The problem with both of these criteria is that they depend very much on the particular tags that are used, and their entropic properties. Note that both of these criteria change if we split tags into a finer-grained set of tags, or group tags together. Thus it is unlikely that they will work either with another, very different set of tags, and still less likely that they will work cross-linguistically.

Klein and Manning (2001) propose using a measure based on the entropy of the context distributions: They argue that the contexts that constituents occur in can vary more widely than those that non constituents can occur in. They recognise the fact that this measure is highly dependent on the tag set used, and propose two rather *ad hoc* scaling algorithms to compensate for this problem and a sparse data problem.

The technique I propose shares some of the same insights; namely that the distributions of non-constituents are constrained in some way, but I characterise this in a different way that is more abstract from the details of the entropy, and is more or less independent of the details of the tag set. In particular, if we take a particular tag and split it into two, assigning the occurrences of the original tag at random to the two new tags, then this will change the raw scores of the entropy, but won't change the mutual information measure. For example suppose that the tag DT is always followed by NN. Then the right conditional entropy of any tag sequence that ends in DT will be zero, and the conditional entropy criterion will say that none of these tag sequences are constituents. If we split the NN tag into a large number of separate classes based on some syntactic or morphological criteria, then the conditional entropy will go up, and these tag sequences will no longer be ruled out. The mutual information will remain unchanged, except for data sparseness effects. In fact the MI between the tags before and after is an estimate of the MI between the words before and after, just as we saw in Equation 5.3 earlier. This won't always be the case though: if we have a sequence, Determiner Adjective, then it will often appear in the context $< Verb, Noun >$. If we split these tags into *semantic* classes then we will see that the MI will increase; for example if a verb that means 'eat' appear before this sequence, then a noun that means 'food' is more likely to appear after it. If we derive the classes based on local distributional evidence as in Chapter 5, or if they are derived based on traditional syntactic criteria this should not happen.

I will now try to describe the relationship more formally. We can define the joint distribution of the tags before and after as $p_{XY}$, and the distributions of the symbols before and after, i.e. the marginal distributions, as $p_X$ and $p_Y$. I will use $H$ for the entropy. So Mori and Nagao (1995)'s criterion is that $H(p_X)$ and $H(p_y)$ must both be high, and Klein and Manning (2001)'s criterion is that $H(p_{XY})$ must be high.

The MI criterion I use can be written as

$$I(X;Y) = H(p_X) + H(p_Y) - H(p_{XY}) \qquad (7.9)$$

So we can see that high mutual information means that the entropy of the joint distribution will be substantially less that it would be under an independence assumption; so in effect the MI criterion directly measures how constrained it is.

### 7.8.1 Palindrome language

It can also help to see the relationship, if one considers an artifical example – the palindrome language. Consider the palindrome language of all odd-length palindromes over two terminal symbols *a* and *b*, generated by the following PCFG:

$$0.4 : S \rightarrow aSa$$
$$0.4 : S \rightarrow bSb$$
$$0.1 : S \rightarrow a$$
$$0.1 : S \rightarrow b$$

The constituents will all be themselves palindromes, but not all occurrences of substrings that are palindromes will in fact be constituents – consider the string *aaaaa* where the first occurrence of *aaa* is not a constituent. Nonetheless, most of the occurrences of a particular palindrome will be constituents, and with the constituents there will be very high mutual information between the symbol before and the symbol after – they must be the same, either *a*, *b* or a sentence boundary. This simple language, that has been studied before in the context of unsupervised language acquisition (Pereira & Schabes, 1992; Keller & Lutz, 1997), thus provides a good set of examples to compare these criteria.

I generated 10000 strings from this language, and calculated the various quantities discussed above using ML estimates on various strings that I chose rather arbitrarily: the results are summarised in Table 7.6. I divided the table into three sections; first we have palindromes of odd length, secondly odd-length strings that are not palindromes, and thirdly some even-length strings. Since the probabilities of the PCFG are symmetric with regard to swapping *a* and *b* and reversing the string, these strings cover most of the possibilities of length 3 and 5. Note first of all that the differences in the entropy $H(p_{XY})$ are rather small in absolute terms, and misleading with respect to the two length-7 strings – the palindrome has lower entropy than the non palindrome. The marginal entropy fares slightly better, though the differences are again small. However, the mutual information criterion shows sizable differences when comparing strings of the same length together. Note finally that with the even length strings, if one uses a Chomsky normal form grammar, the two strings *aa* and *aaaa* will also often be constituents, whereas the string *aabb* will never be one.

| String | $H(p_X)$ | $H(p_Y)$ | $H(p_{XY})$ | $I(X;Y)$ |
|--------|----------|----------|-------------|----------|
| aaa | 1.081820 | 1.081820 | 1.981995 | 0.024931 |
| aba | 1.084536 | 1.084536 | 1.959177 | 0.025950 |
| aaaaa | 1.092265 | 1.092265 | 1.941428 | 0.098320 |
| aabaa | 1.094171 | 1.094171 | 1.898402 | 0.143882 |
| abbba | 1.090961 | 1.090961 | 1.889518 | 0.154027 |
| ababa | 1.082768 | 1.082768 | 1.939730 | 0.079296 |
| aababaa | 1.088019 | 1.088019 | 1.642225 | 0.343747 |
| aab | 1.078490 | 1.083498 | 1.895160 | 0.014412 |
| aaaab | 1.043055 | 1.089997 | 1.867315 | 0.017028 |
| aaabb | 1.057471 | 1.068674 | 1.900353 | 0.016466 |
| baabb | 1.078258 | 1.051931 | 1.871374 | 0.014028 |
| babbb | 1.063206 | 1.079690 | 1.869075 | 0.013831 |
| abbabbb | 1.084602 | 1.063020 | 1.828712 | 0.020685 |
| aa | 1.082876 | 1.082876 | 1.905240 | 0.045166 |
| ab | 1.076238 | 1.082951 | 1.907760 | 0.015266 |
| aaaa | 1.086664 | 1.086664 | 1.886326 | 0.071863 |
| aabb | 1.079312 | 1.056195 | 1.874459 | 0.012771 |

Table 7.6: Comparisons of different criteria, when applied to 10000 string sample from the odd palindrome language. All criteria claim that higher values are related to constituenthood.

This is a rather informal test, but it gives an idea of the sort of circumstances under which the criteria I propose perform well. First of all since it is tag-set invariant it has a better chance of being cross-linguistically valid and valid across different tag sets. Secondly, since it is related to the entropy of the external syntax, there is more possibility of integrating it directly into an induction algorithm rather than using it rather crudely as at present. Thirdly, since it is related to a deep property of the stochastic process, it is maybe not too domain-specific – rather than just being something tuned to produce the desired effect, it is actually quite principled. Regardless of how these various techniques compare in formal elegance, it appears that there are a number of criteria that suffice to separate constituents from non-constituents.

## 7.9   Complete System

This technique can be incorporated into a grammar induction algorithm. We use the clustering algorithm to identify sets of sequences that can be derived from a single non-terminal. The MI cri-terion allows us to find the right places to cut the sentences up; we look for sequences where there are interesting long-range dependencies. Given these potential sequences, we can then hypothesise sets of rules with the same right hand side. This naturally suggests a minimum description length (MDL) or Bayesian approach (Stolcke, 1994; Chen, 1995). Starting with the maximum likelihood grammar, which has one rule for each sentence type in the corpus, and a single non-terminal, at each iteration we cluster all frequent strings, and filter according to the MI criterion discussed above.

Clearly with a corpus of this size, reparsing the whole corpus at each iteration is not possible. However starting from the maximum likelihood grammar means that the Viterbi approximation is very accurate, and remains reliable for some time.

At each iteration, we greedily select the cluster that will give the best immediate reduction in description length, calculated according to a theoretically optimal code. We add a new non-terminal with rules for each sequence in the cluster. If there is a sequence of length 1 with a non-terminal in it, then instead of adding a new non-terminal, we add rules expanding that old non-terminal. Thus, if we have a cluster which consists of the three sequences `NP`, `NP PRP NP` and `NP PRF NP` we would merely add the two rules `NP -> NP PRP NP` and `NP -> NP PRF NP`, rather than three rules with a new non-terminal on the left hand side. This allows the algorithm to learn recursive rules, and thus context-free grammars.

We then perform a partial parse of all the sentences in the corpus, and for each sentence select the path through the chart that provides the shortest description length, using standard dynamic programming techniques. This greedy algorithm is not ideal, but appears to be unavoidable given the computational complexity. Following this, we aggregate rules with the same right hand sides and repeat the operation.

Since the algorithm only considers strings whose frequency is above a fixed threshold, the application of a rule in rewriting the corpus will often result in a large number of strings being rewritten so that they are the same, thus creating a new sequence that will be above the threshold. Then at the next iteration, this sequence will be examined by the algorithm. Thus the algorithm progressively probes deeper into the structure of the corpus as syntactic variation is removed by the partial parse of low level constituents.

Singleton rules require special treatment; I have experimented with various different options, without finding an ideal solution. The results presented here use singleton rules, but they are only applied when the result is necessary for the application of a further rule. This is a natural consequence of the shortest description length choice for the partial parse: using a singleton rule will in general increase the description length of a path using it.

I ran the algorithm for 40 iterations. Beyond this point the algorithm appeared to stop producing plausible constituents. Part of the problem is to do with sparseness: it requires a large number of samples of each string to estimate the distributions reliably. Table 7.7 shows an outline of the algorithm.

There are a number of technical problems that must be solved before this algorithm can be implemented. Estimating the MI of the symbols before and after from the sparse counts available is again difficult. The estimate made by calculating the MI of the ML estimate tends to overstate the MI – (Li, 1990); there are more sophisticated techniques available (Wolf & Wolpert, 1992,

Initialise grammar to ML
**repeat**
    Gather all frequent strings
    Calculate distributions
    Cluster using $k$-means
    **for all** $c$ in the set of clusters **do**
        **if** $c$ satisfies MI criterion **then**
            Calculate MDL gain
        **end if**
    **end for**
    Select cluster that gives greatest reduction in description length
    Add rules corresponding to all the sequences in the cluster
    Parse
    Remove duplicates
**until** The grammar is sufficiently small

Table 7.7: Outline of the grammar induction algorithm

1993).

Sometimes there are clusters that are a combination of two or more rather different subclusters. In this case, each of the sequences may have very low MI, but the average of the distribution may have high MI, because of the different subclusters. In this case, the average MI will be greater than the actual MI. Normally because of the sparseness the MI calculation overestimates the MI, and thus the MI of each of the sequences will tend to be higher than the MI of the cluster, which is of course less sparse.

## 7.10 Evaluation

Evaluation of unsupervised algorithms is difficult. One evaluation scheme that has been used is to compare the constituent structures produced by the grammar induction algorithm against a treebank, and use PARSEVAL scoring metrics, as advocated by (van Zaanen & Adriaans, 2001): i.e. use exactly the same evaluation as is used for *supervised* learning schemes. This proposal fails to take account of the fact that the annotation scheme used in any corpus, does not reflect some theory-independent reality, but is the product of various more or less arbitrary decisions by the annotators (Carroll, Briscoe, & Sanfilippo, 1998). Given a particular annotation scheme, the structures in the corpus are not arbitrary, but the choice of annotation scheme inevitably is. Thus expecting an unsupervised algorithm to converge on one particular annotation scheme out of many possible ones seems overly onerous.

It is at this point that one must question what the point of syntactic structure is: it is not an end in itself, but a precursor to semantics. We need to have syntactic structure so we can abstract over it when we learn the semantic relationships between words. Seen in this context, the suggestion of evaluation based on dependency relationships amongst words (Carroll et al., 1998) seems eminently sensible.

With unsupervised algorithms, there are two aspects to the evaluation; first how good the annotation scheme is, and secondly how good the parsing algorithm is – i.e. how accurately the algorithm assigns the structures. Since we have a very trivial, non-lexicalised, parser here I shall focus on evaluating the sort of structures that are produced, rather than trying to evaluate how well the parser works. To facilitate comparison with other techniques, I shall also present an evaluation on the ATIS corpus.

| Symbol | Description | Number of rules | Most Frequent |
|--------|-------------|-----------------|---------------|
| NP | Noun Phrase | 107 | `AT0 NN1` |
| AVP | Adverb Phrase | 6 | `AV0 AV0` |
| PP | Prep. Phrase | 47 | `PRP NP` |
| S | Clause | 19 | `PNP VVD NP` |
| XPCONJ | Phrase and Conj. | 5 | `PP CJC` |
| N-BAR | | 121 | `AJ0 NN1` |
| S-SUB | Subordinate Clause ? | 58 | `S-SUB PP` |
| NT-NP0AV0 | | 3 | `PNP AV0` |
| NT-VHBVBN | Finite copula phrase | 12 | `VM0 VBI` |
| NT-AV0AJ0 | Adjective Phrase | 11 | `AV0 AJ0` |
| NT-AJ0CJC | | 10 | `AJ0 CJC` |
| NT-PNPVBBVVN | Subject + copula | 21 | `PNP VBD` |

Table 7.8: Non-terminals produced during first 20 iterations of the algorithm.

Pereira and Schabes (Pereira & Schabes, 1992) establish that evaluation according to the bracketing accuracy and evaluation according to perplexity or cross-entropy are very different. In fact, the model trained on the bracketed corpus, although scoring much better on bracketing accuracy, had a *higher* (worse) perplexity than the one trained on the raw data. This means that optimising the likelihood of the model may not lead you to a linguistically plausible grammar.

In Table 7.8 I show the non-terminals produced during the first 20 iterations of the algorithm. Note that there are less than 20 of them, since as mentioned above sometimes we will add more rules to an existing non-terminal. I have taken the liberty of attaching labels such as `NP` to the non-terminals where this is well justified. Where it is not, I leave the symbol produced by the program which starts with `NT-`. Table 7.9 shows the most frequent rules expanding the `NP` non-terminal, and Table 7.10 the `S` rules. Note that there is a good match between these rules and the traditional phrase structure rules.

To facilitate comparison with other unsupervised approaches, I performed an evaluation against the ATIS corpus, the results of which are summarised in Table 7.11. To perform this evaluation, I tagged the ATIS corpus with the CLAWS tags used here, using the freely available CLAWS demo tagger available on the web, removed empty constituents, and adjusted a few tokenisation differences (*at least* is one token in the BNC.) I then corrected a few systematic tagging errors. This might be slightly controversial. For example, "Washington D C" which is three tokens was tagged as `NP0 ZZ0 ZZ0` where `ZZ0` is a tag for alphabetic symbols. In the BNC, that I trained the model on, the DC is tagged as `NP0`, and in the ATIS corpus it is marked up as `(NP (NNP Washington))(NP (NNP D) (NNP C))`, i.e. with the DC as a separate noun phrase. In this case I altered the tags to `NP0 NP0 NP0`, and I performed some similar corrections in a couple of other instances. I did not alter the mark up of flight codes and so on that occur frequently in this corpus and very infrequently in the BNC.

| Count | Right Hand Side |
|---|---|
| 255793 | `AT0 NN1` |
| 104314 | `NP PP` |
| 103727 | `AT0 AJ0 NN1` |
| 73151 | `AT0 NN2` |
| 72686 | `DPS NN1` |
| 52202 | `AJ0 NN2` |
| 51575 | `DT0 NN1` |
| 35473 | `NP NP` |
| 34523 | `DT0 NN2` |
| 34140 | `AV0 NP` |

Table 7.9: Ten most frequent rules expanding NP. Note that three of them are recursive.

| Count | Right Hand Side |
|---|---|
| 14727 | `PNP VVD NP` |
| 10677 | `PNP VVD PP` |
| 7410 | `PNP VBD NP` |
| 6429 | `PNP VBZ NP` |
| 6061 | `PNP VVB NP` |
| 5885 | `PNP VM0 VVI NP` |
| 5336 | `NP VVD NP` |
| 5334 | `NP VBZ NP` |
| 5297 | `EX0 VBZ NP` |
| 4590 | `NP VVZ NP` |

Table 7.10: Ten most frequent rules expanding S. Note that since NPs include sequences of NPs and PPs, these rules parse a wide range of sets of complements.

| Algorithm | Iterations | UR | UP | F-score | CB | 0 CB | ≤ 2 CB |
|-----------|-----------|------|------|---------|------|------|--------|
| EMILE | | 16.8 | 51.6 | 25.4 | 0.84 | 47.4 | **93.4** |
| ABL | | **35.6** | 43.6 | 39.2 | 2.12 | 29.1 | 65.0 |
| CDC | 10 | 23.7 | **57.2** | 33.5 | **0.82** | 57.3 | 90.9 |
| CDC | 20 | 27.9 | 54.2 | 36.8 | 1.10 | 54.9 | 85.0 |
| CDC | 30 | 33.3 | 54.9 | 41.4 | 1.31 | 48.3 | 80.5 |
| CDC | 40 | 34.6 | 53.4 | **42.0** | 1.46 | 45.3 | 78.2 |
| Optimistic | | | | | | | |
| CDC | 10 | 33.3 | 67.3 | 44.6 | 0.82 | 57.3 | 90.9 |
| CDC | 20 | 37.1 | 63.5 | 46.8 | 1.10 | 54.9 | 85.0 |
| CDC | 30 | 40.7 | 61.1 | 48.9 | 1.31 | 48.3 | 80.5 |
| CDC | 40 | 41.8 | 59.5 | 49.1 | 1.46 | 45.3 | 78.2 |

Table 7.11: Results of evaluation on ATIS corpus. UR is unlabelled recall, UP is unlabelled precision, CB is average number of crossing brackets, ≤ 2 CB is percentage with two or fewer crossing brackets. The results for EMILE and ABL are taken from van Zaanen and Adriaans (2001). The optimistic results do not remove the top-level of brackets: thus the algorithm gets credit for the sentence bracket, which is standard in supervised frameworks with labelled precision and recall measures.

It is worth pointing out that the ATIS corpus is a very simple corpus, of radically different structure and markup to the BNC. It consists primarily of short questions and imperatives, and many sequences of letters and numbers such as T W A, A P 5 7 and so on.

A simple sentence like "Show me the meal" has the gold standard parse:

```
(S (VP (VB Show)
       (NP  (PRP me))
       (NP  (DT the)
            (NN meal))))
```

and is parsed by this algorithm as

```
 (ROOT (VVB Show)
       (PNP me)
       (NP (AT0 the)
           (NN1 meal)))
```

According to this evaluation scheme its recall is only 33%, because of the presence of the non-branching rules, though intuitively it has correctly identified the bracketing. However, the crossing brackets measures overvalues these algorithms, since it produces only a partial parse and for some sentences produces a completely flat parse tree which of course has no crossing brackets.

I then performed a partial parse of this data using the SCFG trained on the BNC, and evaluated the results against the gold-standard ATIS parse using the PARSEVAL metrics calculated by the EVALB program. Table 7.11 presents the results of the evaluation on the ATIS corpus, with the results on this algorithm (CDC) compared against two other algorithms, EMILE (Adriaans et al., 2000) and ABL (van Zaanen, 2000). I also include an evaluation with a slightly more optimistic criterion, where I do not remove the top level brackets. Thus the example discussed above would have a recall of 2 out of 5 (40%) rather than 1 out of 3 (33%) according to this more optimistic criterion. The comparison presented here does only allow tentative conclusions. First, there are minor differences in the test sets used. Secondly, the CDC algorithm is not completely unsupervised at the moment as it runs on tagged text, not raw text, though the ATIS corpus has very little lexical ambiguity so the problem is probably quite minor. Thirdly, it is worth reiterating that the CDC algorithm was trained on a radically different and much more complex data set, so these results greatly understate its accuracy. However, we can conclude that the CDC algorithm compares favourably to other unsupervised algorithms.

In particular the CDC algorithm beats the right-branching baseline model which has an F-score of 40.52, with an F-score of 42.0 after 40 iterations.

## 7.11    Experiments with automatically derived tags

I now present the results of some experiments with the automatically derived tags from Chapter 5. These were produced from the same amount of data, 12 million words, with the punctuation *included*, and without using any morphological information. I will use as labels for each class the most frequent word in the class, with one exception: the class whose most frequent element is `)` is predominantly composed of adverbial particles and I will accordingly use the second most frequent member `UP` as it is more representative.

Table 7.12 shows the first twenty non-terminals produced by the algorithm. A good example of the sort of erroneous rules that this algorithm produces is the cluster labelled `ERROR1`. It clustered the following four sequences together:

- `&EQUO ,`

- `, NP ,`

- `OF FACT ,`

- `OF THE AP_CONJ`

`AP_CONJ` is a non terminal that generates adjective phrases followed by a conjunction or a comma. `&EQUO` is a closing (right) quotation mark. Clearly part of the problem is a difficulty with the analysis of punctuation. Tables 7.13 and 7.14 show the ten most frequent rules that expand NP and VP respectively, or rather the symbols that I have labelled as NP and VP. As can be seen, the most frequent rules correspond well to the sorts of strings that are generated by these categories.

I include in an appendix all the rules generated during the first 30 iterations of the algorithm.

## 7.12    Conclusion

### 7.12.1    Discussion

There are a number of weaknesses of this approach. First, we need a lot of counts to get reliable estimates of the mutual information; as previously stated, I have not used sophisticated methods for estimating the mutual information, so it is possible that quite small amounts of data might suffice. But in any event there are always rare sequences, and this model does not capture them well. Perhaps a tag set that spreads more evenly over the set of words might alleviate this problem.

| Symbol | Description | Number of rules | Most Frequent |
|---|---|---|---|
| NP | Noun Phrase | 79 | THE TIME |
| NP_PROPER | Proper Noun Phrase | 6 | JOHN PHILIP |
| AP_CONJ | Adjective followed by conj. | 8 | NEW , |
| PP_CONJ | Prepositional Phrase plus conj. | 12 | OF NP , |
| S_OR_NP | Clause or NP | 7 | BUT NP |
| COMPLEX_AP | Complex Adjective Phrase | 3 | MORE NEW |
| ERROR1 | Unusual Mixture | 4 | , NP , |
| AP | Adjective Phrase | 5 | NOT POSSIBLE |
| NP_CONJ | NP followed by conj. | 7 | NP , |
| PP | Prepositional Phrase | 3 | OF NP |
| NT_GOING_TO_USE | ? | 3 | GOING TO BE |
| VP | Verb Phrase | 8 | 'S NP |
| NP_V | Subject plus verb | 7 | NOT MUCH |
| ADVP_NEG | Adverb phrase | 2 | NOT NOT |
| NT_THE_NEED_TO | ? | 3 | THE NEED TO |
| S | Clause | 4 | IT VP |
| CONJUNCT | conjuncts | 4 | , HOWEVER , |
| NOT_NP | ? | 5 | NOT NP |
| TO_SEE | ? | 2 | TO SEE |
| N1 | N bar | 7 | PARTY GROUPS |

Table 7.12: First 20 non-terminals produced by the algorithm operating on CDC tags.

| Count | Right Hand Side |
|---|---|
| 316191 | THE TIME |
| 156893 | LONDON |
| 108564 | NP OF NP |
| 86400 | NEW PEOPLE |
| 75279 | THE GROUP |
| 73967 | THE PEOPLE |
| 72857 | YOU |
| 71879 | THE NEW TIME |
| 67493 | THE NEW GROUP |
| 58561 | THE NEW PEOPLE |

Table 7.13: Ten most frequent rules expanding NP, together with the number of times each was applied.

| Count | Right Hand Side |
|-------|-----------------|
| 13784 | 'S NP |
| 13463 | IS POSSIBLE |
| 13405 | CAME PP |
| 11329 | WILL BE NP |
| 9832 | IS AP |
| 9441 | ARE NP |
| 8001 | IS ONE PP |
| 6108 | IS NOT NP |

Table 7.14: All rules expanding VP, together with the number of times each was applied.

Secondly, the way that the MI criterion is integrated into the MDL framework is very *ad hoc*. The problem is that I have two criteria: MDL gain and MI. If you have one criterion, then you just choose the best; if you have two criteria you have to have some way of balancing the relative merits, and I do not. I have therefore concealed the problem by turning one of them, the MI criterion, into a binary criterion, and selecting the best according to the other criterion. This is clearly inadequate.

Thirdly, I have presented results on only one language; though this is very standard (van Zaanen and Adriaans (2001) stand as a very honourable exception) it is not acceptable. There are now sufficiently large corpora available in many languages, and in several languages there are also small tree-banks that would be suitable for evaluation - at least Czech, Chinese, German and Dutch. The difficult case of morphologically rich languages with free word order make Czech a particular challenge.

Fourthly, the clustering algorithm is very crude and assumes that each sequence is unambiguously a member of a particular cluster. Of course this is not the case: the sequence Article Noun is often a constituent but often it is the beginning of a larger constituent such as Article Noun Noun.

Fifthly, the greediness of the algorithm causes many problems: though at the moment it is necessary to do this to make the algorithm feasible on the large amounts of data I am currently using, it should be possible to avoid doing this, or at least to apply the rules only selectively.

### 7.12.2    Future Work

There are a number of possible avenues for future research. The most important, in my opinion, is to experiment with other languages, particularly languages with very free word order. There is some evidence that these techniques will work with Chinese (Redington et al., 1995), which has quite fixed word order, but no work has been done in highly inflected languages with free word order.

In terms of exploring modifications to the existing algorithms, the obvious things to do are to explore more sophisticated versions of the various techniques I have used here. First, I will experiment with the use of other metrics, such as information theory derived metrics like the KL divergence. Secondly, I will try to use more sophisticated clustering algorithms, that either induce a hierarchical clustering, or can adjust the number of clusters to the data, using techniques such as those of Figueiredo and Jain (2001). In addition, given that we know that the distributions are made up of linear combinations of distributions with zero mutual information, we can use this fact to derive a more sophisticated soft EM-based clustering algorithm that implicitly divides the clusters into valid and invalid constituents.

More long term projects include looking at modifications to the IO algorithm that will allow the direct incorporation of the MI criterion into the algorithm, and exploring the use of lexicalised grammars. In addition, none of the phenomena I have been modelling here really require the full power of context-free grammars: it is possible that a slightly shallower approach might give good results.

### 7.12.3   Conclusion

These results are clearly very preliminary. Though it has demonstrated the validity of this technique, it is possible that developing it entirely in English means I have subtly encoded my understandings of the nature of English syntax into the search space of the program. It is therefore essential to experiment extensively with other languages. An advantage of unsupervised methods is of course that one is not limited to languages with extant treebanks which are rather hard to come by. This does raise issues of evaluation, though with the crudity of the current analyses this should not be a major problem, since an evaluation on only a few tens or hundreds of sentences should suffice to establish whether or not it is choosing plausible analyses.

# Chapter 8

# Conclusion

## 8.1 Summary

In this thesis I have addressed a number of areas in the unsupervised learning of language. I have discussed various theoretical problems related to this issue, and presented various algorithms that, when combined, can learn a certain amount of language. To recapitulate, the aim of this thesis was to examine the validity of the Argument from the Poverty of the Stimulus (APS) by examining the use of unsupervised machine learning algorithms on a corpus that approximates the primary linguistic data (PLD) available to an infant child. I first examined the requirements and restrictions on such algorithms and discussed some possible counter-arguments. I then discussed some formal and theoretical arguments against the enterprise, and concluded that these were without force. Then in Chapters 5, 6 and 7, I presented three sets of algorithms that could learn the syntactic categories of a language, the morphological relationships between the syntactic categories, and the basic phrase structure of the language, respectively. I also showed how the output from the syntactic category algorithm could serve as the input for the other two algorithms.

## 8.2 Previous work

There is very little work to which this thesis can be compared to directly. In the previous chapters I have tried to compare the individual parts and algorithms, but to my knowledge there has been no attempt to show how these parts might be combined into a complete system. I shall briefly discuss some work that seems relevant, or that spans more than one area of language acquisition.

### 8.2.1 Empiricist work

Several researchers have produced series of papers that cover more than one aspect of unsupervised language acquisition. Brill produced in the early 1990s a sequence of papers (Brill, 1991; Brill & Marcus, 1992) that explored how distributional evidence could be used to learn parts of speech, and syntax. de Marcken (1995, 1996b, 1996a) discussed how to acquire a lexicon from a raw speech signal and how to learn a segmentation of a corpus using an MDL technique. Brent and coworkers have produced work on segmentation (Brent & Cartwright, 1997), acquisition of syntactic categories (Brent, 1997), unsupervised learning of morphology (Snover & Brent, 2001) and learning of lexical syntax (Brent, 1991, 1993). Together these papers span an impressive range of language phenomena, though limited in the range of languages covered.

### 8.2.2 Nativist work

Nativist work has largely been limited in recent years to the problem of parameter setting, in the belief that this provides a solution to the problem of language acquisition. There have been a large number of papers discussing how certain parameters might be learned, many of them merely providing formal models, with no attempt to test them on real data. I shall not attempt a survey.

Niyogi and Berwick (2000) say

> The problem of language acquisition is reduced to learning to set the parameter values for the target language on the basis of sentences from that target.

This is of course not quite the case: the learner must also learn a lexicon, which is a highly non-trivial task. Indeed given that most modern syntactic theories, are highly lexicalised, learning a lexicon is more or less all that one has to do. In the absence of a theory as to how the lexicon is learned it is difficult to see that learning parameters answers anything.

A recent paper that is a computational model of learning in a nativist framework is Villav-icencio (2000). This paper discusses the learning of word-order parameters, using a unification based framework. Here the algorithm is provided with sentences from the CHILDES database that have been fully semantically annotated. Thus the algorithm has been provided with a full lexicon, together with an unambiguous semantic representation for all of the sentences even if they are ambiguous.

## 8.3 A Refutation of the Argument from the Poverty of the Stimulus?

I have presented various algorithms for learning various aspects of natural language. Now we must ask to what extent do these results provide an answer to the scientific question examined in this thesis: does this constitute a refutation of the APS? I will discuss this in two steps; first I examine how much of language these models have 'learned', and second I examine the extent to which flaws in my techniques reduce the significance of the results I have obtained.

First of all the syntactic category induction algorithm relies on the statistical properties of English too much; I hope to redress this in the near future. Nonetheless, I think it is pretty clear that similar techniques will be able to work cross-linguistically with reasonable accuracy particularly when the algorithms are given access to other sources of information. The morphology component seems the most convincing to me: in spite of the fact that it does not handle the full range of morphological transductions that are seen in the world's languages, it can learn, with high accuracy, in a variety of languages. It does not take account of the interactions that exist with other components of the grammar, but this is not too serious at this stage of the enquiry. The syntax algorithm is less satisfactory: though it performs well, and the criterion I use has some independent plausibility, the actual algorithm is rather *ad hoc*, and the results I obtain are not very satisfactory. In particular the algorithm hasn't learnt any of the sorts of complex constructions that have been used in the APS. On the other hand, the results I have, show that the stimulus is not as poor as all that – a large amount of data is a rich source of information, and that the statistical patterns in the input are sufficiently prominent to give substantial clues as to statistical input.

### 8.3.1 Criticisms

I have made numerous domain specific assumptions, of various types, but at all steps I have tried to use standard algorithms. In fact, I find myself in a slight presentational dilemma: on the one hand, I want my work to appear original and complex, but on the other hand I want the algorithms I am using to appear simple and natural. I hope I have managed to get the balance right.

For example, I have used very standard techniques of statistical estimation, primarily Maximum Likelihood estimation together with various applications of the EM algorithm. I have re-used techniques from one chapter to the next, such as distributional clustering, and at each step I have tried to indicate how these algorithms are not domain-dependent, and have pointed out applications in other areas. In addition I have tried to limit the use of *ad hoc* filters, and tunable parameters, though in some cases they have proved to be necessary.

The most significant decision I made was to split the program up into separate components, and model the different levels of language, morphology and syntax, with separate programs. I do not think this should be considered to be anything other than an optimisation; it would certainly be possible to do some of these simultaneously, computing over the ambiguities, and I intend to do just this in future work.

A lot of other assumptions I have made can also be seen as optimisations: for example, I have used a finite-state model for morphology, but I used a context free model for syntax. Of course, I could have used a context-free model for the morphology as well, and that would probably have worked as well, but the computational burden would have gone up sharply.

### 8.3.2 Excuses

There are a number of reasons why these results are less than completely convincing. First, the limited power of current computers has been a constant problem: all of the techniques presented here have been computationally expensive. Basically, all of the algorithms have fairly low polynomial complexity, cubic or better, but the amounts of data have not allowed me to test all of the algorithms as thoroughly as I would like, or to present proper statistical analysis of the significance of some of the results. Secondly, though the growth of publicly available corpora has been enormous in the past few years, we are still a long way from having suitable corpora in a wide range of typologically distinct languages. Thirdly, the study of Machine Learning is still in its infancy, and many of the more advanced techniques are still poorly understood: as a result I have restricted myself to fairly basic techniques. Fourthly, I have perhaps been overly strict in rejecting all other sources of information: clearly, prosody and situational context do provide some information that might well be used by children, but it is difficult to incorporate this without trivialising the problem. Finally, this is a very new field of inquiry – only recently did it become possible to do this sort of computational modelling, and it is thus unrealistic to expect a complete solution to the problem of language acquisition.

In spite of all of the weaknesses of the models presented here, they still compare well to the best nativist models of language acquisition – vacuously, since to the best of my knowledge, there are none that accept cognitively reasonable inputs. Although the algorithm presented here may not be completely adequate, we can see the outlines of an algorithm that *is* adequate, though it may not be possible to construct it at the moment. To repeat a point made earlier, we are trying to demonstrate the existence of a plausible algorithm; it is not necessary that the one we present is perfect for the argument to be convincing.

## 8.4   Cognitive Plausibility

Though as mentioned before, the cognitive plausibility of these results is not necessary for the argument to go through, I will make a few brief comments on the subject. It would be absurd to write the whole thesis and not discuss whether this relates at all to the much more interesting question as to how infants actually *do* learn language.

Saffran et al. (1996), Maye and Gerken (2000) are good examples of the sort of empirical work that is showing that children actually do use these sorts of statistical learning algorithms to learn language. Are these techniques within the computational capabilities of the human brain?

Clearly the human brain doesn't do parameter estimation: this is just a technical device that we use to model what it does. I think the best way of thinking about how the brain actually learns is through the statistical mechanics analysis of learning that has been advanced in the context of neural networks (Hertz, Krogh, & Palmer, 1991; Engel & Van den Broeck, 2001). Statistical mechanics studies the properties of large systems of weakly interacting particles; the ability to learn can be seen as an emergent property of the whole system, in the case of humans, the set of neurons in the human brain. But in any event, it seems clear that the sorts of computation I propose are clearly within the computational grasp of the human brain, and are closely related to models that have been proposed in other areas, notably vision, one of the most productive areas of computational neuroscience.

At this point we can stop and see to what extent these models address the remaining criteria that Pinker (1979, p.219) identified, that I quoted in Section 2.4:

> It is instructive to spell out these conditions one by one and examine the progress that has been made in meeting them. First, since all normal children learn the language of their community, a viable theory will have to posit mechanisms powerful enough to acquire a natural language. This criterion is doubly stringent: though the rules of language are beyond doubt highly intricate and abstract, children uniformly *succeed* at learning them nonetheless, unlike chess, calculus and other complex cognitive skills. Let us say that a theory that can account for the fact that languages can be learned in the first place has met the *Learnability condition*. Second, the theory should not account for the child's success by positing mechanisms narrowly adapted to the acquisition of a particular language. For example, a theory positing an innate grammar for English would fail to meet this criterion, which can be called the *Equipotentiality Condition*. Third, the mechanisms of a viable theory must allow the child to learn his language within the time span normally taken by children, which is in the order of three years for the basic components of language skill. Fourth, the mechanisms must not require as input types of information or amounts of information that are unavailable to the child. Let us call these the *Time* and *Input Conditions*, respectively. Fifth, the theory should make predictions about the intermediate stages of acquisition that agree with empirical findings in the study of child language. Sixth, the mechanisms described by the theory should not be wildly inconsistent with what is known about the cognitive faculties of the child, such as the perceptual discriminations he can make, his conceptual abilities, his memory, attention, and so forth. These can be called the *Developmental* and *Cognitive Conditions*, respectively.

As mentioned previously, I have been trying to construct models that satisfy the first four conditions, the Learnability condition, the Equipotentiality condition, the Time condition and the Input condition. I have just discussed the Cognitive condition; the remaining condition, the Developmental condition requires that the model make the correct predictions about the intermediate stages of language development, and the sorts of errors that the model predicts the child would make. I will not address this fully, as I have previously discussed some of this in Section 2.5. I will just say that while the morphology component seems to be very suitable for this sort of model, the category induction algorithm has a serious problem as it stands. The first categories that the algorithm acquires are the closed-class categories – they really leap out of the data, and the open class categories such as noun and so on are acquired later. This is in direct contrast to the order in which children actually produce the words: first, they go through a phase of so-called telegraphic speech, where there speech is composed almost entirely of open-class content words, and only later do they fill in the closed-class function words. This is not fatal for this theory, since production of speech is presumably restricted by limitations on the semantic content in the very early years of life, so it is possible that they have been learned but are not being used because children haven't worked out what the point of them is, or are limited in production in some other way. But it is still a bit uncomfortable.

## 8.5   Conclusion

We can see here the outlines of an empiricist theory of language acquisition, but regardless of one's theoretical bias I hope the arguments presented here will have clarified some issues. In particular, people should mistrust their intuitions about what can and cannot be learned from raw data. One of the arguments for semantic boot-strapping is that without semantic information, children would not be able to learn the syntactic categories; so even though it is implausible we must assume that they do use it. This sort of argument should clearly not be used.

For those who are convinced by other arguments for innate knowledge, this thesis can be thought of as adjusting one of the bounds for innate knowledge. An upper bound on the innate knowledge we have is the wide range of languages that there are in the world. We cannot hypothesize too much language specific innate knowledge or we will rule out some of these languages. Conversely the APS puts a lower bound on the amount of innate knowledge: if we do not have at least *this* much innate knowledge, we will not be able to learn any language. Seen in this light, all I am doing is lowering this lower bound; perhaps to zero.

# Bibliography

Abney, S. (1996). Statistical methods and linguistics. In Klavans, & Resnik (Klavans & Resnik, 1996), pp. 1–26.

Abney, S., & Light, M. (1999). Hiding a semantic hierarchy in a Markov model. In *Proceedings of the Workshop in Unsupervised Learning in Natural Language Processing at ACL 99* University of Maryland.

Adriaans, P. (1999). Learning shallow context-free languages under simple distributions. Tech. rep. ILLC Report PP-1999-13, Institute for Logic, Language and Computation, Amsterdam.

Adriaans, P., Trautwein, M., & Vervoort, M. (2000). Towards high speed grammar induction on large text corpora. In Hlavac, V., Jeffery, K. G., & Wiedermann, J. (Eds.), *SOFSEM 2000: Theory and Practice of Informatics*, pp. 173–186. Springer Verlag.

Aho, A. V., & Ullman, J. D. (1969a). Properties of syntax-directed translations. *Journal of Computer and System Sciences*, *3*(3), 319–334.

Aho, A. V., & Ullman, J. D. (1969b). Syntax-directed translations and the pushdown assembler. *Journal of Computer and System Sciences*, *3*(1), 37–56.

Allison, L. (1993). Normalization of affine gap costs in optimal sequence alignment. *Journal of Theoretical Biology*, *161*, 263–269.

Allison, L., Powell, D., & Dix, T. I. (1999). Compression and approximate matching. *The Computer Journal*, *42*(1), 1–10.

Allison, L., Wallace, C. S., & Yee, C. N. (1990). Inductive inference over macro-molecules. Tech. rep. 90/148, Department of Computer Science, Monash University.

Allison, L., Wallace, C. S., & Yee, C. N. (1992). Finite-state models in the alignment of macro-molecules. *Journal of Molecular Evolution*, *35*, 77–89.

Allison, L., & Wallace, C. (1994). The posterior probability distribution of alignments and its application to parameter estimation of evolutionary trees and to optimisation of multiple alignments. *Journal of Molecular Evolution*, *39*, 418–430.

Alshawi, H. (1996). Qualitative and quantitative models of speech translation. In Klavans, & Resnik (Klavans & Resnik, 1996), pp. 27–48.

Aston, G., & Burnard, L. (1998). *The BNC Handbook: Exploring the British National Corpus with SARA*. Edinburgh University Press.

Atwell, E. (1987). Constituent-likelihood grammar. In Garside et al. (Garside, Leech, & Sampson, 1987), pp. 57–65.

Baayen, H., Piepenbrock, R., & van Rijn, H. (1993). The CELEX lexical database.. CD-Rom,Linguistic Data Consortium, Philadelphia, PA.

Baayen, H. (1991). Quantitative aspects of morphological productivity. In Booij, G., & van Marle, J. (Eds.), *Yearbook of Morphology 1991*, pp. 109–150. Kluwer.

Baayen, H., & Sproat, R. (1996). Estimating lexical priors for low-frequency morphologically ambiguous forms. *Computational Linguistics*, *22*(2), 155–166.

Baker, J. K. (1979). Trainable grammars for speech recognition. In Klatt, D. H., & Wolf, J. J. (Eds.), *Speech communication papers for the 97th meeting of the Acoustic Society of America*, pp. 547–550.

Baldwin, T., & Tanaka, H. (1999). The applications of unsupervised learning to Japanese grapheme-phoneme alignment. In *Proceedings of ACL Workshop on Unsupervised Learning in Natural Language Processing*, pp. 9–16.

Baldwin, T., & Tanaka, H. (2000). A comparative study of unsupervised grapheme-phoneme alignment methods. In *Proceedings of the 14th Pacific Asia Conference on Language, Information and Computation (PACLIC 14)*, pp. 3–14.

Barvinok, A. I. (1999). Polynomial time algorithms to approximate permanents and mixed discriminants within a simple exponential factor. *Random Structures and Algorithms*, *14*, 29–61.

Bates, E., & Elman, J. (1996). Learning rediscovered. *Science*, *274*(5294), 1849–50.

Bates, E. (1994). Modularity, domain specificity and the development of language. *Discussions in Neuroscience*, *1/2*, 136–149.

Bayes, T. (1763). An essay towards solving a problem in the doctrine of chances.. *Philosophical Transactions of the Royal Society London*, *53*, 370–418.

Beichl, I., & Sullivan, F. (1999). Approximating the permanent via importance sampling with application to the dimer covering problem. *Journal of Computational Physics*, *149*(1), 128–147.

Bengio, Y. amd Ducharme, R., & Vincent, P. (2000). A neural probabilistic language model. Tech. rep. 1178, Université de Montréal.

Bengio, S., & Bengio, Y. (1996). An EM algorithm for Asynchronous Input/Output Hidden Markov Models. In Xu, L. (Ed.), *International Conference on Neural Information Processing*, pp. 328–334 Hong Kong.

Bengio, Y. (1999). Markovian models for sequential data. *Neural Computing Surveys*, *2*, 129–162.

Bengio, Y., & Frasconi, P. (1996). Input/output HMMs for sequence processing. *IEEE Transactions on Neural Networks*, *7*(5).

Bernardo, J. M., & Smith, A. F. M. (1994). *Bayesian Theory*. Wiley.

Bertram, R., Laine, M., Baayen, R. H., Schreuder, R., & Hyönä, J. (2000). Affixal homonymy triggers full-form storage even with inflected words, even in a morphologically rich language. *Cognition*, *972*, 1–13.

Bhatia, R. (1996). *Matrix analysis*. Springer.

Bishop, C. M. (1995). *Neural Networks for Pattern Recognition*. Oxford University Press.

Bishop, D. V. M. (1996). Editorial: A gene for grammar. *Semiotic Review of Books*, *7*(2), 1–2.

Bloom, P. (Ed.). (1994). *Language Acquisition: Core Readings*. MIT Press.

Bod, R. (1993). Using an annotated corpus as a stochastic grammar. In *Proceedings of EACL-93*.

Bod, R. (1998). *Beyond Grammar*. CSLI Publications, Stanford University, California.

Boden, M. A. (1990). Introduction. In Boden, M. A. (Ed.), *The Philosophy of Artificial Intelligence*, pp. 1–21. Oxford University Press.

Bohannon, J. N., MacWhinney, B., & Snow, C. E. (1990). No negative evidence revisited: beyond learnability or who has to prove what to whom. *Developmental Psychology*, *26*(2), 221–226.

Braine, M. D. S. (1988). Review of S. Pinker's "Language Learnability and Language Development". *Journal of Child Language*, *15*, 189–219.

Braine, M. D. S. (1992). What sort of innate structure is needed to bootstrap into syntax?. *Cognition*, *45*, 77–100.

Brakke, K. E., & Savage-Rumbaugh, E. S. (1996). The development of language skills in *Pan* II production. *Language and Communication*, *16*(4), 361–380.

Bregman, L. M. (1967). Proof of convergence of Sheleikhovskii's method for a problem with transportation constraints. *Zh. vychsl. Mat. mat. Fiz.*, *147*(7).

Brent, M. R. (Ed.). (1997). *Computational Approaches to Language Acquisition*. Cognition. MIT Press.

Brent, M. R. (1991). Automatic acquisition of subcategorisation frames from untagged text. In *Proceedings of the 29th Annual Meeting of the ACL*, pp. 209–214.

Brent, M. R. (1993). From grammar to lexicon: unsupervised learning of lexical syntax. *Computational Linguistics*, *19*, 243–262.

Brent, M. R. (1997). Syntactic categorisation in early language acquisition: formalizing the role of distributional analysis. *Cognition*, *63*(2), 121–170.

Brent, M. R., & Cartwright, T. A. (1997). Distributional regularity and phonotactic constraints are useful for segmentation. In Brent (Brent, 1997), pp. 93–125.

Brill, E. (1991). Discovering the lexical features of a language. In *Proceedings of the 29th Annual Meeting of the Association fro Computational Linguistics*, pp. 339–340.

Brill, E. (1992). A simple rule-based part-of-speech tagger. In *Third Conference on Applied Natural Language Processing*.

Brill, E. (1993). Automatic grammar induction and parsing free text: A transformation-based approach. In *Proceedings of the 31st Annual Meeting of the ACL*, pp. 259–265.

Brill, E., & Marcus, M. (1992). Automatically acquiring phrase structure using distributional analysis. In *Proceedings of DARPA workshop on speech and natural language*.

Briscoe, T., & Waegner, N. (1992). Robust stochastic parsing using the inside-outside algorithm. In *AAAI-92 Workshop on Statistically-Based NLP Techniques*.

Broeder, P., & Murre, J. (Eds.). (2000). *Models of Language Acquisition*. Oxford University Press.

Brown, P. F., Della Pietra, V. J., de Souza, P. V., Lai, J. C., & Mercer, R. (1992). Class-based n-gram models of natural language. *Computational Linguistics*, *18*, 467–479.

Burnard, L. (1995). *Users Reference Guide for the British National Corpus*.

Bybee, J. l. (1995). Regular morphology and the lexicon. *Language and Cognitive Processes*, *10*, 425–455.

Bybee, J. L., & Moder, C. L. (1983). Morphological classes as natural categories. *Language*, *59*, 251–270.

Bybee, J. L., & Slobin, D. I. (1982). Rules and schemes in the development and use of the English past tense. *Language*, *58*, 265–289.

Cahill, L. J., & Gazdar, G. (1997). The inflectional phonology of German adjectives, determiners and pronouns. *Linguistics*, *35*(2), 211–245.

Cahill, L. J., & Gazdar, G. (1999). German noun inflection. *Journal of Linguistics*, *35*(1), 1–42.

Carbonell, J. (Ed.). (1990). *Machine Learning: Paradigms and methods*. MIT press.

Carden, G. (1983). The non-finite-stateness of the word formation component. *Linguistic Inquiry*, *14*, 537–541.

Carroll, G., & Charniak, E. (1992). Two experiments on learning probabilistic dependency grammars from corpora. Tech. rep. CS-92-16, Department of Computer Science, Brown University.

Carroll, J., Briscoe, T., & Sanfilippo, A. (1998). Parser evaluation: a survey and a new proposal. In *Proceedings of the 1st International Conference on Language Resources and Evaluation*, pp. 447–454 Granada, Spain.

Casacuberta, F. (1995). Probabilistic estimation of stochastic regular syntax-directed translation schemes. In *Proceedings of the VIth Spanish Symposium on Pattern Recognition and Image Analysis*, pp. 201–207.

Casacuberta, F. (1996). Maximum mutual information and conditional maximum likelihood estimations of stochastic regular syntax-directed translation schemmes. In Miclet, L., & de la Higuera, C. (Eds.), *ICGI-96*, pp. 282–291.

Casacuberta, F., & de la Higuera, C. (2000). Computational complexity of problems on probabilistic grammars and transducers. In Oliveira, A. L. (Ed.), *Grammatical Inference: Algorithms and Applications*, No. 1891 in Lecture Notes in Artificial Intelligence, pp. 15–24. Springer Verlag.

Charness, N. (1992). The impact of chess research on cognitive science. *Psychological Research*, *54*, 4–9.

Charniak, E. (1993). *Statistical Language Learning*. MIT Press.

Charniak, E. (2001). Immediate head parsing for language models. In *Proceedings of the 39th annual meeting of the ACL*, pp. 116–123 Toulouse, France.

Chen, S. (1995). Bayesian grammar induction for language modelling. In *Proceedings of the 33rd Annual Meeting of the ACL*, pp. 228–235.

Chomsky, C. (1979). *The acquisition of syntax in children from 5 to 10*. MIT Press.

Chomsky, N. (1959). A review of B. F. Skinner's "Verbal Behavior". *Language*, *35*(1), 26–58.

Chomsky, N. (1964). Degrees of grammaticalness. In Fodor, & Katz (Fodor & Katz, 1964), pp. 384–389.

Chomsky, N. (1965). *Aspects of the Theory of Syntax*. MIT.

Chomsky, N. (1966). *Topics in the Theory of Generative Grammar*. Mouton.

Chomsky, N. (1967). Recent contributions to the theory of innate ideas. *Synthese*, *17*, 2–11.

Chomsky, N. (1975a). *The Logical Structure of Linguistic Theory*. University of Chicago Press.

Chomsky, N. (1975b). Recent contributions to the theory of innate ideas. In Stich (Stich, 1975), pp. 121–132.

Chomsky, N. (1975c). *Reflections on Language*. Pantheon Books.

Chomsky, N. (1981). *Lectures on Government and Binding*. Foris Publications.

Chomsky, N. (1986). *Knowledge of Language : Its Nature, Origin, and Use*. Praeger.

Chomsky, N. (1988a). *The Generative Enterprise. A discussion with Riny Huybregts and Henk van Riemsdijk*. Foris, Dordrecht.

Chomsky, N. (1988b). *Language and Problems of Knowledge*. MIT Press.

Chomsky, N. (2000). *New Horizons in the Study of Language and Mind*. Cambridge University Press.

Christophe, A., Dupoux, E., Bertoncini, J., & Mehler, J. (1994). Do infants perceive word boundaries? an empirical study of the bootstrapping of lexical acquisition.. *Journal of the Acoustical Society of America*, *95*, 1570–1580.

Church, K., & Mercer, R. (1993). Introduction to the special issue on computational linguistics using large corpora. *Computational Linguistics*, *19*(1), 1–24.

Clahsen, H., Rothweiler, M., Wöst, A., & Marcus, G. M. (1992). Regular and irregular inflection in the acquisition of German noun plurals.. *Cognition*, *45*, 225–255.

Clark, A. (2000). Inducing syntactic categories by context distribution clustering. In *Proceedings of CoNLL-2000 and LLL-2000*, pp. 91–94 Lisbon, Portugal.

Clark, A. (2001a). Learning morphology with Pair Hidden Markov Models. In *Proceedings of the Student Workshop at the 39th Annual Meeting of the Association for Computational Linguistics*, pp. 55–60 Toulouse, France.

Clark, A. (2001b). Partially supervised learning of morphology with stochastic transducers. In *Proceedings of NLPRS 2001* Tokyo, Japan. to appear.

Clark, A. (2001c). Unsupervised induction of stochastic context free grammars with distributional clustering. In *Proceedings of CoNLL 2001*, pp. 105–112 Toulouse, France.

Clark, R., Gleitman, L., & Kroch, A. (1997). Acquiring language. *Science*, *276*(5316), 1177–1181.

Comrie, B. (1987). *The World's Major Languages*. Routledge.

Cover, T. M., & Thomas, J. A. (1991). *Elements of Information Theory*. Wiley Series in Telecommunications. John Wiley & Sons.

Cowie, F. (1999). *What's Within? Nativism Reconsidered*. Oxford University Press.

Crago, M., & Gopnik, M. (1994). From families to phenotypes: theoretical and clinical implications of research into the genetic basis of specific language impairment. In Watkins, R., & Rice, M. (Eds.), *Specific Language Impairments in Children*, pp. 35–51. Paul H. Brookes, Baltimore.

Crain, S., & Pietroski, P. (2001). Nature, nurture and universal grammar. *Linguistics and Philosophy*, *24*, 139–186.

Crain, S., & Thornton, R. (1998). *Investigations in Universal Grammar: A guide to Experiments on the Acquisition of Syntax and Semantics*. Bradford.

Daelemans, W., Berck, P., & Gillis, S. (1997). Data mining as a method for linguistic analysis: Dutch diminutives. *Folia Linguistica*, *XXXI*, 1–2.

Daelemans, W. M. P., & van den Bosch, A. P. J. (1997). Language-independent data-oriented grapheme-to-phoneme conversion. In van Santen, J. P. H., Sproat, R. W., Olive, J. P., & Hirschberg, J. (Eds.), *Progress in Speech Synthesis*, pp. 77–89. Springer, New York.

Dagan, I., Marcus, S., & Markovitch, S. (1993). Contextual word similarity and estimation from sparse data. In *Proceedings of the 31st Annual Meeting of the Association for Computational Linguistics*, pp. 164–171.

de Marcken, C. G. (1995). The unsupervised acquisition of a lexicon from continuous speech. AI memo 1558, MIT AI Lab.

de Marcken, C. G. (1996a). Linguistic structure as composition and perturbation. In Joshi, A., & Palmer, M. (Eds.), *Proceedings of the Thirty-Fourth Annual Meeting of the Association for Computational Linguistics*, pp. 335–341 San Francisco. Morgan Kaufmann Publishers.

de Marcken, C. G. (1996b). *Unsupervised Language Acquisition*. Ph.D. thesis, MIT.

de Marcken, C. G. (1999). On the unsupervised induction of phrase-structure grammars. In Armstrong, S., Church, K., Isabelle, P., Manzi, S., Tzoukermann, E., & Yarowsky, D. (Eds.), *Natural Language Processing Using Very Large Corpora*, pp. 191–208. Kluwer.

Deerwester, S. C., Dumais, S. T., Landauer, T. K., Furnas, G. W., & Harshman, R. A. (1990). Indexing by latent semantic analysis. *Journal of the American Society of Information Science*, *41*(6), 391–407.

Déjean, H. (1998). Morphemes as necessary concept for stuctures discovery from untagged corpora. In *Workshop on Paradigms and Grounding in Natural Language Learning*, pp. 295–299 Adelaide.

Dempster, A. P., Laird, N. M., & Rubin, D. B. (1977). Maximum likelihood from incomplete data via the EM algorithm. *Journal of the Royal Statistical Society Series B*, *39*, 1–38.

Dermatas, E., & Kokkinakis, G. (1995). Automatic stochastic tagging of natural language texts. *Computational Linguistics*, *21*(2), 137–164.

Derwing, B. L., & Skousen, R. (1994). The reality of linguistic rules. In Lima, S. D., & Corrigan, Roberta L. Iverson, G. K. (Eds.), *The Reality of Linguistic Rules*, No. 26 in Studies in Language Companion Series, chap. Productivity and the English Past Tense, pp. 193–218. John Benjamins Publishing Company, Amsterdam.

Divay, M., & Vitale, A. J. (1997). Algorithms for grapheme-phoneme translation for English and French: Applications. *Computational Linguistics*, *23*(4), 495–524.

Durbin, R., Eddy, S., Krogh, A., & Mitchison, G. (1998). *Biological Sequence Analysis: Probabilistic Models of proteins and nucleic acids*. Cambridge University Press.

Eisner, J. (2001). Expectation semi-rings: Flexible EM for learning finite-state transducers. In *Proceedings of the ESSLLI Workshop on Finite-State Methods in NLP* Helsinki.

Elman, J., Bates, E., Johnson, M., Karmiloff-Smith, A., Parisi, D., & Plunkett, K. (1996). *Rethinking Innateness : A connectionist Perspective on Development*. MIT.

Engel, A., & Van den Broeck, C. (2001). *Statistical Mechanics of Learning*. Cambridge.

Feldman, J. A., Lakoff, G., Bailey, D., Narayanan, S., Regier, T., & Stolcke, A. (1996). $l_0$ - the first five years of an automated language acquisition project. *Artificial Intelligence Review*, *10*(1-2), 103–129.

Feldman, J. A., Lakoff, G., Stolcke, A., & Weber, S. H. (1990). Miniature language acquisition: A touchstone for cognitive science. In *Proceedings of the Cognitive Science Society*, pp. 686–693 Cambridge, MA.

Figueiredo, M., & Jain, A. K. (2001). Unsupervised learning of finite mixture models. *IEEE Transactions on Pattern Analysis and Machine Intelligence*. to appear.

Finch, S., & Chater, N. (1992a). Bootstrapping syntactic categories. In *Proceedings of the 14th Annual Meeting of the Cognitive Science Society*, pp. 820–825.

Finch, S., & Chater, N. (1992b). Bootstrapping syntactic categories using statistical methods. In Daelemans, W., & Powers, D. (Eds.), *Background and Experiments in Machine Learning of Natural Language*, pp. 229–235. Tilburg University: Institute for Language Technology and AI.

Finch, S., Chater, N., & Redington, M. (1995). Acquiring syntactic information from distributional statistics. In Levy, J. P., Bairaktaris, D., Bullinaria, J. A., & Cairns, P. (Eds.), *Connectionist Models of Memory and Language*. UCL Press.

Fisher, C., Hall, D. G., Rakowitz, S., & Gleitman, L. R. (1994). When it is better to give than to receive: Syntactic support for verb learning. *Lingua*, *92*(1), 333–375.

Fodor, J. A. (1981). *Representations*. MIT Press.

Fodor, J. A. (1983). *Modularity of Mind*. MIT Press.

Fodor, J. A. (2001). Doing without what's within: Fiona Cowie's critique of nativism. *Mind*, *110*(437), 99–148.

Fodor, J. A., & Katz, J. J. (Eds.). (1964). *The structure of language: Readings in the philosophy of language*. Prentice-Hall.

Freyhof, H., Gruber, H., & Ziegler, A. (1992). Expertise and hierarchical knowledge representation in chess. *Psychological Research*, *54*, 32–37.

Garfield, J. L. (1994). Innateness. In Guttenplan, S. (Ed.), *A companion to the philosophy of mind*, pp. 366–374. Blackwell.

Garside, R., Leech, G., & Sampson, G. (Eds.). (1987). *The Computational Analysis of English*. Longman.

Gaussier, E. (1999). Unsupervised learning of derivational morphology from inflectional lexicons. In *Unsupervised Learning in Natural Language Processing*. ACL.

Gazdar, G. (1996). Paradigm merger in natural language processing. In Milner, R., & Wand, I. (Eds.), *Computing Tomorrow: Future Research Directions in Computer Science*. Cambridge University Press.

Gazdar, G., Klein, E., Pullum, G., & Sag, I. (1985). *Generalised Phrase Structure Grammar*. Basil Blackwell.

Gerken, L., Jusczyk, P. W., & Mandel, D. R. (1994). When prosody fails to cue syntactic structure: 9-month-olds' sensitivity to phonological versus syntactic phrases. *Cognition*, *51*, 237–265.

Geurts, B. (2000). Review of Stephen Crain & Rosalind Thornton, *Investigations in Universal Grammar*. *Linguistics and Philosophy*, *23*, 523–532.

Gildea, D., & Jurafsky, D. (1996). Learning bias and phonological-rule induction. *Computational Linguistics*, *22*(4), 497–530.

Gleitman, L. (1990). The structural sources of verb meaning. *Language Acqusition*, *1*(1), 3–55.

Gleitman, L. (1994). The structural sources of verb meaning. In Bloom (Bloom, 1994), pp. 174–221.

Gleitman, L. R., & Gillette, J. (1995). The role of syntax in verb learning. In Fletcher, P., & MacWhinney, B. (Eds.), *The Handbook of Child Language*, pp. 413–427. Blackwell.

Gold, E. M. (1967). Language identification in the limit. *Information and Control*, *10*, 447–474.

Gold, S., & Rangarajan, A. (1996). Softmax to softassign: Neural network algorithms for combinatorial optimization. *Journal of Artificial Neural Networks*, *2*(4), 381–399.

Golding, A. R., & Thompson, H. S. (1985). A morphology component for language programs. *Linguistics*, *23*, 263–284.

Goldsmith, J. A. (1998). On information theory, entropy, and phonology in the 20th century. draft of a paper read at the Royaumont CTIP II Round table on phonology in the 20th Century.

Goldsmith, J. A. (2000). Unsupervised learning of the morphology of a natural language. Available at http://humanities.uchicago.edu/faculty/goldsmith.

Goldsmith, J. A. (2001). Unsupervised learning of the morphology of a natural language. *Computational Linguistics*, *27*(2), 153–198.

Good, I. J. (1953). The population frequencies of species and the estimation of population parameters.. *Biometrika*, *40*, 237–264.

Goodman, J. (1996). Efficient algorithms for parsing the DOP model. In Brill, E., & Church, K. (Eds.), *Proceedings of the Conference on Empirical Methods in Natural Language Processing*, pp. 143–152. Association for Computational Linguistics, Somerset, New Jersey.

Goodman, J. (1998). *Parsing Inside-Out*. Ph.D. thesis, Harvard University.

Gopnik, M. (1990). Feature blindness: a case study. *Language Acquisition*, *1*, 139–164.

Gopnik, M., & Crago, M. (1991). Familial aggregation of a developmental language disorder. *Cognition*, *39*, 1–50.

Gordon, P. (1985). Level-ordering in lexical development. *Cognition*, *21*, 73–93.

Grefenstette, G. (1994). *Explorations in Automatic Thesaurus Discovery*. Kluwer.

Grenander, U. (1996). *Elements of Pattern theory*. John Hopkins.

Grünwald, P. (1996). A minimum description length approach to grammar inference. In Wermter et al. (Wermter et al., 1996).

Gusfield, D. (1997). *Algorithms on Strings, Trees and Sequences: Computer Science and Computational Biology*. Cambridge University Press.

Halmos, P. R. (1950). *Measure Theory*. Springer-Verlag.

Harris, Z. (1954). Distributional structure. In Fodor, & Katz (Fodor & Katz, 1964), pp. 33–49.

Harris, Z. (1955). From phonemes to morphemes. *Language*, *31*, 190–222.

Hart, B., & Risley, T. R. (1995). *Meaningful differences in the everyday experiences of young children*. Paul H. Brookes, Baltimore.

Hertz, J., Krogh, A., & Palmer, R. G. (1991). *Introduction to the theory of neural computation*. Perseus Books, Reading, MA.

Hobbs, J. R., & Shieber, S. M. (1987). An algorithm for generating quantifier scopings. *Computational Linguistics*, *13*(1-2), 47–63.

Hockett, C. (1958). Two models of grammatical description. In Joos, M. (Ed.), *Readings in Linguistics* (2nd edition). University of Chicago Press.

Hofmann, T., & Puzicha, J. (1998). Statistical models for co-occurrence data. Tech. rep. AI Memo 1625, CBCL Memo 159, Artificial Intelligence Laboratory and Center for Biological and Computational Learning, MIT.

Holt, L. L., Lotto, A. J., & Kluender, K. R. (1998). Incorporating principles of general learning in theories of language acquisition. In Gruber, M., Higgins, C. D., Olson, K. S., & Wysocki, T. (Eds.), *Chicago Linguistic Society, Volume 34*, pp. 253–268. Chicago Linguistic Society.

Horning, J. J. (1969). *A study of grammatical inference*. Ph.D. thesis, Computer Science Department, Stanford University.

IHGMC (2001). A physical map of the human genome. *Nature*, *409*, 934–941. The International Human Genome Mapping Consortium.

Jelinek, F. (1997). *Statistical methods for speech recognition*. Language, Speech and Communication. MIT Press.

Jerrum, M., & Sinclair, A. (1989). Approximating the permanent. *SIAM Journal on Computing*, *18*, 1149–1178.

Jespersen, O. (1924). *The Philosophy of Grammar*. Allen & Unwin. reprinted (1992) by University of Chicago Press.

Johnson, D., & Lappin, S. (1997). A critique of the Minimalist Program. *Linguistics and Philosophy*, *20*, 273–333.

Juola, P. (1998). On psycholinguistic grammars. *Grammars*, *1*(1), 15–31.

Kaplan, R. M., & Kay, M. (1994). Regular models of phonological rule systems. *Computational Linguistics*, *20*(3), 331–378.

Kapur, S., & Bilardi, G. (1992). Language learning from stochastic input. In *Proceedings of the fifth conference on Computational Learning Theory*, pp. 303–310 Pittsburgh.

Kapur, S. (1991). *Computational Learning of Languages*. Ph.D. thesis, Cornell University. Computer Science Department Technical Report 91-1234.

Kapur, S., & Clark, R. (1996). The automatic construction of a symbolic parser via statistical techniques. In Klavans, & Resnik (Klavans & Resnik, 1996), pp. 95–117.

Keller, B., & Lutz, R. (1997). Evolving stochastic context-free grammars from examples using a minimum description length principle. In *Workshop on Automatic Induction, Grammatical Inference and Language Acquisition*.

Kelly, M. H., & Martin, S. (1992). Domain-general abilities applied to domain-specific tasks: Sensitivity to probabilities in perception, cognition and language. *Lingua*, *92*, 105–140.

Kit, C. (1998). A goodness measure for phrase learning via compression with the mdl principle. In *Proceedings of the ESSLLI Student session*, pp. 175–187.

Klavans, J., & Resnik, P. (Eds.). (1996). *The Balancing Act*. MIT Press.

Klein, D., & Manning, C. (2001). Distributional phrase structure induction. In *Proceedings of CoNLL 2001*, pp. 113–121.

Knight, K., & Graehl, J. (1997). Machine transliteration. In *Proceedings of ACL-97*.

Köpcke, K.-M. (1988). Schemas in German plural formation. *Lingua*, *74*, 303–335.

Kornai, A. (1992). Frequency in morphology. In Kenesie, I. (Ed.), *Approaches to Hungarian*, Vol. 4, pp. 246–268. Szeged:JATE.

Kornai, A. (1998). Quantitative comparison of languages. *Grammars*, *2*, 155–165.

Koskenniemi, K. (1983a). Two-level model for morphological analysis. In Bundy, A. (Ed.), *Proceedings of the Eighth International Conference on Artificial Intelligence*, pp. 683–685.

Koskenniemi, K. (1983b). *A Two-level Morphological Processor*. Ph.D. thesis, University of Helsinki.

Koskenniemi, K. (1984). A general computational model for word-form recognition and production. In *Proceedings of COLING-84*, pp. 178–181.

Krogh, A., & Riis, S. K. (1999). Hidden neural networks. *Neural Computation*, *11*(2), 541–563.

Kuich, W., & Salomaa, A. (1986). *Semirings, Automata and Languages*. No. 5 in EATCS Monographs on Theoretical Computer Science. Springer-Verlag.

Lamb, S. M. (1961). On the mechanisation of syntactic analysis. In *1961 Conference on Machine Translation of Languages and Applied Language Analysis*, Vol. 2 of *National Physical Laboratory Symposium No. 13*, pp. 674–685. Her Majesty's Stationery Office, London.

Lance, G. N., & Williams, W. T. (1967). A general theory of classificatory sorting strategies. *Computer Journal*, *9*, 373–380.

Langendoen, O. T. (1981). The generative capacity of word-formation components. *Linguistic Inquiry*, *12*, 320–322.

Lari, K., & Young, S. J. (1990). The estimation of stochastic context-free grammars using the inside-outside algorithm. *Computer Speech and Language*, *4*, 35–56.

Lee, L. (1999). Measures of distributional similarity. In *37th Annual Meeting of the Association for Computational Linguistics*, pp. 25–32.

Leech, G., Garside, R., & Bryant, M. (1994). CLAWS4: the tagging of the British National Corpus. In *Proceedings of the 15th International Conference on Computational Linguistics*, pp. 622–628.

Lerdahl, F., & Jackendoff, R. (1983). *A Generative theory of tonal music*. The MIT Press.

Levin, B. (1993). *English verb classes and alternations: a preliminary investigation*. The Univeristy of Chicago Press.

Li, H., & Abe, N. (1996). Clustering words with the mdl principle. In *Proceedings of COLING '96*, pp. 4–9.

Li, H., & Abe, N. (1998). Word clustering and disambiguation based on co-occurrence data. In *Proceedings of COLING ACL '98*, pp. 749–755.

Li, W. (1990). Mutual information functions versus correlation functions. *Journal of Statistical Physics*, *60*, 823–837.

Lin, J.-h., & Vitter, J. S. (1994). A theory for memory-based learning. *Machine Learning*, *17*(1-26).

Ling, C. X. (1994). Learning the past tense of English verbs: The symbolic pattern associator vs. connectionist models. *Journal of Artifical Intelligence Research*, *1*, 209–229.

Locke, J. (1690). *An Essay Concerning Human Understanding*.

Locke, W. N., & Booth, A. D. (Eds.). (1955). *Machine Translation of Languages*. MIT Press.

MacWhinney, B., & Leinbach, J. (1991). Implementations are not conceptualizations: Revising the verb model.. *Cognition*, *40*, 121–157.

MacWhinney, B. (1995). *The CHILDES Project: Tools for analyzing talk*. Lawrence Erlbaum, Hillsdale, New Jersey.

Magerman, D. M., & Marcus, M. P. (1990). Parsing a natural language using mutual information statistics. In *Proceedings of the Eighth National Conference on Artificial Intelligence*.

Manandhar, S., Dzeroski, S., & Erjavec, T. (1998). Learning multi-lingual morphology with clog. In Page, C. D. (Ed.), *Proc. of the 8th International Workshop on Inductive Logic Programming (ILP-98)*. Springer Verlag.

Manning, C. D., & Schütze, H. (1999). *Foundations of Statistical Natural Language Processing*. MIT Press.

Manning, C. D. (1998). The segmentation problem in morphology learning. In Powers, D. M. W. (Ed.), *NeMLaP3/CoNLL98 Workshop on Paradigms and Grounding in Language Learning*, pp. 299–305. ACL.

Marcus, G. F., Brinkmann, U., Clahsen, H., Wiese, R., & Pinker, S. (1995). German inflection: The exception that proves the rule. *Cognitive Psychology*, *29*, 189–256.

Marcus, M. P., Santorini, B., & Marcinkiewicz, M. A. (1993). Building a large annotated corpus of English: the Penn Treebank. *Computational Linguistics*, *19*.

Marr, D. (1982). *Vision*. W. H. Freeman.

Matthews, R. J. (1989). The plausibility of rationalism. In Matthews, R. J., & Demopoulos, W. (Eds.), *Learnability and Linguistic Theory*, pp. 51–76. Dordrecht.

Maye, J., & Gerken, L. (2000). Learning phonemes without minimal pairs. In *Proceedings of the 24th Annual Boston University Conference on Language Development*.

McCarthy, J., & Prince, A. (1990). Foot and word in prosodic morphology: The Arabic broken plural. *Natural Language and Linguistic Theory*, *8*, 209–284.

McCawley, J. D. (1988). *The Syntactic Phenomena of English*. The University of Chicago Press.

Miller, G. A., & Chomsky, N. (1963). Finitary models of language users. In Luce, R. D., Bush, R., & Galanter, E. (Eds.), *Handbook of Mathematical Psychology*, Vol. 2, pp. 419–491. Wiley, New York.

Miller, M. I., & O' Sullivan, J. A. (1992). Entropies and combinatorics of random branching processes and context-free languages. *IEEE Transactions on Information Theory*, *38*(4), 1292–1310.

Miller, S., Stallard, D., Bobrow, R., & Schwartz, R. (1996). A fully statistical approach to natural language interfaces. In *Proceedings of the 34th Annual Meeting of the Association for Computational Linguistics*, pp. 55–61 Santa Cruz, California.

Mohri, M., Pereira, F. C. N., & Riley, M. (2000). The design principles of a weighted finite-state transducer library. *Theoretical Computer Science*, *231*(1), 17–32.

Moll, R. N., Arbib, M. A., & Kfoury, A. J. (1988). *An introduction to formal language theory*. Springer-Verlag.

Mooney, R. J., & Califf, M. E. (1995). Induction of first-order decision lists: Results on learning the past tense of English verbs. *Journal of Artificial Intelligence Research*, *3*, 1–24.

Mori, A., & Nagao, M. (1995). Parsing without grammar. In *Proceedings of the 4th International Workshop on Parsing Technologies*, pp. 174–185.

Mori, S., & Nagao, M. (1998). A stochastic language model using dependency and its improvement by word clustering. In *Proceedings of COLING ACL '98*, pp. 898–904.

Mori, S., Nishimura, M., & Ito, N. (1997). Word clustering for class-based language models. *Transactions of Information Processing Society of Japanese*, *38*(11), 2200–2208. in Japanese.

Muggleton, S. (1999). Inductive Logic Programming: issues, results and the LLL challenge. *Artificial Intelligence*, *114*(1-2), 283–296.

Muggleton, S., & Bain, M. (1999). Analogical prediction. In *Proceedings of the 9th International Workshop on Inductive Logic Programming (ILP-99)* Berlin. Springer-Verlag.

Muggleton, S. (Ed.). (1997). *Inductive Logic Programming*. Springer-Verlag.

Murakami, J., Yamatomo, H., & Sagayama, S. (1993). The possibility for acquisition of statistical network grammar using ergodic HMM. In *Proceedings of Eurospeech 93*, pp. 1327–1330.

Neidle, C., Kegl, J., MacLaughlin, D., Bahan, B., & Lee, R. G. (1999). *The Syntax of American Sign Language*. MIT Press.

Newmeyer, Frederick, J. (1983). *Grammatical Theory: Its Limits and Its Possibilities*. The University of Chicago Press.

Newport, E. H., Gleitman, H., & Gleitman, E. (1977). Mother I'd rather do it myself: some effects and non-effects of maternal speech style. In Snow, C. E., & Ferguson, C. A. (Eds.), *Talking to Children: language input and acquisition*, pp. 109–149. Cambridge University Press.

Newton, I. (1687). Philosophiae naturalis principia mathematica. In Cohen, I. B., & Whitman, A. (Eds.), *Isaac Newton: The Principia: A new translation*. University of California Press. (1999).

Ney, H., Essen, U., & Kneser, R. (1994). On structuring probabilistic dependencies in stochastic language modelling. *Computer Speech and Language*, *8*, 1–38.

Niesler, T. (1997). *Category-based statistical language models*. Ph.D. thesis, University of Cambridge.

Niyogi, P., & Berwick, R. C. (2000). Formal models for learning in the principle and parameters framework. In Broeder, & Murre (Broeder & Murre, 2000), pp. 225–243.

Normandin, Y. (1996). Maximum mutual information estimation of Hidden Markov Models. In Lee, C.-H., Soong, F. K., & Paliwal, K. K. (Eds.), *Automatic Speech and Speaker Recognition*, pp. 58–81 Norwell, MA. Klewer Academic Publishers.

Oflazer, K. (1996). Error-tolerant finite-state recognition with applications to morphological analysis and spelling correction. *Computational Linguistics*, *22*(1), 73–89.

Osherson, D. N., Stob, M., & Weinstein, S. (1986). *Systems that Learn: An introduction to Learning Theory for Cognitive and Computer Scientists* (First edition). MIT Press.

Pereira, F. (2000). Formal grammar and information theory: together again?. *Philosophical Transactions of the Royal Society Series A*, *358*, 1239–1253.

Pereira, F., & Lee, L. (1999). Distributional similarity models: Clustering vs. nearest neighbours. In *Proceedings of the 37th annual meeting of the Association for Computational Linguistics*, pp. 33–40.

Pereira, F., & Schabes, Y. (1992). Inside-outside reestimation from partially bracketed corpora. In *Proceedings of the 30th annual meeting of the Association for Computational Linguistics*, pp. 128–135.

Pereira, F., Tishby, N., & Lee, L. (1993). Distributional clustering of English words. In *Proceedings of the 31st annual meeting of the Association for Computational Linguistics*.

Piattelli-Palmarini, M. (Ed.). (1980). *Language and Learning: The Debate between Jean Piaget and Noam Chomsky*. Routledge and Kegan Paul.

Piattelli-Palmarini, M. (1994). Ever since language and learning: Afterthoughts on the Piaget-Chomsky debate. *Cognition*, *50*, 315–346.

Pico, D., & Casacuberta, F. (2001). Some statistical-estimation methods for Stochastic Finite-State Transducers. *Machine Learning*, *44*, 12–141.

Pinker, S., & Prince, A. (1988). On language and connectionism: Analysis of a parallel distributed processing model of language acquisition. In Pinker, S., & Mehler, J. (Eds.), *Connections and Symbols*, pp. 73–193. MIT Press, Cambridge, MA.

Pinker, S. (1979). Formal models of language learning. *Cognition*, *7*, 217–282.

Pinker, S. (1989). *Learnability and Cognition*. MIT Press.

Pinker, S. (1991). Rules of language. *Science*, *153*, 530–535.

Pinker, S. (1994). *The Language Instinct*. Allen Lane.

Pinker, S. (1995). Language acquisition. In Osherson, D., Gleitman, L. R., & Liberman, M. (Eds.), *An invitation to cognitive science: Volume 1 Language* (2nd edition)., chap. 6, pp. 135–182. MIT Press.

Pinker, S. (1996). *Language Learnability and Language Development* (Second edition). Harvard University Press.

Plunkett, K., & Marchman, V. (1990). From rote learning to system building. Tech. rep. 9020, Center for Research in Language, UCSD.

Plunkett, K., & Nakisa, R. C. (1997). A connectionist model of the Arabic plural system. *Language and Cognitive Processes*, *12*(5/6), 807–836.

Pollard, C., & Sag, I. (1994). *Head Driven Phrase Structure Grammar*. University of Chicago Press.

Prasada, S., & Pinker, S. (1993). Generalisation of regular and irregular morphological patterns. *Language and Cognitive Processes*, *8*(1), 1–56.

Pullum, G. K. (1996). Learnability, hyperlearning and the argument from the poverty of the stimulus. In *Parasession on Learnability, 22nd Annual Meeting of the Berkeley Linguistics Society* Berkeley, California.

Pullum, G. K. (1997). The morpho-lexical nature of *to*-contraction. *Language*, *73*, 79–102.

Pullum, G. K., & Scholz, B. C. (2001). Empirical assessment of stimulus poverty arguments. *The Linguistic Review*, to appear.

Pylyshyn, Z., & Burkell, J. A. (1997). Searching through subsets: a test of the visual indexing hypothesis. *Spatial Vision*, *11*(2), 225–258.

Quine, W. V. (1969). Linguistics and philosophy. In Hook, S. (Ed.), *Language and Philosophy*, pp. 95–98. New York University Press.

Quine, W. V. O. (1975). Linguistics and philosophy. In Stich (Stich, 1975), pp. 200–202.

Quinlan, J. R. (1993). *C4.5 Programs for machine learning*. Morgan Kauffman.

Rabiner, L. R. (1989). A tutorial on Hidden Markov Models and selected applications in speech recognition. *Proceedings of the IEEE*, *77*(2), 257–285.

Raimy, E. (2000). *The Phonology and Morphology of reduplication*. No. 52 in Studies in Generative Grammar. Mouton de Gruyter.

Ramsey, W., & Stich, S. P. (1991). Connectionism and three levels of nativism. In Ramsey, W., Stich, S. P., & Rumelhart, D. E. (Eds.), *Philosophy and Connectionist Theory*, pp. 287–310. Lawrence Erlbaum Associates.

Redington, M., Chater, N., Huang, C., Chang, L., Finch, S., & Chen, K. (1995). The universality of simple distributional methods: Identifying syntactic categories in chinese. In *Proceedings of the Cognitive Science of Natural Language Processing* Dublin. Dublin City University.

Reilly, R., & Sharkey, N. (Eds.). (1989). *Connectionism and Language*. Lawrence Erbaum.

Rentzepopoulos, P. A., & Kokkinakis, G. K. (1996). Efficient multi-lingual phoneme-to-grapheme conversion based on HMM. *Computational Linguistics*, *22*(3), 351–376.

Rissanen, J. (1978). Modeling by shortest data description. *Automatica*, *14*, 465–471.

Ristad, E. S., & Yianilos, P. N. (1998). Learning string-edit distance. *IEEE Transactions on Pattern Analysis and Machine Intelligence*, *20*(5), 522–532.

Ristad, E. S. (1997). Finite growth models. Tech. rep. CS-TR-533-96, Department of Computer Science, Princeton University. revised in 1997.

Robbins, H., & Monro, S. (1951). A stochastic approximation method. *Annals of Mathematical Statistics*, *22*, 400–407.

Rosenblatt, M. (1974). *Random Processes* (2nd edition). Springer-Verlag.

Rumelhart, D. E., & McClelland, J. L. (1986a). On learning past tenses of English verbs. In Rumelhart, & McClelland (Rumelhart & McClelland, 1986b), pp. 216–271.

Rumelhart, D. E., & McClelland, J. L. (Eds.). (1986b). *Parallel Distributed Processing*, Vol. 2. MIT Press, Cambridge, MA.

Saffran, J. R., Aslin, R. N., & Newport, E. L. (1996). Statistical learning by eight month old infants. *Science*, *274*, 1926–1928.

Sakakibara, Y., Brown, M., Hughey, R., Mian, I., Sjolander, D., Underwood, R. D., & Haussler, D. (1994). Stochastic context-free grammars for tRNA modeling. *Nucleic Acids Research*, *22*(23), 5112–5120.

Sampson, G. (1995). *English for the Computer: The SUSANNE Corpus and Analytic Scheme*. Clarendon Press.

Sampson, G. (1987). Probabilistic models of analysis. In Garside et al. (Garside et al., 1987), pp. 16–29.

Sampson, G. (1989). Language acquisition: growth or learning?. *Philosophical papers*, *18*, 203–240.

Sampson, G. (1997). *Educating Eve*. Cassell.

Savitch, W. J. (Ed.). (1987). *The Formal Complexity of Natural Language*. D. Reidel.

Schone, P., & Jurafsky, D. (2000). Knowledge-free induction of morphology using latent semantic analysis. In *Proceedings of CoNLL-2000 and LLL-2000*, pp. 67–72 Lisbon, Portugal.

Schütze, C. T. (1996). *The Empirical Basis of Linguistics*. University of Chicago Press.

Schütze, H. (1993). Part of speech induction from scratch. In *Proceedings of the 31st annual meeting of the Association for Computational Linguistics*, pp. 251–258.

Schütze, H. (1997). *Ambiguity Resolution in Language Learning*. CSLI Publications.

Scott, D. W. (1992). *Multivariate density estimation: Theory, Practice and Visualization*. Wiley Interscience.

Shames, G. H., Wiig, E. H., & Secord, W. A. (Eds.). (1998). *Human Communication Disorders: An introduction* (5th edition). Allyn and Bacon.

Silvey, S. D. (1975). *Statistical Inference*. Chapman and Hall.

Simon, H. (1962). The architecture of complexity. *Proceedings of the American Philosophy Society*, *106*, 476–482.

Sinkhorn, R. (1964). A relation between arbitrary positive matrices and doubly stochastic matrices. *Annals of Mathematical Statistics*, *35*(2).

Skinner, B. F. (1957). *Verbal Behavior*. Appleton Century Crofts.

Smolensky, P. (1996). On the comprehension/production dilemma in child language. Tech. rep. JHU-COGSCI-96-1, Department of Cognitive Science, John Hopkins University.

Snover, M. G., & Brent, M. R. (2001). A Bayesian model for morpheme and paradigm identification. In *Proceedings of ACL 2001*, pp. 482–490.

Soules, G. W. (1991). The rate of convergence of Sinkhorn balancing. *Linear Algebra and Its Applications*, *150*(3).

Spencer, A. (1991). *Morphological Theory*. Basil Blackwell.

Steedman, M. (1990). Syntax and intonational structure in a combinatory grammar. In Altmann, G. T. M. (Ed.), *Cognitive Models of Speech Processing: Psycholinguistic and Computational Perspectives*, pp. 457–482. MIT Press.

Steedman, M. (1999). Connectionist sentence processing in perspective. *Cognitive Science*, *23*, 615–634.

Steedman, M. (2000). Information structure and the syntax-phonology interface. *Linguistic Inquiry*, *31*(4), 649–689.

Stich, S. P. (Ed.). (1975). *Innate Ideas*. University of California Press.

Stolcke, A. (1994). *Bayesian Learning of Probabilistic Language Models*. Ph.D. thesis, Dept. of Electrical Engineering and Computer Science, University of California at Berkeley.

Stolcke, A. (1995). An efficient probabilistic context-free parsing algorithm that computes prefix probabilities. *Computational Linguistics*, *21*(2), 165–202.

Theeramunkong, T., & Okumura, M. (1997). Grammar acquisition based on clustering analysis and its application to statistical parsing. In *Proceedings of the 5th Workshop on Very Large Corpora*, pp. 31–40.

Theron, P., & Cloete, I. (1997). Automatic acquisition of two-level morphological rules. In *Proceedings of the Fifth Conference on Applied Natural Language Processing*, pp. 103–110 Washington.

Tjong Kim Sang, E. (1998). *Machine learning of Phonotactics*. Ph.D. thesis, University of Groningen.

Torkkola, K. (1993). An efficient way to learn English grapheme-to-phoneme rules automatically. In *ICASSP-93. 1993 IEEE International Conference on Acoustics, Speech, and Signal Processing (Cat. No. 92CH3252-4)*, Vol. 2, pp. 199–202 New York, NY, USA. IEEE.

Tyack, P. L. (1986). Population biology, social behavior and communication in whales and dolphins. *Trends in Ecology and Evolution*, 144–150.

Valiant, L. G. (1979). The complexity of computing the permanent. *Theoretical Computer Science*, *8*(2), 189–201.

van den Bosch, A., & Daelemans, W. (1999). Memory-based morphological analysis. In *Proceedings of the 37th Annual Meeting of the Association for Computational Linguistics*, pp. 285–292.

van Everbroeck, E. (1999). Could Sarah read the Wall Street Journal?. *The Newsletter of the Center for Research in Language*, *11*(7), 3–14.

van Noord, G., & Gerdemann, D. (2001). Finite state transducers with predicates and identities. *Grammars*.

van Zaanen, M. (2000). ABL: Alignment-based learning. In *COLING 2000 - Proceedings of the 18th International Conference on Computational Linguistics*.

van Zaanen, M., & Adriaans, P. (2001). Comparing two unsupervised grammar induction systems: Alignment-based learning vs. Emile. Research report series 2001.05, School of Computing, University of Leeds.

Vapnik, V. N. (1998). *Statistical Learning Theory*. John Wiley.

Vargha-Khadem, F., & Passingham, R. E. (1990). Speech and language deficits. *Nature*, *346*, 226.

Versteegh, K. (1997). *The Arabic Language*. Edinburgh University Press.

Villavicencio, A. (2000). The acquisition of word order by a computational learning system. In *Proceedings of CoNLL-2000 and LLL-2000*, pp. 209–218 Lisbon, Portugal.

Wallace, C. S., & Boulton, D. M. (1968). An information measure for classification. *The Computer Journal*, *11*(2), 185–194.

Weaver, W. (1949). Translation. In Locke, & Booth (Locke & Booth, 1955), pp. 15–23.

Wehr, H. (1979). *Arabic-English Dictionary* (Fourth edition). Spoken Language Services Inc. Edited by J. Milton Cowan.

Wenner, A. M., & Wells, P. H. (1990). *Anatomy of a Controversy: The Question of a "language" among Bees*. Columbia University Press.

Wermter, S., Riloff, E., & Scheler, G. (Eds.). (1996). *Connectionist, Statistical and Symbolic Approaches to Learning for Natural Language Processing*. Springer-Verlag.

Westermann, G., & Goebel, R. (1995). Connectionist rules of language. In *Proceedings of the 17th annual conference of the Cognitive Science Society*, pp. 236–241.

Wexler, K., & Culicover, P. W. (1980). *Formal Principles of Language Acquisition*. MIT Press.

Wolf, D. R., & Wolpert, D. H. (1992). Estimating functions from a finite set of samples, part 1: Bayes estimators and the shannon entropy. Tech. rep. LA-UR-92-4369, Los Alamos National Laboratory.

Wolf, D. R., & Wolpert, D. H. (1993). Estimating functions from a finite set of samples, part 2: Bayes estimators for mutual information, chi-squared etc.. Tech. rep. LA-UR-93-833, Los Alamos National Laboratory.

Wolff, J. G. (1977). The discovery of segments in natural language. *British Journal of Psychology*, *68*, 97–106.

Wolff, J. G. (1980). Language acquisition and the discovery of phrase structure. *Language and Speech*, *23*(3), 255–269.

Wolff, J. G. (1988). Learning syntax and meanings through optimization and distributional analysis. In Schelsinger, I. M., Levy, Y., & Braine, M. D. S. (Eds.), *Categories and Processes in Language Acquisition*, pp. 85–98. Lawrence Erlbaum.

Wolff, J. G. (1991). *Towards a theory of cognition and computing*. Ellis Horwood.

Wolpert, D. H., & Macready, W. G. (1997). No free lunch theorems for optimization. *IEEE Transactions on Evolutionary Computation*, *1*(1), 67–82.

Wright, W. (1967). *A Grammar of the Arabic Language* (Third edition). Cambridge University Press.

Wu, D. (1995). Stochastic inversion transduction grammars, with application to segmentation, bracketing, and alignment of parallel corpora. In *IJCAI-95*, pp. 1328–1335 Montreal.

Wu, D. (1997). Stochastic inversion transduction grammars and bilingual parsing of parallel corpora. *Computational Linguistics*, *23*(3), 377–403.

Yang, C. D. (1999). A selectionist theory of language acquisition. In *Proceedings of ACl 1999*, pp. 429–430.

Yarowsky, D., & Wicentowski, R. (2000). Minimally supervised morphological analysis by multimodal alignment. In *Proceedings of ACL 2000*, pp. 207–216 Hong Kong.

Zavrel, J., & Daelemans, W. (1999). Recent advances in memory-based part-of-speech tagging. Tech. rep. 9903, ILK, Tilburg University.

# Index

# Appendix A

# CLAWS-5 tag set

**AJ0** Adjective

**AJC** Comparative adjective

**AJS** Superlative adjective

**AT0** Article

**AV0** Adverb

**AVP** Adverb particle

**AVQ** *wh*-adverb

**CJC** Coordinating Conjunction

**CJS** subordinating conjunction

**CJT** *that*

**CRD** cardinal number

**DPS** possessive determiner

**DT0** determiner e.g. *this*

**DTQ** *wh*-determiner

**EX0** Expletive *there*

**ITJ** Interjection

**NN0** noun without number marking

**NN1** singular noun

**NN2** plural noun

**NP0** proper noun

**ORD** ordinal e.g.*first*

**PNI** indefinite pronoun

**PNP** personal pronoun

**PNQ** *wh*-pronoun

**PNX** reflexive pronoun

**POS** possessive particle *'s*

**PRF** *of*

**PRP** preposition

**TO0** *to* as infinitive particle

**UNC** unclassified

**VBB VBD VBG VBI VBN VBZ** forms of verb *be*

**VDB VDD VDG VDI VDN VDZ**  forms of verb *do*

**VHB VHD VHG VHI VHN VHZ**  forms of verb *have*

**VM0**  modal auxiliary

**VVB VVD VVG VVI VVN VVZ**  forms of other verbs

**XX0**  negative particle

# Appendix B

# Syntactic rules

This is the list of all syntactic rules produced by the unsupervised induction algorithm presented in Chapter 7, together with the number of times each rules was applied in the corpus.

```
316191 NP → THE TIME
196481 PP → OF NP
156893 NP → LONDON
108564 NP → NP OF NP
 86400 NP → NEW PEOPLE
 75279 NP → THE GROUP
 73967 NP → THE PEOPLE
 72857 NP → YOU
 71879 NP → THE NEW TIME
 67493 NP → THE NEW GROUP
 63296 NP_CONJ → NP ,
 58561 NP → THE NEW PEOPLE
 57518 NP → MORE YEARS
 55748 NP_CONJ → NP AND
 55689 NP → ONE OF NP
 54868 S_OR_NP → BUT NP
 50926 NP → NEW GROUP
 49379 NP → NP AND NP
 48093 NP → THE PARTY
 46667 NP → NP , NP
 38420 NP → ALL NP
 38198 PP_CONJ → OF NP ,
 35611 NP → NP UP
 35249 NP → NP NP
 33347 NP → NP GROUP
 32750 NP → NP OF LONDON
 29651 NT-ONEOF → ONE OF
 28390 PP → OF LONDON
 28064 PP_CONJ → OF NP_CONJ
 26637 AP_CONJ → NEW ,
 25809 AP_CONJ → NEW AND
 24878 NT-&BQUO → HOWEVER ,
 23975 N1 → PARTY GROUP
 22295 N1 → NEW TIME
 20934 NP_PROPER → JOHN PHILIP
 20775 S_OR_NP → &BQUO NP
 20218 NP → NP OF PEOPLE
 19261 PP_CONJ → UP ,
 19135 NP_CONJ → NP BUT
 18929 NP → PEOPLE OF NP
```

```
18828 NT-&BQUO → S_OR_NP ,
18494 NP → LONDON , NP
18425 NT-TOSEE → TO SEE
18129 NP → THE PARTY GROUP
17793 NP_PROPER → JOHN THATCHER
17257 NP → NP VP
16079 NP_PROPER → MR PHILIP
15680 NP → THE TIME OF THE TIME
15667 NP → MORE NP
15553 NP → NP OF NEW PEOPLE
14968 S_OR_NP → HE TOOK NP
14802 NP → NP &EQUO
14801 N1_CONJ → PEOPLE AND
14774 N1_CONJ → PEOPLE ,
14640 S → IT VP
14432 NP → NP OF NP OF NP
14156 N1 → LAST TIME
13994 NP → NP PEOPLE
13905 NP → ALL THE TIME
13860 ADVP_NEG → NOT NOT
13845 NP → THE TIME OF LONDON
13815 AP_CONJ → ONE OF NP ,
13784 VP → 'S NP
13717 NT-HESAID → HE SAID
13713 NP → NP TO NP
13463 VP → IS POSSIBLE
13405 VP → CAME PP
13379 NP → NP AND PEOPLE
13289 NT-GOINGTOUSE → GOING TO BE
13091 NP → NP , BUT NP
13084 NP → THE ONE OF NP
13046 AP_CONJ → POSSIBLE ,
12992 AP → NOT POSSIBLE
12980 NP → NP ONE OF NP
12688 NP → THE LAST TIME
12620 NP → THE NP
12361 NP → THE GROUP OF LONDON
12215 NP → THE USE
12127 NP_V → NOT MUCH
11956 NOT_NP → NOT NP
11802 NP → THE NEW PARTY
11760 NP_CONJ → NP , BUT
```

```
11738 NT-THENEEDTO → THE NEED TO
11622 S_OR_NP → HOWEVER , NP
11597 NP_CONJ → NP , AND
11512 PP_CONJ → TO NP ,
11494 NP → NP , OF NP
11399 COMPLEX_AP → MORE NEW
11329 VP → WILL BE NP
11197 NP → NP IS NP
11185 NP → THE YEARS
11045 NP → THE FACT
10925 NP → SOMETHING
10501 NT-NOTHING → NOT AP_CONJ
10095 NP → ALL PEOPLE
10089 NOT_NP → NT-GOINGTOUSE NP
9966 S_OR_NP → IT IS NP
9966 NT-GROUP, → GROUP ,
9907 S → HE VP
9876 NP → NP , AND NP
9850 PP_CONJ → UP PP_CONJ
9832 VP → IS AP
9798 NP → NP OF THE PARTY
9659 NP → LONDON OF NP
9611 NP_V → HE THOUGHT
9574 NP → MORE OF NP
9463 NT-THENEEDTO → NP_CONJ NOT
9441 VP → ARE NP
9436 AP_CONJ → MADE ,
9331 NP → NP 'S TIME
9282 NP_CONJ → LONDON ,
9135 CONJUNCT → , HOWEVER ,
9106 NP → THE NEW PARTY GROUP
9013 ERROR1 → , NP ,
8908 NP → NEW PEOPLE OF NP
8683 NP → THE TIME UP
8581 AP_CONJ → NOT ,
8578 S_OR_NP → NP TOOK NP
8504 COMPLEX_AP → &BQUO NEW
8469 NP → NP TIME
8453 NOT_NP → GOING TO NP
8371 COMPLEX_AP → AP_CONJ NEW
8328 AP → MORE POSSIBLE
8254 NT-THENEEDTO → NP NOT
```

```
8248 NP_CONJ → LONDON AND
8235 PP_CONJ → UP AND
8216 NP → THE NEW NEW GROUP
8197 NP → THE NEW ONE OF NP
8001 VP → IS ONE PP
7867 PP_CONJ → OF NP , BUT
7860 PP_CONJ → OF LONDON ,
7761 S → NP_V NP
7725 NP_PROPER → MR THATCHER
7720 NP_V → NP MADE
7717 NP_V → HE TOOK
7613 NP → NP LATER
7611 PP_CONJ → OF NP , AND
7152 NP → NEW NEW PEOPLE
7051 NT-ONEOF → PEOPLE OF
7010 N1 → ONE OF PEOPLE
6952 NP → THE GROUP OF THE TIME
6923 PP → OF YOU
6877 AP → NOT NEW
6762 N1_CONJ → N1 AND
6741 N1 → PARTY PEOPLE
6705 NP → THE LONDON GROUP
6591 NP_V → NP IS MADE
6558 NT-GROUP, → GROUP PP
6418 NP → THE NEW GROUP OF LONDON
6403 N1 → NEW PARTY GROUP
6385 NP → PHILIP 'S TIME
6274 NOT_NP → NOT ONE PP
6255 S_OR_NP → THERE IS NP
6233 PP_CONJ → PP THAT
6137 NP → THE PEOPLE OF LONDON
6108 VP → IS NOT NP
6099 NP → YOU UP
6073 NT-GOINGTOUSE → MORE THAN
6034 NT-&BQUO → S_OR_NP AND
5999 PP_CONJ → TO BE NP ,
5975 NOT_NP → NOT PP
5939 NP → THE NEW NEW TIME
5920 AP_CONJ → MADE NP ,
5812 NP → THE NEW LONDON
5772 PP_CONJ → PP_CONJ BUT
5768 AP_CONJ → POSSIBLE AND
```

```
5718 NT-NOTHING → NOT NP_CONJ
5710 NP_V → HE MADE
5705 NP → THE MORE PEOPLE
5674 ERROR1 → OF FACT ,
5669 NT-GOINGTOUSE → GOING TO USE
5584 NP → THE MORE YEARS
5578 NP → THE NEW NEW PEOPLE
5570 NP_PROPER → JOHN PEOPLE
5460 NP → THAT TIME
5444 NP_V → HE BE
5436 NP → THE PEOPLE OF THE TIME
5345 NT-ONEOF → YEARS ,
5295 ERROR1 → OF THE AP_CONJ
5226 NP → MORE PEOPLE
5185 ERROR1 → &EQUO ,
5185 CONJUNCT → , OF
5126 NP → THE ONE OF LONDON
5113 NP → ALL YEARS
4965 CONJUNCT → , PP_CONJ
1665 AP → BETTER
```